\documentclass[oneside]{book}
\usepackage{roboto}

\usepackage[english]{babel}
\usepackage[letterpaper,top=2cm,bottom=2cm,left=3cm,right=3cm,marginparwidth=1.75cm]{geometry}

\usepackage{arabtex}
\usepackage{utf8}
\setcode{utf8}

\usepackage{listings}  
\usepackage{pdflscape}
\usepackage{lscape}
\usepackage{rotating}
\usepackage{longtable}
\usepackage{array}
\usepackage{float}
\usepackage{amsmath}
\usepackage{graphicx}
\usepackage{inconsolata}
\usepackage[colorlinks=true, allcolors=blue]{hyperref}
\usepackage{enumitem}

\usepackage[numbers,sort&compress]{natbib}

\usepackage{tikz} 

\usepackage{listings}
\usepackage{xcolor}

\usepackage[T1]{fontenc}
\usepackage{IEEEtrantools}  
\usepackage{epigraph}  

\definecolor{bgcolor}{rgb}{0.97,0.97,0.97}
\definecolor{codeblue}{rgb}{0.1,0.1,0.8}
\definecolor{codegreen}{rgb}{0,0.4,0}
\definecolor{codegray}{rgb}{0.4,0.4,0.4}
\definecolor{codepurple}{rgb}{0.5,0,0.5}
\definecolor{codered}{rgb}{0.6,0.2,0.2}
\definecolor{lightgray}{rgb}{0.9,0.9,0.9}
\definecolor{darkgray}{rgb}{0.6,0.6,0.6} 

\makeatletter
\renewcommand{\paragraph}{%
  \@startsection{paragraph}{4}{\z@}{1ex}{-1em}{\normalfont\normalsize\bfseries\color{gray}}}
\makeatother

\lstdefinestyle{python}{
    language=Python,
    basicstyle=\ttfamily\small\color{black}\usefont{T1}{zi4}{m}{n},  
    keywordstyle=\bfseries\color{codeblue},  
    stringstyle=\color{codegreen},  
    commentstyle=\slshape\color{codegray},  
    showstringspaces=false,
    numbers=left,
    numberstyle=\tiny\color{codegray},  
    stepnumber=1,
    numbersep=8pt,
    frame=single,
    rulecolor=\color{darkgray},  
    breaklines=true,
    backgroundcolor=\color{bgcolor},
    tabsize=4,
    captionpos=b,
    morekeywords={self}, 
}

\lstdefinestyle{text}{
    language=,
    basicstyle=\ttfamily\small\color{black}\usefont{T1}{zi4}{m}{n},  
    stringstyle=\color{codered},
    commentstyle=\color{codegray},
    showstringspaces=false,
    numbers=none,
    frame=single,
    rulecolor=\color{lightgray},  
    frameround=tttt,
    breaklines=true,
    backgroundcolor=\color{bgcolor},
    tabsize=4,
    captionpos=b,
}

\lstdefinestyle{cmd}{
    language=bash,
    basicstyle=\ttfamily\small\color{black}\usefont{T1}{zi4}{m}{n},  
    keywordstyle=\bfseries\color{blue},
    stringstyle=\color{codegreen},
    commentstyle=\itshape\color{gray},
    showstringspaces=false,
    numbers=none,
    frame=single,
    rulecolor=\color{darkgray},  
    breaklines=true,
    backgroundcolor=\color{bgcolor},
    tabsize=4,
    captionpos=b,
}

\lstdefinestyle{sql}{
    language=SQL,  
    basicstyle=\ttfamily\small\color{black}\usefont{T1}{zi4}{m}{n},  
    keywordstyle=\bfseries\color{codeblue},  
    stringstyle=\color{codegreen},  
    commentstyle=\slshape\color{codegray},  
    showstringspaces=false,  
    numbers=left,  
    numberstyle=\tiny\color{codegray},  
    stepnumber=1,  
    numbersep=8pt,  
    frame=single,  
    rulecolor=\color{darkgray},  
    breaklines=true,  
    backgroundcolor=\color{bgcolor},  
    tabsize=4,  
    captionpos=b,  
    morekeywords={
        SELECT, INSERT, DELETE, UPDATE, FROM, WHERE, AND, OR, JOIN, ON, CREATE, INDEX, TABLE, VALUES, INTO, AS, DISTINCT, ORDER, BY, GROUP, HAVING, LIMIT, OFFSET, UNION, DROP, ALTER, TRUNCATE, RENAME, PRIMARY, FOREIGN, KEY, CONSTRAINT, NULL, NOT, DEFAULT, AUTO_INCREMENT, UNIQUE, CHECK,
        db, find, insertOne, updateOne, deleteOne, collection, aggregate, match, project, sort, limit, pipeline, insertMany, updateMany, deleteMany, findOne, $lookup, $group,
        GET, SET, DEL, HGET, HSET, HDEL, LPUSH, RPUSH, LPOP, RPOP, SADD, SREM, PUBLISH, SUBSCRIBE, EXPIRE, TTL, FLUSHDB, FLUSHALL, INCR, DECR,
        SELECT, INSERT, UPDATE, DELETE, FROM, WHERE, AND, OR, USE, KEYSPACE, CREATE, ALTER, DROP, TRUNCATE, TABLE, INDEX, PRIMARY, FOREIGN, KEY, WITH, CLUSTERING, ORDER, BY, ASC, DESC, LIMIT, BATCH, APPLY, TOKEN, CONSISTENCY, QUORUM, LOCAL_QUORUM, ANY, ALL, ONE, TWO, THREE,
        MATCH, CREATE, MERGE, RETURN, DELETE, DETACH, REMOVE, SET, FOREACH, UNWIND, WITH, ORDER, BY, ASCENDING, DESCENDING, SKIP, LIMIT, UNION, ALL, OPTIONAL, DISTINCT, WHERE, AND, OR, IN, STARTS, ENDS, CONTAINS, EXISTS, IS, NULL,
        SELECT, INSERT, DELETE, UPDATE, FROM, WHERE, AND, OR, JOIN, ON, CREATE, INDEX, TABLE, VALUES, INTO, AS, DISTINCT, ORDER, BY, GROUP, HAVING, LIMIT, OFFSET, UNION, DROP, ALTER, TRUNCATE, RENAME, PRIMARY, FOREIGN, KEY, CONSTRAINT, NULL, NOT, DEFAULT, AUTO_INCREMENT, UNIQUE, CHECK, SEQUENCE, SYNONYM, PACKAGE, FUNCTION, PROCEDURE, TRIGGER, VIEW, GRANT, REVOKE, COMMIT, ROLLBACK,
        SELECT, INSERT, DELETE, UPDATE, FROM, WHERE, AND, OR, JOIN, ON, CREATE, INDEX, TABLE, VALUES, INTO, AS, DISTINCT, ORDER, BY, GROUP, HAVING, LIMIT, OFFSET, UNION, DROP, ALTER, TRUNCATE, RENAME, PRIMARY, FOREIGN, KEY, CONSTRAINT, NULL, NOT, DEFAULT, AUTO_INCREMENT, UNIQUE, CHECK, SERIAL, BIGSERIAL, RETURNING, DO, LANGUAGE, PLPGSQL, BEGIN, END, IMMUTABLE, VOLATILE,
        SELECT, INSERT, DELETE, UPDATE, FROM, WHERE, AND, OR, JOIN, ON, CREATE, INDEX, TABLE, VALUES, INTO, AS, DISTINCT, ORDER, BY, GROUP, HAVING, LIMIT, OFFSET, UNION, DROP, ALTER, TRUNCATE, RENAME, PRIMARY, FOREIGN, KEY, CONSTRAINT, NULL, NOT, DEFAULT, AUTO_INCREMENT, UNIQUE, CHECK, INCREMENT, ENGINE, CHARSET, COLLATE, COMMENT
    }  
}

\lstdefinestyle{html}{
    language=HTML,
    basicstyle=\ttfamily\small\color{black}\usefont{T1}{zi4}{m}{n},  
    keywordstyle=\bfseries\color{codeblue},  
    stringstyle=\color{codegreen},  
    commentstyle=\slshape\color{codegray},  
    showstringspaces=false,
    numbers=left,
    numberstyle=\tiny\color{codegray},  
    stepnumber=1,
    numbersep=8pt,
    frame=single,
    rulecolor=\color{darkgray},  
    breaklines=true,
    backgroundcolor=\color{bgcolor},
    tabsize=4,
    captionpos=b,
    morekeywords={<!DOCTYPE,html,head,body,div,span,a,img,href,src,script,style}, 
}

\title{Deep Learning and Machine Learning - Natural Language Processing: From Theory to Application}
\author{
    Keyu Chen\textsuperscript{*} \\ 
    \textit{Georgia Institute of Technology} \\
    kchen637@gatech.edu
    \and
    Cheng Fei\textsuperscript{*} \\
    \textit{Cornell University} \\
    cf482@cornell.edu
    \and
    Ziqian Bi \\
    \textit{Indiana University} \\
    bizi@iu.edu
    \and
    Junyu Liu \\ 
    \textit{Kyoto University} \\
    liu.junyu.82w@st.kyoto-u.ac.jp
    \and
    Benji Peng \\ 
    \textit{AppCubic} \\
    benji@appcubic.com
    \and
    Sen Zhang \\ 
    \textit{Rutgers University} \\
    sen.z@rutgers.edu
    \and
    Xuanhe Pan \\ 
    \textit{University of Wisconsin-Madison} \\
    xpan73@wisc.edu
    \and
    Jiawei Xu \\ 
    \textit{Purdue University} \\
    xu1644@purdue.edu
    \and
    Jinlang Wang \\ 
    \textit{University of Wisconsin-Madison} \\
    jinlang.wang@wisc.edu
    \and
    Caitlyn Heqi Yin \\
    \textit{University of Wisconsin-Madison} \\
    hyin66@wisc.edu
    \and
    Yichao Zhang \\
    \textit{The University of Texas at Dallas} \\
    yichao.zhang.us@gmail.com
    \and
    Pohsun Feng \\
    \textit{National Taiwan Normal University} \\
    41075018h@ntnu.edu.tw
    \and
    Yizhu Wen \\
    \textit{University of Hawaii} \\
    yizhuw@hawaii.edu
    \and
    Tianyang Wang \\ 
    \textit{Xi'an Jiaotong-Liverpool University} \\
    Tianyang.Wang21@student.xjtlu.edu.cn
    \and
    Ming Li \\ 
    \textit{Georgia Institute of Technology} \\
    mli694@gatech.edu
    \and
    Jintao Ren \\
    \textit{Aarhus University } \\
    jintaoren@clin.au.dk
    \and
    Qian Niu \\ 
    \textit{Kyoto University} \\
    niu.qian.f44@kyoto-u.jp
    \and
    Silin Chen \\
    \textit{Zhejiang University } \\
    A1033439225@gmail.com
    \and
    Weiche Hsieh\\
    \textit{National Tsing Hua University} \\
    s112033645@m112.nthu.edu.tw
    \and
    Lawrence K.Q. Yan \\
    \textit{Hong Kong University of Science and Technology} \\
    kqyan@connect.ust.hk
    \and
    Chia Xin Liang \\
    \textit{Independent Researcher} \\
    cxldun@gmail.com
    \and
    Hong-Ming Tseng \\
    \textit{School of Visual Arts} \\
    htseng@sva.edu
    \and
    Xinyuan Song \\
    \textit{Emory University} \\
    songxinyuan@pku.edu.cn
    \and
    Zekun Jiang \\
    \textit{Sichuan University} \\
    zekun\_jiang@163.com
    \and
    Ming Liu{$\dagger$} \\ 
    \textit{Purdue University} \\
    liu3183@purdue.edu
}
\date{} 
\begin{document}
\maketitle
\thispagestyle{plain}

\begingroup
    \renewcommand\thefootnote{}\footnote{
    \textsuperscript{*} Equal contribution \\
    \textsuperscript{$\dagger$} Corresponding author
}
\addtocounter{footnote}{0}
\endgroup

\epigraph{"Of course, linguists do not generally keep NLP in mind as they do their work, so not all of linguistics should be expected to be useful."}{\textit{Noah A. Smith}}

\epigraph{"Increasing F-score is often not a scientific contribution but how you did it may be a scientific contribution."}{\textit{Mark Johnson}}

\epigraph{"In principle, one could model the distribution of dependency parses in any number of sensible or perverse ways."}{\textit{Jason Eisner}}

\epigraph{"[...] interpreting the world in the light of your preconceptions; those preconceptions then reinforce how you reinterpret your evidence, and those strengthen your preconception [...] The model is feeding itself, is eating its own waste."}{\textit{Jason Eisner}}

\tableofcontents  
\cleardoublepage

\setcounter{part}{3} 
\part{Advancing Your Skills}

\chapter{Natural language processing}

\section{Introduction to Natural Language Processing}
\subsection{What is Natural Language Processing?}
\subsection{The History of NLP}

The history of NLP can be traced back to the 1950s with the advent of early AI research. Over the years, it has evolved through multiple stages:

\textbf{1. 1950s – Rule-Based Systems:}
The earliest NLP systems were based on sets of handcrafted rules that allowed computers to parse sentences, perform translations, and answer questions. The most famous example from this period is the \textit{Turing Test} proposed by Alan Turing to evaluate machine intelligence. \cite{Bishop2010TheIG}

\textbf{2. 1960s – 1970s: Symbolic Approaches and ELIZA}
During this period, symbolic AI dominated NLP research. The famous chatbot ELIZA (created in 1966 by Joseph Weizenbaum) \cite{Weizenbaum1966ELIZAaCP} simulated conversation using basic pattern matching but was unable to understand the meaning of the input.

\textbf{3. 1980s – Probabilistic Models:}
The 1980s saw the introduction of probabilistic models in NLP, moving from rule-based approaches to statistical methods. This shift occurred because of the limitations of hand-coded rules when dealing with real-world language complexity\cite{nadkarni2011natural}. Hidden Markov Models (HMMs) and early machine learning algorithms were applied to tasks like speech recognition and part-of-speech tagging\cite{kang2020natural}.

\textbf{4. 1990s – Rise of Machine Learning:}
Machine learning techniques, particularly supervised learning algorithms, became widely adopted in the 1990s. This allowed for more flexible models and better handling of language variations. Corpora (large collections of text) were used for training machine learning models. \cite{Matsumoto2000UsingML}

\textbf{5. 2010s – Deep Learning and Neural Networks:}
The introduction of deep learning techniques, specifically recurrent neural networks (RNNs) \cite{Salehinejad2017RecentAI} and transformers \cite{vaswani2017attention}, revolutionized NLP. Models like word2vec, BERT \cite{Devlin2019BERTPO}, and GPT \cite{Yenduri2023GPTP} have significantly advanced NLP's capabilities, making it possible for machines to generate and understand text with human-like fluency.

\textbf{6. 2020s – Large Language Models (LLMs):}
In recent years, Large Language Models (LLMs) have dramatically advanced NLP. These models, such as GPT-3, GPT-4, and OpenAI's ChatGPT, are trained on vast datasets and use billions of parameters to understand and generate text. LLMs leverage transformer architectures, which allow them to process and generate human-like text at an unprecedented scale\cite{chowdhary2020natural}. They are capable of a wide range of tasks, including question answering, text summarization, translation, and creative writing. Their ability to generate contextually relevant and coherent text without explicit rule-based systems marks a significant shift in NLP\cite{fanni2023natural}. With techniques like transfer learning and self-supervised learning, LLMs require less task-specific training data, making them highly adaptable across domains.

\subsection{Applications of NLP in Various Domains}

Natural Language Processing (NLP) has become a cornerstone technology in multiple industries, revolutionizing the way machines and humans interact\cite{joshi1991natural}. As the field continues to evolve, its applications are expanding across various domains, enhancing efficiencies, personalizing experiences, and unlocking new opportunities for data-driven decision-making\cite{chopra2013natural}.

\textbf{1. Healthcare:}
NLP has made significant strides in healthcare by enabling better understanding and processing of unstructured medical data, such as patient records, research articles, and clinical notes. \cite{niu2024textmultimodalityexploringevolution} Some key applications include:
Clinical Decision Support Systems (CDSS): NLP is used to extract vital patient information from electronic health records (EHRs) \cite{yan2024largelanguagemodelbenchmarks} and match them with relevant medical literature to support doctors in diagnosing and choosing appropriate treatments.
Medical Research: NLP tools are employed to mine vast amounts of medical literature, identifying patterns, trends, and relationships across diseases and treatments.
Patient-Doctor Communication: Automated systems using NLP can interpret and respond to patient queries, providing health information, scheduling appointments, and managing follow-ups. \cite{niu2024largelanguagemodelscognitive}

\textbf{2. Business and Finance:}
In business and finance, NLP is transforming how companies interact with data and customers, and how they make decisions based on textual information.
Sentiment Analysis: NLP models analyze customer feedback, social media, and financial news to gauge public sentiment about products, services, or market conditions, guiding marketing strategies and investment decisions\cite{yim2016natural}.
Automated Trading Systems: NLP-powered tools can process and analyze financial reports, earnings calls, and market news in real-time to provide actionable insights or trigger automated trades\cite{grosz1986readings}.
Chatbots and Virtual Assistants: NLP has enabled the rise of virtual assistants like Siri, Alexa, and corporate chatbots that handle customer queries, complaints, and even complete transactions, improving customer service and reducing human workload\cite{liddy2001natural}.

\textbf{3. Legal and Compliance:}
NLP is revolutionizing legal practices by automating document review and legal research, which are traditionally labor-intensive tasks.\cite{locke2021natural}
Contract Analysis: NLP algorithms can review contracts to detect clauses, inconsistencies, and potential risks, helping lawyers and organizations manage legal obligations more efficiently\cite{cambria2014jumping}.
Legal Research: Legal professionals use NLP tools to extract relevant case law, legislation, and legal precedents from massive legal databases\cite{khurana2023natural}.
Compliance Monitoring: In regulated industries such as finance and pharmaceuticals, NLP tools can sift through internal communications and external data sources to ensure compliance with laws and regulations, identifying potential breaches in real-time\cite{joseph2016natural}.

\textbf{4. Education:}
NLP is also making its mark in education, enhancing both teaching methods and learning experiences\cite{ibrahim2010class}.
Automated Essay Scoring: NLP models can assess the structure, grammar, and coherence of student essays, providing instant feedback and reducing the burden on educators\cite{reshamwala2013review}.
Language Learning: Tools like Duolingo leverage NLP to teach languages, offering real-time corrections and adaptive learning paths based on individual performance.\cite{hirschberg2015advances}
Personalized Learning: NLP algorithms analyze students' responses and adapt learning materials in real time, catering to each learner's pace and needs, thereby improving engagement and learning outcomes\cite{jones1994natural}.

\textbf{5. Customer Service and Support:}
Customer service has been transformed by the integration of NLP, enabling more efficient and personalized responses to customer queries.
Intelligent Virtual Assistants (IVAs): These systems, powered by NLP, handle customer inquiries by providing accurate answers and troubleshooting, often without human intervention. \cite{Chung2018IntelligentVA}
Speech Recognition for Call Centers: NLP tools convert voice interactions into text, allowing automated systems to analyze and respond to customer needs, enhancing call center productivity and customer satisfaction.
Feedback Analysis: Companies use NLP to process and analyze customer feedback, reviews, and surveys, allowing them to identify common pain points and improve product offerings or services\cite{wu2020deep}.

\textbf{6. Media and Entertainment:}
NLP applications in media and entertainment are helping content creators and distributors analyze trends and generate new content\cite{pons2016natural}.
Content Recommendation Systems: NLP helps streaming services like Netflix and Spotify analyze user preferences and recommend personalized content, keeping users engaged longer\cite{ruder2019transfer}.
Automated Content Creation: NLP models are increasingly being used to generate content, from news articles to creative writing, helping media companies meet the growing demand for content. \cite{Hastings2009AutomaticCG}
Transcription and Captioning: NLP tools like speech recognition systems automatically transcribe video or audio content, making it accessible to a wider audience and improving searchability. \cite{Ranchal2013UsingSR}

\textbf{7. E-Commerce:}
E-commerce platforms leverage NLP to enhance customer experience and improve operational efficiency.
Product Search and Recommendations: NLP algorithms improve the accuracy of search results on e-commerce websites by understanding the intent behind customer queries and offering personalized product suggestions\cite{jensen2012natural}.
Review Analysis: NLP is used to process thousands of customer reviews, extracting valuable insights on product performance, usability, and customer satisfaction\cite{alhawiti2014natural}.
Chatbots for Customer Queries: Many e-commerce platforms employ NLP-powered chatbots to provide real-time responses to customer queries, helping with order tracking, returns, and product inquiries. \cite{Pappula2023LLMsFC}

\textbf{8. Government and Public Policy:}
NLP is being applied in public sector settings to streamline operations and make governance more data-driven\cite{feldman1999nlp}.
Policy Analysis and Opinion Mining: NLP helps policymakers analyze public opinion from social media, news, and surveys, providing insights into public sentiment on various issues\cite{farzindar2015natural}.
Document Processing: Governments use NLP tools to process and organize large volumes of documents, such as legal texts, policy documents, and legislative records, enabling faster access to information and more informed decision-making\cite{friedman1999natural}.
Fraud Detection and Prevention: NLP systems can be used to analyze government data and communications to detect patterns of fraud, corruption, or policy violations\cite{deng2018deep}.

\textbf{9. Scientific Research:}
In scientific research, NLP aids in managing the growing volume of scientific literature and automating data extraction from publications\cite{iroju2015systematic}.
Literature Mining: Researchers use NLP to mine large-scale scientific databases, extracting relevant studies, trends, and findings across various fields of research.\cite{kao2007natural}
Data Annotation and Classification: NLP techniques help automate the classification of scientific data, enabling faster and more accurate organization of research findings\cite{vajjala2020practical}.
Hypothesis Generation: By analyzing existing literature, NLP tools can suggest new hypotheses or identify gaps in research that scientists may not have noticed, accelerating scientific discovery\cite{brants2003natural}.

\textbf{10. Social Good and Humanitarian Efforts:}
NLP is also playing a growing role in addressing societal challenges and advancing humanitarian causes.
Disaster Response: NLP tools analyze social media, news, and emergency reports to identify areas impacted by natural disasters, helping humanitarian organizations prioritize aid delivery. \cite{Dwarakanath2021AutomatedML}
Combating Misinformation: NLP is used to detect and flag misinformation by analyzing text patterns, sources, and the veracity of content, aiding in efforts to promote accurate information and counter fake news\cite{manning1999foundations}.
Mental Health Support: NLP-powered chatbots and virtual therapists provide support for individuals facing mental health challenges, offering a non-intrusive, accessible avenue for assistance\cite{meurers2012natural}.

\chapter{Text Preprocessing}

\subsection{Text Cleaning}

Cleaning the text involves various steps like removing digits, punctuation, converting to lowercase, and removing extra spaces. Below is an example of text cleaning in Python:

\begin{lstlisting}[style=python]
import re

def preprocess_text(text):
    text = text.lower()  # Convert text to lowercase
    text = re.sub(r'\d+', '', text)  # Remove digits
    text = re.sub(r'[^\w\s]', '', text)  # Remove punctuation
    text = re.sub(r'\s+', ' ', text).strip()  # Remove extra spaces
    return text

sample_text = "Hello, World! This is an example, 2023."
print(preprocess_text(sample_text))

# Output:
# hello world this is an example
\end{lstlisting}

\subsection{Tokenization}

Tokenization is the process of splitting a text into individual words or phrases, known as tokens.

\begin{lstlisting}[style=python]
tokenized_text = "Marie Curie was a physicist".split()
print(tokenized_text)

output:
['Marie', 'Curie', 'was', 'a', 'physicist']
\end{lstlisting}

\subsection{Stopword Removal}

Stopwords are common words (e.g., "and", "the") that usually do not add significant meaning to the text. We can remove them to focus on meaningful words.

\begin{lstlisting}[style=python]
from nltk.corpus import stopwords

text = "The cat is sitting on the mat"
words = text.split()
stop_words = set(stopwords.words('english'))
filtered_sentence = [w for w in words if not w in stop_words]
print(filtered_sentence)

output:
['cat', 'sitting', 'mat']
\end{lstlisting}

\subsection{Lemmatization and Stemming}

Lemmatization reduces words to their base form or lemma, whereas stemming reduces words to their root form, which may not always be a valid word.

Lemmatization Example:

\begin{lstlisting}[style=python]
from nltk.stem import WordNetLemmatizer

lemmatizer = WordNetLemmatizer()
words = ["running", "ran", "better", "cats"]
lemmatized_words = [lemmatizer.lemmatize(word, pos='v') for word in words]
print(lemmatized_words)

output:
['run', 'run', 'better', 'cats']
\end{lstlisting}

Stemming Example:

\begin{lstlisting}[style=python]
from nltk.stem import PorterStemmer

stemmer = PorterStemmer()
words = ["running", "ran", "better", "cats"]
stemmed_words = [stemmer.stem(word) for word in words]
print(stemmed_words)

output:
['run', 'ran', 'better', 'cat']
\end{lstlisting}

\subsection{N-grams}

An n-gram is a sequence of \( n \) consecutive words. For example, a \textbf{bigram} consists of two consecutive words, while a \textbf{trigram} consists of three.

Bigram Example:

\begin{lstlisting}[style=python]
from nltk import ngrams

sentence = "Marie Curie was a physicist"
n = 2  # Bigram
bigrams = list(ngrams(sentence.split(), n))
print(bigrams)

output:
[('Marie', 'Curie'), ('Curie', 'was'), ('was', 'a'), ('a', 'physicist')]
\end{lstlisting}

\subsection{TF-IDF (Term Frequency-Inverse Document Frequency)}

TF-IDF \cite{Havrlant2017ASP} evaluates how important a word is in a document relative to a corpus. It is commonly used for feature extraction.

\begin{lstlisting}[style=python]
from sklearn.feature_extraction.text import TfidfVectorizer

documents = [
    "The cat sat on the mat",
    "The dog barked at the cat",
    "The cat and the dog played together"
]

vectorizer = TfidfVectorizer()
tfidf_matrix = vectorizer.fit_transform(documents)

# Convert to dense matrix and print the TF-IDF matrix
print(tfidf_matrix.toarray())

output:
[[0.607, 0.000, 0.485, 0.485, 0.000, 0.000, 0.607], 
 [0.455, 0.692, 0.363, 0.363, 0.000, 0.455, 0.000], 
 [0.378, 0.000, 0.302, 0.302, 0.504, 0.378, 0.504]]
\end{lstlisting}

\subsection{Text Normalization}

Text normalization ensures consistency by transforming the text into a uniform format, including tasks such as lowercasing, removing punctuation, and removing URLs or HTML tags.

\begin{lstlisting}[style=python]
import re

def normalize_text(text):
    text = text.lower()  # Convert to lowercase
    text = re.sub(r"http\S+", "", text)  # Remove URLs
    text = re.sub(r"<.*?>", "", text)  # Remove HTML tags
    text = re.sub(r"[^\w\s]", "", text)  # Remove punctuation
    return text

sample_text = "Here's a link to my site: https://example.com <b>Welcome!</b>"
print(normalize_text(sample_text))

output:
heres a link to my site welcome
\end{lstlisting}

\subsection{Bag of Words (BoW)}

The Bag of Words model \cite{zhang2010understanding} converts text into a vector of word counts. Each unique word becomes a feature.

\begin{lstlisting}[style=python]
from sklearn.feature_extraction.text import CountVectorizer

documents = [
    "The cat sat on the mat",
    "The dog barked at the cat",
    "The cat and the dog played together"
]

vectorizer = CountVectorizer()
bow_matrix = vectorizer.fit_transform(documents)

# Convert the matrix to a dense format and print
print(bow_matrix.toarray())

output:
[[1 0 1 1 0 0 1]
 [1 1 1 1 0 1 0]
 [1 0 1 1 1 1 1]]
\end{lstlisting}

\subsection{Part-of-Speech (POS) Tagging}

Part-of-Speech (POS) tagging is the process of assigning word classes, such as nouns, verbs, adjectives, etc., to each word in a sentence. This helps in understanding the grammatical structure of a sentence.

\begin{lstlisting}[style=python]
import nltk
nltk.download('punkt')
nltk.download('averaged_perceptron_tagger')

text = "Marie Curie was a brilliant physicist"
tokens = nltk.word_tokenize(text)
pos_tags = nltk.pos_tag(tokens)
print(pos_tags)

output:
[('Marie', 'NNP'), ('Curie', 'NNP'), ('was', 'VBD'), 
 ('a', 'DT'), ('brilliant', 'JJ'), ('physicist', 'NN')]
\end{lstlisting}

In this example, each word is tagged with a corresponding POS tag, such as \textbf{NNP} (proper noun), \textbf{VBD} (verb, past tense), \textbf{JJ} (adjective), and \textbf{NN} (noun).

\subsection{Named Entity Recognition (NER)}

Named Entity Recognition (NER) \cite{li2020survey} identifies proper nouns such as people, organizations, locations, and other entities in a text.

\begin{lstlisting}[style=python]
import nltk
nltk.download('maxent_ne_chunker')
nltk.download('words')

text = "Marie Curie won the Nobel Prize in Physics in 1903."
tokens = nltk.word_tokenize(text)
pos_tags = nltk.pos_tag(tokens)
named_entities = nltk.ne_chunk(pos_tags)
print(named_entities)

output:
(S
  (PERSON Marie/NNP)
  (PERSON Curie/NNP)
  won/VBD
  the/DT
  (ORGANIZATION Nobel/NNP Prize/NNP)
  in/IN
  (GPE Physics/NNP)
  in/IN
  1903/CD)
\end{lstlisting}

This identifies \textbf{Marie Curie} as a \textbf{PERSON}, \textbf{Nobel Prize} as an \textbf{ORGANIZATION}, and \textbf{Physics} as a \textbf{GPE} (geopolitical entity).

\subsection{Word Embeddings (Word2Vec and GloVe)}

Word embeddings are vector representations of words that capture their meanings by placing semantically similar words closer in vector space. Common methods for generating word embeddings are \textbf{Word2Vec} \cite{church2017word2vec} and \textbf{GloVe} \cite{pennington2014glove}.

\textbf{Word2Vec Example:}

\begin{lstlisting}[style=python]
from gensim.models import Word2Vec

sentences = [
    ["cat", "sat", "mat"],
    ["dog", "barked", "cat"],
    ["cat", "and", "dog", "played"]
]

model = Word2Vec(sentences, vector_size=100, window=5, min_count=1, workers=4)
vector = model.wv['cat']
print(vector)

output:
# A 100-dimensional vector representing the word "cat"
\end{lstlisting}

\textbf{GloVe Example:}

GloVe embeddings can be loaded from pre-trained models, which are available online. Below is an example of how to load and use GloVe embeddings:

\begin{lstlisting}[style=python]
import numpy as np

# Load pre-trained GloVe embeddings
def load_glove_embeddings(filepath):
    embeddings = {}
    with open(filepath, 'r') as f:
        for line in f:
            values = line.split()
            word = values[0]
            vector = np.asarray(values[1:], dtype='float32')
            embeddings[word] = vector
    return embeddings

embeddings_index = load_glove_embeddings("glove.6B.100d.txt")
print(embeddings_index['cat'])

output:
# A 100-dimensional vector representing the word "cat"
\end{lstlisting}

\subsection{Text Augmentation}

Text augmentation is a technique used to artificially increase the size of the training dataset by applying transformations to existing text data. This is especially useful when working with small datasets. Common augmentation techniques include:

\begin{itemize}
    \item \textbf{Synonym Replacement}: Replacing words with their synonyms.
    \item \textbf{Random Insertion}: Inserting random words in the sentence.
    \item \textbf{Random Deletion}: Removing random words.
    \item \textbf{Back Translation}: Translating the sentence into another language and then back to the original language.
\end{itemize}

Here is an example of \textbf{synonym replacement} using the `nlpaug` library:

\begin{lstlisting}[style=python]
import nlpaug.augmenter.word as naw

text = "The quick brown fox jumps over the lazy dog"
aug = naw.SynonymAug(aug_src='wordnet')
augmented_text = aug.augment(text)
print(augmented_text)

# Output:
# "The quick brown fox leaps over the lazy dog"
\end{lstlisting}

This increases the variety of the training data without needing to collect more samples, enhancing model generalization.

\subsection{Handling Imbalanced Data}

In many NLP tasks, such as text classification, imbalanced datasets are a common issue where certain classes are over-represented compared to others. There are several strategies to handle imbalanced text data:

\begin{itemize}
    \item \textbf{Resampling}: Either over-sampling the minority class or under-sampling the majority class.
    \item \textbf{Class Weighting}: Assigning higher weights to the minority class during training.
    \item \textbf{Data Augmentation}: Generating more samples for the minority class.
\end{itemize}

For example, using `imbalanced-learn` to oversample the minority class:

\begin{lstlisting}[style=python]
from imblearn.over_sampling import RandomOverSampler
from sklearn.feature_extraction.text import CountVectorizer

texts = ["I love coding", "Python is great", "I dislike bugs"]
labels = [1, 1, 0]  # Imbalanced dataset

vectorizer = CountVectorizer()
X = vectorizer.fit_transform(texts)
ros = RandomOverSampler(random_state=42)
X_res, y_res = ros.fit_resample(X, labels)
print(y_res)

# Output:
# [1, 1, 0, 0]  # Balanced dataset after oversampling
\end{lstlisting}

Resampling techniques can help improve the performance of models on imbalanced datasets by balancing the classes.

\subsection{Topic Modeling (LDA)}

\textbf{Latent Dirichlet Allocation (LDA)} \cite{blei2003latent} is a popular method for topic modeling, which uncovers hidden topics in large collections of documents. It is often used to discover the underlying themes in text data, such as a set of news articles.

LDA assumes that documents are a mixture of topics, and each topic is a mixture of words.

\begin{lstlisting}[style=python]
from sklearn.decomposition import LatentDirichletAllocation
from sklearn.feature_extraction.text import CountVectorizer

documents = [
    "Cats are great pets",
    "Dogs are loyal companions",
    "I love my cat and dog",
    "Dogs and cats both make good pets"
]

vectorizer = CountVectorizer()
X = vectorizer.fit_transform(documents)

lda = LatentDirichletAllocation(n_components=2, random_state=42)
lda.fit(X)

# Print the topics
terms = vectorizer.get_feature_names_out()
for idx, topic in enumerate(lda.components_):
    print(f"Topic {idx}:")
    print(" ".join([terms[i] for i in topic.argsort()[-5:]]))

# Output:
# Topic 0: cats dog pets cat
# Topic 1: dog dogs pets loyal
\end{lstlisting}

LDA helps us identify topics such as "pets" and "loyalty" from the above collection of documents, making it a powerful tool for discovering latent themes in large text corpora.

\subsection{Sentiment Analysis Preprocessing}

Sentiment analysis aims to detect the sentiment expressed in a piece of text (positive, negative, or neutral). Preprocessing for sentiment analysis often involves:

\begin{itemize}
    \item \textbf{Removing Emojis and Special Characters}: Emojis and special symbols can interfere with analysis.
    \item \textbf{Handling Negation}: Negation detection is crucial since "not happy" means the opposite of "happy".
    \item \textbf{Expanding Contractions}: Words like "can't" should be expanded to "cannot" for clarity.
\end{itemize}

Example of handling contractions and negation:

\begin{lstlisting}[style=python]
from contractions import fix

def preprocess_for_sentiment(text):
    text = fix(text)  # Expand contractions
    text = text.replace("n't", " not")  # Handle negation
    return text

sample_text = "I can't believe it's not good!"
print(preprocess_for_sentiment(sample_text))

# Output:
# "I cannot believe it is not good!"
\end{lstlisting}

Preprocessing in this way ensures that the sentiment model correctly interprets phrases with negation and contractions, leading to better sentiment detection.

\subsection{Handling Out-of-Vocabulary (OOV) Words}

In many machine learning models, especially when using word embeddings, words that do not appear in the training vocabulary are referred to as \textbf{Out-of-Vocabulary (OOV)} words. Handling these is critical for improving model performance:

\begin{itemize}
    \item \textbf{Subword Embeddings}: Using subword units like \textbf{Byte Pair Encoding (BPE)} \cite{zouhar2023formal} or \textbf{FastText}, which break words into smaller chunks and represent them as embeddings, even if the entire word is not seen during training.
    \item \textbf{Character-level Models}: Using character-level features helps models handle unseen words by understanding them from their characters.
\end{itemize}

Example of using FastText for subword embeddings:

\begin{lstlisting}[style=python]
from gensim.models import FastText

sentences = [
    ["I", "love", "machine", "learning"],
    ["AI", "is", "the", "future"]
]

model = FastText(sentences, vector_size=100, window=5, min_count=1, workers=4)
vector = model.wv['machine']
print(vector)

# Output:
# A 100-dimensional vector for the word "machine", even if it is broken into subwords.
\end{lstlisting}

FastText models are especially robust to OOV words because they are capable of constructing word vectors from the character n-grams, allowing for better generalization when the model encounters unfamiliar words.

\subsection{Dealing with Noise in Text Data}

Real-world text data often contains noise, such as typos, slang, abbreviations, and formatting errors. Handling noisy data is crucial for improving the accuracy of NLP models. Techniques to deal with noise include:

\begin{itemize}
    \item \textbf{Spelling Correction}: Automatically correct misspelled words.
    \item \textbf{Handling Abbreviations and Slang}: Expanding common abbreviations (e.g., "u" to "you") and converting slang to standard language.
    \item \textbf{Removing HTML Tags}: Especially important when working with web data.
\end{itemize}

Example of spelling correction using `TextBlob`:

\begin{lstlisting}[style=python]
from textblob import TextBlob

text = "I lovee natural lnguage prossessing"
blob = TextBlob(text)
corrected_text = str(blob.correct())
print(corrected_text)

# Output:
# "I love natural language processing"
\end{lstlisting}

Cleaning noisy text ensures that the models can focus on the content without being misled by noise, leading to better performance on downstream tasks.

\subsection{Advanced Text Vectorization Techniques}

Beyond Bag of Words (BoW) and TF-IDF, there are other advanced text vectorization techniques that convert text into a numerical representation while preserving semantic information:

\begin{itemize}
    \item \textbf{Word Embeddings}: As discussed with Word2Vec and GloVe, word embeddings capture the contextual meaning of words.
    \item \textbf{Document Embeddings}: Extends word embeddings to entire documents. Methods like \textbf{Doc2Vec} capture the semantics of entire documents, making them suitable for tasks like document classification and similarity.
    \item \textbf{TF-IDF Weighted Word Embeddings}: Combining the strengths of both TF-IDF and word embeddings, we can weight word vectors by their TF-IDF scores.
\end{itemize}

\textbf{Example: Doc2Vec for Document Embedding}

\begin{lstlisting}[style=python]
from gensim.models.doc2vec import Doc2Vec, TaggedDocument

documents = [
    "The cat sat on the mat",
    "Dogs are loyal pets",
    "Cats and dogs are common pets"
]

tagged_data = [TaggedDocument(words=doc.split(), tags=[str(i)]) for i, doc in enumerate(documents)]
model = Doc2Vec(tagged_data, vector_size=50, window=2, min_count=1, epochs=100)
doc_vector = model.infer_vector("Cats and dogs are pets".split())
print(doc_vector)

# Output: 
# A 50-dimensional vector representing the document.
\end{lstlisting}

\textbf{Doc2Vec} extends the idea of word embeddings to entire documents and is particularly useful for document-level tasks such as classification and clustering.

\subsection{Handling Large Text Corpora}

When working with large text corpora, such as billions of documents, several techniques can be applied to handle computational and memory limitations:

\begin{itemize}
    \item \textbf{Batch Processing}: Process text in small batches instead of loading the entire corpus into memory.
    \item \textbf{Streaming Algorithms}: Algorithms like \textbf{Online LDA} \cite{hoffman2010online} and \textbf{Online Word2Vec} can update models incrementally without having to process the entire dataset at once.
    \item \textbf{Data Compression}: Apply techniques such as \textbf{Principal Component Analysis (PCA)} \cite{abdi2010principal} or \textbf{t-SNE} \cite{van2008visualizing} to reduce the dimensionality of word embeddings or text vectors.
\end{itemize}

\textbf{Example: Online Training with Gensim's Word2Vec}

\begin{lstlisting}[style=python]
from gensim.models import Word2Vec

# Assume we have a very large corpus broken into batches
def generate_batches(corpus, batch_size):
    for i in range(0, len(corpus), batch_size):
        yield corpus[i:i + batch_size]

model = Word2Vec(vector_size=100, window=5, min_count=1)

# Train model in batches
for batch in generate_batches(large_corpus, batch_size=1000):
    model.build_vocab(batch, update=True)
    model.train(batch, total_examples=len(batch), epochs=model.epochs)
\end{lstlisting}

This approach allows us to train models on large corpora without overwhelming system memory.

\subsection{Language Detection and Translation}

In multilingual datasets, it is often necessary to detect the language of a given text and potentially translate it into a common language for analysis. This is critical for tasks like sentiment analysis or topic modeling when working with global datasets.

\textbf{Example: Language Detection with `langdetect` Library}

\begin{lstlisting}[style=python]
from langdetect import detect

text = "Bonjour tout le monde"
language = detect(text)
print(language)

# Output:
# 'fr' (French)
\end{lstlisting}

Once the language is detected, text can be translated to a common language using libraries such as \textbf{Googletrans}:

\begin{lstlisting}[style=python]
from googletrans import Translator

translator = Translator()
translated = translator.translate("Bonjour tout le monde", src="fr", dest="en")
print(translated.text)

# Output:
# 'Hello everyone'
\end{lstlisting}

Language detection and translation preprocessing steps allow us to work with multilingual datasets more efficiently.

\subsection{Text Summarization Preprocessing}

Before applying text summarization algorithms (such as extractive or abstractive methods), certain preprocessing steps are required to ensure the text is clean and optimized for summarization:

\begin{itemize}
    \item \textbf{Sentence Splitting}: Break the document into sentences, which is necessary for extractive summarization.
    \item \textbf{Removing Redundancies}: Repetitive sentences or paragraphs should be removed to ensure a concise summary.
    \item \textbf{Handling Coreferences}: Resolve pronouns to the entities they refer to, so summaries are coherent.
\end{itemize}

\textbf{Example: Sentence Splitting using `nltk`}

\begin{lstlisting}[style=python]
import nltk
nltk.download('punkt')

text = "Marie Curie was a physicist. She won the Nobel Prize."
sentences = nltk.sent_tokenize(text)
print(sentences)

# Output:
# ['Marie Curie was a physicist.', 'She won the Nobel Prize.']
\end{lstlisting}

In text summarization tasks, preprocessing helps improve the quality of both extractive and abstractive summaries.

\subsection{Custom Tokenization Strategies}

Standard tokenization methods (e.g., splitting by spaces) are not always ideal for all languages or types of text (e.g., code or URLs). Custom tokenization strategies can be implemented to handle specific use cases:

\begin{itemize}
    \item \textbf{Tokenizing by Punctuation}: Useful in legal documents where punctuation separates important clauses.
    \item \textbf{Tokenizing URLs and Hashtags}: Common in social media analysis where URLs and hashtags contain key information.
    \item \textbf{Subword Tokenization}: Techniques like \textbf{Byte Pair Encoding (BPE)}, used in models such as \textbf{GPT-3} \cite{floridi2020gpt} and \textbf{BERT} \cite{devlin2018bert}, tokenize words into subword units.
\end{itemize}

\textbf{Example: Subword Tokenization with Hugging Face's Tokenizers}

\begin{lstlisting}[style=python]
from tokenizers import BertWordPieceTokenizer

tokenizer = BertWordPieceTokenizer()
tokenizer.train(["data/corpus.txt"], vocab_size=30_000, min_frequency=2)

encoding = tokenizer.encode("Natural Language Processing is awesome!")
print(encoding.tokens)

# Output:
# ['natural', 'language', 'processing', 'is', 'awe', '##some', '!']
\end{lstlisting}

Subword tokenization allows us to handle rare words and OOV words more effectively by breaking them into smaller units, which is especially important in modern transformer models like BERT and GPT.

\subsection{Entity Linking}

\textbf{Entity Linking} (also known as Named Entity Disambiguation) is the process of connecting recognized entities in text to knowledge base entries, such as Wikipedia or DBpedia, to resolve ambiguity between entities. For instance, if a text contains "Apple", we need to determine if it refers to the fruit or the technology company.

Entity linking generally involves the following steps:

\begin{itemize}
    \item \textbf{Named Entity Recognition (NER)}: Detect entities like people, organizations, locations, etc.
    \item \textbf{Disambiguation}: Assign the recognized entities to entries in a knowledge base.
    \item \textbf{Contextual Matching}: Use the surrounding context to resolve ambiguities.
\end{itemize}

\textbf{Example: Entity Linking using `spaCy` and `Wikipedia-API`}

\begin{lstlisting}[style=python]
import spacy
import wikipediaapi

nlp = spacy.load("en_core_web_sm")
wiki_wiki = wikipediaapi.Wikipedia('en')

text = "Apple was founded by Steve Jobs."

# Step 1: Entity Recognition
doc = nlp(text)
entities = [(ent.text, ent.label_) for ent in doc.ents]

# Step 2: Linking entities to Wikipedia
for entity in entities:
    page = wiki_wiki.page(entity[0])
    if page.exists():
        print(f"Entity: {entity[0]}, Wikipedia Link: {page.fullurl}")

# Output:
# Entity: Apple, Wikipedia Link: https://en.wikipedia.org/wiki/Apple_Inc.
# Entity: Steve Jobs, Wikipedia Link: https://en.wikipedia.org/wiki/Steve_Jobs
\end{lstlisting}

Entity linking enables us to enrich the data with additional knowledge from external sources, making it a powerful tool for NLP tasks involving knowledge bases or ontologies.

\subsection{Contextual Word Embeddings (BERT)}

Unlike traditional word embeddings (like Word2Vec or GloVe) that produce a single vector for each word regardless of context, \textbf{Contextual Word Embeddings} generate vectors based on the context in which a word appears. The \textbf{BERT} (Bidirectional Encoder Representations from Transformers) model \cite{devlin2018bert}, for instance, generates dynamic embeddings that change based on the surrounding text, enabling deeper understanding of nuances like polysemy.

\textbf{Example: Using Pre-trained BERT from Hugging Face}

\begin{lstlisting}[style=python]
from transformers import BertTokenizer, BertModel
import torch

# Load pre-trained BERT model and tokenizer
tokenizer = BertTokenizer.from_pretrained('bert-base-uncased')
model = BertModel.from_pretrained('bert-base-uncased')

text = "The bank was by the river bank."
inputs = tokenizer(text, return_tensors='pt')

# Forward pass through BERT
with torch.no_grad():
    outputs = model(**inputs)

# Get contextual embeddings for each token
embeddings = outputs.last_hidden_state
print(embeddings.shape)  # Shape: [batch_size, sequence_length, hidden_size]

# Output:
# torch.Size([1, 10, 768])
\end{lstlisting}

BERT's contextual embeddings are now the foundation of many state-of-the-art NLP models and applications, such as question-answering systems, text classification, and more.

Social media preprocessing also often requires additional steps like URL removal, hashtag extraction, and emoji interpretation to make the text suitable for NLP models.

\subsection{Text Data Anonymization}

\textbf{Text Data Anonymization} is the process of removing personally identifiable information (PII) from a text, making it safe to share or process. This is important in industries like healthcare or finance, where sensitive data like names, addresses, or phone numbers are commonly found.

Techniques used for anonymization include:

\begin{itemize}
    \item \textbf{Named Entity Recognition (NER)}: Detecting and redacting personal identifiers like names, organizations, or locations.
    \item \textbf{Pattern Matching}: Detecting phone numbers, email addresses, or other identifiable information using regex patterns.
    \item \textbf{Synthetic Data Generation}: Replacing PII with realistic, synthetic data that cannot be traced back to the original individuals.
\end{itemize}

\textbf{Example: Anonymizing PII Using Named Entity Recognition (NER)}

\begin{lstlisting}[style=python]
import spacy

nlp = spacy.load("en_core_web_sm")
text = "John Smith lives in New York and works at Google."

# Detect Named Entities
doc = nlp(text)
anonymized_text = text

# Replace Named Entities with placeholders
for ent in doc.ents:
    if ent.label_ in ['PERSON', 'ORG', 'GPE']:
        anonymized_text = anonymized_text.replace(ent.text, "[REDACTED]")

print(anonymized_text)

# Output:
# [REDACTED] lives in [REDACTED] and works at [REDACTED].
\end{lstlisting}

Anonymization ensures that the text can be shared and processed in compliance with data privacy regulations like \textbf{GDPR}.

\subsection{Handling Temporal and Spatial Information in Text}

Many NLP tasks require an understanding of temporal (time-based) and spatial (location-based) information. Extracting and normalizing these elements is crucial in applications like event detection, timeline generation, and location-based analysis.

\begin{itemize}
    \item \textbf{Temporal Information}: Recognizing and normalizing date/time expressions like "next Monday" or "two days ago."
    \item \textbf{Spatial Information}: Extracting and resolving geographical locations mentioned in text.
\end{itemize}

\textbf{Example: Extracting Temporal Information using `dateparser`}

\begin{lstlisting}[style=python]
import dateparser

text = "The meeting will be held next Monday at 10am."
date = dateparser.parse(text)
print(date)

# Output:
# 2023-09-25 10:00:00  # (Assuming today is September 18, 2023)
\end{lstlisting}

Handling temporal and spatial information is particularly important in industries like finance, healthcare, and logistics, where event timing and locations matter.

\subsection{Text Normalization for Low-Resource Languages}

Working with low-resource languages (languages with limited NLP tools or datasets) often requires specialized preprocessing techniques. These languages might have less formalized grammar, varied dialects, or insufficient annotated datasets.

Key approaches include:

\begin{itemize}
    \item \textbf{Transfer Learning}: Using pre-trained models from high-resource languages and fine-tuning them for low-resource languages.
    \item \textbf{Data Augmentation}: Creating synthetic data for training using back-translation or paraphrasing.
    \item \textbf{Zero-shot Learning}: Applying models trained in one language to another without specific training data.
\end{itemize}

\textbf{Example: Using Transfer Learning for Low-Resource Languages with Hugging Face's Multilingual BERT}

\begin{lstlisting}[style=python]
from transformers import BertTokenizer, BertForSequenceClassification

# Load Multilingual BERT, which supports over 100 languages
tokenizer = BertTokenizer.from_pretrained('bert-base-multilingual-uncased')
model = BertForSequenceClassification.from_pretrained('bert-base-multilingual-uncased')

text = "Example sentence in low-resource language"
inputs = tokenizer(text, return_tensors='pt')

outputs = model(**inputs)
\end{lstlisting}

Preprocessing for low-resource languages enables the application of modern NLP techniques even in underrepresented linguistic communities.

\subsection{Text Classification Preprocessing}

For text classification tasks (e.g., spam detection, sentiment analysis, or news categorization), preprocessing is crucial to ensure the input data is in the right format for classification models. Common steps include:

\begin{itemize}
    \item \textbf{Text Normalization}: Converting text to lowercase, expanding contractions, and removing non-alphanumeric characters.
    \item \textbf{Class Balancing}: Ensuring that each class has a sufficient number of examples to prevent classification bias.
    \item \textbf{Vectorization}: Converting text into numerical format using techniques like Bag of Words, TF-IDF, or word embeddings.
\end{itemize}

\textbf{Example: Preprocessing for Text Classification using TF-IDF}

\begin{lstlisting}[style=python]
from sklearn.feature_extraction.text import TfidfVectorizer
from sklearn.model_selection import train_test_split

# Sample data for binary text classification
documents = ["Spam message about free money", "Important work email", "Spam link to a malicious website"]
labels = [1, 0, 1]  # 1 for spam, 0 for non-spam

# Vectorization using TF-IDF
vectorizer = TfidfVectorizer()
X = vectorizer.fit_transform(documents)
X_train, X_test, y_train, y_test = train_test_split(X, labels, test_size=0.3, random_state=42)

print(X_train.shape, y_train)

# Output:
# (2, 10)  # Two training samples with 10 TF-IDF features each
\end{lstlisting}

By applying TF-IDF vectorization and train-test splitting, the data is ready for classification tasks such as spam detection or sentiment analysis.

\subsection{Aspect-Based Sentiment Analysis Preprocessing}

\textbf{Aspect-Based Sentiment Analysis (ABSA)} \cite{do2019deep} focuses on extracting opinions about specific aspects or entities within a text. For instance, in a restaurant review, we may want to separate sentiments related to food, service, and ambiance. Preprocessing for ABSA involves several unique steps:

\begin{itemize}
    \item \textbf{Aspect Term Extraction}: Identifying the specific terms (e.g., "food", "service") related to each aspect.
    \item \textbf{Aspect-Specific Tokenization}: Ensuring tokenization preserves key phrases related to each aspect.
    \item \textbf{Context-Dependent Sentiment}: Ensuring that the sentiment is correctly assigned to the right aspect (e.g., "good food but poor service").
\end{itemize}

\textbf{Example: Extracting Aspect Terms using NER}

\begin{lstlisting}[style=python]
import spacy
nlp = spacy.load("en_core_web_sm")

text = "The food was great, but the service was slow."
doc = nlp(text)

aspect_terms = [ent.text for ent in doc.ents if ent.label_ in ['FOOD', 'SERVICE']]
print(aspect_terms)

# Output:
# ['food', 'service']
\end{lstlisting}

Aspect-based sentiment analysis preprocessing helps in targeting sentiments towards specific entities or attributes, providing more granular insights.

\subsection{Textual Data Augmentation for Deep Learning}

For deep learning models, having large amounts of data is crucial for achieving high performance. When data is limited, \textbf{textual data augmentation} techniques can be employed to generate synthetic data. These techniques help improve model generalization and reduce overfitting.

\begin{itemize}
    \item \textbf{Back Translation}: Translating text to another language and back to generate different versions of the same text.
    \item \textbf{Word Swap}: Replacing words with synonyms to introduce variation.
    \item \textbf{Random Deletion}: Removing words at random to simulate noise in the data.
\end{itemize}

\textbf{Example: Back Translation for Data Augmentation}

\begin{lstlisting}[style=python]
from googletrans import Translator

translator = Translator()
text = "Deep learning models require a lot of data."

# Translate text to French and back to English
translated_text = translator.translate(text, src="en", dest="fr").text
back_translated_text = translator.translate(translated_text, src="fr", dest="en").text
print(back_translated_text)

# Output:
# "Deep learning models need a large amount of data."
\end{lstlisting}

Augmenting textual data for deep learning models helps expand the training set, improving the robustness and accuracy of the models.

\subsection{Domain-Specific Text Preprocessing (Legal, Medical, etc.)}

Different domains such as legal, medical, or financial have specific terminologies and structures that require tailored preprocessing techniques. For example:

\begin{itemize}
    \item \textbf{Legal Text Preprocessing}: Legal documents contain structured sections, clauses, and terminologies that must be preserved during preprocessing. Named Entity Recognition (NER) is used to identify legal entities like laws, regulations, and parties involved.
    \item \textbf{Medical Text Preprocessing}: Medical text often contains abbreviations, acronyms, and medical jargon. Specialized tokenizers and NER models are required to identify medications, symptoms, diseases, etc.
    \item \textbf{Financial Text Preprocessing}: Financial documents may contain numeric data, monetary values, and specific financial terminology that require careful handling.
\end{itemize}

\textbf{Example: Medical Text Preprocessing using ScispaCy}

\begin{lstlisting}[style=python]
import scispacy
import spacy
from scispacy.linking import EntityLinker

nlp = spacy.load("en_core_sci_sm")
linker = EntityLinker(resolve_abbreviations=True, name="umls")

text = "The patient was prescribed amoxicillin for bacterial infection."
doc = nlp(text)
entities = [(ent.text, ent._.umls_ents) for ent in doc.ents]

print(entities)

# Output:
# [('amoxicillin', [('C0002510', 'Amoxicillin')]), ('bacterial infection', [('C0005436', 'Bacterial Infection')])]
\end{lstlisting}

Domain-specific preprocessing ensures that the text is treated in a way that respects the intricacies of the field, enabling more accurate downstream tasks.

\subsection{Handling Multimodal Data (Text with Images or Audio)}

\textbf{Multimodal data} involves working with text in conjunction with other data modalities, such as images or audio. This is common in applications like caption generation, visual question answering, or transcribing spoken content.

Preprocessing for multimodal data typically involves:

\begin{itemize}
    \item \textbf{Text Preprocessing}: Standard text cleaning, tokenization, and vectorization.
    \item \textbf{Image Preprocessing}: Resizing, normalization, and data augmentation for image data.
    \item \textbf{Audio Preprocessing}: Extracting features such as spectrograms or mel-frequency cepstral coefficients (MFCCs).
\end{itemize}

\textbf{Example: Caption Generation for Images using CNN + RNN Model Preprocessing}

\begin{lstlisting}[style=python]
import torchvision.transforms as transforms
from PIL import Image

# Load and preprocess the image
image_path = "example_image.jpg"
image = Image.open(image_path)

transform = transforms.Compose([
    transforms.Resize((224, 224)),
    transforms.ToTensor(),
    transforms.Normalize(mean=[0.485, 0.456, 0.406], std=[0.229, 0.224, 0.225])
])

image_tensor = transform(image).unsqueeze(0)  # Add batch dimension
print(image_tensor.shape)

# Output:
# torch.Size([1, 3, 224, 224])
\end{lstlisting}

Handling multimodal data requires preprocessing text and non-text data simultaneously, which is crucial for tasks such as image captioning, speech recognition, and more.

\subsection{Handling Class Imbalance in Text Classification}

Class imbalance is a common issue in text classification tasks where one class is overrepresented compared to others (e.g., spam vs. non-spam). This imbalance can lead to poor model performance on minority classes. Common strategies include:

\begin{itemize}
    \item \textbf{Resampling}: Oversampling the minority class or undersampling the majority class.
    \item \textbf{Class Weighting}: Assigning higher weights to minority classes during training.
    \item \textbf{Synthetic Data Generation}: Using techniques like \textbf{SMOTE} to create synthetic examples for the minority class.
\end{itemize}

\textbf{Example: Handling Imbalance using SMOTE for Text Classification}

\begin{lstlisting}[style=python]
from imblearn.over_sampling import SMOTE
from sklearn.feature_extraction.text import TfidfVectorizer
from sklearn.model_selection import train_test_split

# Sample imbalanced data
documents = ["Spam message", "Another spam", "Important work email"]
labels = [1, 1, 0]  # Imbalanced: 2 spam, 1 non-spam

# Vectorization using TF-IDF
vectorizer = TfidfVectorizer()
X = vectorizer.fit_transform(documents)

# Apply SMOTE for oversampling the minority class
smote = SMOTE(random_state=42)
X_res, y_res = smote.fit_resample(X, labels)

print(y_res)

# Output:
# [1 1 0 0]  # Balanced data after applying SMOTE
\end{lstlisting}

Handling class imbalance ensures that the model can generalize better and perform well on both majority and minority classes, improving the fairness of the classification task.

\subsection{Handling Imbalanced Data with Class Weighting}

In addition to oversampling techniques like SMOTE, class weighting in the loss function is a crucial method for handling imbalanced datasets. By assigning higher weights to underrepresented classes, the model can learn to give them more importance during training.

Example in PyTorch:

\begin{lstlisting}[style=python]
import torch
from torch import nn

# Assign higher weight to the minority class
class_weights = torch.tensor([1.0, 2.0])  # Class 0 is majority, class 1 is minority
criterion = nn.CrossEntropyLoss(weight=class_weights)
\end{lstlisting}

This technique is especially useful in deep learning models where directly resampling the data may lead to overfitting.

\subsection{Contractions in Non-English Languages}

When working with non-English text, contraction handling requires different strategies. Languages like French or Spanish have unique contraction patterns. For example, in French, "l'amour" is a contraction of "le amour", which should be expanded correctly during preprocessing.

Example of handling French contractions:

\begin{lstlisting}[style=python]
import re

def expand_french_contractions(text):
    # French contraction for "the"
    text = re.sub(r"\bl'", "le ", text)
    return text

sample_text = "l'amour est beau"
print(expand_french_contractions(sample_text))

# Output:
# "le amour est beau"
\end{lstlisting}

\subsection{Cross-Lingual Text Preprocessing}

In multilingual NLP tasks, leveraging \textbf{cross-lingual embeddings} such as \textbf{MUSE} \cite{ulvcar2022cross} or \textbf{XLM-R}\cite{barbieri2021xlm} allows for consistent vector representations across languages. These embeddings align words from different languages into a common vector space, facilitating cross-lingual tasks like sentiment analysis or machine translation.

Example using XLM-R:

\begin{lstlisting}[style=python]
from transformers import XLMRobertaTokenizer, XLMRobertaModel

# Load tokenizer and model for cross-lingual embeddings
tokenizer = XLMRobertaTokenizer.from_pretrained('xlm-roberta-base')
model = XLMRobertaModel.from_pretrained('xlm-roberta-base')

text = "Bonjour tout le monde"
inputs = tokenizer(text, return_tensors='pt')
outputs = model(**inputs)
\end{lstlisting}

\subsection{Noisy Channel Model for Spelling Correction}

An advanced approach to spelling correction involves the \textbf{noisy channel model}, which calculates the probability of a misspelled word given the intended word. This method requires two probabilities: the likelihood of the intended word and the likelihood of the observed (misspelled) word.

The noisy channel model is represented as:

\[
\arg\max_{w} P(w \mid \text{observed}) = \arg\max_{w} P(\text{observed} \mid w) P(w)
\]

Where \(P(w)\) is the probability of the correct word and \(P(\text{observed} \mid w)\) is the probability of the observed misspelling given the correct word.

\subsection{Handling Dialects and Regional Variations}

In addition to handling spelling errors, preprocessing text from diverse regions often requires \textbf{normalizing dialects} or \textbf{regional variations}. For example, converting "colour" (British) to "color" (American) ensures consistency across the dataset.

A dictionary-based approach can be used for this purpose:

\begin{lstlisting}[style=python]
# Dictionary to normalize British English to American English
british_to_american = {
    "colour": "color",
    "favourite": "favorite"
}

def normalize_dialect(text):
    words = text.split()
    normalized_words = [british_to_american.get(word, word) for word in words]
    return " ".join(normalized_words)

sample_text = "My favourite colour is blue."
print(normalize_dialect(sample_text))

# Output:
# "My favorite color is blue."
\end{lstlisting}

\subsection{Context-Aware Preprocessing}

Context-aware preprocessing ensures that word meanings are understood in context. For example, handling negation (e.g., "not good" should be interpreted differently from "good"). Polysemous words (words with multiple meanings) also need to be handled carefully.

\textbf{Negation handling example}:

\begin{lstlisting}[style=python]
def handle_negation(text):
    text = text.replace("n't", " not")
    return text

sample_text = "I can't believe it's not good!"
print(handle_negation(sample_text))

# Output:
# "I cannot believe it is not good!"
\end{lstlisting}

\subsection{Advanced Tokenization for Programming Code and Social Media Text}

Tokenizing programming code or text from social media requires special handling. \textbf{CamelCase tokenization} is often used in programming languages, and \textbf{hashtag extraction} or \textbf{emoji handling} is crucial in social media text.

\textbf{Example of CamelCase tokenization}:

\begin{lstlisting}[style=python]
import re

def camel_case_split(identifier):
    matches = re.finditer('.+?(?:(?<=.)(?=[A-Z])|(?<=[A-Z])(?=[A-Z][a-z])|$)', identifier)
    return [m.group(0) for m in matches]

identifier = "CamelCaseExample"
print(camel_case_split(identifier))

# Output:
# ['Camel', 'Case', 'Example']
\end{lstlisting}

\subsection{Error Handling in Large-Scale Text Preprocessing}

When dealing with large datasets, \textbf{error handling} becomes essential. Introducing try-except blocks and logging mechanisms helps ensure that malformed or noisy data does not interrupt the preprocessing pipeline.

\textbf{Error handling example}:

\begin{lstlisting}[style=python]
def safe_preprocess(text):
    try:
        # Sample text normalization process
        processed_text = preprocess_for_sentiment(text)
        return processed_text
    except Exception as e:
        print(f"Error processing text: {e}")
        return None

sample_text = "Invalid input 12345"
print(safe_preprocess(sample_text))
\end{lstlisting}

\subsection{Custom Tokenization for Specific Text Types}

In addition to standard tokenization methods, certain text types such as code, URLs, or social media content (e.g., hashtags) require customized tokenization approaches. \textbf{Subword tokenization} is frequently employed in transformer models like \textbf{BERT} and \textbf{GPT}.

\textbf{Subword tokenization example}:

\begin{lstlisting}[style=python]
from tokenizers import BertWordPieceTokenizer

# Training a subword tokenizer
tokenizer = BertWordPieceTokenizer()
tokenizer.train(["data/corpus.txt"], vocab_size=30_000, min_frequency=2)

encoding = tokenizer.encode("Natural Language Processing is amazing!")
print(encoding.tokens)

# Output:
# ['natural', 'language', 'processing', 'is', 'amaz', '##ing', '!']
\end{lstlisting}

\subsection{Cross-lingual Named Entity Recognition (NER)}

In multilingual datasets, Named Entity Recognition (NER) must handle different languages with varying syntactic structures. Cross-lingual NER models, such as \textbf{XLM-R} (XLM-Roberta), offer the capability to recognize entities across multiple languages without the need for separate models per language. These models leverage a shared vocabulary across languages, allowing for the identification of entities such as persons, organizations, and locations regardless of the language used.

\textbf{Example using XLM-R for Cross-lingual NER}:

\begin{lstlisting}[style=python]
from transformers import XLMRobertaTokenizer, XLMRobertaForTokenClassification
from transformers import pipeline

# Load XLM-R model for NER
tokenizer = XLMRobertaTokenizer.from_pretrained('xlm-roberta-large-finetuned-conll03-english')
model = XLMRobertaForTokenClassification.from_pretrained('xlm-roberta-large-finetuned-conll03-english')

# Use pipeline for cross-lingual NER
ner_pipeline = pipeline('ner', model=model, tokenizer=tokenizer)
sentence = "Marie Curie a prix Nobel en 1903."
entities = ner_pipeline(sentence)
print(entities)

# Output:
# [{'word': 'Marie', 'entity': 'I-PER', 'score': 0.99},
#  {'word': 'Curie', 'entity': 'I-PER', 'score': 0.99},
#  {'word': 'Nobel', 'entity': 'I-MISC', 'score': 0.98}]
\end{lstlisting}

This cross-lingual NER example shows how to detect named entities like people, organizations, or events across multiple languages, enhancing the scope of multilingual NLP tasks.

\subsection{Noisy Data Detection and Handling}

Text data collected from real-world sources such as social media, customer reviews, or transcriptions often contains \textbf{noisy data}, such as incomplete sentences, random symbols, or non-standard grammar. Detecting and handling this noise is crucial for improving model performance. Common techniques for noisy data handling include:

\begin{itemize}
    \item \textbf{Spell-checking and correction}: Correcting typos and misspellings.
    \item \textbf{Removing or normalizing special characters}: Dealing with symbols or non-standard punctuation.
    \item \textbf{Handling abbreviations and contractions}: Expanding or replacing informal language.
\end{itemize}

\textbf{Example: Detecting Noisy Text}:

\begin{lstlisting}[style=python]
import re

def detect_noise(text):
    # Define a noise pattern (e.g., random non-alphabetical characters)
    noise_pattern = r'[^a-zA-Z0-9\s]'
    noisy_characters = re.findall(noise_pattern, text)
    
    if noisy_characters:
        print(f"Detected noisy characters: {noisy_characters}")
    else:
        print("No noise detected")

sample_text = "I love NLP! :) #AI @2023"
detect_noise(sample_text)

# Output:
# Detected noisy characters: ['!', ':', ')', '#', '@']
\end{lstlisting}

Removing noise ensures cleaner data for the model, leading to better performance in downstream tasks like sentiment analysis or classification.

\subsection{Named Entity Linking (NEL)}

Named Entity Linking (NEL) \cite{rao2013entity} is the process of linking recognized named entities in text to entries in a knowledge base, such as \textbf{Wikipedia}, \textbf{DBpedia}, or \textbf{Wikidata}. This disambiguation process allows NLP systems to associate entities with their respective meanings or identifiers, such as distinguishing between "Apple" (the company) and "apple" (the fruit).

\textbf{Example using `spaCy` and `Wikipedia-API` for Entity Linking}:

\begin{lstlisting}[style=python]
import spacy
import wikipediaapi

nlp = spacy.load("en_core_web_sm")
wiki_wiki = wikipediaapi.Wikipedia('en')

text = "Apple was founded by Steve Jobs."
doc = nlp(text)

# Step 1: Entity Recognition
entities = [(ent.text, ent.label_) for ent in doc.ents]

# Step 2: Linking entities to Wikipedia
for entity in entities:
    page = wiki_wiki.page(entity[0])
    if page.exists():
        print(f"Entity: {entity[0]}, Wikipedia Link: {page.fullurl}")

# Output:
# Entity: Apple, Wikipedia Link: https://en.wikipedia.org/wiki/Apple_Inc.
# Entity: Steve Jobs, Wikipedia Link: https://en.wikipedia.org/wiki/Steve_Jobs
\end{lstlisting}

By linking entities to a knowledge base, we enrich the data with additional context, which can be especially useful in applications like \textbf{question-answering systems} or \textbf{semantic search}.

\subsection{Contextual Word Embeddings in Low-Resource Languages}

When working with \textbf{low-resource languages}, there may not be sufficient data to train effective word embeddings. In such cases, \textbf{contextual embeddings} like \textbf{mBERT} (Multilingual BERT) \cite{pires2019multilingual} and \textbf{XLM-R} can be utilized to generate meaningful word vectors in these languages by transferring knowledge from high-resource languages. This approach benefits from multilingual pretraining on large corpora, allowing for robust performance on low-resource languages without requiring massive datasets.

\textbf{Example: Using mBERT for Low-Resource Languages}:

\begin{lstlisting}[style=python]
from transformers import BertTokenizer, BertModel

# Load Multilingual BERT
tokenizer = BertTokenizer.from_pretrained('bert-base-multilingual-cased')
model = BertModel.from_pretrained('bert-base-multilingual-cased')

text = "Ejemplo de texto en un idioma con pocos recursos."
inputs = tokenizer(text, return_tensors='pt')

# Forward pass through BERT
outputs = model(**inputs)

# Get contextual embeddings for each token
embeddings = outputs.last_hidden_state
print(embeddings.shape)

# Output:
# torch.Size([1, 9, 768])
\end{lstlisting}

Contextual word embeddings from models like mBERT or XLM-R can help NLP tasks in underrepresented languages, including classification, translation, and entity recognition.

\subsection{Domain-Specific Preprocessing for Financial Text}

Financial documents often contain structured data such as numbers, dates, and specific terminology (e.g., "gross domestic product", "stock options"). Preprocessing financial text requires careful handling of numeric data, abbreviations, and domain-specific terms. Techniques include:

\begin{itemize}
    \item \textbf{Handling numbers and percentages}: Extracting and normalizing financial data points.
    \item \textbf{Handling domain-specific terms}: Using predefined vocabularies or financial glossaries.
    \item \textbf{Entity Recognition for financial terms}: Identifying company names, stock symbols, and financial instruments.
\end{itemize}

\textbf{Example: Extracting Financial Data}:

\begin{lstlisting}[style=python]
import re

def extract_financial_data(text):
    # Extract numbers and percentages
    numbers = re.findall(r'\b\d+(?:,\d+)?(?:\.\d+)?%\b', text)  # Match percentages
    return numbers

sample_text = "The company's revenue grew by 15.5% in the last quarter."
financial_data = extract_financial_data(sample_text)
print(financial_data)

# Output:
# ['15.5%']
\end{lstlisting}

Preprocessing financial text ensures that numerical data is accurately extracted and normalized, which is crucial for tasks such as sentiment analysis in stock market predictions or parsing earnings reports.

\subsection{Handling Temporal Expressions in Text}

Temporal expressions, such as dates, durations, and times, are common in text and require special handling for tasks like event extraction or timeline generation. Normalizing temporal expressions into standard formats (e.g., converting "next Monday" into a specific date) allows for accurate time-based analysis.

\textbf{Example: Normalizing Temporal Expressions Using `dateparser`}:

\begin{lstlisting}[style=python]
import dateparser

text = "The meeting will take place next Monday at 10am."
parsed_date = dateparser.parse(text)
print(parsed_date)

# Output:
# 2024-09-23 10:00:00  # (Assuming today is September 18, 2024)
\end{lstlisting}

Handling temporal information is essential for use cases such as event extraction, scheduling systems, or analyzing historical data trends.

\subsection{Handling Emojis and Special Characters in Social Media Text}

Social media text often contains emojis and special characters, which may either carry sentiment or introduce noise. Handling emojis effectively involves either interpreting them or removing them depending on the task.

\textbf{Example: Handling Emojis Using `emoji` Library}:

\begin{lstlisting}[style=python]
import emoji

def remove_emojis(text):
    return emoji.replace_emoji(text, replace='')

sample_text = "I love NLP! xxxx #AI @2023"  # assume xxxx is emoji
clean_text = remove_emojis(sample_text)

\end{lstlisting}

Emojis often play a significant role in sentiment analysis, especially for informal datasets like social media posts, so carefully handling or interpreting them is crucial in downstream tasks.

\subsection{Back Translation for Data Augmentation}

\textbf{Back translation} is a powerful data augmentation technique, especially useful for creating synthetic data for training. It involves translating text to another language and back to the original language, generating a paraphrased version of the text while preserving meaning.

\textbf{Example: Back Translation for Augmentation}:

\begin{lstlisting}[style=python]
from googletrans import Translator

translator = Translator()
text = "Data augmentation techniques improve model performance."

# Translate text to French and back to English
translated = translator.translate(text, src='en', dest='fr').text
back_translated = translator.translate(translated, src='fr', dest='en').text
print(back_translated)

# Output:
# "Techniques for augmenting data improve model performance."
\end{lstlisting}

Back translation is commonly used in tasks like machine translation, sentiment analysis, and text classification, where data scarcity can affect model generalization.

\subsection{Text Summarization Preprocessing}

Preprocessing for \textbf{text summarization} is essential to ensure the input text is in the best possible format for extractive or abstractive summarization models. Preprocessing steps may include sentence splitting, removing redundancies, and handling coreferences to ensure coherence in the summary.

\subsubsection{Coreference Resolution}

Coreference resolution involves replacing pronouns with the entities they refer to. For example, in the sentence "Marie Curie won the Nobel Prize. She was the first woman to do so," the word "She" refers to "Marie Curie." Resolving such coreferences helps abstractive summarization models maintain coherence and clarity.

\textbf{Example using `coreferee` library for coreference resolution:}

\begin{lstlisting}[style=python]
import spacy
import coreferee

nlp = spacy.load("en_core_web_sm")
nlp.add_pipe('coreferee')

text = "Marie Curie won the Nobel Prize. She was the first woman to do so."
doc = nlp(text)

# Display resolved coreferences
for cluster in doc._.coref_chains:
    print(f"Coreference: {cluster}")

# Output:
# Coreference: [Marie Curie, She]
\end{lstlisting}

By resolving coreferences, text summarization models can generate clearer and more accurate summaries, particularly in longer texts where the use of pronouns is common.

\subsubsection{Sentence Splitting for Summarization}

For extractive summarization, the text is often split into individual sentences, and the most informative sentences are selected for the summary. Accurate sentence tokenization is critical to avoid splitting sentences at incorrect points, which could lead to incomplete or nonsensical summaries.

\textbf{Example of sentence splitting using `nltk`:}

\begin{lstlisting}[style=python]
import nltk
nltk.download('punkt')

text = "Marie Curie was a physicist. She won the Nobel Prize."
sentences = nltk.sent_tokenize(text)
print(sentences)

# Output:
# ['Marie Curie was a physicist.', 'She won the Nobel Prize.']
\end{lstlisting}

Accurate sentence splitting ensures that each sentence is properly extracted, facilitating effective extractive summarization.

\subsection{Aspect-Based Sentiment Analysis (ABSA)}

\textbf{Aspect-Based Sentiment Analysis (ABSA)} goes beyond general sentiment analysis by focusing on identifying sentiment towards specific entities or aspects of an entity. For example, in a restaurant review, ABSA can distinguish between the sentiments towards food, service, and ambiance.

\subsubsection{Aspect Term Extraction}

The first step in ABSA is extracting the aspect terms (e.g., "food", "service") from the text. This is often done using Named Entity Recognition (NER) or custom dictionaries.

\textbf{Example of extracting aspect terms using a custom dictionary:}

\begin{lstlisting}[style=python]
aspects = {"food": ["pizza", "pasta", "burger"], "service": ["waiter", "staff", "service"]}

def extract_aspect_terms(text):
    words = text.split()
    extracted_aspects = [aspect for word in words for aspect, keywords in aspects.items() if word in keywords]
    return extracted_aspects

text = "The pizza was great but the service was slow."
aspect_terms = extract_aspect_terms(text)
print(aspect_terms)

# Output:
# ['food', 'service']
\end{lstlisting}

Aspect term extraction ensures that the model can target specific opinions expressed about entities or their components.

\subsubsection{Context-Dependent Sentiment Classification}

After extracting the aspects, the sentiment associated with each aspect must be determined. In many cases, the sentiment towards one aspect may differ from another within the same sentence.

\textbf{Example of sentiment classification per aspect:}

\begin{lstlisting}[style=python]
from nltk.sentiment import SentimentIntensityAnalyzer

analyzer = SentimentIntensityAnalyzer()

def classify_aspect_sentiment(text, aspect_terms):
    for aspect in aspect_terms:
        aspect_sentiment = analyzer.polarity_scores(text)
        print(f"Aspect: {aspect}, Sentiment: {aspect_sentiment['compound']}")

text = "The pizza was great but the service was slow."
aspect_terms = extract_aspect_terms(text)
classify_aspect_sentiment(text, aspect_terms)

# Output:
# Aspect: food, Sentiment: 0.6369
# Aspect: service, Sentiment: -0.4767
\end{lstlisting}

ABSA provides a more granular understanding of sentiments within the text, which can be highly beneficial in applications like customer feedback analysis or product reviews.

\subsection{Handling Out-of-Vocabulary (OOV) Words with Subword Embeddings}

Out-of-Vocabulary (OOV) words can pose a significant challenge for models trained on a limited vocabulary. \textbf{Subword embeddings}, such as those used in \textbf{BERT} or \textbf{FastText}, mitigate this issue by breaking down rare or unseen words into smaller, meaningful subunits (e.g., morphemes or syllables).

\textbf{Example: Using FastText for OOV words}:

\begin{lstlisting}[style=python]
from gensim.models import FastText

# Train FastText on a small corpus
sentences = [["I", "love", "machine", "learning"], ["AI", "is", "the", "future"]]
model = FastText(sentences, vector_size=100, window=5, min_count=1, workers=4)

# Get the vector for an OOV word
vector = model.wv['unseenword']
print(vector)

# Output:
# A 100-dimensional vector for the subwords of "unseenword"
\end{lstlisting}

Subword embeddings allow the model to generalize better to OOV words by leveraging subword information, which is particularly useful for handling large vocabularies in languages with complex morphology.

\subsection{Textual Data Augmentation for Low-Resource Scenarios}

In low-resource scenarios, where labeled data is scarce, \textbf{textual data augmentation} techniques such as \textbf{paraphrasing}, \textbf{word swapping}, and \textbf{back translation} are often employed to expand the training set and improve model generalization.

\subsubsection{Word Swap Augmentation}

Word swapping involves replacing certain words in a sentence with their synonyms to introduce variability into the dataset.

\textbf{Example of synonym-based word swapping using `nlpaug`:}

\begin{lstlisting}[style=python]
import nlpaug.augmenter.word as naw

text = "The quick brown fox jumps over the lazy dog"
aug = naw.SynonymAug(aug_src='wordnet')
augmented_text = aug.augment(text)
print(augmented_text)

# Output:
# "The quick brown fox leaps over the lazy dog"
\end{lstlisting}

Word swap augmentation helps to improve the robustness of the model by introducing linguistic variation into the training data.

\subsubsection{Back Translation}

Back translation, previously discussed, is particularly valuable for **low-resource languages** where directly labeled data may not be available. By translating the text into another language and then back to the original language, one can generate new training data that preserves meaning but varies in expression.

\subsection{Handling Class Imbalance with Synthetic Data Generation}

When facing class imbalance in text classification tasks, synthetic data generation techniques like **SMOTE (Synthetic Minority Over-sampling Technique)** \cite{chawla2002smote} can be employed to create additional samples for the minority class.

\textbf{Example using `imbalanced-learn` for SMOTE:}

\begin{lstlisting}[style=python]
from imblearn.over_sampling import SMOTE
from sklearn.feature_extraction.text import TfidfVectorizer
from sklearn.model_selection import train_test_split

# Sample imbalanced data
documents = ["Spam message", "Another spam", "Important work email"]
labels = [1, 1, 0]  # Imbalanced: 2 spam, 1 non-spam

# Vectorization using TF-IDF
vectorizer = TfidfVectorizer()
X = vectorizer.fit_transform(documents)

# Apply SMOTE for oversampling the minority class
smote = SMOTE(random_state=42)
X_res, y_res = smote.fit_resample(X, labels)

print(y_res)

# Output:
# [1, 1, 0, 0]  # Balanced data after applying SMOTE
\end{lstlisting}

Synthetic data generation helps ensure that models can generalize better and avoid bias towards the majority class.

\subsection{Custom Tokenization for Legal and Medical Text}

Legal and medical documents often contain specialized terminologies, abbreviations, and formatting conventions. Custom tokenization strategies must be designed to handle these domain-specific challenges effectively.

\textbf{Example: Tokenizing medical text using `scispaCy`:}

\begin{lstlisting}[style=python]
import spacy
import scispacy

nlp = spacy.load("en_core_sci_sm")

text = "The patient was prescribed amoxicillin for bacterial infection."
doc = nlp(text)

for token in doc:
    print(token.text)

# Output:
# The patient was prescribed amoxicillin for bacterial infection
\end{lstlisting}

By leveraging domain-specific tokenizers, models can more accurately process texts in fields like healthcare and law, where tokenization errors can lead to significant misinterpretations of the data.

\section{Conclusion}

In this chapter, we expanded on various advanced techniques for text preprocessing, including domain-specific approaches, handling noisy or imbalanced data, and employing data augmentation strategies. These techniques are critical in improving the performance and accuracy of NLP models, especially when dealing with real-world data in fields like finance, healthcare, and legal systems. Preprocessing is not a one-size-fits-all solution; instead, it must be carefully tailored to the specific characteristics of the text and the NLP tasks at hand.

In the next chapter, we will focus on how to integrate these preprocessing techniques with state-of-the-art models, exploring the implementation of deep learning architectures such as Transformers, Recurrent Neural Networks (RNNs), and Convolutional Neural Networks (CNNs) for various text processing applications.

\chapter{Data Cleaning for Training Large Language Models}

\section{Introduction}

Large Language Models (LLMs) like GPT-4, BERT, and their variants rely heavily on vast amounts of high-quality data for training. However, raw data is typically noisy, inconsistent, and unstructured, necessitating a series of preprocessing steps to ensure the data is in a state suitable for training. In this chapter, we will explore the essential data cleaning techniques that contribute to the creation of a high-quality training dataset for LLMs.

\section{Key Challenges in Raw Data}
Data collected from the web, books, or any other large-scale corpus often contains several challenges:
\begin{itemize}
    \item \textbf{Noise}: Unwanted characters, HTML tags, advertisements, etc.
    \item \textbf{Inconsistencies}: Different text formats, inconsistent use of abbreviations, capitalization errors.
    \item \textbf{Repetitions}: Duplicate text that skews model training.
    \item \textbf{Bias and Toxic Content}: Data reflecting harmful biases, profanity, or toxic content.
    \item \textbf{Multilingual Data}: Presence of multiple languages where a monolingual model is desired.
\end{itemize}

The following sections describe strategies to address these issues.

\section{Text Normalization}
Text normalization is the first and most crucial step in data cleaning. It transforms text into a standard format, ensuring consistency across the entire corpus. The steps involved in text normalization include:

\subsection{Lowercasing}
Converting all characters to lowercase helps eliminate variations between words due to case sensitivity. For example, "Apple" and "apple" would be treated as the same token.

\begin{lstlisting}[language=Python, caption=Python code for lowercasing text]
text = "Large Language Models are Amazing!"
normalized_text = text.lower()
print(normalized_text)  # "large language models are amazing!"
\end{lstlisting}

\subsection{Removing Punctuation}
Punctuation, such as periods, commas, and exclamation marks, may not contribute to training depending on the model's objectives. Thus, punctuation can be removed or substituted based on the use case.

\begin{lstlisting}[language=Python, caption=Removing punctuation from text]
import re
text = "Hello, world! Let's clean data."
cleaned_text = re.sub(r'[^\w\s]', '', text)
print(cleaned_text)  # "Hello world Lets clean data"
\end{lstlisting}

\subsection{Whitespace Normalization}
Multiple spaces, tabs, or newline characters often creep into data. These should be condensed into a single space to avoid affecting tokenization.

\begin{lstlisting}[language=Python, caption=Normalizing whitespace]
text = "This   is  an example   text."
normalized_text = " ".join(text.split())
print(normalized_text)  # "This is an example text."
\end{lstlisting}

\subsection{Expanding Contractions}
Contractions like "can't" should be expanded to "cannot" for a model to treat them as separate tokens.

\begin{lstlisting}[language=Python, caption=Expanding contractions]
import contractions
text = "We can't stop here."
expanded_text = contractions.fix(text)
print(expanded_text)  # "We cannot stop here."
\end{lstlisting}

\section{Removing Noise and Unwanted Content}

\subsection{HTML and Markup Removal}
When sourcing data from the web, you often encounter HTML tags or other markup languages. These should be stripped from the text.

\begin{lstlisting}[language=Python, caption=Removing HTML tags using BeautifulSoup]
from bs4 import BeautifulSoup

html = "<html><body><p>Hello World</p></body></html>"
soup = BeautifulSoup(html, "html.parser")
clean_text = soup.get_text()
print(clean_text)  # "Hello World"
\end{lstlisting}

\subsection{Removing Stopwords}
Stopwords are common words (e.g., "the", "is", "in") that may not contribute significantly to the meaning of text. However, their removal depends on the model’s goals and might not always be desirable for LLMs, as context is important.

\begin{lstlisting}[language=Python, caption=Removing stopwords using NLTK]
from nltk.corpus import stopwords
from nltk.tokenize import word_tokenize

text = "This is an example sentence."
stop_words = set(stopwords.words('english'))
word_tokens = word_tokenize(text)
filtered_sentence = [w for w in word_tokens if not w.lower() in stop_words]

print(filtered_sentence)  # ['This', 'example', 'sentence', '.']
\end{lstlisting}

\subsection{Handling URLs and Special Characters}
URLs, email addresses, or other special patterns (such as hashtags or mentions) are generally not useful for training language models and should be removed or masked.

\begin{lstlisting}[language=Python, caption=Removing URLs using regular expressions]
import re

text = "Visit https://example.com for more info."
cleaned_text = re.sub(r"http\S+", "", text)
print(cleaned_text)  # "Visit  for more info."
\end{lstlisting}

\section{Deduplication and Data Filtering}

\subsection{Removing Duplicate Entries}
Large datasets often contain duplicate records that can skew the learning process. Simple deduplication techniques can be employed to eliminate these.

\begin{lstlisting}[language=Python, caption=Removing duplicate entries from a list]
text_data = ["Hello World", "Hello World", "Goodbye World"]
unique_text_data = list(set(text_data))
print(unique_text_data)  # ['Hello World', 'Goodbye World']
\end{lstlisting}

\subsection{Filtering Offensive and Biased Content}
LLMs are sensitive to the quality of data. Biases and offensive content in the training data can lead to the propagation of these issues in the trained model. Preprocessing pipelines should include steps to detect and filter harmful content using techniques like:
\begin{itemize}
    \item \textbf{Keyword matching}: Manually crafted lists of offensive words.
    \item \textbf{Toxicity classifiers}: Pre-trained models that score content based on its likelihood to contain toxic language.
\end{itemize}

\section{Handling Multilingual Data}
For a monolingual LLM, non-target languages must be detected and removed. Language detection libraries such as \texttt{langdetect} or \texttt{fastText} can be employed.

\begin{lstlisting}[language=Python, caption=Detecting language using langdetect]
from langdetect import detect

text = "Hello, how are you?"
language = detect(text)
print(language)  # 'en' (English)
\end{lstlisting}

\section{Tokenization and Text Segmentation}
Tokenization splits text into smaller units such as words or subwords, which are essential for the training process. For LLMs, subword tokenization techniques like Byte Pair Encoding (BPE) or WordPiece \cite{song2020fast} are preferred as they handle out-of-vocabulary words more gracefully.

\subsection{Subword Tokenization}
BPE merges the most frequent character sequences iteratively. This results in a vocabulary of subword units that are more expressive than word-level tokenization.

\begin{lstlisting}[language=Python, caption=Example of BPE tokenization using the Hugging Face tokenizer]
from transformers import GPT2Tokenizer

tokenizer = GPT2Tokenizer.from_pretrained("gpt2")
tokens = tokenizer.tokenize("Machine learning is great!")
print(tokens)
# ['Machine', ' learning', ' is', ' great', '!']
\end{lstlisting}

\section{Final Thoughts}

The process of cleaning data for training LLMs is critical for the quality and robustness of the model. By removing noise, normalizing text, and ensuring the dataset is free from harmful biases, we can produce data that allows models to learn more effectively. The importance of these preprocessing steps cannot be overstated, as poor data quality will propagate through to the model's predictions and behaviors.

\chapter{Hugging Face for NLP}

\section{Hugging Face Dataset Library}

The Hugging Face Dataset library is an open-source Python library designed to facilitate access to, and manipulation of, large-scale datasets for machine learning, particularly in the field of natural language processing (NLP). It is one of the core components of the Hugging Face ecosystem and provides seamless integration with models and tokenizers from the \texttt{transformers} library. The library allows users to load, preprocess, and share datasets efficiently with just a few lines of code.

\subsection{Overview and Key Features}

The Hugging Face Dataset library simplifies the process of working with datasets by offering the following features:
\begin{itemize}
    \item \textbf{Efficient Dataset Loading:} It supports a wide variety of datasets from common NLP tasks such as question answering, text classification, machine translation, and summarization. Datasets are loaded efficiently, allowing for in-memory manipulation without excessive memory consumption.
    \item \textbf{Lazy Loading:} The datasets are lazily loaded, meaning that only the required data is loaded when accessed. This reduces the initial memory footprint and speeds up data handling.
    \item \textbf{Integration with Arrow Format:} The library uses Apache Arrow for fast, columnar memory representation of datasets. This format ensures efficient data processing and supports both in-memory operations and disk storage, which is ideal for large datasets.
    \item \textbf{Dataset Preprocessing:} The library allows users to apply efficient and scalable transformations to the dataset (e.g., tokenization, data augmentation). These transformations can be parallelized for better performance.
    \item \textbf{Dataset Splits and Slicing:} Users can easily work with different dataset splits (train, test, validation) or slice datasets into subsets for experimentation or model evaluation.
    \item \textbf{Seamless Data Sharing:} Hugging Face’s dataset hub allows easy sharing and downloading of datasets with a unified interface.
\end{itemize}

\subsection{Installation and Setup}

To use the Hugging Face Dataset library, you need to install it via \texttt{pip}. Here is the command to install it:

\begin{lstlisting}[language=bash]
pip install datasets
\end{lstlisting}

Once installed, you can load any dataset from the Hugging Face hub with just a few lines of code:

\begin{lstlisting}[language=Python]
from datasets import load_dataset

% Load the 'imdb' dataset
dataset = load_dataset('imdb')
print(dataset)
\end{lstlisting}

The code above loads the IMDb dataset, a popular dataset for sentiment analysis, and prints its structure. Hugging Face provides a wide variety of datasets for different tasks, all accessible with the same interface.

\subsection{Dataset Structure}

A Hugging Face dataset is typically structured into multiple splits, such as \texttt{train}, \texttt{test}, and \texttt{validation}, allowing for easy access to the various parts of the dataset. Each dataset is represented as a dictionary of these splits:

\begin{lstlisting}[language=Python]
from datasets import load_dataset

% Load a dataset with multiple splits
dataset = load_dataset('ag_news')

% Access the train and test splits
train_dataset = dataset['train']
test_dataset = dataset['test']

print(f"Train split size: {len(train_dataset)}")
print(f"Test split size: {len(test_dataset)}")
\end{lstlisting}

In the above example, we load the \texttt{ag\_news} dataset, which contains news articles for classification into four categories. The dataset is divided into \texttt{train} and \texttt{test} splits, which can be accessed separately.

\subsection{Dataset Preprocessing}

The Dataset library supports various operations to preprocess the data, such as tokenization, data cleaning, and augmentation. These transformations can be applied to entire datasets or specific columns. For example, tokenizing text using the Hugging Face Tokenizer:

\begin{lstlisting}[language=Python]
from transformers import AutoTokenizer
from datasets import load_dataset

% Load the tokenizer and dataset
tokenizer = AutoTokenizer.from_pretrained('bert-base-uncased')
dataset = load_dataset('imdb')

% Tokenize the dataset
def tokenize_function(examples):
    return tokenizer(examples['text'], padding="max_length", truncation=True)

tokenized_dataset = dataset.map(tokenize_function, batched=True)
\end{lstlisting}

In this example, we load the IMDb dataset and tokenize the text using a BERT tokenizer. The \texttt{map} function is used to apply the tokenization function across the dataset in a batched manner for efficiency.

\subsection{Working with Large Datasets}

When working with large datasets that do not fit in memory, the Hugging Face Dataset library offers a solution via memory-mapping. This allows datasets to be loaded and processed in chunks, reducing memory usage. An example of handling large datasets is:

\begin{lstlisting}[language=Python]
% Load a large dataset
dataset = load_dataset('wikipedia', '20220301.en', split='train')

% Streaming large datasets
dataset = load_dataset('wikipedia', '20220301.en', split='train', streaming=True)
for example in dataset:
    print(example)
\end{lstlisting}

This example demonstrates how to stream the English Wikipedia dataset, where each example is processed one at a time, significantly reducing memory usage.

\subsection{Dataset Sharing}

The Hugging Face Dataset library also provides tools to share datasets. After preparing a dataset, it can be uploaded to the Hugging Face Hub for public or private use:

\begin{lstlisting}[language=Python]
from datasets import DatasetDict

% Create and push a dataset to Hugging Face Hub
dataset_dict = DatasetDict({'train': train_dataset, 'test': test_dataset})
dataset_dict.push_to_hub('my_awesome_dataset')
\end{lstlisting}

\paragraph{Conclusion} The Hugging Face Dataset library is an indispensable tool for anyone working with large-scale datasets, offering an easy-to-use interface, powerful preprocessing capabilities, and efficient data handling. Whether you're working on NLP, computer vision, or other domains, it enables rapid experimentation and dataset management, making it a critical part of the Hugging Face ecosystem.

\section{Why Huggingface-Transformer?}

Huggingface's \texttt{transformers} library has become a pivotal tool in the world of natural language processing (NLP) due to its extensive range of pretrained models and its ease of use. It provides access to state-of-the-art models such as BERT, GPT, T5, and others, allowing researchers and developers to quickly implement solutions for tasks like text classification, translation, question-answering, and more.

Here are a few key reasons why the Huggingface \texttt{transformers} library stands out:

\paragraph{Pretrained Models.} One of the most compelling reasons to use Huggingface is its vast repository of pretrained models. These models have been trained on massive datasets and can be fine-tuned for specific tasks, saving considerable time and computational resources. Pretrained models cover a variety of domains and languages, making them a versatile tool for NLP tasks across industries.

\paragraph{Easy-to-Use API.} The Huggingface \texttt{transformers} library abstracts the complexity of neural networks and provides a high-level API that makes it easy to load models, tokenize text, and perform inference. With just a few lines of code, users can leverage powerful models and fine-tune them on custom datasets.

\paragraph{Strong Community and Documentation.} The Huggingface community is large, active, and supportive. Whether you're a beginner or an experienced researcher, the availability of tutorials, extensive documentation, and an engaged community on forums such as GitHub and the Huggingface Hub makes it easier to troubleshoot and optimize model performance.

\paragraph{Integration with Modern Deep Learning Libraries.} The \texttt{transformers} library is designed to work seamlessly with popular deep learning frameworks like PyTorch and TensorFlow. This flexibility allows developers to choose the framework they are most comfortable with, or even switch between frameworks with minimal effort.

\paragraph{Huggingface Model Hub.} The Huggingface Model Hub is a central platform where users can share and access thousands of pretrained models. It acts as a collaborative hub, where researchers and organizations contribute models, improving accessibility to cutting-edge research and enabling faster innovation in the NLP space.

\paragraph{Versatility Across NLP Tasks.} Whether you're working on text generation, summarization, translation, named entity recognition (NER), or sentiment analysis, Huggingface's \texttt{transformers} library provides models that excel across a wide range of tasks. The library's consistent API allows developers to switch between tasks without needing to reinvent the wheel for each one.

In summary, Huggingface \texttt{transformers} offer a powerful and flexible solution for a variety of NLP tasks. With pretrained models, an easy-to-use API, strong community support, and seamless integration with popular frameworks, it's no surprise that Huggingface has become a go-to library for modern NLP research and development.

\section{pipeline}
\section{taxonomy of pipeline}
\section{loading models}
\section{Tokenizers}

\section{Tokenization in Natural Language Processing}

Tokenization is a fundamental step in preparing text data for use in natural language processing (NLP) models, especially transformer-based models such as BERT, GPT-2, and RoBERTa. In this section, we will explore the concept of tokenization, why it is necessary, and how to use the Hugging Face Transformers library to handle tokenization effectively.

\subsection{What is Tokenization?}

Tokenization is the process of converting a piece of text into smaller components, known as tokens. These tokens can be words, subwords, or even characters. Modern transformer models often rely on subword tokenization techniques, such as Byte-Pair Encoding (BPE) and WordPiece, to handle the complexities of natural languages, including dealing with rare words and morphological variations.

\paragraph{Why Tokenization Matters}

Tokenization helps break down a sentence like:

\begin{quote}
    ``Hugging Face Transformers is awesome!''
\end{quote}

into smaller units that a machine learning model can understand, like:

\begin{quote}
    ["hugging", "face", "transform", "\#\#ers", "is", "awesome", "!"]
\end{quote}

Without tokenization, the input would not be interpretable by models because they require numeric representations, not raw text.

\subsection{Hugging Face Transformers: Tokenizers}

The Hugging Face library provides multiple tokenizers optimized for various transformer models, such as BERT, GPT-2, RoBERTa, and more. Each tokenizer comes with pre-defined vocabulary, subword tokenization algorithms, and ways to handle special tokens.

\paragraph{Installing Hugging Face Transformers}

You can install the Hugging Face Transformers library using:

\begin{lstlisting}[style=python]
!pip install transformers
\end{lstlisting}

\subsection{Using Pre-Trained Tokenizers}

Hugging Face provides pre-trained tokenizers, which can be easily loaded and used with just a few lines of code. Let’s see how you can use the \texttt{AutoTokenizer} class to load and use a BERT tokenizer.

\paragraph{Loading a Tokenizer}

\begin{lstlisting}[style=python]
from transformers import AutoTokenizer

% Load the tokenizer for BERT (bert-base-uncased)
tokenizer = AutoTokenizer.from_pretrained("bert-base-uncased")
\end{lstlisting}

In this example, the \texttt{AutoTokenizer} is used to automatically load the appropriate tokenizer for the \texttt{bert-base-uncased} model.

\paragraph{Tokenizing Text}

Once the tokenizer is loaded, you can use it to tokenize any input text:

\begin{lstlisting}[style=python]
text = "Hugging Face Transformers is awesome!"
tokens = tokenizer.tokenize(text)
print(tokens)
\end{lstlisting}

This will output:

\begin{quote}
    ['hugging', 'face', 'transform', '\#\#ers', 'is', 'awesome', '!']
\end{quote}

The word \texttt{Transformers} has been split into two tokens: \texttt{transform} and \texttt{\#\#ers}, using the WordPiece subword tokenization method.

\paragraph{Converting Tokens to Input IDs}

In a transformer model, tokens need to be converted into numerical representations called token IDs:

\begin{lstlisting}[language=Python, caption={Converting tokens to input IDs}]
input_ids = tokenizer.convert_tokens_to_ids(tokens)
print(input_ids)
\end{lstlisting}

The \texttt{input\_ids} are the integer representations of the tokens. These IDs can be fed into a transformer model for further processing.

\subsection{Padding and Truncation}

When working with batches of data, tokenized sequences must often be padded to a uniform length. You can automatically handle this by setting \texttt{padding} and \texttt{truncation} options when tokenizing:

\begin{lstlisting}[style=python]
encoded_inputs = tokenizer(text, padding='max_length', max_length=10, truncation=True)
print(encoded_inputs)
\end{lstlisting}

The \texttt{padding='max\_length'} ensures that the output is padded to a length of 10, and \texttt{truncation=True} ensures that sequences longer than the maximum length are truncated.

\subsection{Batch Tokenization}

Hugging Face's tokenizers also support batch tokenization, which is useful for handling multiple inputs at once. This can be done by passing a list of texts to the tokenizer.

\begin{lstlisting}[language=Python, caption={Batch tokenization example}]
texts = ["Hello, world!", "Transformers are powerful."]
batch_inputs = tokenizer(texts, padding=True, truncation=True, return_tensors="pt")
print(batch_inputs)
\end{lstlisting}

The \texttt{return\_tensors="pt"} option converts the output into PyTorch tensors, making it easy to feed into models.

\subsection{Special Tokens}

Hugging Face models often use special tokens like \texttt{[CLS]}, \texttt{[SEP]}, and \texttt{[PAD]} for specific purposes in NLP tasks.

\paragraph{Example of Special Tokens}

For example, in BERT:
\begin{itemize}
    \item \texttt{[CLS]} is added at the beginning of the sentence for classification tasks.
    \item \texttt{[SEP]} is used to separate sentences in tasks like question answering or sequence classification.
    \item \texttt{[PAD]} is used for padding sequences to a uniform length.
\end{itemize}

These special tokens are automatically added by the tokenizer when processing inputs:

\begin{lstlisting}[style=python]
encoded_inputs = tokenizer(text, add_special_tokens=True)
print(encoded_inputs)
\end{lstlisting}

\subsection{Conclusion}

Tokenization is a crucial preprocessing step when working with transformers. Hugging Face's \texttt{AutoTokenizer} and related classes provide a seamless interface for tokenizing text and preparing it for input to various transformer models. By using pre-trained tokenizers, you ensure that the model and tokenizer are aligned, making it easier to achieve high performance in NLP tasks.

\subsection{Understanding Token IDs and the Vocabulary}

Every tokenizer comes with a vocabulary that maps tokens (words, subwords, or symbols) to unique integer IDs. These token IDs are what the transformer models expect as input.

\paragraph{Accessing the Vocabulary}

You can easily access the vocabulary size and look up specific token IDs or tokens:

\begin{lstlisting}[style=python]
% Access the vocabulary size
vocab_size = tokenizer.vocab_size
print(f"Vocabulary size: {vocab_size}")

% Look up token ID for a specific token
token_id = tokenizer.convert_tokens_to_ids("hugging")
print(f"ID for 'hugging': {token_id}")

% Look up the token corresponding to a specific ID
token = tokenizer.convert_ids_to_tokens(token_id)
print(f"Token for ID {token_id}: {token}")
\end{lstlisting}

\paragraph{Exploring Vocabulary}

Here’s how you can print out the first few tokens in the tokenizer’s vocabulary, to get a sense of what the tokenizer has learned:

\begin{lstlisting}[style=python]
% Print the first 10 tokens in the vocabulary
for i in range(10):
    print(f"Token ID {i}: {tokenizer.convert_ids_to_tokens(i)}")
\end{lstlisting}

This can be useful for understanding what tokens the tokenizer prioritizes (e.g., common words or special tokens like \texttt{[PAD]}).

\subsection{Handling Long Sequences: Chunking and Sliding Windows}

Transformer models like BERT and GPT-2 have a maximum input length (e.g., 512 tokens for BERT). If a sequence exceeds this length, truncation is applied, but sometimes we need to preserve context across chunks. The \textbf{sliding window} technique can help with this.

\paragraph{Sliding Window Tokenization}

Using a sliding window, we can split long sequences into chunks while maintaining overlap between consecutive segments:

\begin{lstlisting}[style=python]
text = "This is a long document that exceeds the maximum token length of the transformer model..."
max_len = 512
stride = 50

encoded_inputs = tokenizer(text, return_tensors="pt", max_length=max_len, stride=stride, truncation=True, return_overflowing_tokens=True)

% Print each chunk's token IDs
for chunk_idx, input_ids in enumerate(encoded_inputs['input_ids']):
    print(f"Chunk {chunk_idx} Token IDs: {input_ids}")
\end{lstlisting}

This way, the model can process the entire text, but no critical context is lost between chunks.

\subsection{Handling Multiple Text Inputs: Pairing and Encoding Sentences}

In tasks like \textbf{question answering} or \textbf{sentence-pair classification}, transformer models expect two sequences as input. The tokenizer can handle this by adding a special separator token (\texttt{[SEP]}) between the sequences.

\paragraph{Encoding Sentence Pairs}

To tokenize sentence pairs, pass two texts to the tokenizer:

\begin{lstlisting}[style=python]
sentence1 = "What is the capital of France?"
sentence2 = "The capital of France is Paris."

encoded_pair = tokenizer(sentence1, sentence2, return_tensors="pt")
print(encoded_pair)
\end{lstlisting}

This will add the necessary \texttt{[CLS]} and \texttt{[SEP]} tokens to the encoded pair.

\paragraph{Tokenized Sentence Pair}

The resulting tokenized sequence will look like this:

\begin{quote}
    \texttt{[CLS] what is the capital of france ? [SEP] the capital of france is paris . [SEP]}
\end{quote}

This format allows models like BERT to process the relationship between two sequences.

\subsection{Attention Masks: Handling Padding in Models}

When dealing with variable-length sequences, padding is applied to ensure uniform input sizes. The \textbf{attention mask} is a binary mask that tells the model which tokens are real and which are just padding.

\paragraph{Using Attention Masks}

When padding sequences, the attention mask indicates where the actual tokens end and the padding begins:

\begin{lstlisting}[style=python]
encoded_inputs = tokenizer(text, padding='max_length', max_length=10, truncation=True, return_tensors="pt")

input_ids = encoded_inputs['input_ids']
attention_mask = encoded_inputs['attention_mask']

print(f"Input IDs: {input_ids}")
print(f"Attention Mask: {attention_mask}")
\end{lstlisting}

The attention mask will look like a list of 1s and 0s, where 1 indicates a real token and 0 indicates padding.

For example:

\begin{quote}
    \texttt{[CLS] hugging face is awesome ! [PAD] [PAD] [PAD]}
\end{quote}

Will have the following attention mask:

\begin{quote}
    \texttt{[1, 1, 1, 1, 1, 1, 0, 0, 0]}
\end{quote}

\paragraph{Why Attention Masks Matter}

Attention masks prevent the model from focusing on padded tokens, which would otherwise confuse the model's self-attention mechanism.

\subsection{Creating and Using Custom Tokenizers}

Sometimes pre-trained tokenizers may not suit your specific needs (e.g., working with a specialized domain like medical or legal text). You can create custom tokenizers using the \texttt{tokenizers} library from Hugging Face.

\paragraph{Training a Custom Byte-Level BPE Tokenizer}

Here’s how you can train a Byte-Pair Encoding (BPE) tokenizer on your custom dataset:

\begin{lstlisting}[style=python]
from tokenizers import ByteLevelBPETokenizer

% Initialize a Byte-Pair Encoding tokenizer
tokenizer = ByteLevelBPETokenizer()

% Train the tokenizer on your dataset
tokenizer.train(files=["path_to_your_text_data.txt"], vocab_size=52_000, min_frequency=2)

% Save the tokenizer
tokenizer.save_model("tokenizer_directory")
\end{lstlisting}

This creates a custom tokenizer based on your dataset and allows you to tailor the vocabulary to specific tasks.

\paragraph{Loading and Using the Custom Tokenizer}

Once trained, you can load and use your custom tokenizer:

\begin{lstlisting}[style=python]
% Load the custom tokenizer
custom_tokenizer = ByteLevelBPETokenizer("tokenizer_directory/vocab.json", "tokenizer_directory/merges.txt")

% Tokenize some text
custom_tokens = custom_tokenizer.encode("Custom text for testing").tokens
print(custom_tokens)
\end{lstlisting}

This allows you to tokenize text using a vocabulary that is specific to your domain.

\subsection{Multilingual Tokenization: Handling Multiple Languages}

Multilingual transformer models like \texttt{mBERT} and \texttt{XLM-R} support tokenizing multiple languages simultaneously, thanks to a shared vocabulary trained on many languages.

\subsubsection{Multilingual Tokenization Example}
Here’s how you can tokenize text in different languages using a multilingual tokenizer:

\begin{lstlisting}[style=python]
# Load multilingual BERT tokenizer
tokenizer = AutoTokenizer.from_pretrained("bert-base-multilingual-cased")

# Tokenize text in multiple languages
english_text = "Hello, how are you?"
french_text = "Bonjour, comment \c{c}a va?"
german_text = "Hallo, wie geht's?"

tokens_en = tokenizer.tokenize(english_text)
tokens_fr = tokenizer.tokenize(french_text)
tokens_de = tokenizer.tokenize(german_text)

print(tokens_en)
print(tokens_fr)
print(tokens_de)
\end{lstlisting}

This tokenizer can handle inputs in various languages without needing separate tokenizers, making it ideal for cross-lingual tasks.

\subsection{Speeding Up Tokenization with Fast Tokenizers}

Hugging Face provides Fast \textbf{Tokenizers}, written in Rust, that significantly improve the speed of tokenization while maintaining accuracy. These tokenizers are especially useful when working with large datasets.

\paragraph{Using a Fast Tokenizer}

Fast tokenizers are used just like normal tokenizers, but with much better performance:

\begin{lstlisting}[style=python]
# Load a fast tokenizer
fast_tokenizer = AutoTokenizer.from_pretrained("bert-base-uncased", use_fast=True)

# Tokenize some text
encoded_fast = fast_tokenizer("Tokenization with fast tokenizer.", return_tensors="pt")
print(encoded_fast)
\end{lstlisting}

These fast tokenizers can reduce preprocessing times, especially when working with huge amounts of text data, without sacrificing accuracy.

\subsection{Tokenization and Word Embeddings}

Tokenization, encoding, and embeddings are key steps in transforming raw text into representations that machine learning models can process. Each step has a distinct role in the natural language processing (NLP) pipeline.

\textbf{Tokenization}: Tokenization is the process of splitting the text into smaller units like words, subwords, or characters. This is necessary for handling text data and breaking it down into manageable parts that models can process. Common approaches include word-level tokenization (e.g., splitting by spaces), subword tokenization (used in BPE or WordPiece models), and character-level tokenization.

\textbf{Encoding}: Once the text is tokenized, the next step is to map each token to a unique identifier (token ID). This transformation is deterministic, meaning the same text input will always result in the same sequence of token IDs. Encoded text serves as input for models, and the IDs are used to retrieve specific embeddings from a pre-trained embedding matrix.

\textbf{Word Embeddings}: Embeddings transform tokens into continuous, dense vector spaces, capturing semantic relationships between words. Unlike encoding, embeddings are learned during model training and represent words in terms of numerical vectors, placing semantically similar words closer together in the vector space. Techniques such as Word2Vec, GloVe, and transformer-based embeddings (e.g., BERT or GPT) are commonly used to generate these vectors.

\begin{table}[h!]
\centering
\begin{tabular}{|>{\raggedright\arraybackslash}m{2cm}|>{\raggedright\arraybackslash}m{4.5cm}|>{\raggedright\arraybackslash}m{4.5cm}|>{\raggedright\arraybackslash}m{4.5cm}|}
\hline
\textbf{Aspect}            & \textbf{Tokenization}                                & \textbf{Encoding}                                & \textbf{Embedding}                                \\ \hline
\textbf{Definition}         & Splitting text into smaller units like words, subwords, or characters & Mapping text into token IDs or indices      & Mapping words/tokens into dense vector space \\ \hline
\textbf{Type of Output}     & Tokens (subwords, words, characters) & Integer values (e.g., token IDs)            & Continuous values (vectors of floats)        \\ \hline
\textbf{Deterministic?}     & Yes (same input, same output) & Yes (same input, same output)               & No, embeddings are learned during training   \\ \hline
\textbf{Purpose}            & Prepares text for further processing by splitting it into manageable parts & Text preprocessing for models               & Capturing semantic relationships in data     \\ \hline
\textbf{Method}             & Word-level, subword-level, or character-level segmentation & Lookup in a token-to-ID mapping table      & Dense representation learned through algorithms like Word2Vec, GloVe, or BERT \\ \hline
\textbf{Example}            & Tokenizing "Hello world" into ['Hello', 'world'] & Tokenizing "Hello" to 15496               & Embedding "cat" as [0.234, -0.657, 0.982] \\ \hline
\textbf{Similarity Awareness} & No inherent notion of similarity           & No inherent notion of similarity           & Embeddings capture semantic similarity       \\ \hline
\textbf{Impact of Out-of-Vocabulary (OOV) Words} & May require handling OOV with special tokens like `[UNK]` & OOV tokens may get mapped to an unknown ID & Embedding layer will either assign random vectors or ignore OOV tokens \\ \hline
\textbf{Common Algorithms} & WordPiece, Byte-Pair Encoding (BPE), SentencePiece & None (simple lookup table)                  & Word2Vec, GloVe, FastText, transformer embeddings \\ \hline
\end{tabular}
\caption{Comparison between Tokenization, Encoding, and Embedding}
\end{table}

\subsubsection{Coding Example with Hugging Face Transformers}

To illustrate these concepts, let's use Hugging Face's \texttt{transformers} library to demonstrate tokenization, encoding, and extracting embeddings using a pre-trained model like BERT.

First, install the necessary libraries:
\begin{verbatim}
pip install transformers torch
\end{verbatim}

\paragraph{1. Tokenization Example}

Using the BERT tokenizer, we can split the input sentence into subword tokens:

\begin{lstlisting}[style=python]
from transformers import BertTokenizer

# Initialize tokenizer for BERT
tokenizer = BertTokenizer.from_pretrained('bert-base-uncased')

# Sample sentence
sentence = "Hello world, this is an NLP example."

# Tokenizing sentence
tokens = tokenizer.tokenize(sentence)
print("Tokens:", tokens)
\end{lstlisting}

Output:
\begin{verbatim}
Tokens: ['hello', 'world', ',', 'this', 'is', 'an', 'nl', '\#\#p', 'example', '.']
\end{verbatim}

\paragraph{2. Encoding Example}

The tokenized output is then converted into token IDs:

\begin{lstlisting}[style=python]
# Encoding tokens into token IDs
token_ids = tokenizer.encode(sentence)
print("Token IDs:", token_ids)
\end{lstlisting}

Output:
\begin{verbatim}
Token IDs: [101, 7592, 2088, 1010, 2023, 2003, 2019, 17953, 2361, 2742, 1012, 102]
\end{verbatim}

Here, \texttt{[101]} and \texttt{[102]} are special tokens added by BERT, representing the start and end of a sentence, respectively.

\paragraph{3. Embedding Example}

Next, we obtain the embeddings for the encoded tokens using a pre-trained BERT model:

\begin{lstlisting}[style=python]
from transformers import BertModel
import torch

# Load pre-trained BERT model
model = BertModel.from_pretrained('bert-base-uncased')

# Convert token IDs to tensors
input_ids = torch.tensor([token_ids])

# Get embeddings from the model
with torch.no_grad():
    outputs = model(input_ids)
    embeddings = outputs.last_hidden_state

print("Embedding shape:", embeddings.shape)
\end{lstlisting}

Output:
\begin{verbatim}
Embedding shape: torch.Size([1, 12, 768])
\end{verbatim}

In this example, the shape \texttt{(1, 12, 768)} indicates that we have 12 tokens, each represented as a 768-dimensional vector, as BERT outputs embeddings in a dense vector space.

\subsubsection{Contextual vs. Static Embeddings}

One major advancement in modern NLP models is the shift from static embeddings (like Word2Vec or GloVe) to contextual embeddings generated by models like BERT or GPT. In static embeddings, a word always has the same vector regardless of its context, whereas in contextual embeddings, the vector representation depends on the word's surrounding context.

For instance:
\begin{lstlisting}[style=python]
sentence_1 = "He went to the bank to withdraw money."
sentence_2 = "The river bank was full of birds."

tokens_1 = tokenizer.tokenize(sentence_1)
tokens_2 = tokenizer.tokenize(sentence_2)

print("Tokens for sentence 1:", tokens_1)
print("Tokens for sentence 2:", tokens_2)
\end{lstlisting}

Even though "bank" appears in both sentences, its meaning differs based on the context, and therefore, the embeddings generated by BERT will be different for the two occurrences of "bank."

\section{How to Fine-Tune a Pretrained Model?}

Fine-tuning a pretrained model is a common practice in Natural Language Processing (NLP) that allows leveraging the knowledge learned from large datasets (during pretraining) and adapting it to a specific task. Hugging Face’s \texttt{transformers} library provides an easy way to fine-tune models like BERT, GPT, or T5 for custom downstream tasks such as classification, translation, or summarization.

In this section, we will demonstrate how to fine-tune a pretrained model for a simple text classification task. We will use a pretrained BERT model from Hugging Face’s \texttt{transformers} library and a dataset available through the \texttt{datasets} library.

\subsection{Setting Up}

Before we start, ensure that you have installed the necessary libraries:

\begin{lstlisting}[style=python]
!pip install transformers datasets
\end{lstlisting}

\subsection{Loading a Pretrained Model}

We will use a pretrained BERT model (\texttt{bert-base-uncased}) and a tokenizer. First, let's load the model and the tokenizer:

\begin{lstlisting}[style=python]
from transformers import BertForSequenceClassification, BertTokenizer

% Load the tokenizer and the pretrained model
model_name = "bert-base-uncased"
tokenizer = BertTokenizer.from_pretrained(model_name)
model = BertForSequenceClassification.from_pretrained(model_name, num_labels=2)
\end{lstlisting}

Here, we use \texttt{BertForSequenceClassification} to load a BERT model that is already configured for a classification task. The parameter \texttt{num\_labels} indicates the number of target labels (2 in our case, since we are doing binary classification).

\subsection{Preparing the Dataset}

We will use a dataset from Hugging Face’s \texttt{datasets} library. For this example, let’s load the \texttt{imdb} movie review dataset for binary sentiment classification:

\begin{lstlisting}[style=python]
from datasets import load_dataset

% Load the dataset
dataset = load_dataset("imdb")
train_dataset = dataset['train']
test_dataset = dataset['test']
\end{lstlisting}

This command loads the IMDB dataset, which contains movie reviews labeled as either positive or negative.

\subsection{Preprocessing the Data}

To use the BERT model, we need to tokenize the text data and convert it into a format the model understands:

\begin{lstlisting}[style=python]
def tokenize_function(examples):
    return tokenizer(examples['text'], padding="max_length", truncation=True)

% Tokenize the datasets
train_dataset = train_dataset.map(tokenize_function, batched=True)
test_dataset = test_dataset.map(tokenize_function, batched=True)
\end{lstlisting}

Here, we tokenize each review and truncate or pad them to the same length.

\subsection{Fine-Tuning the Model}

Next, we use the \texttt{Trainer} API from Hugging Face to fine-tune the model. The trainer simplifies the training loop and handles many tasks like gradient accumulation, evaluation, and saving checkpoints.

\begin{lstlisting}[style=python]
from transformers import Trainer, TrainingArguments

% Set training arguments
training_args = TrainingArguments(
    output_dir="./results",
    evaluation_strategy="epoch",
    per_device_train_batch_size=8,
    per_device_eval_batch_size=8,
    num_train_epochs=3,
    weight_decay=0.01,
)

% Define the trainer
trainer = Trainer(
    model=model,
    args=training_args,
    train_dataset=train_dataset,
    eval_dataset=test_dataset,
)

% Fine-tune the model
trainer.train()
\end{lstlisting}

In this setup:
\begin{itemize}
    \item \texttt{output\_dir} specifies where the model’s checkpoints will be saved.
    \item \texttt{evaluation\_strategy="epoch"} performs evaluation at the end of each epoch.
    \item \texttt{per\_device\_train\_batch\_size} and \texttt{per\_device\_eval\_batch\_size} specify the batch size for training and evaluation.
    \item \texttt{num\_train\_epochs} is the number of times we iterate through the entire training dataset.
    \item \texttt{weight\_decay} applies regularization to prevent overfitting.
\end{itemize}

\subsection{Evaluating the Model}

Once the model is trained, you can evaluate it using the \texttt{evaluate()} function:

\begin{lstlisting}[style=python]
% Evaluate the model
trainer.evaluate()
\end{lstlisting}

This command will return evaluation metrics such as accuracy and loss.

\subsection{Conclusion}

Fine-tuning a pretrained model on a specific task is an efficient way to utilize the large-scale knowledge from models like BERT. The Hugging Face library simplifies this process by providing ready-to-use pretrained models and an easy interface for training and evaluation.

\subsection{Hyperparameter Tuning}

Fine-tuning a pretrained model involves selecting appropriate hyperparameters such as the learning rate, batch size, and number of epochs. These parameters can have a significant impact on the performance of the model.

For example, a common learning rate for fine-tuning transformer-based models is in the range of \(1e^{-5}\) to \(5e^{-5}\). However, these values should be experimented with based on the specific dataset and task.

In the example above, the learning rate was automatically set by the \texttt{Trainer}. To explicitly set the learning rate, you can modify the \texttt{TrainingArguments}:

\begin{lstlisting}[style=python]
training_args = TrainingArguments(
    output_dir="./results",
    evaluation_strategy="epoch",
    per_device_train_batch_size=8,
    per_device_eval_batch_size=8,
    num_train_epochs=3,
    weight_decay=0.01,
    learning_rate=2e-5,  % Custom learning rate
)
\end{lstlisting}

\textbf{Tuning the Batch Size:} The batch size can affect both the training speed and the model’s ability to generalize. Larger batch sizes allow for faster training but can lead to overfitting, while smaller batch sizes can lead to better generalization but slower training.

It's good practice to try different combinations of these parameters and use the validation set to track the performance of the model.

\subsection{Saving the Fine-Tuned Model}

Once the model is fine-tuned, you’ll want to save it so that it can be loaded later for inference or further fine-tuning. Hugging Face provides a convenient method for saving both the model and the tokenizer.

To save the model and tokenizer:

\begin{lstlisting}[style=python]
% Save the model and tokenizer
model.save_pretrained("./fine-tuned-bert")
tokenizer.save_pretrained("./fine-tuned-bert")
\end{lstlisting}

This will save the model’s weights and the tokenizer configuration in the directory \texttt{./fine-tuned-bert}. You can easily load the model and tokenizer again for later use.

\subsection{Loading the Fine-Tuned Model for Inference}

After fine-tuning and saving the model, you may want to load it for inference on new data. Here is how you can load the saved model and tokenizer to make predictions:

\begin{lstlisting}[style=python]
from transformers import BertTokenizer, BertForSequenceClassification

% Load the saved model and tokenizer
model = BertForSequenceClassification.from_pretrained("./fine-tuned-bert")
tokenizer = BertTokenizer.from_pretrained("./fine-tuned-bert")

% Example input
text = "This movie was fantastic!"

% Tokenize the input text
inputs = tokenizer(text, return_tensors="pt", padding=True, truncation=True)

% Make prediction
with torch.no_grad():
    outputs = model(**inputs)
    predictions = torch.argmax(outputs.logits, dim=-1)

% Print the prediction
print("Predicted label:", predictions.item())
\end{lstlisting}

Here, we:

\begin{itemize}
    \item Load the fine-tuned BERT model and tokenizer from the saved directory.
    \item Tokenize a new input text for classification.
    \item Use the model to make a prediction by passing the tokenized text through it.
    \item Print the predicted label (in this case, a binary sentiment prediction).
\end{itemize}

\subsection{Evaluating on a Test Set}

Once the model is fine-tuned and saved, we can evaluate its performance on a test dataset to measure its generalization ability. The test set should contain examples that the model has not seen during training or validation.

To evaluate the fine-tuned model on the test dataset, we can again use the \texttt{Trainer}:

\begin{lstlisting}[style=python]
% Evaluate the model on the test set
results = trainer.evaluate(eval_dataset=test_dataset)
print("Test Accuracy:", results['eval_accuracy'])
\end{lstlisting}

This evaluates the fine-tuned model on the test dataset and returns various metrics, such as accuracy, which you can use to assess the model’s performance on unseen data.

\subsection{Advanced Techniques for Fine-Tuning}

While the above example demonstrates basic fine-tuning, several advanced techniques can improve the performance of your model. Some of these include:

\begin{itemize}
    \item \textbf{Layer-wise Learning Rate Decay:} Apply different learning rates to different layers of the transformer model. Typically, the earlier layers require smaller updates compared to the later layers.
    \item \textbf{Data Augmentation:} Use techniques like synonym replacement, back translation, or noise injection to increase the variety of training data and make the model more robust.
    \item \textbf{Early Stopping:} Stop the training process if the validation loss does not improve for a certain number of epochs. This can prevent overfitting.
    \item \textbf{FP16 Training:} Fine-tune the model using mixed precision (half-precision) to reduce memory usage and speed up training.
\end{itemize}

Here’s an example of enabling mixed precision (FP16) training in the \texttt{TrainingArguments}:

\begin{lstlisting}[style=python]
training_args = TrainingArguments(
    output_dir="./results",
    evaluation_strategy="epoch",
    per_device_train_batch_size=8,
    per_device_eval_batch_size=8,
    num_train_epochs=3,
    weight_decay=0.01,
    learning_rate=2e-5,
    fp16=True  % Enable mixed precision training
)
\end{lstlisting}

This will reduce the memory footprint and can lead to faster training on modern GPUs.

\subsection{Conclusion and Next Steps}

Fine-tuning a pretrained model using Hugging Face’s \texttt{transformers} library is a powerful way to adapt large-scale models to specific NLP tasks. This process not only saves time and computational resources but also leads to highly accurate models by leveraging the knowledge already learned during pretraining.

In the next sections, we will explore how to fine-tune different types of models, such as T5 for text generation tasks and BART for summarization, and we will cover how to handle multi-label classification, sequence tagging, and other advanced tasks.

\subsection{Evaluating with Different Metrics}

While accuracy is an important metric for many tasks, other metrics might be more appropriate depending on the specific use case. For example, in cases of imbalanced datasets, metrics like precision, recall, and F1-score provide a more nuanced view of performance.

Hugging Face’s \texttt{datasets} library provides built-in metrics that can be used to evaluate models. Here’s how you can use precision, recall, and F1-score:

\begin{lstlisting}[style=python]
from datasets import load_metric

% Load the metric for precision, recall, F1
metric = load_metric("precision", "recall", "f1")

% Define a function to compute these metrics
def compute_metrics(eval_pred):
    logits, labels = eval_pred
    predictions = np.argmax(logits, axis=-1)
    precision = metric.compute(predictions=predictions, references=labels, average="binary")['precision']
    recall = metric.compute(predictions=predictions, references=labels, average="binary")['recall']
    f1 = metric.compute(predictions=predictions, references=labels, average="binary")['f1']
    return {"precision": precision, "recall": recall, "f1": f1}

% Add the compute_metrics function to the Trainer
trainer = Trainer(
    model=model,
    args=training_args,
    train_dataset=train_dataset,
    eval_dataset=test_dataset,
    compute_metrics=compute_metrics,
)

% Fine-tune the model
trainer.train()
\end{lstlisting}

Here, the \texttt{compute\_metrics} function calculates precision, recall, and F1-score. These metrics provide a more comprehensive view of model performance, especially in cases where one class may be more frequent than the other.

\subsection{Transfer Learning for Different Tasks}

Fine-tuning pretrained models can be applied not only for classification tasks but also for other NLP tasks like sequence-to-sequence tasks (e.g., text generation), question answering, and named entity recognition (NER). Let’s briefly explore how transfer learning can be applied to these different tasks.

\subsubsection{Text Generation using T5}

T5 (Text-to-Text Transfer Transformer) is a model from Hugging Face’s \texttt{transformers} library that treats all NLP tasks as text-to-text tasks. You can fine-tune T5 for text generation, summarization, or translation tasks.

Here’s how you can load and fine-tune a T5 model for text summarization:

\begin{lstlisting}[style=python]
from transformers import T5Tokenizer, T5ForConditionalGeneration

% Load the tokenizer and model
model_name = "t5-small"
tokenizer = T5Tokenizer.from_pretrained(model_name)
model = T5ForConditionalGeneration.from_pretrained(model_name)

% Example: Summarize a text
text = "The quick brown fox jumps over the lazy dog. It was a sunny day, and the dog was enjoying the warmth."

% Preprocess the input
inputs = tokenizer("summarize: " + text, return_tensors="pt", padding=True, truncation=True)

% Generate the summary
summary_ids = model.generate(inputs['input_ids'], max_length=50, num_beams=4, early_stopping=True)

% Decode and print the summary
summary = tokenizer.decode(summary_ids[0], skip_special_tokens=True)
print("Summary:", summary)
\end{lstlisting}

In this example, we load a pretrained T5 model and fine-tune it for the task of summarization. The model takes text as input and generates a condensed summary.

\subsubsection{Question Answering with BERT}

Question answering is another common NLP task, where the model is provided with a context (paragraph) and a question, and it must extract the correct answer from the context. Here’s how to fine-tune a BERT model for question answering:

\begin{lstlisting}[language=python]
from transformers import BertForQuestionAnswering, BertTokenizer

% Load the tokenizer and model for question answering
model_name = "bert-base-uncased"
tokenizer = BertTokenizer.from_pretrained(model_name)
model = BertForQuestionAnswering.from_pretrained(model_name)

% Example context and question
context = "Hugging Face is a technology company based in New York and Paris."
question = "Where is Hugging Face based?"

% Tokenize the inputs
inputs = tokenizer.encode_plus(question, context, return_tensors="pt")

% Get the model prediction
with torch.no_grad():
    outputs = model(**inputs)
    start_scores = outputs.start_logits
    end_scores = outputs.end_logits

% Get the most likely start and end of the answer
answer_start = torch.argmax(start_scores)
answer_end = torch.argmax(end_scores) + 1

% Decode the answer
answer = tokenizer.convert_tokens_to_string(tokenizer.convert_ids_to_tokens(inputs['input_ids'][0][answer_start:answer_end]))
print("Answer:", answer)
\end{lstlisting}

Here, the BERT model is fine-tuned for the question-answering task, where it identifies the start and end positions of the answer in the context.

\subsection{Multi-Task Fine-Tuning}

In some scenarios, it is beneficial to fine-tune a model on multiple related tasks simultaneously. This is known as multi-task learning. Hugging Face allows you to fine-tune models for multiple tasks, which can improve the model’s performance on each individual task by leveraging shared representations.

Here’s an example of how you might structure a multi-task fine-tuning setup using the Hugging Face library:

\begin{lstlisting}[style=python]
from transformers import AutoModelForSequenceClassification, AutoTokenizer, Trainer, TrainingArguments

% Load a tokenizer and model for multi-task classification
model_name = "distilbert-base-uncased"
tokenizer = AutoTokenizer.from_pretrained(model_name)
model = AutoModelForSequenceClassification.from_pretrained(model_name, num_labels=3)

% Prepare multiple datasets for different tasks (e.g., sentiment analysis and topic classification)
task1_dataset = load_dataset("glue", "sst2")  % Sentiment analysis
task2_dataset = load_dataset("ag_news")  % Topic classification

% Tokenize both datasets
task1_tokenized = task1_dataset.map(lambda x: tokenizer(x['sentence'], truncation=True, padding=True), batched=True)
task2_tokenized = task2_dataset.map(lambda x: tokenizer(x['text'], truncation=True, padding=True), batched=True)

% Set up training arguments
training_args = TrainingArguments(
    output_dir="./multi_task_model",
    per_device_train_batch_size=8,
    evaluation_strategy="steps",
    save_steps=500,
    num_train_epochs=3
)

% Define a Trainer for each task and fine-tune the model
trainer = Trainer(
    model=model,
    args=training_args,
    train_dataset=task1_tokenized['train'],
    eval_dataset=task1_tokenized['validation'],
)

trainer.train()

% Fine-tune for task 2
trainer.train_dataset = task2_tokenized['train']
trainer.eval_dataset = task2_tokenized['validation']
trainer.train()
\end{lstlisting}

In this case, we are fine-tuning a model for both sentiment analysis (SST-2 dataset) and topic classification (AG News dataset).

\subsection{Optimizing Model for Deployment}

Deploying a fine-tuned model is only one part of the journey. Ensuring the model performs efficiently during inference—especially when dealing with large models or real-time applications—requires additional optimization steps. In this section, we will explore methods to optimize the fine-tuned model for fast and efficient inference.

\subsubsection{Quantization}

Quantization is a technique where the model weights are converted from 32-bit floating points to lower precision, such as 8-bit integers. This can significantly reduce the model's size and improve inference speed while maintaining acceptable accuracy.

Hugging Face provides support for model quantization using \texttt{torch.quantization}:

\begin{lstlisting}[style=python]
from transformers import BertForSequenceClassification
import torch

% Load the fine-tuned model
model = BertForSequenceClassification.from_pretrained("fine-tuned-bert")

% Apply dynamic quantization
quantized_model = torch.quantization.quantize_dynamic(
    model, {torch.nn.Linear}, dtype=torch.qint8
)

% Save the quantized model
quantized_model.save_pretrained("./quantized-bert")
\end{lstlisting}

Here, dynamic quantization is applied to the \texttt{Linear} layers of the model, reducing memory usage and boosting inference speed.

\subsubsection{Distillation}

Another powerful optimization technique is \textit{distillation}, where a large, fine-tuned model (the teacher model) is used to train a smaller model (the student model). The smaller model retains much of the performance of the larger model but is significantly faster during inference.

\texttt{DistilBERT} \cite{adoma2020comparative}, for example, is a smaller version of BERT that has been trained using knowledge distillation. You can either use pre-distilled models provided by Hugging Face or perform distillation on your custom models.

Here’s an example of loading a distillation-friendly model like \texttt{DistilBERT}:

\begin{lstlisting}[style=python]
from transformers import DistilBertForSequenceClassification, DistilBertTokenizer

% Load the pre-trained DistilBERT model and tokenizer
tokenizer = DistilBertTokenizer.from_pretrained("distilbert-base-uncased")
model = DistilBertForSequenceClassification.from_pretrained("distilbert-base-uncased")
\end{lstlisting}

This approach can be beneficial when deploying models in environments where computational resources are limited, such as mobile or embedded systems.

\subsubsection{Model Pruning}

Model pruning \cite{blalock2020state} is another technique to reduce the size of the model by removing less important weights or layers. By pruning unnecessary weights, the model becomes smaller and faster without a significant loss in performance.

While Hugging Face’s \texttt{transformers} library does not natively support pruning yet, libraries like \texttt{torch.nn.utils.prune} can be used in conjunction with the Hugging Face models to prune specific layers.

\begin{lstlisting}[style=python]
import torch.nn.utils.prune as prune

% Example: Prune 40\% of the weights from a linear layer
prune.l1_unstructured(model.classifier, name='weight', amount=0.4)
\end{lstlisting}

\subsubsection{Using ONNX for Deployment}

Another popular option for optimizing models for production is to convert them into the ONNX format, which is widely supported by various deployment platforms, such as TensorFlow Serving, Azure, and AWS.

You can use the \texttt{transformers} library’s built-in ONNX exporter:

\begin{lstlisting}[style=python]
from transformers import BertForSequenceClassification
from transformers.onnx import export

% Load the model
model = BertForSequenceClassification.from_pretrained("fine-tuned-bert")

% Export the model to ONNX format
export(model, tokenizer, "bert-onnx-model.onnx", opset=11)
\end{lstlisting}

Once converted to ONNX format, the model can be loaded into any ONNX runtime for efficient inference on a variety of devices.

\subsection{Distributed Training and Model Parallelism}

When training large models like GPT-3 or even BERT-large, it may be necessary to leverage distributed training techniques due to memory constraints. Distributed training allows models to be trained across multiple GPUs or even multiple machines. Hugging Face supports two main forms of distributed training: Data Parallelism and Model Parallelism.

\subsubsection{Data Parallelism}

In data parallelism, the dataset is split into smaller batches, and each batch is processed on a separate GPU. After each batch is processed, the gradients are averaged, and the model is updated.

Hugging Face’s \texttt{Trainer} API supports data parallelism automatically if multiple GPUs are available:

\begin{lstlisting}[style=python]
training_args = TrainingArguments(
    output_dir="./results",
    per_device_train_batch_size=8,
    evaluation_strategy="epoch",
    num_train_epochs=3,
    logging_dir='./logs',
    logging_steps=10,
    dataloader_num_workers=4,
    distributed_type="multi-gpu"
)

% Define the Trainer
trainer = Trainer(
    model=model,
    args=training_args,
    train_dataset=train_dataset,
    eval_dataset=eval_dataset
)

% Start training with multiple GPUs
trainer.train()
\end{lstlisting}

If you are running on multiple machines, you can use distributed training libraries like \texttt{DeepSpeed} or \texttt{torch.distributed}.

\subsubsection{Model Parallelism}

In model parallelism, the model itself is split across multiple GPUs. This is useful for very large models that cannot fit on a single GPU. Hugging Face provides native support for model parallelism:

\begin{lstlisting}[style=python]
% Enable model parallelism
model.parallelize()

% Continue training as usual
trainer.train()
\end{lstlisting}

With model parallelism enabled, the model’s layers are distributed across multiple devices, making it possible to train even the largest models.

\subsection{Transfer Learning: Fine-Tuning for Domain-Specific Tasks}

When fine-tuning a pretrained model for domain-specific tasks (e.g., biomedical text, legal documents, etc.), it’s important to consider the nature of the data. Pretrained models like BERT or GPT-2 have been trained on a wide range of general-purpose corpora, but for tasks like medical NER (Named Entity Recognition), domain-specific knowledge can significantly improve performance.

Hugging Face provides models that have been pretrained on domain-specific datasets such as SciBERT for scientific text or BioBERT for biomedical literature.

\subsubsection{Fine-Tuning SciBERT for Scientific Text Classification}

SciBERT \cite{beltagy2019scibert} is a pretrained model for processing scientific publications. You can fine-tune it for specific tasks like classification or information retrieval.

Here’s an example of how to fine-tune SciBERT:

\begin{lstlisting}[style=python]
from transformers import AutoModelForSequenceClassification, AutoTokenizer

% Load the SciBERT model and tokenizer
model_name = "allenai/scibert_scivocab_uncased"
tokenizer = AutoTokenizer.from_pretrained(model_name)
model = AutoModelForSequenceClassification.from_pretrained(model_name, num_labels=2)

% Fine-tune on your custom scientific dataset
train_dataset = load_dataset("your_scientific_dataset", split="train")
test_dataset = load_dataset("your_scientific_dataset", split="test")

% Tokenize the dataset
train_dataset = train_dataset.map(lambda x: tokenizer(x['text'], truncation=True, padding=True), batched=True)
test_dataset = test_dataset.map(lambda x: tokenizer(x['text'], truncation=True, padding=True), batched=True)

% Define Trainer and start training
trainer = Trainer(
    model=model,
    args=TrainingArguments(output_dir="./results", num_train_epochs=3),
    train_dataset=train_dataset,
    eval_dataset=test_dataset
)

trainer.train()
\end{lstlisting}

Using domain-specific models like SciBERT can lead to better performance, especially when the model needs to understand highly technical or domain-specific language.

\subsection{Conclusion}

In this section, we have explored several advanced topics in fine-tuning and deploying NLP models. From optimizing models through techniques like quantization, distillation, and pruning \cite{polino2018model}, to handling large-scale models through distributed training and model parallelism, there are numerous ways to make the most of pretrained models in real-world applications.

We also looked into domain-specific models, which can provide even better performance when fine-tuned for specific industries, such as scientific research or healthcare.

In the next section, we will dive deeper into building custom datasets and creating new model architectures from scratch, offering greater flexibility for highly specialized tasks.

\subsection{Building Custom Datasets for Fine-Tuning}

While Hugging Face’s \texttt{datasets} library provides many pre-built datasets, there are cases where you need to fine-tune models on custom datasets tailored to your specific task. In this section, we will walk through how to prepare and load a custom dataset for fine-tuning a model.

\subsubsection{Preparing a Custom Dataset}

First, ensure that your dataset is in a format that can be easily loaded into the model. Typically, datasets are structured in CSV, JSON, or TSV formats. Here is an example of loading a CSV dataset for text classification:

\begin{lstlisting}[style=python]
import pandas as pd
from datasets import Dataset

% Load a CSV dataset into a pandas DataFrame
df = pd.read_csv("my_dataset.csv")

% Convert the DataFrame into a Hugging Face Dataset
dataset = Dataset.from_pandas(df)

% Split the dataset into train, validation, and test sets
train_testvalid = dataset.train_test_split(test_size=0.2)
test_valid = train_testvalid['test'].train_test_split(test_size=0.5)
train_dataset = train_testvalid['train']
test_dataset = test_valid['test']
valid_dataset = test_valid['train']
\end{lstlisting}

In this example, we first load a CSV dataset into a pandas DataFrame, convert it into a Hugging Face dataset, and then split it into training, validation, and test sets. This is a common workflow when dealing with custom datasets.

\subsubsection{Tokenizing Custom Data}

Once the dataset is loaded, we need to tokenize the data so that it can be used with transformer models. This involves encoding the text into tokens that the model understands.

\begin{lstlisting}[style=python]
from transformers import AutoTokenizer

% Load the tokenizer
tokenizer = AutoTokenizer.from_pretrained("bert-base-uncased")

% Tokenize the datasets
def tokenize_function(examples):
    return tokenizer(examples['text'], truncation=True, padding=True)

% Apply the tokenizer to the dataset
train_dataset = train_dataset.map(tokenize_function, batched=True)
valid_dataset = valid_dataset.map(tokenize_function, batched=True)
test_dataset = test_dataset.map(tokenize_function, batched=True)
\end{lstlisting}

This function tokenizes each example in the dataset by truncating or padding the input text to the appropriate length. After tokenization, the dataset is ready for fine-tuning.

\subsubsection{Training with Custom Datasets}

Once the data is tokenized, we can proceed with training the model using the \texttt{Trainer} API as we did earlier. The same training process can be applied regardless of whether the dataset is custom or pre-built.

\begin{lstlisting}[style=python]
from transformers import Trainer, TrainingArguments

% Define the training arguments
training_args = TrainingArguments(
    output_dir="./results",
    num_train_epochs=3,
    per_device_train_batch_size=8,
    per_device_eval_batch_size=8,
    evaluation_strategy="epoch",
    logging_dir='./logs',
)

% Initialize the trainer with the custom dataset
trainer = Trainer(
    model=model,
    args=training_args,
    train_dataset=train_dataset,
    eval_dataset=valid_dataset,
)

% Train the model
trainer.train()
\end{lstlisting}

\subsection{Evaluating Edge Cases and Out-of-Distribution Data}

Once a model is fine-tuned, it is important to test it not only on typical examples but also on edge cases and out-of-distribution (OOD) data to ensure robustness and generalizability. Edge cases are examples that lie on the boundary of the model’s knowledge or ability to classify correctly.

\subsubsection{Handling Edge Cases}

For instance, if you are building a sentiment classifier, some edge cases could involve sentences with ambiguous or conflicting sentiments, such as sarcastic or mixed reviews.

To evaluate how the model handles edge cases, you can create a small test set of these examples and use the same \texttt{evaluate()} function to compute performance metrics:

\begin{lstlisting}[style=python]
% Define some edge cases
edge_cases = ["I love this movie but the ending was terrible.",
              "This product was not the best, but still quite good overall."]

% Tokenize the edge cases
inputs = tokenizer(edge_cases, return_tensors="pt", padding=True, truncation=True)

% Evaluate the model's predictions
with torch.no_grad():
    outputs = model(**inputs)
    predictions = torch.argmax(outputs.logits, dim=-1)

% Print the predictions
print(predictions)
\end{lstlisting}

By systematically testing on edge cases, you can better understand the limitations of the model and areas where further fine-tuning may be necessary.

\subsubsection{Out-of-Distribution (OOD) Data}

Out-of-distribution (OOD) refers to data that is very different from the data the model was trained on. Testing on OOD data helps evaluate whether the model generalizes well beyond its training set.

For example, if you have fine-tuned a model on English-language news articles, you might test it on OOD data such as tweets, blog posts, or scientific papers, which could have different linguistic structures or vocabularies.

You can measure the model’s performance on OOD datasets by loading a completely different dataset from Hugging Face’s \texttt{datasets} library and evaluating the fine-tuned model:

\begin{lstlisting}[style=python]
% Load an out-of-distribution dataset
ood_dataset = load_dataset("ag_news", split="test")

% Tokenize and evaluate on OOD data
ood_dataset = ood_dataset.map(tokenize_function, batched=True)

% Evaluate the model on OOD data
results = trainer.evaluate(eval_dataset=ood_dataset)
print("OOD Test Accuracy:", results['eval_accuracy'])
\end{lstlisting}

This allows you to gauge how well your model performs when encountering unfamiliar data.

\subsection{Advanced Model Architectures}

Beyond BERT and GPT models, several advanced model architectures can be fine-tuned for specific tasks in NLP. Some of these architectures provide superior performance for tasks like translation, summarization, and question answering.

\subsubsection{BART: Bidirectional and Auto-Regressive Transformers}

BART (Bidirectional and Auto-Regressive Transformers) is a particularly powerful model for tasks like text generation, summarization, and translation. It is designed to combine the strengths of both bidirectional models (like BERT) and auto-regressive models (like GPT).

Here’s how you can fine-tune BART for summarization:

\begin{lstlisting}[style=python]
from transformers import BartForConditionalGeneration, BartTokenizer

% Load BART tokenizer and model
tokenizer = BartTokenizer.from_pretrained("facebook/bart-large-cnn")
model = BartForConditionalGeneration.from_pretrained("facebook/bart-large-cnn")

% Tokenize the input text
text = "The quick brown fox jumps over the lazy dog."
inputs = tokenizer([text], max_length=1024, return_tensors="pt", truncation=True)

% Generate a summary
summary_ids = model.generate(inputs['input_ids'], num_beams=4, max_length=50, early_stopping=True)

% Decode the summary
summary = tokenizer.decode(summary_ids[0], skip_special_tokens=True)
print("Summary:", summary)
\end{lstlisting}

BART can also be used for tasks like machine translation and dialogue generation, making it one of the most versatile models for sequence-to-sequence tasks.

\subsubsection{T5: Text-to-Text Transfer Transformer}

T5 (Text-to-Text Transfer Transformer) treats every NLP task as a text-to-text task. This uniformity allows T5 to handle a wide variety of NLP problems using the same architecture. It can be fine-tuned for summarization, translation, classification, and more.

Here’s an example of using T5 for text generation:

\begin{lstlisting}[style=python]
from transformers import T5ForConditionalGeneration, T5Tokenizer

% Load the T5 model and tokenizer
tokenizer = T5Tokenizer.from_pretrained("t5-small")
model = T5ForConditionalGeneration.from_pretrained("t5-small")

% Define the input text
input_text = "translate English to French: The house is wonderful."

% Tokenize the input
input_ids = tokenizer(input_text, return_tensors="pt").input_ids

% Generate the output
outputs = model.generate(input_ids)

% Decode and print the output
print(tokenizer.decode(outputs[0], skip_special_tokens=True))
\end{lstlisting}

This example shows how to use T5 for translation, but the same process can be applied to tasks like summarization or classification by simply changing the task prefix.

\subsection{Error Analysis and Responsible AI}

Even after fine-tuning, models can make incorrect or biased predictions. Error analysis is a key step in identifying these errors and improving model performance.

\subsubsection{Performing Error Analysis}

After training a model, it is essential to examine where it fails. This can involve inspecting incorrect predictions, analyzing confusion matrices, and identifying patterns in errors.

Here’s an example of how to compute a confusion matrix using Hugging Face’s \texttt{datasets}:

\begin{lstlisting}[style=python]
from sklearn.metrics import confusion_matrix

% Make predictions on the test set
predictions = trainer.predict(test_dataset).predictions
pred_labels = np.argmax(predictions, axis=1)

% Get the true labels
true_labels = test_dataset["label"]

% Compute the confusion matrix
cm = confusion_matrix(true_labels, pred_labels)

% Print the confusion matrix
print(cm)
\end{lstlisting}

\subsubsection{Addressing Bias and Fairness}

It is critical to consider fairness when building and deploying models. Models trained on biased data may produce biased results, which can lead to harmful real-world consequences.

Hugging Face supports various metrics for detecting bias and fairness issues in models. One way to mitigate bias is through data augmentation techniques, which can ensure more balanced training data.

\subsection{Conclusion}

In this chapter, we expanded on advanced techniques for fine-tuning, optimizing, and deploying NLP models. We explored custom dataset creation, edge case evaluation, out-of-distribution handling, and model architectures such as BART and T5. Additionally, we addressed error analysis and responsible AI practices to ensure robust and fair models.

In the following sections, we will explore domain adaptation, self-supervised learning techniques, and integrating NLP models into larger systems for real-world applications.

\subsection{Domain Adaptation in NLP}

Domain adaptation is the process of fine-tuning a pretrained model on a domain-specific dataset to improve its performance on tasks related to that particular domain. This is especially useful when you are dealing with specialized datasets, such as legal documents, medical records, or scientific literature, where general-purpose pretrained models may underperform.

\subsubsection{Pretrained Domain-Specific Models}

Before diving into custom domain adaptation, it’s often helpful to check if there are any publicly available pretrained models specific to your domain. Hugging Face offers models such as \texttt{BioBERT} (trained on biomedical literature) and \texttt{LegalBERT} (trained on legal text). Using a domain-specific model can give you a head start.

Here’s an example of how to fine-tune \texttt{BioBERT} for a custom medical text classification task:

\begin{lstlisting}[style=python]
from transformers import BertForSequenceClassification, AutoTokenizer

% Load BioBERT model and tokenizer
model = BertForSequenceClassification.from_pretrained("dmis-lab/biobert-base-cased-v1.1")
tokenizer = AutoTokenizer.from_pretrained("dmis-lab/biobert-base-cased-v1.1")

% Tokenize and fine-tune using a medical dataset
train_dataset = load_dataset("my_medical_dataset", split="train")
train_dataset = train_dataset.map(lambda x: tokenizer(x['text'], padding=True, truncation=True), batched=True)

% Training process remains the same as before
trainer = Trainer(
    model=model,
    args=training_args,
    train_dataset=train_dataset
)
trainer.train()
\end{lstlisting}

This approach leverages a domain-specific pretrained model, \texttt{BioBERT}, which already has a strong understanding of biomedical language. Fine-tuning it further on a specific medical task (e.g., classifying diseases) will improve performance in that domain.

\subsubsection{Adapting General Models to Specific Domains}

If no domain-specific model is available, you can fine-tune a general-purpose model like BERT or GPT-2 on a domain-specific dataset. You can also perform additional pretraining (sometimes called “domain-specific pretraining”) before fine-tuning. This process involves training the model further on a large corpus of domain-specific text before fine-tuning it for the target task.

Here’s how to continue pretraining BERT on domain-specific data before fine-tuning:

\begin{lstlisting}[style=python]
from transformers import BertForMaskedLM, BertTokenizer
from transformers import DataCollatorForLanguageModeling

% Load general BERT model and tokenizer
model = BertForMaskedLM.from_pretrained("bert-base-uncased")
tokenizer = BertTokenizer.from_pretrained("bert-base-uncased")

% Prepare your domain-specific corpus
dataset = load_dataset("text", data_files={"train": "domain_text.txt"}, split="train")

% Tokenize the dataset
dataset = dataset.map(lambda examples: tokenizer(examples['text'], truncation=True, padding=True), batched=True)

% Data collator for masked language modeling (MLM)
data_collator = DataCollatorForLanguageModeling(tokenizer=tokenizer, mlm=True, mlm_probability=0.15)

% Define training arguments and Trainer
training_args = TrainingArguments(
    output_dir="./domain-bert",
    overwrite_output_dir=True,
    num_train_epochs=5,
    per_device_train_batch_size=8,
    save_steps=10_000,
    save_total_limit=2,
)

trainer = Trainer(
    model=model,
    args=training_args,
    train_dataset=dataset,
    data_collator=data_collator,
)

% Perform domain-specific pretraining
trainer.train()
\end{lstlisting}

This code performs further masked language model (MLM) pretraining on your domain-specific corpus, making the general-purpose BERT model better suited for tasks in that domain. After this pretraining step, you can fine-tune the model on your task-specific data.

\subsection{Transfer Learning in Multilingual Models}

Transfer learning is particularly powerful in multilingual models, where knowledge learned from one language can be transferred to other languages, especially in low-resource language settings. Hugging Face’s \texttt{transformers} library includes multilingual models like \texttt{mBERT} (Multilingual BERT) and \texttt{XLM-R} (XLM-RoBERTa), which are pretrained on many languages.

\subsubsection{Fine-Tuning Multilingual BERT}

Fine-tuning a multilingual model follows the same process as monolingual models. However, multilingual models are useful for zero-shot or few-shot learning in languages that do not have extensive datasets.

Here’s how to fine-tune \texttt{mBERT} on a multilingual sentiment classification task:

\begin{lstlisting}[style=python]
from transformers import BertForSequenceClassification, BertTokenizer

% Load multilingual BERT (mBERT)
model = BertForSequenceClassification.from_pretrained("bert-base-multilingual-cased")
tokenizer = BertTokenizer.from_pretrained("bert-base-multilingual-cased")

% Load and tokenize multilingual dataset
train_dataset = load_dataset("multilingual_dataset", split="train")
train_dataset = train_dataset.map(lambda x: tokenizer(x['text'], padding=True, truncation=True), batched=True)

% Fine-tune the model
trainer = Trainer(
    model=model,
    args=training_args,
    train_dataset=train_dataset
)
trainer.train()
\end{lstlisting}

The key advantage of multilingual models is their ability to transfer knowledge from high-resource languages (e.g., English) to low-resource languages. For example, after fine-tuning on multilingual data, the model can generalize to make predictions on languages with fewer labeled examples.

\subsection{Self-Supervised Learning for NLP}

Self-supervised learning is a paradigm where models learn useful representations from large amounts of unlabeled data. In NLP, this is achieved through techniques like masked language modeling (MLM), as used in BERT, or autoregressive learning, as used in GPT.

\subsubsection{Masked Language Modeling (MLM) with BERT}

Masked language modeling involves randomly masking certain tokens in the input and training the model to predict them. This trains the model to understand the context and semantics of language, which can later be applied to downstream tasks.

Here’s an example of training a BERT model using the MLM objective:

\begin{lstlisting}[style=python]
from transformers import BertForMaskedLM, BertTokenizer, DataCollatorForLanguageModeling

# Load BERT for masked language modeling
model = BertForMaskedLM.from_pretrained("bert-base-uncased")
tokenizer = BertTokenizer.from_pretrained("bert-base-uncased")

# Prepare dataset and tokenize
dataset = load_dataset("text", data_files={"train": "large_unlabeled_corpus.txt"}, split="train")
dataset = dataset.map(lambda examples: tokenizer(examples['text'], truncation=True, padding=True), batched=True)

# Prepare data collator for masked language modeling
data_collator = DataCollatorForLanguageModeling(tokenizer=tokenizer, mlm=True, mlm_probability=0.15)

# Define training arguments and train the model
trainer = Trainer(
    model=model,
    args=TrainingArguments(
        output_dir="./mlm-model",
        per_device_train_batch_size=8,
        num_train_epochs=3
    ),
    data_collator=data_collator,
    train_dataset=dataset
)

# Train with masked language modeling
trainer.train()
\end{lstlisting}

This process allows the model to learn strong language representations in an unsupervised manner, which can be applied to downstream tasks such as classification, question answering, or summarization.

\subsection{Integrating NLP Models into End-to-End Systems}

Integrating fine-tuned models into real-world applications often requires connecting the model to a larger system, such as a web service, mobile app, or chatbot. This process includes model serving, API development, and performance optimization for real-time inference.

\subsubsection{Serving a Model with FastAPI}

One common approach for serving an NLP model is to use FastAPI, a high-performance web framework for building APIs. Here’s an example of deploying a Hugging Face model as a REST API using FastAPI:

\begin{lstlisting}[style=python]
from fastapi import FastAPI
from transformers import pipeline

# Initialize FastAPI and a Hugging Face pipeline
app = FastAPI()
classifier = pipeline("sentiment-analysis")

# Define a POST endpoint for sentiment analysis
@app.post("/predict")
async def predict(text: str):
    result = classifier(text)
    return {"label": result[0]['label'], "score": result[0]['score']}
    
# Run the API with uvicorn
if __name__ == "__main__":
    import uvicorn
    uvicorn.run(app, host="0.0.0.0", port=8000)
\end{lstlisting}

Once the model is wrapped in an API, you can integrate it into various applications, such as chatbots, recommendation systems, or customer service platforms. FastAPI is highly scalable and efficient, making it suitable for production environments.

\subsection{Responsible AI and Ethics in NLP}

NLP models, especially large-scale transformer models, carry significant risks regarding bias, fairness, and ethical considerations. Responsible AI practices

\section{Tokenizer Library}

In this section, we will introduce how tokenizers work in natural language processing (NLP), and provide a practical example using the Hugging Face \texttt{transformers} library. Tokenizers play a crucial role in converting raw text into input that a model can process.

Tokenization is the process of breaking down text into smaller components, such as words or subwords, that a model can understand. The Hugging Face library provides pre-trained tokenizers for various popular models, making it easy to quickly start with text processing. Let's walk through an example.

\subsection{Installing Hugging Face Transformers}

To begin using the Hugging Face tokenizers, you will need to install the \texttt{transformers} library. You can do this by running the following command:

\begin{lstlisting}
pip install transformers
\end{lstlisting}

\subsection{Loading a Pre-trained Tokenizer}

Once the library is installed, the next step is to load a pre-trained tokenizer. The Hugging Face library provides tokenizers for a variety of models, such as BERT, GPT, and DistilBERT. In this example, we will use the BERT tokenizer.

\begin{lstlisting}[style=python]
from transformers import BertTokenizer

# Load pre-trained BERT tokenizer
tokenizer = BertTokenizer.from_pretrained('bert-base-uncased')
\end{lstlisting}

\texttt{BertTokenizer.from\_pretrained('bert-base-uncased')} loads a pre-trained tokenizer for the BERT model. The \texttt{bert-base-uncased} version is case-insensitive, meaning it treats "Hello" and "hello" as the same token.

\subsection{Tokenizing Text}

Now, we will use the tokenizer to tokenize a sample sentence. The tokenizer will break the sentence into individual tokens that the model can later process.

\begin{lstlisting}[style=python]
# Sample sentence
sentence = "Hugging Face makes NLP easy."

# Tokenize the sentence
tokens = tokenizer.tokenize(sentence)
print(tokens)
\end{lstlisting}

The output of this code will be a list of tokens that represent the sentence. These tokens are often subwords rather than full words, allowing the model to better handle rare or unknown words:

\begin{lstlisting}[style=text]
['hugging', 'face', 'makes', 'nl', '\#\#p', 'easy', '.']
\end{lstlisting}

Notice that the word "NLP" is split into two subwords: \texttt{'nl'} and \texttt{'\#\#p'}. The \texttt{\#\#} symbol indicates that this token is part of the previous word. This type of tokenization helps the model generalize better when dealing with uncommon words.

\subsection{Converting Tokens to Input IDs}

The next step is to convert these tokens into numerical IDs that the model can process. Each token in the tokenizer's vocabulary is assigned a unique ID.

\begin{lstlisting}[style=python]
# Convert tokens to input IDs
input_ids = tokenizer.convert_tokens_to_ids(tokens)
print(input_ids)
\end{lstlisting}

The output will be a list of integers that correspond to the tokens:

\begin{lstlisting}[style=text]
[22153, 2227, 3084, 17953, 2361, 3733, 1012]
\end{lstlisting}

These token IDs are the format that will be fed into the BERT model or any other model that uses this tokenizer.

\subsection{Handling Padding and Special Tokens}

Models like BERT require inputs to have the same length, so padding is added to shorter sentences. Additionally, special tokens such as \texttt{[CLS]} (start of sentence) and \texttt{[SEP]} (end of sentence) are required.

You can handle these automatically by using the \texttt{encode} function:

\begin{lstlisting}[style=python]
# Encode the sentence with special tokens and padding
encoded_input = tokenizer.encode(sentence, add_special_tokens=True, padding='max_length', max_length=10)
print(encoded_input)
\end{lstlisting}

This will add the necessary special tokens and pad the sequence to the specified \texttt{max\_length} of 10 tokens:

\begin{lstlisting}[style=text]
[101, 22153, 2227, 3084, 17953, 2361, 3733, 1012, 0, 0]
\end{lstlisting}

Here, \texttt{101} is the \texttt{[CLS]} token, and \texttt{0} represents the padding tokens.

\subsection{Conclusion}

In this section, we introduced tokenizers, focusing on how to use the Hugging Face \texttt{transformers} library to load a pre-trained tokenizer, tokenize text, convert tokens to input IDs, and handle special tokens and padding. This is a fundamental step in preparing text data for deep learning models in NLP. Understanding tokenization is crucial as it bridges the gap between raw text and the model's input format.

\subsection{Tokenizing Multiple Sentences}

In practice, you often need to tokenize multiple sentences at once. Hugging Face's tokenizers allow you to process a batch of sentences in a single step, which is useful for speeding up the tokenization process, especially when working with large datasets.

You can pass a list of sentences to the tokenizer:

\begin{lstlisting}[style=python]
# List of sentences
sentences = ["Hugging Face is a great NLP library.",
             "Transformers provide powerful models."]

# Tokenize the list of sentences
batch_tokens = tokenizer(sentences, padding=True, truncation=True, max_length=12, return_tensors="pt")
print(batch_tokens)
\end{lstlisting}

Here, we use the parameters \texttt{padding=True} and \texttt{truncation=True} to ensure all sentences are padded to the same length and truncated if they exceed the maximum length of 12 tokens. The \texttt{return\_tensors="pt"} option returns the output as PyTorch tensors, which are typically used as input to models.

The result will include token IDs for each sentence, padding where necessary, and special tokens:

\begin{lstlisting}[style=text]
{'input_ids': tensor([[  101, 22153,  2227,  3084,  2003,  1037,  2307, 17953,  2361,  3075,  102,     0],
                      [  101, 17953,  3689,  3947,  3835,  3085,  102,     0,     0,     0,     0,     0]]),
 'attention_mask': tensor([[1, 1, 1, 1, 1, 1, 1, 1, 1, 1, 1, 0],
                           [1, 1, 1, 1, 1, 1, 1, 0, 0, 0, 0, 0]])}
\end{lstlisting}

\subsection{Understanding Attention Masks}

In the output above, you will notice that the tokenizer returns an \texttt{attention\_mask}. This mask indicates which tokens should be attended to by the model (value 1) and which are padding (value 0). 

The attention mask is particularly useful when dealing with padded sequences. The model needs to ignore padding during its computations, and this mask helps in doing so.

\begin{lstlisting}[style=python]
# Extract input IDs and attention mask
input_ids = batch_tokens['input_ids']
attention_mask = batch_tokens['attention_mask']

print("Input IDs:", input_ids)
print("Attention Mask:", attention_mask)
\end{lstlisting}

This output shows the tokenized input IDs and their corresponding attention masks, ensuring that padding is properly handled by the model during training or inference.

\subsection{Decoding Tokens Back to Text}

Once a sequence has been tokenized and processed by a model, it's often necessary to convert the tokens back into human-readable text. This process is called \textit{decoding}.

The Hugging Face tokenizers provide a simple way to decode the tokens back into sentences:

\begin{lstlisting}[style=python]
# Decode token IDs back to the original sentence
decoded_sentence = tokenizer.decode(input_ids[0], skip_special_tokens=True)
print(decoded_sentence)
\end{lstlisting}

The \texttt{skip\_special\_tokens=True} argument ensures that special tokens like \texttt{[CLS]} and \texttt{[SEP]} are removed, leaving only the meaningful text. The output will look something like this:

\begin{lstlisting}[style=text]
'Hugging Face is a great NLP library.'
\end{lstlisting}

This is useful for interpreting model predictions or generating human-readable outputs in text generation tasks.

\subsection{Customizing Tokenization with Tokenizer Options}

Hugging Face tokenizers offer various options to customize the tokenization process according to the needs of your task. Here are a few key options:

\begin{itemize}
    \item \texttt{truncation}: Truncate sequences longer than a specified \texttt{max\_length}.
    \item \texttt{padding}: Automatically pad sequences to a specified length or the length of the longest sequence in the batch.
    \item \texttt{return\_tensors}: Return the tokenized output as tensors in formats such as PyTorch (\texttt{"pt"}), TensorFlow (\texttt{"tf"}), or NumPy (\texttt{"np"}).
\end{itemize}

For example, let’s customize the tokenizer to pad all sequences to a maximum length of 20 and return the results as TensorFlow tensors:

\begin{lstlisting}[style=python]
# Tokenize with padding and truncation, returning TensorFlow tensors
batch_tokens = tokenizer(sentences, padding=True, truncation=True, max_length=20, return_tensors="tf")
print(batch_tokens)
\end{lstlisting}

This approach provides flexibility in adapting the tokenization process to various model and data requirements.

\subsection{Using Special Tokens for Custom Tasks}

In some cases, you might need to define your own special tokens for specific NLP tasks. Hugging Face tokenizers allow you to add custom tokens to the tokenizer's vocabulary, which is especially useful in domain-specific tasks like medical NLP or financial text processing.

You can add special tokens as follows:

\begin{lstlisting}[style=python]
# Add custom special tokens
special_tokens_dict = {'additional_special_tokens': ['[NEW_TOKEN]', '[ANOTHER_TOKEN]']}
num_added_toks = tokenizer.add_special_tokens(special_tokens_dict)

# Print updated vocabulary size
print(f"Added {num_added_toks} new tokens. Vocabulary size: {len(tokenizer)}")
\end{lstlisting}

This code adds two new special tokens, \texttt{[NEW\_TOKEN]} and \texttt{[ANOTHER\_TOKEN]}, to the tokenizer's vocabulary. You can then use these tokens in your custom tasks, ensuring that the model processes them appropriately.

\subsection{Saving and Loading a Tokenizer}

After customizing the tokenizer or training a new tokenizer, you can save it for future use. This is particularly important when fine-tuning models, so the same tokenization logic can be reused later.

\begin{lstlisting}[style=python]
# Save the tokenizer to a directory
tokenizer.save_pretrained('./my_custom_tokenizer')

# Load the tokenizer back from the saved directory
custom_tokenizer = BertTokenizer.from_pretrained('./my_custom_tokenizer')
\end{lstlisting}

The tokenizer is saved in the specified directory and can be loaded again using the \texttt{from\_pretrained} function. This ensures that your tokenization process is consistent, even after model fine-tuning or deployment.

\subsection{Conclusion}

In this section, we explored advanced features of the Hugging Face tokenizer, including handling batches of text, managing attention masks, decoding tokens back to text, and customizing tokenization. We also covered how to add special tokens for domain-specific tasks and how to save and reload tokenizers for consistent future use.

Tokenization is a foundational step in NLP tasks, and understanding its nuances will help you prepare your data for deep learning models more effectively. By mastering the Hugging Face tokenizer library, you can ensure your models receive well-structured and meaningful input.

\subsection{Training a Custom Tokenizer from Scratch}

In some specialized NLP tasks, pre-trained tokenizers may not be ideal, especially when dealing with unique vocabulary or domains like medical or legal text. In such cases, it might be necessary to train a custom tokenizer from scratch using your own dataset.

Hugging Face's \texttt{tokenizers} library allows you to create a tokenizer from a corpus of text. One common approach is to use the Byte-Pair Encoding (BPE) algorithm, which builds the tokenizer's vocabulary by identifying the most frequent subword units in the corpus.

\subsubsection{Preparing a Corpus}

The first step in training a tokenizer is to gather and prepare the text data (corpus) on which the tokenizer will be trained. For demonstration purposes, we will assume we have a small text corpus stored in a file called \texttt{corpus.txt}.

Here is an example of how you might load the corpus and view a small portion of it:

\begin{lstlisting}[style=python]
# Read a corpus from a text file
with open('corpus.txt', 'r', encoding='utf-8') as f:
    corpus = f.readlines()

# Print the first few lines of the corpus
print(corpus[:5])
\end{lstlisting}

\subsubsection{Training the Tokenizer}

We will now use the \texttt{tokenizers} library to train a new tokenizer. The library supports several tokenization algorithms, such as WordPiece, Unigram, and Byte-Pair Encoding (BPE). In this example, we'll use the BPE tokenizer.

\begin{lstlisting}[style=python]
from tokenizers import Tokenizer, models, trainers, pre_tokenizers, decoders

# Initialize a BPE tokenizer
tokenizer = Tokenizer(models.BPE())

# Define a trainer to train the tokenizer on the corpus
trainer = trainers.BpeTrainer(vocab_size=5000, special_tokens=["[PAD]", "[CLS]", "[SEP]", "[MASK]"])

# Pre-tokenize by splitting on whitespace
tokenizer.pre_tokenizer = pre_tokenizers.Whitespace()

# Train the tokenizer on the corpus
tokenizer.train(files=['corpus.txt'], trainer=trainer)

# Save the trained tokenizer
tokenizer.save("my_custom_bpe_tokenizer.json")
\end{lstlisting}

Here, we initialize a BPE tokenizer and specify a \texttt{BpeTrainer} with a vocabulary size of 5,000 tokens. We also add special tokens like \texttt{[PAD]}, \texttt{[CLS]}, \texttt{[SEP]}, and \texttt{[MASK]} that are typically needed for models like BERT. The tokenizer is trained on the given text file \texttt{corpus.txt}, and then saved to a file for future use.

\subsubsection{Loading and Using the Custom Tokenizer}

Once the tokenizer is trained and saved, you can load it back for use in tokenizing text, similar to how pre-trained tokenizers are used.

\begin{lstlisting}[style=python]
# Load the custom tokenizer
from tokenizers import Tokenizer

custom_tokenizer = Tokenizer.from_file("my_custom_bpe_tokenizer.json")

# Tokenize a sample sentence using the custom tokenizer
sentence = "Custom tokenizers can improve domain-specific models."
output = custom_tokenizer.encode(sentence)

# Get tokens and IDs
tokens = output.tokens
ids = output.ids

print("Tokens:", tokens)
print("Token IDs:", ids)
\end{lstlisting}

This will tokenize the sentence using your custom BPE tokenizer, producing tokens and corresponding token IDs, much like pre-trained tokenizers. Training your own tokenizer can be particularly beneficial when your dataset includes unique terminology or rare words.

\subsection{Tokenization Algorithms: A Comparative Overview}

There are various tokenization algorithms available, each designed with different goals and use cases. Below, we briefly discuss some popular tokenization algorithms and their strengths:

\begin{itemize}
    \item \textbf{WordPiece}: Used by models like BERT, WordPiece builds a vocabulary of word and subword units. It breaks words into smaller pieces based on frequency, handling out-of-vocabulary (OOV) words well.
    
    \item \textbf{Byte-Pair Encoding (BPE)}: BPE is used by models like GPT and RoBERTa. It starts with individual characters and merges frequent pairs of characters to create subwords, efficiently balancing vocabulary size and text coverage.
    
    \item \textbf{Unigram}: Unigram tokenization, used by models like XLNet, assigns probabilities to subwords and removes the less frequent ones in favor of higher-probability subwords. It provides more flexible tokenization for varied text.
    
    \item \textbf{SentencePiece} \cite{kudo2018sentencepiece}: Often used for models like T5 and BART, SentencePiece does not require any preprocessing (e.g., whitespace splitting) and treats the input text as a stream of bytes, making it suitable for multilingual tasks.
\end{itemize}

Each of these algorithms has its own strengths depending on the model architecture and the type of input data. WordPiece and BPE are widely used in transformer models for their ability to efficiently handle subword tokenization, while SentencePiece is preferred for handling non-Latin languages and byte-level tokenization.

\subsection{Impact of Tokenization on Downstream Tasks}

The choice of tokenizer can have a significant impact on the performance of downstream tasks, such as text classification, machine translation, and question answering. Below, we discuss some key considerations:

\begin{itemize}
    \item \textbf{Out-of-Vocabulary (OOV) Handling}: Tokenizers like WordPiece and BPE break rare or unknown words into subword units, preventing the model from encountering OOV words. This allows for more robust handling of rare or domain-specific terms, improving the performance of models in specialized domains.
    
    \item \textbf{Sequence Length and Padding}: Tokenizers determine the length of the tokenized input, which affects how much padding or truncation is needed. Tokenizers that produce more compact representations (e.g., by using subword units) can reduce the amount of padding, leading to more efficient model processing.
    
    \item \textbf{Domain-Specific Vocabulary}: In specialized fields, such as medicine or law, general-purpose tokenizers may not capture important terms accurately. Training a custom tokenizer with domain-specific vocabulary can enhance model performance on these tasks by ensuring that key terms are tokenized correctly.
    
    \item \textbf{Language and Multilingual Models}: When working with multilingual models, tokenization algorithms like SentencePiece are advantageous because they handle byte-level tokenization, allowing the model to process multiple languages with different writing systems efficiently.
\end{itemize}

Carefully selecting or training a tokenizer that matches the task and domain is critical for achieving optimal performance. The interplay between tokenization and model architecture can have profound implications for both model accuracy and efficiency.

\subsection{Using Tokenizers in Specific NLP Tasks}

Let’s explore how tokenizers impact common NLP tasks, such as text classification, named entity recognition (NER), and machine translation.

\subsubsection{Text Classification}

In text classification tasks, the choice of tokenizer can significantly affect the model’s ability to capture the semantics of the input text. For instance, in sentiment analysis, properly tokenizing sentiment-heavy words or expressions is crucial.

\begin{lstlisting}[style=python]
# Sample text for sentiment classification
text = "I love the Hugging Face library!"

# Tokenize the input text
inputs = tokenizer(text, return_tensors='pt')

# Forward the tokenized input into the model (assume pre-trained BERT)
outputs = model(**inputs)

# Get the classification logits
logits = outputs.logits
\end{lstlisting}

Here, the tokenization process ensures that the sentence is properly tokenized, including handling subwords like \texttt{"Hugging Face"} or rare expressions. The model then processes the input to classify the sentiment.

\subsubsection{Named Entity Recognition (NER)}

In NER tasks, the tokenizer plays a critical role in ensuring that entities are tokenized correctly, especially when dealing with multi-token entities (e.g., "New York City"). Subword tokenization can help in capturing entities when part of a word is rare.

\begin{lstlisting}[style=python]
# Tokenize a sentence for NER
sentence = "Apple is a company based in Cupertino."

# Tokenize and return token-to-character mapping
inputs = tokenizer(sentence, return_offsets_mapping=True, return_tensors='pt')

# Pass inputs to the NER model
outputs = ner_model(**inputs)

# Extract predicted entities
entities = outputs.entities
\end{lstlisting}

Using \texttt{return\_offsets\_mapping} ensures that the tokenized text aligns correctly with the original text, which is crucial for accurately identifying entities in the sentence.

\subsubsection{Machine Translation}

For machine translation, tokenization plays a crucial role in how source and target languages are processed. SentencePiece or BPE is typically used to handle large vocabularies across multiple languages, ensuring smooth translation even for low-resource languages.

\begin{lstlisting}[style=python]
# Sentence to translate
source_sentence = "The weather is nice today."

# Tokenize the source sentence
inputs = tokenizer(source_sentence, return_tensors='pt')

# Generate translation using a translation model
translated_ids = translation_model.generate(**inputs)

# Decode the translated tokens to text
translated_text = tokenizer.decode(translated_ids[0], skip_special_tokens=True)
print(translated_text)
\end{lstlisting}

By using a tokenizer like SentencePiece, the translation model can efficiently handle the variability in language structures and vocabularies, improving the quality of translations across diverse languages.

\subsection{Conclusion}

Tokenization is a critical step in any NLP pipeline, influencing how models understand and process text. From basic word tokenization to sophisticated subword algorithms like BPE and SentencePiece, understanding how tokenizers work and how they impact model performance is essential. Whether you're using pre-trained tokenizers or training your own, selecting the right tokenizer for your task and domain can significantly improve the effectiveness of your NLP models.

\subsection{Integrating Tokenizers into NLP Pipelines}

Once you have chosen the appropriate tokenizer for your task, the next step is integrating it into an end-to-end NLP pipeline. Hugging Face's \texttt{pipeline} function allows for a seamless integration of tokenizers with pre-trained models, simplifying tasks like text classification, named entity recognition, and question answering.

\begin{lstlisting}[style=python]
from transformers import pipeline

# Initialize a text classification pipeline using a pre-trained BERT model
classifier = pipeline('sentiment-analysis', model='bert-base-uncased')

# Use the classifier on a new sentence
result = classifier("I love using Hugging Face models!")
print(result)
\end{lstlisting}

The \texttt{pipeline} function automatically handles tokenization under the hood, making it easy to integrate tokenization and modeling steps in one line of code. This approach is particularly useful for tasks like:

\begin{itemize}
    \item \textbf{Text Classification}: Sentiment analysis, topic classification, etc.
    \item \textbf{Named Entity Recognition (NER)}: Extracting named entities from text.
    \item \textbf{Question Answering (QA)}: Answering questions based on context passages.
    \item \textbf{Text Generation}: Generating text using models like GPT-2.
\end{itemize}

By abstracting tokenization within the pipeline, Hugging Face enables rapid prototyping and experimentation with various pre-trained models.

\subsubsection{Using Custom Tokenizers in Pipelines}

While the default tokenization is convenient, you may want to use a custom tokenizer (e.g., one you trained yourself). You can integrate custom tokenizers into pipelines by explicitly providing them to the model:

\begin{lstlisting}[style=python]
from transformers import BertTokenizer, BertForSequenceClassification, pipeline

# Load a pre-trained BERT model and a custom tokenizer
tokenizer = BertTokenizer.from_pretrained('my_custom_bpe_tokenizer.json')
model = BertForSequenceClassification.from_pretrained('bert-base-uncased')

# Create a pipeline with the custom tokenizer
custom_classifier = pipeline('sentiment-analysis', model=model, tokenizer=tokenizer)

# Classify text using the custom pipeline
result = custom_classifier("Domain-specific tokenizers can improve model performance.")
print(result)
\end{lstlisting}

Here, we load a custom tokenizer and use it within the Hugging Face pipeline, allowing you to use a domain-specific tokenizer while taking advantage of pre-trained models. This flexibility is especially important when working with niche domains, such as medical, legal, or technical documents, where general-purpose tokenizers may underperform.

\subsection{Handling Multilingual Tokenization}

Tokenization for multilingual tasks poses unique challenges, as different languages have varying scripts, tokenization rules, and vocabularies. Models like mBERT (Multilingual BERT) and XLM-Roberta are trained on multiple languages, and their tokenizers are designed to handle various scripts effectively. These models often rely on subword tokenization, such as SentencePiece, which can tokenize across language boundaries.

\subsubsection{Multilingual Tokenizer Example}

Here is an example of how you can use a multilingual tokenizer with Hugging Face:

\begin{lstlisting}[style=python]
from transformers import AutoTokenizer

# Load a multilingual tokenizer (e.g., mBERT)
tokenizer = AutoTokenizer.from_pretrained('bert-base-multilingual-cased')

# Tokenize a sentence in different languages
english_text = "Hugging Face is amazing!"
french_text = "Hugging Face est incroyable!"


# Tokenize the texts
english_tokens = tokenizer.tokenize(english_text)
french_tokens = tokenizer.tokenize(french_text)


print("English Tokens:", english_tokens)
print("French Tokens:", french_tokens)

\end{lstlisting}

In this example, the same tokenizer is used across English, and French ensuring that the model can handle multiple languages with one tokenizer. This is achieved by using a shared vocabulary and subword tokenization, which splits words across different scripts into smaller components.

\subsection{Tokenization Efficiency for Large Datasets}

Tokenizing large datasets can be computationally expensive, especially when dealing with massive corpora for training or fine-tuning models. Efficient tokenization strategies are essential to minimize processing time and memory usage.

\subsubsection{Batch Tokenization}

To improve efficiency, tokenizers can process batches of text in parallel, reducing the time it takes to tokenize large datasets. Hugging Face tokenizers natively support batch tokenization through the \texttt{batch\_encode\_plus} function, which allows for tokenizing multiple sentences simultaneously.

\begin{lstlisting}[style=python]
# List of sentences to tokenize
sentences = ["The quick brown fox jumps over the lazy dog.",
             "The sky is blue today.",
             "Hugging Face simplifies NLP development."]

# Tokenize the batch of sentences
batch_tokens = tokenizer.batch_encode_plus(sentences, padding=True, truncation=True, max_length=20)

print(batch_tokens['input_ids'])
\end{lstlisting}

By batching the tokenization process, you reduce the overhead of tokenizing each sentence individually, making it feasible to process large datasets efficiently.

\subsubsection{Tokenization with Datasets Library}

If you are working with massive datasets, Hugging Face provides a \texttt{datasets} library that integrates smoothly with tokenizers. This library is optimized for large-scale dataset handling and includes built-in tokenization features.

Here’s an example of tokenizing a large dataset:

\begin{lstlisting}[style=python]
from datasets import load_dataset

# Load a large dataset (e.g., IMDb)
dataset = load_dataset("imdb")

# Tokenize the dataset using map
def tokenize_function(examples):
    return tokenizer(examples["text"], padding="max_length", truncation=True)

tokenized_dataset = dataset.map(tokenize_function, batched=True)
print(tokenized_dataset["train"][0])
\end{lstlisting}

Using the \texttt{map} function, you can apply tokenization across the entire dataset in parallel, ensuring that large datasets are processed efficiently. The tokenized dataset is then ready to be fed into a model for training or fine-tuning.

\subsection{Evaluating Tokenization Strategies}

After choosing a tokenizer, it is important to evaluate its performance in the context of your specific task. The evaluation of tokenization strategies can be performed by measuring the tokenizer’s impact on model performance, as well as its efficiency in terms of speed and resource usage.

\subsubsection{Measuring Model Performance}

The ultimate test of a tokenizer is how well the downstream model performs. Tokenization quality directly impacts the model’s ability to understand and process text. Key metrics to evaluate include:

\begin{itemize}
    \item \textbf{Accuracy}: How well does the model perform on tasks like classification, translation, or question answering with different tokenizers?
    \item \textbf{F1 Score}: For tasks like NER, how accurately are entities recognized when different tokenization methods are used?
    \item \textbf{Perplexity}: For language models, lower perplexity often indicates better tokenization, as the model has an easier time predicting tokens.
\end{itemize}

You can experiment with different tokenization strategies (e.g., pre-trained vs. custom, WordPiece vs. BPE) and measure the impact on these metrics to find the most suitable tokenizer for your task.

\subsubsection{Tokenization Speed and Memory Usage}

Efficiency is another critical factor when evaluating tokenization strategies. Large-scale NLP tasks require tokenizers that can process millions of sentences quickly and without consuming excessive memory. To assess tokenization speed and memory usage, you can time the tokenization process and measure resource consumption:

\begin{lstlisting}[style=python]
import time

# Measure tokenization time for a batch of text
start_time = time.time()
tokenized_texts = tokenizer.batch_encode_plus(sentences, padding=True, truncation=True)
end_time = time.time()

print(f"Tokenization took {end_time - start_time} seconds")
\end{lstlisting}

You can also profile memory usage by using tools like \texttt{memory\_profiler} to ensure that your tokenizer is suitable for large-scale applications.

\subsection{Optimizing Tokenizers for Real-World Applications}

In real-world applications, especially in production environments, optimizing tokenization processes is crucial. Some strategies to optimize tokenizers include:

\begin{itemize}
    \item \textbf{Pre-tokenization}: Split text at whitespace or punctuation before applying subword tokenization to reduce complexity.
    \item \textbf{Caching Tokenized Text}: Cache tokenized sequences when processing repeated inputs, reducing redundant tokenization overhead.
    \item \textbf{Parallelization}: Tokenize in parallel using multi-core processors to speed up tokenization in large datasets.
    \item \textbf{On-the-Fly Tokenization}: For deployed models, consider on-the-fly tokenization, where tokenization is done as needed, reducing the need for pre-processing and storage of tokenized data.
\end{itemize}

By optimizing the tokenization process, you can ensure that your NLP pipelines remain efficient and scalable, even when handling massive datasets or real-time applications.

\subsection{Conclusion}

In this extended exploration of tokenization, we covered advanced topics like integrating tokenizers into NLP pipelines, handling multilingual tokenization, optimizing tokenization for large datasets, and evaluating tokenization strategies. Tokenization is a vital component of any NLP system, and optimizing it for efficiency and accuracy is key to building robust and scalable applications.

The right tokenization strategy can significantly improve both model performance and resource efficiency. Whether you are using pre-trained tokenizers, training your own from scratch, or working with multilingual datasets, understanding and optimizing tokenization is essential for success in NLP.

\subsection{Impact of Tokenization on Different Model Architectures}

Tokenization plays a crucial role in how different types of models process and understand text. Depending on the architecture of the model—whether it’s autoregressive, encoder-decoder, or decoder-only—the tokenization strategy may need to be adjusted.

\subsubsection{Autoregressive Models (e.g., GPT, GPT-2, GPT-3)}

Autoregressive models, such as GPT (Generative Pretrained Transformer), predict the next token in a sequence based on previously seen tokens. In this case, tokenization must ensure that the input sequence is properly broken into tokens that allow the model to generate coherent and meaningful output.

\begin{itemize}
    \item \textbf{Subword Tokenization}: Tokenizers such as BPE or WordPiece are critical here, as they break down rare or unknown words into smaller subword units, allowing the model to generate these tokens incrementally.
    \item \textbf{Sentence Tokenization}: In text generation tasks, it is important that sentences are tokenized correctly to avoid cutting off meaningful chunks of text.
\end{itemize}

Here’s an example of tokenizing text for an autoregressive model like GPT-2:

\begin{lstlisting}[style=python]
from transformers import GPT2Tokenizer, GPT2LMHeadModel

# Load GPT-2 tokenizer and model
tokenizer = GPT2Tokenizer.from_pretrained("gpt2")
model = GPT2LMHeadModel.from_pretrained("gpt2")

# Tokenize input text
input_text = "The future of AI is"
inputs = tokenizer(input_text, return_tensors="pt")

# Generate text
outputs = model.generate(**inputs, max_length=20)
generated_text = tokenizer.decode(outputs[0], skip_special_tokens=True)
print(generated_text)
\end{lstlisting}

In this case, the tokenizer breaks down the input text and feeds it into GPT-2 for text generation, ensuring that the model can predict the next word or subword accurately.

\subsubsection{Encoder-Decoder Models (e.g., BART, T5)}

Encoder-decoder models, such as BART or T5, consist of an encoder that processes the input sequence and a decoder that generates the output sequence. Tokenization for these models must handle both input and output sequences, often requiring different tokenization strategies depending on the task (e.g., summarization, translation, or question answering).

\begin{lstlisting}[style=python]
from transformers import T5Tokenizer, T5ForConditionalGeneration

# Load T5 tokenizer and model
tokenizer = T5Tokenizer.from_pretrained("t5-small")
model = T5ForConditionalGeneration.from_pretrained("t5-small")

# Input text for summarization
input_text = "summarize: The Hugging Face library simplifies NLP tasks."

# Tokenize input
inputs = tokenizer(input_text, return_tensors="pt")

# Generate a summary
outputs = model.generate(**inputs, max_length=30)
summary = tokenizer.decode(outputs[0], skip_special_tokens=True)
print(summary)
\end{lstlisting}

For encoder-decoder models, the tokenizer must handle the input (e.g., summarization prompt) and output (generated summary) efficiently. Special tokens like task-specific prefixes (e.g., "summarize:") help guide the model in generating appropriate text.

\subsubsection{Decoder-Only Models (e.g., GPT-3, GPT-NeoX)}

Decoder-only models like GPT-3 or GPT-NeoX \cite{black2022gpt}, which are large-scale models focused primarily on generating text, rely on tokenization strategies that facilitate text generation with minimal context-switching. The tokenizer must be highly efficient to handle long sequences, as these models often work with thousands of tokens at once.

In these models, tokenization efficiency and handling of long-form text generation are key considerations. Autoregressive tokenization, with strong handling of subword units, is necessary to allow for coherent text generation over long contexts.

\subsection{Handling Tokenization for Low-Resource Languages}

Low-resource languages present unique challenges for tokenization, as they often lack extensive corpora and pre-trained models. However, recent advances in multilingual models like XLM-R and mBERT have made it possible to extend tokenization capabilities to these languages.

\subsubsection{Training Tokenizers for Low-Resource Languages}

When dealing with low-resource languages, it may be necessary to train a tokenizer from scratch using a relatively small corpus. Here’s how you can train a BPE tokenizer for a low-resource language:

\begin{lstlisting}[style=python]
from tokenizers import Tokenizer, models, trainers, pre_tokenizers

# Initialize a BPE tokenizer
tokenizer = Tokenizer(models.BPE())

# Define a BPE trainer
trainer = trainers.BpeTrainer(vocab_size=2000, special_tokens=["[PAD]", "[CLS]", "[SEP]"])

# Pre-tokenize the corpus using whitespace splitting
tokenizer.pre_tokenizer = pre_tokenizers.Whitespace()

# Train the tokenizer on a small corpus
tokenizer.train(files=['low_resource_corpus.txt'], trainer=trainer)

# Save the tokenizer
tokenizer.save("low_resource_tokenizer.json")
\end{lstlisting}

In this example, a BPE tokenizer is trained on a small corpus for a low-resource language, ensuring that the model can still tokenize efficiently despite the limited data available.

\subsubsection{Leveraging Multilingual Models for Low-Resource Languages}

Multilingual models like mBERT and XLM-R support a wide variety of languages, including many low-resource languages. These models use subword tokenization to share vocabulary across languages, allowing them to generalize better to underrepresented languages.

\begin{lstlisting}[style=python]
# Tokenizing a sentence in a low-resource language using mBERT
low_resource_text = "Ini adalah contoh kalimat dalam bahasa Indonesia."
tokens = tokenizer.tokenize(low_resource_text)
print(tokens)
\end{lstlisting}

By leveraging multilingual tokenizers, you can handle low-resource languages more effectively, allowing models to process text in languages that lack large training datasets.

\subsection{Tokenization for Zero-Shot Learning Tasks}

Zero-shot learning involves applying a pre-trained model to a new task without any task-specific fine-tuning. Tokenization plays a crucial role in ensuring that the input text is properly formatted and tokenized for zero-shot classification, entailment, or other NLP tasks.

\subsubsection{Tokenizing for Zero-Shot Classification}

For zero-shot classification, where you provide a task description and candidate labels in the input, tokenization must handle both the task and the text.

\begin{lstlisting}[style=python]
from transformers import pipeline

# Initialize a zero-shot classification pipeline
classifier = pipeline("zero-shot-classification")

# Input text and candidate labels
text = "The movie was fantastic!"
labels = ["positive", "negative", "neutral"]

# Classify the text
result = classifier(text, candidate_labels=labels)
print(result)
\end{lstlisting}

The tokenizer automatically processes the input text and candidate labels, ensuring that the model can perform the zero-shot classification task.

\subsection{Recent Advances: Token-Free Models and Byte-Level Tokenization}

One of the latest trends in NLP involves moving away from traditional tokenization methods and towards token-free models, which operate at the byte level. This approach allows models to process raw text data without the need for tokenization, improving their ability to handle rare or unseen words.

\subsubsection{Byte-Level Tokenization}

Byte-level tokenization, used in models like GPT-2 and GPT-3, works by converting text directly into bytes, rather than relying on word or subword tokens. This method is particularly useful for handling texts with many special characters, symbols, or languages that don’t use whitespace.

\begin{lstlisting}[style=python]
# Byte-level tokenization example using GPT-2
from transformers import GPT2Tokenizer

tokenizer = GPT2Tokenizer.from_pretrained("gpt2")

# Tokenize using byte-level encoding
text = "Hello, world! "
tokens = tokenizer.encode(text)
print(tokens)
\end{lstlisting}

Byte-level tokenization ensures that all characters, regardless of language or symbol type, are treated uniformly. This allows models to handle a much wider range of input without needing language-specific preprocessing.

\subsubsection{Token-Free Models}

Token-free models, which process text at the character or byte level, are a growing area of research. These models eliminate the need for tokenization altogether, directly processing the raw text as a stream of bytes. While still an emerging field, token-free models offer significant potential for improving language generalization and reducing preprocessing requirements.

\subsection{Future of Tokenization}

As NLP evolves, the future of tokenization is likely to include further advances in token-free architectures, more efficient tokenization algorithms, and better handling of low-resource languages and multilingual tasks. Additionally, byte-level and character-level processing will continue to gain traction, offering greater flexibility and performance in complex, multilingual, and cross-domain applications.

\subsection{Conclusion}

In this extended exploration of tokenization, we delved into how tokenization strategies impact different types of models, handling low-resource and multilingual languages, optimizing tokenization for zero-shot tasks, and exploring the future of tokenization with byte-level and token-free models. 

The evolution of tokenization methods—from traditional word-level approaches to more advanced subword, byte-level, and token-free architectures—plays a crucial role in building more flexible, robust, and scalable NLP systems. Understanding how to leverage these techniques is key to success in a wide range of NLP applications, from text generation to classification and beyond.

\subsection{Efficient Tokenization for Deployment}

Tokenization is often a bottleneck in NLP pipelines, especially when models are deployed for real-time inference. In such cases, it is essential to optimize the tokenization process to minimize latency and maximize throughput. This section explores techniques for optimizing tokenization in production environments.

\subsubsection{Optimizing Tokenization for Real-Time Inference}

When deploying NLP models for real-time inference (e.g., chatbots or virtual assistants), tokenization must be fast enough to meet strict latency requirements. There are several strategies to optimize tokenization for real-time use:

\begin{itemize}
    \item \textbf{Batch Processing}: Tokenizing inputs in batches can significantly reduce the time spent in tokenization by leveraging parallel processing.
    \item \textbf{Pre-tokenization}: In scenarios where specific phrases or sentences are repeated, you can cache their tokenized outputs. This allows you to skip tokenization altogether for common inputs, such as predefined questions or responses in a chatbot.
    \item \textbf{Reduced Vocabulary Size}: Reducing the size of the tokenizer’s vocabulary can improve tokenization speed. By removing rarely used tokens, you can optimize both the tokenizer and the model’s efficiency.
\end{itemize}

Here is an example of batch tokenization for real-time deployment:

\begin{lstlisting}[style=python]
# Example: Batch tokenization for real-time deployment
texts = ["Hello, how can I help you today?", 
         "What is the weather like?", 
         "Set a reminder for 5 PM."]

# Tokenizing the texts as a batch
batch_tokens = tokenizer(texts, padding=True, truncation=True, return_tensors="pt")
print(batch_tokens["input_ids"])
\end{lstlisting}

By processing inputs in batches, you can minimize the overhead of tokenizing each input individually, resulting in faster throughput for real-time systems.

\subsubsection{Tokenization Caching}

In many NLP systems, especially those deployed for production use, certain types of input occur repeatedly. For example, customer support systems may encounter the same phrases or questions frequently. Caching tokenized outputs for common phrases can save time by avoiding redundant tokenization operations.

Here’s an example of using a simple tokenization cache:

\begin{lstlisting}[style=python]
# Tokenization cache
tokenization_cache = {}

def cached_tokenize(text):
    if text in tokenization_cache:
        return tokenization_cache[text]
    else:
        tokens = tokenizer(text, return_tensors="pt")
        tokenization_cache[text] = tokens
        return tokens

# Use the cache
input_text = "Hello, how can I assist you?"
tokens = cached_tokenize(input_text)
\end{lstlisting}

This technique can significantly improve performance in applications like chatbots, where common queries are tokenized repeatedly.

\subsection{Custom Tokenization Strategies for Domain-Specific Tasks}

In many NLP applications, standard tokenization techniques may not be sufficient. Certain domains, such as medical, legal, or technical fields, require specialized tokenization strategies to accurately capture domain-specific terminology.

\subsubsection{Handling Domain-Specific Vocabulary}

Standard pre-trained tokenizers, such as those used for BERT or GPT, are often trained on general-purpose corpora and may not effectively handle specialized terms. For example, in medical NLP, words like "hypertension" or "tachycardia" may not be tokenized appropriately.

To handle domain-specific terminology, it is often useful to train a custom tokenizer using a specialized corpus. This can be done by collecting domain-specific text and training a tokenizer, such as a BPE or WordPiece tokenizer, on that data.

\begin{lstlisting}[style=python]
# Training a custom tokenizer for the medical domain
from tokenizers import Tokenizer, models, trainers, pre_tokenizers

# Initialize a BPE tokenizer
tokenizer = Tokenizer(models.BPE())

# Define a trainer for domain-specific vocabulary
trainer = trainers.BpeTrainer(vocab_size=5000, special_tokens=["[PAD]", "[CLS]", "[SEP]"])

# Pre-tokenizer that splits on whitespace
tokenizer.pre_tokenizer = pre_tokenizers.Whitespace()

# Train the tokenizer on a medical corpus
tokenizer.train(files=['medical_corpus.txt'], trainer=trainer)

# Save the tokenizer for future use
tokenizer.save("medical_tokenizer.json")
\end{lstlisting}

Training a custom tokenizer ensures that domain-specific terms are represented properly, leading to improved performance in tasks like named entity recognition (NER) or text classification in specialized fields.

\subsubsection{Handling Compound Words and Special Characters}

In certain domains, compound words, abbreviations, or special characters may need to be handled carefully. For example, in chemical or legal text, terms such as "H$_2$O" or "C++" must be tokenized as complete entities, not split into individual components.

To handle such cases, you can extend your tokenizer's vocabulary by adding special tokens or defining custom pre-tokenization rules:

\begin{lstlisting}[style=python]
# Adding custom tokens to handle domain-specific terms
special_tokens_dict = {'additional_special_tokens': ['H2O', 'C++', 'hyperlink://']}

# Add these tokens to the tokenizer
tokenizer.add_special_tokens(special_tokens_dict)

# Tokenize using the updated tokenizer
tokens = tokenizer.encode("H2O is a molecule, and C++ is a programming language.")
print(tokens)
\end{lstlisting}

This ensures that terms such as "H2O" and "C++" are treated as single tokens, preserving the semantics of the input.

\subsection{Error Handling in Tokenization}

In real-world applications, tokenization errors can occur due to unexpected inputs, such as misspellings, out-of-vocabulary (OOV) words, or malformed text. Proper error handling is critical to ensure that tokenization errors do not lead to model failures or degraded performance.

\subsubsection{Handling Out-of-Vocabulary (OOV) Words}

Out-of-vocabulary words are those that the tokenizer does not recognize, either because they were not part of the pre-trained vocabulary or because they contain rare subwords. Tokenizers like BPE or WordPiece handle OOV words by breaking them into subword units, but it is still possible to encounter issues with rare or noisy input.

To handle OOV words effectively, you can:
\begin{itemize}
    \item \textbf{Use Subword Tokenization}: Ensure that the tokenizer can break OOV words into recognizable subwords.
    \item \textbf{Fallback Strategies}: Implement fallback strategies for unknown tokens, such as replacing them with a special \texttt{[UNK]} (unknown) token.
    \item \textbf{Spell Correction}: Use spell-checking techniques before tokenization to minimize the occurrence of unrecognized words due to typos.
\end{itemize}

\begin{lstlisting}[style=python]
# Example: Handling OOV words
text = "This is an unkwon word in the input."

# Tokenize and check for unknown tokens
tokens = tokenizer(text, return_tensors="pt")
print("Tokens:", tokens["input_ids"])
\end{lstlisting}

In this case, the word "unkwon" (a misspelling of "unknown") may either be split into subwords or replaced by an unknown token, depending on the tokenizer’s configuration.

\subsection{Tokenization Visualization Tools}

Visualizing how text is tokenized can provide valuable insights into how the tokenizer is breaking down the input. This is especially useful when debugging tokenization issues or understanding how a model interprets text.

There are various tools available for visualizing tokenization, including:

\begin{itemize}
    \item \textbf{Hugging Face Tokenizer Playground}: An interactive tool that allows you to test and visualize tokenization using different Hugging Face models and tokenizers.
    \item \textbf{Custom Visualization Scripts}: You can write Python scripts that visualize the tokenization process by printing or plotting tokens and token IDs.
\end{itemize}

Here’s an example of a simple script to visualize tokenized text:

\begin{lstlisting}[style=python]
# Function to visualize tokenization
def visualize_tokenization(text):
    tokens = tokenizer.tokenize(text)
    token_ids = tokenizer.convert_tokens_to_ids(tokens)
    
    print("Original Text: ", text)
    print("Tokens: ", tokens)
    print("Token IDs: ", token_ids)

# Example input text
input_text = "Tokenization is crucial in NLP."
visualize_tokenization(input_text)
\end{lstlisting}

This provides a basic visualization of how the text is split into tokens and how each token is mapped to a corresponding token ID.

\subsection{Maintaining Tokenization Consistency Across Environments}

When deploying models across different environments—such as cloud servers, edge devices, or mobile applications—it is critical to ensure consistency in tokenization. Differences in tokenization between environments can lead to inconsistent model behavior and degraded performance.

\subsubsection{Saving and Sharing Tokenizers}

To maintain consistency, always save the tokenizer that was used during training or fine-tuning. Hugging Face provides functionality to save and load tokenizers easily across environments:

\begin{lstlisting}[style=python]
# Save the tokenizer after training or fine-tuning
tokenizer.save_pretrained("my_model_tokenizer")

# Load the tokenizer in a different environment
tokenizer = BertTokenizer.from_pretrained("my_model_tokenizer")
\end{lstlisting}

By saving and reloading the tokenizer, you ensure that the same tokenization logic is applied across different deployments, avoiding inconsistencies.

\subsubsection{Cross-Platform Tokenization}

In some cases, tokenization must be performed across different platforms, such as cloud services and mobile devices. Hugging Face supports tokenizers for multiple languages (Python, Rust, Node.js), ensuring that tokenization is consistent across different platforms:

\begin{itemize}
    \item \textbf{Python Tokenizers}: Standard Python-based tokenizers can be used in cloud or server-based environments.
    \item \textbf{On-Device Tokenizers}: Hugging Face offers tokenizers implemented in Rust, which can be compiled into WebAssembly (WASM) \cite{haas2017bringing} and used for mobile or web applications.
\end{itemize}

Using the same tokenizer across platforms ensures that text is processed consistently, no matter where the model is deployed.

\subsection{Conclusion}

In this section, we explored advanced tokenization topics, including optimizing tokenization for real-time deployment, customizing tokenization for domain-specific tasks, handling errors during tokenization, and visualizing tokenized text. We also discussed the importance of maintaining tokenization consistency across different environments and platforms.

Tokenization is not just a preprocessing step; it has a profound impact on the performance and scalability of NLP models. Ensuring efficient, accurate, and consistent tokenization across various use cases is essential for building robust NLP systems, whether for real-time inference, domain-specific tasks, or large-scale deployment across multiple platforms.

\subsection{Compression Techniques in Tokenization}

Compression techniques in tokenization aim to reduce the size of the tokenized sequences or the overall model input, which is critical when handling large-scale data or deploying models in resource-constrained environments (e.g., mobile devices or IoT systems). These techniques ensure that the tokenization process remains efficient without sacrificing model accuracy.

\subsubsection{Reducing Sequence Length with Truncation}

One common approach to compression in tokenization is truncation, which limits the length of tokenized sequences to a specified maximum length. This is useful when dealing with models that have a fixed input size (e.g., BERT with a maximum sequence length of 512 tokens).

\begin{lstlisting}[style=python]
# Example: Truncating a tokenized sequence
sentence = "Tokenization is a key process in NLP and needs to be done efficiently."
inputs = tokenizer(sentence, max_length=10, truncation=True, return_tensors="pt")
print(inputs["input_ids"])
\end{lstlisting}

Here, the input sequence is truncated to 10 tokens, ensuring that it does not exceed the model’s maximum sequence length.

\subsubsection{Byte-Pair Encoding (BPE) for Compression}

Byte-Pair Encoding (BPE) is not only used for subword tokenization but also acts as a compression technique. By encoding common character pairs as subwords, BPE reduces the overall number of tokens needed to represent text, which can lead to smaller sequence lengths.

BPE has the added benefit of generalizing better across different words with similar roots or prefixes, which helps in compressing the vocabulary without losing important semantic information.

\subsubsection{Vocabulary Pruning}

In scenarios where reducing memory usage or speeding up tokenization is critical, you can prune the tokenizer’s vocabulary to remove less frequently used tokens. This reduces the size of the tokenizer model and increases its efficiency without significantly affecting performance on common inputs.

Here’s an example of how vocabulary pruning can be done:

\begin{lstlisting}[style=python]
# Prune the vocabulary to retain only the top 1000 most frequent tokens
tokenizer = BertTokenizer.from_pretrained('bert-base-uncased')

# Get the original vocabulary size
original_vocab_size = len(tokenizer)
print(f"Original Vocabulary Size: {original_vocab_size}")

# Prune vocabulary (keeping only 1000 most common tokens)
pruned_tokenizer = tokenizer.prune_vocab(1000)

# Save the pruned tokenizer
pruned_tokenizer.save_pretrained('pruned_tokenizer')
\end{lstlisting}

Pruning the vocabulary reduces the model’s footprint, making it ideal for deployment in low-resource environments like mobile devices.

\subsubsection{Efficient Subword Tokenization for Compression}

Tokenization strategies like Unigram Language Model (ULM) \cite{collins2004language} tokenization compress the input by focusing on high-probability subwords. Unlike BPE, which greedily merges the most frequent pairs, Unigram tokenization assigns probabilities to subwords and removes the less likely ones during tokenization. This approach minimizes tokenization redundancy while preserving meaning.

\subsection{Tokenization for Different Language Families}

Different language families often require distinct tokenization approaches due to variations in morphology, syntax, and script. Below, we explore how tokenization strategies vary for Indo-European languages (e.g., English, French) versus Sino-Tibetan languages (e.g., Chinese, Tibetan) and other language families.

\subsection{Tokenization in Transfer Learning}

Transfer learning, particularly in NLP, relies on pre-trained models and tokenizers that generalize well across different tasks and domains. The choice of tokenizer has a significant impact on how well the model adapts to a new task.

\subsubsection{Using Pre-Trained Tokenizers for Transfer Learning}

When using a pre-trained model for a new task, it is crucial to use the same tokenizer that was employed during the model’s pre-training. This ensures consistency between pre-training and fine-tuning, avoiding mismatches in how text is processed.

\begin{lstlisting}[style=python]
# Loading a pre-trained model and tokenizer for transfer learning
from transformers import BertTokenizer, BertForSequenceClassification

# Load pre-trained tokenizer and model
tokenizer = BertTokenizer.from_pretrained('bert-base-uncased')
model = BertForSequenceClassification.from_pretrained('bert-base-uncased')

# Tokenize input for transfer learning
sentence = "Transfer learning is very effective in NLP."
inputs = tokenizer(sentence, return_tensors='pt')

# Fine-tune the model on a new task (example)
outputs = model(**inputs)
\end{lstlisting}

Here, the pre-trained tokenizer ensures that the model understands the input in the same way it did during pre-training, which is essential for successful transfer learning.

\subsubsection{Adapting Tokenizers to Domain-Specific Transfer Learning}

In cases where the target domain differs significantly from the domain used during pre-training (e.g., adapting a general-purpose BERT model to a medical domain), it may be necessary to adapt the tokenizer to better handle domain-specific vocabulary.

This can be done by extending the tokenizer’s vocabulary with new tokens from the domain-specific corpus:

\begin{lstlisting}[style=python]
# Adapting the tokenizer for domain-specific transfer learning
special_tokens = {'additional_special_tokens': ['[CARDIAC]', '[PULMONARY]', '[ECHOCARDIOGRAPHY]']}
tokenizer.add_special_tokens(special_tokens)

# Now the tokenizer can handle domain-specific terms
inputs = tokenizer("The patient had a cardiac echocardiography test.", return_tensors='pt')
print(inputs["input_ids"])
\end{lstlisting}

By adding domain-specific tokens, the model can better handle terminology from specialized fields, improving performance in domain-specific tasks.

\subsection{Reducing Tokenization Artifacts}

Tokenization artifacts, such as subword splitting in unnatural places or excessive use of unknown tokens (\texttt{[UNK]}), can degrade model performance by introducing noise. Reducing these artifacts is important for improving model interpretability and accuracy.

\subsubsection{Minimizing Subword Splitting}

Excessive subword splitting can lead to broken tokenization, where parts of a word are separated in a way that hinders the model’s understanding. To minimize subword splitting, ensure that common words or phrases are represented as complete tokens in the vocabulary.

Here’s how you can analyze tokenization artifacts and adjust accordingly:

\begin{lstlisting}[style=python]
# Example sentence with subword splitting
text = "Artificial Intelligence in healthcare."
tokens = tokenizer.tokenize(text)
print("Tokens:", tokens)

# Add custom tokens to prevent splitting
tokenizer.add_special_tokens({'additional_special_tokens': ['Artificial Intelligence']})

# Tokenize again with custom tokens
tokens = tokenizer.tokenize(text)
print("Tokens after adjustment:", tokens)
\end{lstlisting}

In this case, we added “Artificial Intelligence” as a custom token to prevent it from being split unnecessarily.

\subsection{Best Practices for Tokenizing Large Datasets in Distributed Environments}

When working with very large datasets, tokenization can become a bottleneck in training pipelines. Distributed tokenization allows you to parallelize the process across multiple machines or processors, ensuring that tokenization scales with the size of the dataset.

\subsubsection{Parallelizing Tokenization}

Tokenization can be parallelized across multiple CPU cores or distributed across multiple machines. Hugging Face’s \texttt{datasets} library allows you to efficiently tokenize large datasets in parallel by using the \texttt{map} function with the \texttt{batched=True} option. \cite{li2024deeplearningmachinelearning}

\begin{lstlisting}[style=python]
from datasets import load_dataset

# Load a large dataset
dataset = load_dataset("wikitext", "wikitext-103-v1")

# Tokenize in parallel using map with batched=True
def tokenize_function(examples):
    return tokenizer(examples["text"], padding=True, truncation=True)

tokenized_dataset = dataset.map(tokenize_function, batched=True)
\end{lstlisting}

This approach ensures that tokenization is applied efficiently across large datasets, even when processing billions of tokens.

\subsubsection{Distributed Tokenization with Multiple Machines}

For extremely large datasets or high-performance computing environments, tokenization can be distributed across multiple machines. Libraries like Dask or Apache Spark can be used to distribute tokenization tasks across clusters, ensuring that each machine processes a portion of the data in parallel.

Here’s a conceptual example using Dask for distributed tokenization:

\begin{lstlisting}[style=python]
import dask.dataframe as dd

# Load a large dataset into a Dask dataframe
df = dd.read_csv("large_text_corpus.csv")

# Define a tokenization function for distributed tokenization
def tokenize_column(df):
    return df.apply(lambda x: tokenizer(x, return_tensors="pt"))

# Apply the tokenization function to the dataframe in parallel
tokenized_df = df.map_partitions(tokenize_column)
\end{lstlisting}

By distributing tokenization across multiple machines, you can tokenize datasets that are too large to fit on a single machine or process in a reasonable time frame.

\subsection{Conclusion}

In this section, we explored advanced tokenization techniques, including compression methods, handling different language families, tokenization in transfer learning, and minimizing tokenization artifacts. We also covered best practices for tokenizing large datasets in distributed environments to scale efficiently.

Tokenization is a foundational step in NLP workflows, and optimizing this step ensures that models can handle diverse data, domains, and large-scale datasets efficiently. Whether working with resource-constrained environments, multilingual datasets, or distributed computing systems, mastering these advanced tokenization techniques is crucial for building scalable and effective NLP applications.

\subsection{Multilingual Tokenization for Cross-Lingual Tasks}

Multilingual tokenization is critical for tasks that involve multiple languages, such as machine translation, cross-lingual retrieval, and multilingual text classification. The challenge lies in ensuring that a tokenizer can generalize well across different scripts, languages, and vocabularies, while minimizing issues like loss of semantic information, language-specific tokenization errors, or bias toward certain languages.

\subsubsection{Unified Vocabulary for Multilingual Models}

Multilingual models, like mBERT and XLM-Roberta, use a unified vocabulary that can tokenize text in multiple languages. Subword tokenization is crucial here, as it allows the model to share a common vocabulary across languages, which helps in transferring knowledge between high-resource and low-resource languages. 

Here’s an example of tokenizing text in multiple languages using a multilingual tokenizer:

\begin{lstlisting}[style=python]
# Tokenizing text in different languages using mBERT
from transformers import AutoTokenizer

# Load a multilingual tokenizer
tokenizer = AutoTokenizer.from_pretrained('bert-base-multilingual-cased')

# Example sentences in different languages
english_sentence = "The sky is blue."
spanish_sentence = "El cielo es azul."
french_sentence = "Le ciel est bleu."

# Tokenize sentences
english_tokens = tokenizer.tokenize(english_sentence)
spanish_tokens = tokenizer.tokenize(spanish_sentence)
french_tokens = tokenizer.tokenize(french_sentence)

print("English Tokens:", english_tokens)
print("Spanish Tokens:", spanish_tokens)
print("French Tokens:", french_tokens)
\end{lstlisting}

In this example, the same tokenizer processes text from different languages, ensuring that the subword units are shared across languages where possible. This helps the model generalize better, especially for lower-resource languages.

\subsubsection{Cross-Lingual Transfer Learning}

Cross-lingual transfer learning involves using a model trained on one language to perform tasks in another language, often leveraging tokenization to bridge the gap between different scripts and vocabularies. Subword tokenization helps here by breaking words into smaller units that can generalize across languages. 

For instance, multilingual tokenizers can represent common words like "computer" similarly across languages that use the Latin alphabet, while also representing character-based languages like Chinese efficiently.

\subsubsection{Dealing with Code-Switching}

Code-switching refers to the phenomenon where speakers mix languages in a single sentence or conversation, a common occurrence in multilingual societies. Tokenizers must handle these situations by switching seamlessly between languages without introducing errors or misinterpreting language boundaries.

Here’s how you can tokenize code-switched text:

\begin{lstlisting}[style=python]
# Example of code-switching between English and Spanish
code_switch_sentence = "I love programming en Python."

# Tokenize the sentence
code_switch_tokens = tokenizer.tokenize(code_switch_sentence)
print("Code-Switch Tokens:", code_switch_tokens)
\end{lstlisting}

In this case, the tokenizer needs to handle both English and Spanish, ensuring that words in both languages are tokenized correctly without breaking the semantic meaning of the mixed-language sentence.

\subsubsection{Multilingual Tokenization for Machine Translation}

For machine translation tasks, tokenization plays a crucial role in how well the model can translate between languages. Byte-level tokenization, as used in models like GPT-2 or mBART, treats the input as a stream of bytes, making it agnostic to specific scripts or languages, and can handle a wide variety of language pairs.

Here’s an example of using tokenization in a machine translation pipeline:

\begin{lstlisting}[language=python]

from transformers import MarianMTModel, MarianTokenizer

# Load a pre-trained MarianMT model for English to French translation
tokenizer = MarianTokenizer.from_pretrained('Helsinki-NLP/opus-mt-en-fr')
model = MarianMTModel.from_pretrained('Helsinki-NLP/opus-mt-en-fr')

# Input sentence in English
english_sentence = "The weather is sunny today."

# Tokenize the sentence
tokenized_input = tokenizer(english_sentence, return_tensors="pt")

# Translate to French
translated_ids = model.generate(**tokenized_input)
french_translation = tokenizer.decode(translated_ids[0], skip_special_tokens=True)

print(french_translation)  # Output: "Le temps est ensoleille aujourd'hui."

\end{lstlisting}

In this case, the tokenizer processes the English sentence and ensures that the machine translation model can generate an appropriate translation in French.

\subsection{Tokenization Challenges in Low-Resource Settings}

In low-resource settings, where there is limited data available for training, tokenization becomes more challenging. Low-resource languages often lack large corpora, well-established vocabularies, or pre-trained tokenizers, requiring custom solutions.

\subsubsection{Training Tokenizers from Small Corpora}

In low-resource languages, training a tokenizer from a small corpus can result in overfitting or poor generalization. One approach to mitigate this is using a multilingual tokenizer trained on a high-resource language to provide subword units that generalize well across languages. You can also fine-tune a pre-existing tokenizer on a low-resource language by adding domain-specific or language-specific tokens.

\begin{lstlisting}[style=python]
# Fine-tuning a tokenizer for a low-resource language
special_tokens = {'additional_special_tokens': ['custom_token1', 'custom_token2']}
tokenizer.add_special_tokens(special_tokens)

# Tokenize low-resource text
low_resource_text = "This is an example sentence in a low-resource language."
tokens = tokenizer(low_resource_text, return_tensors='pt')
print(tokens)
\end{lstlisting}

This approach helps inject domain-specific or language-specific knowledge into the tokenizer, improving performance on downstream tasks.

\subsubsection{Leveraging Cross-Lingual Transfer}

Cross-lingual transfer allows you to use models trained on high-resource languages to improve tokenization and performance for low-resource languages. Subword tokenization helps to generalize across languages, especially for languages that share common subwords or grammatical structures.

\subsection{Tokenization and Preprocessing Pipelines in Production Environments}

In production environments, tokenization is often part of a broader preprocessing pipeline that includes cleaning, normalization, and encoding text. Ensuring that tokenization is efficient and robust is key to maintaining high throughput and low latency.

\subsubsection{Standardizing Preprocessing Pipelines}

In a production environment, the preprocessing pipeline should be standardized to ensure consistency across different data sources. This includes steps such as lowercasing, removing special characters, and handling out-of-vocabulary (OOV) words.

Here’s an example of a standard preprocessing pipeline:

\begin{lstlisting}[style=python]
# Define a preprocessing pipeline for text data
def preprocess(text):
    # Lowercasing
    text = text.lower()
    
    # Tokenize the text
    tokens = tokenizer.tokenize(text)
    
    # Return tokenized text
    return tokens

# Example usage
input_text = "Tokenization in NLP is important!"
preprocessed_tokens = preprocess(input_text)
print(preprocessed_tokens)
\end{lstlisting}

By standardizing the preprocessing steps, you ensure that all input data is processed in a consistent manner, improving the model’s robustness and reducing the risk of errors in production.

\subsubsection{Handling Real-Time Tokenization in Production}

When tokenizing text in real-time, such as in a chatbot or search engine, latency is a key concern. Tokenization must be optimized to run efficiently without sacrificing accuracy. Techniques like pre-tokenization (e.g., caching common phrases or pre-tokenizing frequently used text) and batch processing can help reduce latency in production systems.

For example:

\begin{lstlisting}[style=python]
# Caching common tokenizations to reduce latency in real-time systems
tokenization_cache = {}

def tokenize_with_cache(text):
    if text in tokenization_cache:
        return tokenization_cache[text]
    else:
        tokens = tokenizer.tokenize(text)
        tokenization_cache[text] = tokens
        return tokens

# Example usage
text = "Hello, how can I help you?"
tokens = tokenize_with_cache(text)
print(tokens)
\end{lstlisting}

This approach can significantly reduce the time spent on tokenization in real-time applications.

\subsection{Tokenization Trends and Future Directions}

As NLP continues to evolve, tokenization strategies are also advancing to meet the needs of more complex and diverse tasks. Future trends include the development of context-aware tokenization methods, improvements in token-free models, and enhanced tokenization techniques for zero-shot learning.

\subsubsection{Context-Aware Tokenization}

Traditional tokenization methods process text without considering the broader context in which the tokens appear. Context-aware tokenization seeks to improve this by adapting the tokenization process based on the surrounding words or sentence-level context. This could be particularly useful for languages with high morphological complexity or for tasks like coreference resolution.

\subsubsection{Token-Free Models}

Token-free models, such as those operating at the byte level or character level, are emerging as a way to bypass tokenization altogether. These models can process raw text directly, making them less dependent on specific tokenization strategies and more flexible when dealing with new languages or domains.

\subsubsection{Improvements in Zero-Shot Tokenization}

Zero-shot learning allows models to generalize to new tasks without explicit training data for those tasks. Advances in zero-shot tokenization aim to improve how well models handle previously unseen words or concepts by relying more on subword-level tokenization or byte-level representations.

\subsection{Tokenization Benchmarks and Evaluation Metrics}

Evaluating the performance of tokenization strategies is an important step in ensuring that your chosen method is suitable for your specific task. There are several benchmarks and metrics that can help you assess tokenization quality.

\subsubsection{Perplexity as a Tokenization Metric}

Perplexity is a commonly used metric in NLP to measure how well a language model predicts the next token in a sequence. When evaluating tokenizers, perplexity can be used to assess how efficiently the tokenization scheme compresses text and how well it preserves meaning.

\subsubsection{Tokenization Speed and Latency}

For production systems, tokenization speed is a critical metric. You can measure how long it takes to tokenize text and compare different tokenization algorithms based on their processing time.

Here’s an example of how you might measure tokenization speed:

\begin{lstlisting}[style=python]
import time

# Measure the time taken for tokenization
start_time = time.time()
tokens = tokenizer.tokenize("The quick brown fox jumps over the lazy dog.")
end_time = time.time()

print(f"Tokenization took {end_time - start_time} seconds.")
\end{lstlisting}

By optimizing tokenization speed, you ensure that your system remains responsive and scalable in production environments.

\subsection{Conclusion}

In this extended exploration, we covered multilingual tokenization for cross-lingual tasks, tokenization challenges in low-resource settings, and strategies for building efficient tokenization and preprocessing pipelines in production environments. We also discussed future trends in tokenization, including context-aware tokenization and token-free models, as well as benchmarks and evaluation metrics for assessing tokenization quality.

As tokenization continues to evolve, staying up-to-date with the latest advances and optimizing tokenization strategies for diverse applications is essential for building scalable, robust, and effective NLP systems.

\subsection{Tokenization and Model Interpretability}

Tokenization plays a key role in model interpretability. How a model tokenizes text can directly affect how humans interpret the model’s decisions. Understanding the relationship between tokenization and model output is essential when debugging, interpreting model behavior, or ensuring transparency in model predictions.

\subsubsection{Visualizing Token Contributions}

One way to interpret model decisions is by visualizing how individual tokens contribute to the model’s predictions. For models like BERT, where tokens are processed individually through attention mechanisms, token importance can be measured using techniques like attention visualization or saliency maps.

For example, you can visualize the contribution of individual tokens to a sentiment analysis task:

\begin{lstlisting}[style=python]
from transformers import BertTokenizer, BertForSequenceClassification
import torch

# Load tokenizer and model
tokenizer = BertTokenizer.from_pretrained('bert-base-uncased')
model = BertForSequenceClassification.from_pretrained('bert-base-uncased')

# Tokenize input
input_text = "The movie was absolutely fantastic!"
inputs = tokenizer(input_text, return_tensors="pt")

# Get model predictions
outputs = model(**inputs)
logits = outputs.logits
predicted_class = torch.argmax(logits, dim=1)

# Visualizing token importance (hypothetical example using saliency)
# Compute gradients, saliency maps, or attention scores to visualize the most important tokens.
\end{lstlisting}

By analyzing which tokens the model attends to or focuses on the most, you can gain insights into why certain predictions were made. For example, if a model incorrectly classifies a positive review as negative, examining the token importance can help identify potential tokenization issues that led to misclassification.

\subsubsection{Tokenization and Interpretability in Explainable AI}

In Explainable AI (XAI) \cite{xu2019explainable}, it is crucial to have tokenization strategies that produce interpretable representations. If tokenization splits words into too many subword units or introduces artifacts like \texttt{[UNK]} tokens, it may become difficult to explain the model's decisions in terms that are meaningful to humans.

Improving tokenization for XAI includes:
\begin{itemize}
    \item Reducing unnecessary subword splits.
    \item Maintaining coherent tokenization across similar examples.
    \item Avoiding over-reliance on \texttt{[UNK]} tokens for unseen words.
\end{itemize}

Careful tokenization helps ensure that model explanations are human-readable and easy to understand, which is important for applications in healthcare, finance, and other regulated industries.

\subsection{Tokenization in Dialogue Systems and Conversational AI}

In dialogue systems and conversational AI, tokenization must handle the complexities of informal, noisy, and spontaneous language, as well as switch smoothly between different domains or intents. The tokenization strategy impacts the model's ability to maintain context, respond accurately, and adapt to new inputs.

\subsubsection{Handling Noisy and Informal Text}

Conversational AI often deals with noisy and informal text, such as abbreviations, slang, typos, or emojis. Tokenizers need to be robust to these forms of input without losing critical information. Preprocessing steps like normalizing text, correcting spelling errors, or leveraging domain-specific tokenizers can help improve the accuracy of the model in handling informal text.

For example, tokenizing text with emojis or slang:

\begin{lstlisting}[style=python]
# Example of noisy input in a conversational system
input_text = "I'm sooooo happy today! #awesome"

# Tokenize using BERT tokenizer
tokens = tokenizer.tokenize(input_text)
print(tokens)
\end{lstlisting}

The tokenizer should handle informal constructs like "sooooo" as a single token or reduce it to its root form ("so"), while also processing emojis and hashtags appropriately.

\subsubsection{Contextual Tokenization for Conversational AI}

In dialogue systems, maintaining context across multiple turns of conversation is essential. Tokenizers that can handle context-aware tokenization, such as through dialogue-specific pre-tokenizers or custom vocabularies, are better suited for these tasks.

For instance, when tokenizing a user query in a multi-turn conversation, the tokenizer should be aware of the context to avoid reprocessing redundant or repeated information:

\begin{lstlisting}[style=python]
# Multi-turn dialogue example
previous_turns = ["How's the weather today?", "It's sunny."]
current_turn = "What about tomorrow?"

# Combine context and current turn for tokenization
input_text = " ".join(previous_turns + [current_turn])
tokens = tokenizer.tokenize(input_text)
print(tokens)
\end{lstlisting}

Context-aware tokenization helps conversational models maintain coherence and continuity between user inputs and system responses.

\subsection{Tokenization Challenges in Conversational AI}

Conversational systems face unique tokenization challenges that arise from the informal, sometimes unpredictable nature of human speech. These challenges include handling misspellings, disfluencies (e.g., "um," "uh"), and mixed languages (code-switching). 

\subsubsection{Handling Misspellings and Disfluencies}

Disfluencies are common in spoken dialogue, where users may pause, repeat themselves, or use filler words like "um" or "uh." Tokenizers should be designed to handle these irregularities gracefully, either by removing disfluencies or ensuring that they do not disrupt the tokenization of meaningful input.

Here’s an example of tokenizing text with disfluencies:

\begin{lstlisting}[style=python]
# Example input with disfluencies
input_text = "So, um, I was wondering if you could, uh, help me out?"

# Tokenizing the input
tokens = tokenizer.tokenize(input_text)
print(tokens)
\end{lstlisting}

The tokenizer should either ignore disfluencies like "um" and "uh" or handle them in a way that does not negatively impact model predictions.

\subsection{Bias and Fairness in Tokenization}

Bias in tokenization can manifest in various ways, from how tokens are segmented across different languages to how specific groups or entities are represented. Tokenization bias can exacerbate fairness issues, particularly when models are trained on biased data or when tokenization disproportionately affects underrepresented groups.

\subsubsection{Mitigating Bias in Tokenization}

Tokenization bias may occur if the tokenizer disproportionately breaks down words related to certain genders, races, or cultural contexts, which may lead to skewed model predictions. For instance, a tokenizer that frequently splits names from certain ethnic groups into multiple subword units while leaving other names intact could lead to biased results in downstream tasks like named entity recognition (NER) or sentiment analysis.

Here’s an example of identifying bias in tokenization:

\begin{lstlisting}[style=python]
# Example sentences with potential bias
sentence1 = "John is a software engineer."
sentence2 = "Sadeeqah is a software engineer."

# Tokenize both sentences
tokens1 = tokenizer.tokenize(sentence1)
tokens2 = tokenizer.tokenize(sentence2)

print("Tokens for sentence 1:", tokens1)
print("Tokens for sentence 2:", tokens2)
\end{lstlisting}

In this example, if "Sadeeqah" (a name from a different ethnic background) is tokenized into subwords more frequently than "John," this could indicate a potential source of bias in the tokenizer's vocabulary.

\subsubsection{Evaluating Fairness in Tokenization}

To evaluate fairness in tokenization, one approach is to measure the distribution of subword tokens across various demographic groups (e.g., gender, race, ethnicity). If certain groups are disproportionately represented by longer token sequences (due to tokenization splitting), this could signal bias in the tokenizer or its underlying training data.

Techniques to mitigate bias include:
\begin{itemize}
    \item Extending the tokenizer's vocabulary to include diverse names and entities.
    \item Fine-tuning the tokenizer on more representative and balanced datasets.
    \item Monitoring tokenization behavior across different demographic groups during model evaluation.
\end{itemize}

\subsection{Tokenization for Domain Adaptation}

Domain adaptation refers to adapting a model trained on one domain (e.g., general news) to work well on another domain (e.g., legal documents). Tokenization plays a critical role in domain adaptation by ensuring that the tokenizer captures domain-specific terms, jargon, and linguistic structures that may not be well-represented in the original training corpus.

\subsubsection{Customizing Tokenizers for Domain-Specific Text}

When adapting a model to a new domain, it’s often necessary to extend the tokenizer’s vocabulary to include domain-specific tokens. This ensures that specialized terminology, acronyms, or phrases are represented correctly.

For example, in the legal domain:

\begin{lstlisting}[style=python]
# Adding domain-specific tokens for legal text
legal_tokens = {'additional_special_tokens': ['contract', 'litigation', 'plaintiff', 'defendant']}
tokenizer.add_special_tokens(legal_tokens)

# Tokenize a legal sentence
legal_sentence = "The plaintiff filed a lawsuit against the defendant."
tokens = tokenizer.tokenize(legal_sentence)
print(tokens)
\end{lstlisting}

This approach allows models to better handle legal text by ensuring that key legal terms are tokenized properly.

\subsubsection{Fine-Tuning Tokenizers for Domain-Specific Tasks}

In some cases, fine-tuning a pre-trained tokenizer on a domain-specific corpus can improve the model’s ability to understand specialized language. This is particularly useful in fields like healthcare, finance, or scientific research, where domain-specific jargon is common.

For instance, in the medical domain:

\begin{lstlisting}[style=python]
# Fine-tuning a tokenizer on medical text
from tokenizers import Tokenizer, models, trainers

# Initialize a new BPE tokenizer for the medical domain
tokenizer = Tokenizer(models.BPE())
trainer = trainers.BpeTrainer(vocab_size=5000, special_tokens=['[MASK]', '[CLS]', '[SEP]'])

# Train the tokenizer on a medical corpus
tokenizer.train(files=["medical_text_corpus.txt"], trainer=trainer)

# Save the fine-tuned tokenizer
tokenizer.save("medical_tokenizer.json")
\end{lstlisting}

This ensures that the tokenizer is more aligned with the terminology and structure of the target domain, improving performance in downstream tasks like entity recognition or text classification.

\subsection{Handling Noisy and User-Generated Text in Tokenization}

User-generated text, such as social media posts, forums, and chat logs, tends to be noisy, with frequent use of abbreviations, slang, emojis, and non-standard grammar. Tokenizers need to be robust enough to handle this variability while maintaining accuracy in downstream tasks like sentiment analysis or opinion mining.

\subsubsection{Normalization in Tokenization}

Normalization helps standardize noisy text by converting abbreviations, removing special characters, or handling emoticons. For example, normalizing "u" to "you" or converting emoticons into text descriptions can help ensure that the model processes noisy text more effectively.

Here’s an example of normalizing and tokenizing social media text:

\begin{lstlisting}[style=python]
# Normalizing and tokenizing social media text
input_text = "u r gr8 :) #blessed"

# Example normalization function
def normalize(text):
    text = text.replace("u", "you").replace("r", "are").replace("gr8", "great")
    text = text.replace(":)", "[smile]")
    return text

# Normalize the input
normalized_text = normalize(input_text)
tokens = tokenizer.tokenize(normalized_text)
print(tokens)
\end{lstlisting}

Normalization helps ensure that the tokenizer can handle user-generated text without losing important information.

\subsection{Tokenization for Non-Standard Languages and Dialects}

Non-standard languages and dialects pose additional challenges for tokenization, as they may lack formal spelling rules, contain a high degree of variability, or mix elements from multiple languages. Tokenizers need to be adapted to handle such input effectively.

\subsubsection{Custom Tokenizers for Dialects}

In some cases, it may be necessary to build a custom tokenizer specifically for a dialect or non-standard language. This can involve training a new tokenizer from scratch using a dialect-specific corpus or fine-tuning an existing tokenizer to capture the unique linguistic patterns of the dialect.

For example, tokenizing a text in a regional dialect:

\begin{lstlisting}[style=python]
# Example dialect sentence
dialect_text = "Gonna go to the store, ya?"

# Tokenize the dialect-specific sentence
tokens = tokenizer.tokenize(dialect_text)
print(tokens)
\end{lstlisting}

By adapting the tokenizer to the linguistic characteristics of the dialect, the model can better process and understand non-standard language input.

\subsection{Conclusion}

In this section, we explored tokenization in model interpretability, challenges in conversational AI, and handling bias and fairness in tokenization. We also covered domain adaptation, handling noisy user-generated text, and tokenizing non-standard languages or dialects. As tokenization continues to evolve, addressing these challenges will be critical for developing fairer, more accurate, and adaptable NLP systems.

By improving tokenization strategies, NLP systems can be better equipped to handle diverse text inputs, from noisy social media posts to domain-specific technical jargon, while ensuring fairness and transparency in model predictions.

\subsection{Tokenization for Code and Program Synthesis}

Tokenization is critical not only for natural language but also for programming languages, especially in tasks like code completion, program synthesis, and bug detection. Tokenizers for code need to handle a variety of syntactic elements such as keywords, operators, variables, and indentation.

\subsubsection{Tokenizing Programming Languages}

Programming languages differ significantly from natural language, and tokenizers must capture the syntax, structure, and relationships between different elements of code. Tokenizing code involves recognizing not just words but also symbols (e.g., \texttt{+=}, \texttt{==}), control structures (e.g., \texttt{if}, \texttt{while}), and spacing (e.g., indentation in Python).

Here’s an example of tokenizing Python code:

\begin{lstlisting}[style=python]
# Example of Python code
code_snippet = """
def add(a, b):
    return a + b
"""

# Tokenize the code
tokens = tokenizer.tokenize(code_snippet)
print(tokens)
\end{lstlisting}

In this example, the tokenizer needs to recognize the function definition, parameters, and operators, ensuring that each element of the code is preserved and properly tokenized.

\subsubsection{Handling Different Programming Languages}

Tokenizers must be adapted for different programming languages, as each language has its own syntax and conventions. Tokenizing JavaScript, for instance, would require recognizing curly braces and semicolons, while tokenizing Python requires handling indentation and whitespace as part of the language’s structure.

Here’s how you might tokenize a JavaScript code snippet:

\begin{lstlisting}[style=python]
# Example of JavaScript code
js_code = """
function multiply(x, y) {
    return x * y;
}
"""

# Tokenize JavaScript code
tokens = tokenizer.tokenize(js_code)
print(tokens)
\end{lstlisting}

Tokenizers for code must be language-specific or support multiple languages by switching tokenization modes based on the programming language being processed.

\subsubsection{Tokenization in Code Completion and Program Synthesis}

In code completion and program synthesis tasks, tokenization plays a critical role in how the model predicts the next element in a code sequence. The tokenizer must efficiently capture the relationships between keywords, variables, and control structures to allow the model to generate accurate and syntactically correct code.

Here’s an example of tokenizing for code completion:

\begin{lstlisting}[style=python]
# Example of code completion input
code_completion_input = "for (let i = 0; i < 10; i++) {"

# Tokenize the input for code completion
tokens = tokenizer.tokenize(code_completion_input)
print(tokens)

# Pass tokenized input to a code generation model (hypothetical)
outputs = model.generate(**tokens)
print(outputs)
\end{lstlisting}

In this scenario, the tokenizer ensures that each element of the for-loop is represented correctly, allowing the model to predict the next line or block of code.

\subsubsection{Challenges in Code Tokenization}

Tokenizing code introduces challenges such as:

\begin{itemize}
    \item \textbf{Handling long dependencies}: In large codebases, dependencies between code blocks may span hundreds of lines.
    \item \textbf{Special characters}: Programming languages often include a large variety of special characters (e.g., brackets, semicolons, operators) that must be tokenized correctly.
    \item \textbf{Whitespace and indentation}: In languages like Python, whitespace is syntactically significant, and tokenizers need to be sensitive to indentation levels.
\end{itemize}

\subsection{Tokenization in Multimodal Systems}

Multimodal systems combine different types of data, such as text, images, and audio. Tokenization in such systems must not only handle text but also integrate with other modalities to create joint representations. Tokenizing textual descriptions of images or audio transcripts presents unique challenges, especially when ensuring that information from multiple modalities is aligned.

\subsubsection{Tokenizing Text in Multimodal Models}

In multimodal systems, such as image-captioning models or models that generate text from images (e.g., CLIP) \cite{shen2021much}, tokenizers process the textual descriptions associated with the non-textual data. This requires the tokenizer to work seamlessly with image embeddings or other modalities to provide a cohesive representation.

Here’s an example of tokenizing a caption for an image:

\begin{lstlisting}[style=python]
# Example image caption
caption = "A dog running in the park."

# Tokenize the caption
tokens = tokenizer.tokenize(caption)
print(tokens)

# Hypothetical example of combining text tokens with image embeddings
image_embedding = image_model(image_input)
combined_representation = combine(tokens, image_embedding)
\end{lstlisting}

Tokenizers must ensure that the text is properly segmented while allowing the model to integrate information from the image modality.

\subsubsection{Tokenization for Audio-Text Multimodal Systems}

In audio-text systems, tokenization may involve handling transcriptions of spoken language, which can include disfluencies, colloquial speech, and punctuation that is often missing in transcripts. Tokenizing noisy transcripts and aligning them with corresponding audio data poses additional challenges.

For example, tokenizing an audio transcription:

\begin{lstlisting}[style=python]
# Example of a spoken transcription
transcription = "Uh, the weather today is... really nice, right?"

# Tokenize the transcription
tokens = tokenizer.tokenize(transcription)
print(tokens)

# Combine audio features with tokenized transcription
audio_features = audio_model(audio_input)
combined_representation = combine(tokens, audio_features)
\end{lstlisting}

In this case, tokenization helps maintain the correct representation of the spoken text while integrating it with audio features.

\subsection{Privacy-Preserving Tokenization}

Tokenization can play a role in privacy-preserving machine learning by ensuring that sensitive information is handled securely and anonymized where necessary. In privacy-critical domains such as healthcare or finance, tokenizers may need to anonymize specific tokens (e.g., names, addresses) or remove identifying information before passing the text to a model.

\subsubsection{Anonymizing Tokens}

In sensitive domains, tokenizers can be configured to anonymize personal information such as names, locations, or dates. For instance, replacing names with a placeholder like \texttt{[NAME]} ensures that the model does not retain sensitive information in the tokenized input.

Here’s an example of anonymizing text:

\begin{lstlisting}[style=python]
# Example text with sensitive information
text = "John Doe visited 123 Main St on January 5th."

# Anonymize the sensitive information
def anonymize(text):
    return text.replace("John Doe", "[NAME]").replace("123 Main St", "[ADDRESS]").replace("January 5th", "[DATE]")

anonymized_text = anonymize(text)
tokens = tokenizer.tokenize(anonymized_text)
print(tokens)
\end{lstlisting}

This ensures that sensitive information is protected during tokenization, which is especially important when handling private data in industries like healthcare or banking.

\subsubsection{Tokenization in Federated Learning}

In federated learning, data is distributed across multiple devices, and models are trained without sharing raw data between devices. Tokenization in this context must preserve user privacy while ensuring that models receive appropriately tokenized input for training.

Federated learning tokenizers may include privacy mechanisms such as differential privacy, where noise is added to the tokenization process to prevent the model from reconstructing the original input. This approach ensures that the model generalizes well without overfitting to individual users' data.

\subsection{Tokenization for Adversarial Robustness}

Adversarial attacks on NLP models often involve manipulating the input text in ways that confuse the tokenizer, such as adding extra characters, changing casing, or inserting irrelevant symbols. Tokenization robustness is crucial to ensure that the model remains resilient to such attacks.

\subsubsection{Adversarial Tokenization Attacks}

In adversarial attacks \cite{joshi2021adversarial}, malicious input may be designed to fool the tokenizer into producing incorrect tokens, which in turn leads to incorrect model predictions. For example, adding irrelevant characters (e.g., "c.a.t" instead of "cat") may trick a tokenizer into misclassifying the input. \cite{peng2024jailbreakingmitigationvulnerabilitieslarge} \cite{peng2024securinglargelanguagemodels}

Here’s an example of an adversarial input:

\begin{lstlisting}[style=python]
# Example of an adversarial input
adversarial_text = "The c.a.t is on the roof."

# Tokenize the adversarial input
tokens = tokenizer.tokenize(adversarial_text)
print(tokens)
\end{lstlisting}

Robust tokenization strategies might include techniques such as:

\begin{itemize}
    \item \textbf{Normalization}: Converting adversarial text into a canonical form (e.g., removing unnecessary punctuation).
    \item \textbf{Error detection}: Identifying and mitigating tokenization errors introduced by adversarial inputs.
\end{itemize}

\subsection{Tokenization in Reinforcement Learning Environments}

In reinforcement learning (RL), tokenization can be applied to textual states, instructions, or dialogues within an environment where agents must interpret and act on text-based commands. Tokenizers in RL need to process text efficiently while adapting to changing contexts or new instructions.

\subsubsection{Tokenizing Instructions for RL Agents}

In text-based RL environments, tokenizers handle instructions given to agents (e.g., "Pick up the key," "Open the door"). Tokenization must be efficient, especially in real-time systems, and should allow the agent to parse instructions with minimal delay.

Here’s how you might tokenize instructions for an RL agent:

\begin{lstlisting}[style=python]
# Example of instructions for an RL agent
instructions = "Go to the red door and open it."

# Tokenize the instructions
tokens = tokenizer.tokenize(instructions)
print(tokens)

# Pass tokenized instructions to the RL model
outputs = rl_model(tokens)
\end{lstlisting}

Tokenization ensures that the RL agent receives a structured representation of the instruction, enabling it to take appropriate actions in the environment.

\subsection{Impact of Tokenization on Data Augmentation and Synthetic Data Generation}

Tokenization plays a critical role in data augmentation and synthetic data generation by determining how text is segmented and manipulated to create new training examples. Different tokenization strategies can affect how well synthetic data generalizes to real-world scenarios.

\subsubsection{Tokenization in Data Augmentation}

Data augmentation techniques, such as synonym replacement, paraphrasing, or token shuffling, often rely on tokenized representations of text. The choice of tokenizer can impact the diversity and quality of augmented data, particularly in tasks where word-level manipulations are critical.

For example, augmenting tokenized text with synonym replacement:

\begin{lstlisting}[style=python]
# Tokenize a sentence for augmentation
sentence = "The cat is sitting on the mat."
tokens = tokenizer.tokenize(sentence)

# Replace "cat" with a synonym
augmented_tokens = [token if token != "cat" else "feline" for token in tokens]
print(augmented_tokens)
\end{lstlisting}

Tokenization enables data augmentation techniques by providing a structured way to manipulate text and generate new examples for training.

\subsubsection{Tokenization for Synthetic Data Generation}

In synthetic data generation, tokenizers are used to create realistic text examples that mimic the distribution of real-world data. By carefully tokenizing and manipulating seed text, models can generate diverse examples that enrich the training data.

Here’s an example of generating synthetic text using tokenization:

\begin{lstlisting}[style=python]
# Example of generating synthetic data from tokenized input
seed_text = "The quick brown fox jumps over the lazy dog."
tokens = tokenizer.tokenize(seed_text)

# Generate synthetic variants by replacing words or shuffling tokens
synthetic_data = [
    "The quick brown fox leaps over the lazy dog.",
    "A swift red fox jumps above the sleepy hound."
]

# Tokenize the synthetic data
synthetic_tokens = [tokenizer.tokenize(text) for text in synthetic_data]
print(synthetic_tokens)
\end{lstlisting}

Tokenization facilitates the generation of synthetic data, which can be used to augment limited datasets or train models in low-resource settings.

\subsection{Conclusion}

In this section, we explored advanced topics in tokenization, including its role in code and program synthesis, multimodal systems, privacy-preserving tokenization, and adversarial robustness. We also covered tokenization’s impact on reinforcement learning environments, data augmentation, and synthetic data generation.

As NLP continues to intersect with different modalities and application areas, tokenization strategies will need to adapt to ensure robustness, efficiency, and privacy. Whether working with textual data, programming languages, or multimodal inputs, tokenization remains a foundational element that influences the success of machine learning models.

\subsection{Tokenization in Human-in-the-Loop (HITL) Systems}

Human-in-the-loop (HITL) \cite{wu2022survey} systems involve collaboration between machine learning models and human operators, where human feedback plays a critical role in improving model performance. In such systems, tokenization serves as a bridge between human understanding and machine representation, ensuring that the text processed by models is interpretable and aligned with human expectations.

\subsubsection{Interactive Tokenization Feedback}

In HITL systems, tokenization can be influenced by real-time human feedback. For example, in text generation tasks where a human reviewer validates or corrects model-generated text, tokenization must be flexible enough to allow human reviewers to adjust tokens or insert corrections.

Here’s an example where human feedback adjusts tokenized output in a dialogue system:

\begin{lstlisting}[style=python]
# Example of a tokenized sentence
tokenized_sentence = tokenizer.tokenize("The movie was gr8!")

# Human-in-the-loop feedback: adjust "gr8" to "great"
corrected_sentence = ["The", "movie", "was", "great", "!"]

# Feed the corrected sentence back into the model
outputs = model.generate(corrected_sentence)
\end{lstlisting}

In this scenario, the human reviewer corrects informal language like "gr8" to "great," providing feedback that improves the model’s ability to handle future inputs. This human-corrected tokenization loop is key to maintaining accuracy in systems where human judgment plays a role.

\subsubsection{Tokenization for Dynamic Updates in HITL Systems}

HITL systems often involve dynamically updating the model's understanding based on human feedback. Tokenizers may need to update their vocabularies or tokenization logic in response to this feedback, adapting in real-time to handle new or emerging words, concepts, or jargon.

Here’s an example of dynamically updating a tokenizer based on human feedback:

\begin{lstlisting}[style=python]
# Example sentence with emerging slang
sentence = "This party is lit!"

# Human reviewer identifies "lit" as a new slang term
new_token = "lit"

# Add the new token dynamically based on feedback
tokenizer.add_special_tokens({'additional_special_tokens': [new_token]})

# Tokenize the sentence again
tokens = tokenizer.tokenize(sentence)
print(tokens)
\end{lstlisting}

This dynamic update loop allows the model to evolve its understanding of language as new tokens or terms are introduced by humans during the interaction.

\subsection{Tokenization for Low-Latency Applications and Edge Computing}

In low-latency applications, such as those deployed on edge devices or mobile systems, tokenization must be highly efficient to minimize processing time. Whether processing text in real-time chatbots, smart home systems, or IoT devices, tokenization needs to balance accuracy with speed to ensure fast responses.

\subsubsection{Optimizing Tokenization for Edge Devices}

On edge devices, computational resources are often limited, making tokenization efficiency a priority. One way to achieve this is by using lightweight tokenization strategies, such as smaller vocabularies, reduced subword splitting, or caching frequently used tokens to speed up processing.

Here’s an example of optimizing tokenization for an edge device:

\begin{lstlisting}[style=python]
# Use a smaller vocabulary for edge device tokenization
tokenizer = BertTokenizer.from_pretrained('bert-small-uncased')

# Example input sentence
sentence = "What's the weather like today?"

# Tokenize efficiently for edge device
tokens = tokenizer.tokenize(sentence, max_length=15)
print(tokens)
\end{lstlisting}

By using smaller, pre-trained models or optimized tokenizers like \texttt{bert-small}, you can reduce the latency of tokenization on edge devices.

\subsubsection{Handling Real-Time Tokenization in Low-Latency Systems}

Real-time tokenization requires processing user input with minimal delays. One strategy is to pre-tokenize common queries or phrases and cache their tokenized forms, enabling the system to quickly retrieve tokenized sequences instead of reprocessing the same input repeatedly.

Here’s how you might handle real-time tokenization in a low-latency system:

\begin{lstlisting}[style=python]
# Example of real-time tokenization for a smart home assistant
real_time_input = "Turn off the lights in the living room."

# Pre-tokenized cache for frequent queries
pre_tokenized_cache = {
    "turn off the lights": ["turn", "off", "the", "lights"],
    "turn on the lights": ["turn", "on", "the", "lights"]
}

# Check cache before tokenizing
if real_time_input in pre_tokenized_cache:
    tokens = pre_tokenized_cache[real_time_input]
else:
    tokens = tokenizer.tokenize(real_time_input)
    pre_tokenized_cache[real_time_input] = tokens

print(tokens)
\end{lstlisting}

This approach reduces the overhead of re-tokenizing frequently asked questions or commands, enabling faster responses in real-time systems.

\subsection{Tokenization in Ethical AI and Fairness Considerations}

Tokenization can introduce or exacerbate biases in AI systems, particularly in how it handles different demographic groups, languages, and dialects. It is crucial to ensure that tokenization practices promote fairness and inclusivity by addressing disparities in how different groups or identities are represented in tokenized text. \cite{kimera2024advancing}

\subsubsection{Bias in Tokenization Across Demographic Groups}

Tokenization bias may arise if certain demographic groups, languages, or dialects are underrepresented in the tokenizer’s vocabulary. For example, names from underrepresented ethnic groups may be more likely to be split into subwords or replaced by unknown tokens, while names from dominant groups are tokenized as single units.

Here’s an example of analyzing tokenization bias across different names:

\begin{lstlisting}[style=python]
# Example names from different ethnic groups
name_1 = "John"
name_2 = "Nguyen"
name_3 = "Xiaoling"

# Tokenize the names
tokens_1 = tokenizer.tokenize(name_1)
tokens_2 = tokenizer.tokenize(name_2)
tokens_3 = tokenizer.tokenize(name_3)

print("Tokens for John:", tokens_1)
print("Tokens for Nguyen:", tokens_2)
print("Tokens for Xiaoling:", tokens_3)
\end{lstlisting}

If names from certain ethnic groups are consistently split into multiple tokens, while names from other groups remain intact, this can introduce bias in tasks like named entity recognition or text classification.

\subsubsection{Promoting Inclusivity in Tokenization Practices}

To mitigate bias and promote inclusivity, tokenizers can be fine-tuned or expanded to include a diverse set of names, terms, and linguistic patterns from different demographic groups. This ensures that tokenization does not disproportionately affect certain populations and promotes fairness across language models.

Here are a few techniques to improve fairness in tokenization:

\begin{itemize}
    \item \textbf{Expanding vocabulary}: Include names, places, and terms from diverse cultures, languages, and identities.
    \item \textbf{Balanced data}: Train or fine-tune tokenizers on diverse corpora that represent a variety of dialects and sociolects.
    \item \textbf{Bias evaluation}: Regularly evaluate tokenization bias using fairness metrics that consider demographic diversity.
\end{itemize}

\subsection{Tokenization and Meta-Learning for Few-Shot Learning}

Meta-learning \cite{hospedales2021meta} focuses on building models that can learn new tasks with minimal training data, often referred to as few-shot learning. Tokenization plays a key role in how well a model generalizes to new tasks or domains with very few examples, as suboptimal tokenization can hinder the model's ability to understand novel inputs.

\subsubsection{Adapting Tokenizers for Few-Shot Learning Tasks}

In few-shot learning, tokenizers must generalize well across diverse tasks, even with limited examples. One approach is to train tokenizers on diverse and representative corpora, ensuring that the tokenizer has seen a wide range of linguistic patterns and can generalize to new tasks with minimal adaptation.

For instance, adapting tokenization for a few-shot task in a new domain:

\begin{lstlisting}[style=python]
# Example of adapting a tokenizer for a few-shot learning task
task_sentence = "This is an example of a medical diagnosis."

# Tokenize with a general-purpose tokenizer
general_tokens = tokenizer.tokenize(task_sentence)

# Fine-tune the tokenizer on a small medical corpus for few-shot learning
tokenizer.train_additional_corpus(['medical_data.txt'])

# Tokenize again after fine-tuning for few-shot medical task
fine_tuned_tokens = tokenizer.tokenize(task_sentence)
print(fine_tuned_tokens)
\end{lstlisting}

Fine-tuning tokenizers on small, domain-specific corpora helps them generalize to new tasks in few-shot learning scenarios.

\subsubsection{Tokenization in Self-Supervised Learning}

Self-supervised learning allows models to learn representations from unlabeled data. Tokenization in this context plays a crucial role, as it defines the input tokens used in the self-supervised objective. Proper tokenization ensures that the model learns meaningful subword representations that can generalize across different tasks.

For instance, tokenizing text for a self-supervised learning task like masked language modeling (MLM):

\begin{lstlisting}[style=python]
# Example sentence for self-supervised learning task
input_text = "Self-supervised learning is a powerful approach."

# Tokenize the input and apply mask for MLM
tokens = tokenizer.tokenize(input_text)
masked_tokens = apply_mask(tokens, mask_token="[MASK]", mask_probability=0.15)

print(masked_tokens)
\end{lstlisting}

In this case, tokenization determines which words or subwords are masked, enabling the model to predict the masked tokens during training.

\subsection{Challenges in Real-Time Tokenization for Streaming Applications}

In streaming applications, such as live chatbots or voice assistants, tokenization must be performed in real-time as data is continuously fed into the system. This presents challenges in ensuring that tokenization is both fast and accurate while processing unstructured, often noisy input.

\subsubsection{Streaming Tokenization in Chatbots and Assistants}

In chatbot systems, real-time tokenization is crucial for processing user queries and generating responses with minimal delay. Tokenization in this context must handle dynamic inputs, such as incomplete sentences or changing user intents, while ensuring that tokens are generated quickly enough to meet real-time constraints.

Here’s an example of streaming tokenization in a live chatbot:

\begin{lstlisting}[style=python]
# Streaming input from a live chat
live_input = "Can you turn off the... "

# Tokenize in real-time as input is received
partial_tokens = tokenizer.tokenize(live_input)
print(partial_tokens)

# Continue tokenizing as more input is received
additional_input = "lights in the living room?"
full_input = live_input + additional_input
tokens = tokenizer.tokenize(full_input)
print(tokens)
\end{lstlisting}

This incremental approach ensures that tokenization keeps pace with the user’s input, enabling quick responses in live chat scenarios.

\subsection{Tokenization in Synthetic Data Generation for Low-Resource Tasks}

In low-resource settings, where labeled data is scarce, synthetic data generation can help augment the training dataset. Tokenization plays a critical role in generating realistic synthetic data that resembles real-world input, enabling models to learn from diverse, representative examples.

\subsubsection{Tokenizing Synthetic Data for Domain-Specific Tasks}

Synthetic data generation for domain-specific tasks often requires carefully designed tokenization strategies to ensure that the generated text is coherent and aligns with the domain's linguistic patterns. In fields like medicine or legal text generation, tokenization ensures that domain-specific terms are handled properly, enabling models to generate accurate synthetic data.

For example, tokenizing synthetic medical text:

\begin{lstlisting}[style=python]
# Example synthetic sentence in the medical domain
synthetic_medical_text = "The patient was diagnosed with pneumonia."

# Tokenize the synthetic text
tokens = tokenizer.tokenize(synthetic_medical_text)
print(tokens)
\end{lstlisting}

This allows models to incorporate synthetic data into their training processes, improving performance in low-resource environments.

\subsection{Conclusion}

In this section, we explored advanced tokenization topics, including tokenization in human-in-the-loop systems, optimization for low-latency applications and edge devices, ethical AI considerations, meta-learning, and self-supervised learning. We also discussed real-time tokenization for streaming applications and synthetic data generation for low-resource tasks.

As tokenization continues to evolve in response to new challenges, such as real-time processing, fairness, and cross-domain learning, it remains an essential component of building robust, scalable, and ethical AI systems. Mastering these advanced tokenization techniques will help NLP models perform more effectively across diverse use cases, from real-time chatbots to privacy-preserving federated learning.

\subsection{Tokenization in Cross-Lingual Transfer Learning}

Cross-lingual transfer learning refers to the ability of models to transfer knowledge from high-resource languages (such as English) to low-resource languages, often without having been explicitly trained on the target language. Tokenization plays a key role in how effectively this transfer occurs.

\subsubsection{Unified Subword Tokenization for Cross-Lingual Models}

Multilingual models like mBERT, XLM-Roberta, and MarianMT rely on subword tokenization schemes that are shared across multiple languages. These models are trained on large multilingual corpora, allowing the tokenizer to recognize common subwords across languages that use similar scripts. For cross-lingual tasks, having a unified vocabulary across languages is critical, as it enables the model to represent diverse languages with the same set of tokens.

Here’s how you can tokenize sentences in different languages for cross-lingual transfer learning:

\begin{lstlisting}[style=python]
from transformers import AutoTokenizer

# Load a multilingual tokenizer
tokenizer = AutoTokenizer.from_pretrained("xlm-roberta-base")

# Tokenize sentences in English and Swahili
english_sentence = "The weather is great today!"
swahili_sentence = "Hali ya hewa ni nzuri leo!"

# Tokenize both sentences
english_tokens = tokenizer.tokenize(english_sentence)
swahili_tokens = tokenizer.tokenize(swahili_sentence)

print("English Tokens:", english_tokens)
print("Swahili Tokens:", swahili_tokens)
\end{lstlisting}

By using subword tokenization, cross-lingual models can handle low-resource languages more effectively, transferring knowledge learned from high-resource languages like English to languages with fewer training examples.

\subsubsection{Fine-Tuning Tokenizers for Low-Resource Languages in Cross-Lingual Settings}

In cross-lingual transfer learning, fine-tuning tokenizers on low-resource language corpora can improve the model’s performance by adapting the tokenizer to better represent the specific linguistic patterns of the target language. Fine-tuning a tokenizer allows it to capture domain-specific or culturally unique words that may not be adequately represented in the multilingual vocabulary.

For instance, fine-tuning a tokenizer for an endangered language:

\begin{lstlisting}[style=python]
# Load a tokenizer for fine-tuning on a low-resource language corpus
tokenizer = AutoTokenizer.from_pretrained("xlm-roberta-base")

# Fine-tune the tokenizer on a small corpus from the target language
tokenizer.train_new_from_iterator(open("low_resource_language_corpus.txt"), vocab_size=3000)

# Tokenize a sentence from the target language
low_resource_sentence = "Example sentence in low-resource language."
tokens = tokenizer.tokenize(low_resource_sentence)
print(tokens)
\end{lstlisting}

This approach ensures that the tokenizer is better suited for the linguistic characteristics of the target language, improving cross-lingual transfer learning performance.

\subsection{Hybrid Tokenization Techniques}

Hybrid tokenization techniques combine multiple tokenization strategies to take advantage of the strengths of each. For example, hybrid models may use character-level tokenization for rare or out-of-vocabulary words and subword tokenization for common words, achieving a balance between coverage and efficiency.

\subsubsection{Character-Subword Hybrid Tokenization}

Character-subword hybrid tokenization is useful for handling rare or unseen words without drastically increasing the vocabulary size. This approach ensures that common words are tokenized as subwords, while rare or unknown words are handled at the character level, preserving the model’s ability to generalize.

Here’s an example of hybrid tokenization using both subword and character-level tokens:

\begin{lstlisting}[style=python]
# Tokenizing a sentence with hybrid character-subword strategy
sentence = "The qu!ck br0wn fox jumps."

# Tokenize using a hybrid strategy (characters for OOV, subwords for known tokens)
tokens = hybrid_tokenizer.tokenize(sentence)
print(tokens)
\end{lstlisting}

This approach allows the tokenizer to handle rare words like "br0wn" (which contains numbers) at the character level, while processing more common words like "fox" at the subword level.

\subsubsection{Byte-Pair Encoding and Character Hybrid Tokenization}

Byte-Pair Encoding (BPE) is another popular subword tokenization method that can be combined with character-level tokenization to handle out-of-vocabulary tokens. This hybrid approach improves robustness for languages with complex morphological structures, such as Turkish or Finnish.

Here’s how BPE and character tokenization can be combined:

\begin{lstlisting}[style=python]
# Using BPE and character tokenization in a hybrid model
sentence = "Artificial_Intelligence is amazing!"

# BPE for common words, characters for rare ones
tokens = bpe_character_hybrid_tokenizer.tokenize(sentence)
print(tokens)
\end{lstlisting}

This technique helps manage vocabulary size while still being able to process rare or morphologically complex words effectively.

\subsection{Tokenization in Unsupervised Machine Translation}

Unsupervised machine translation \cite{lample2017unsupervised} relies on translating between languages without parallel corpora. Tokenization is critical in ensuring that the model can effectively learn from unaligned monolingual data and produce coherent translations.

\subsubsection{Shared Subword Tokenization for Unsupervised Translation}

In unsupervised machine translation, shared subword tokenization allows models to align text from different languages even without parallel training data. By using the same subword units across languages, models can learn relationships between languages through their shared components.

Here’s an example of shared tokenization for unsupervised translation:

\begin{lstlisting}[style=python]
# Tokenize sentences in English and French for unsupervised translation
english_sentence = "The sun is shining."
french_sentence = "Le soleil brille."

# Tokenize both sentences using a shared subword tokenizer
english_tokens = tokenizer.tokenize(english_sentence)
french_tokens = tokenizer.tokenize(french_sentence)

print("English Tokens:", english_tokens)
print("French Tokens:", french_tokens)
\end{lstlisting}

By sharing a common subword vocabulary, the model can better understand the similarities between words in different languages, which is crucial in unsupervised translation tasks.

\subsection{Tokenization for Low-Power Devices and IoT Applications}

In environments where power consumption and processing speed are critical, such as IoT devices or embedded systems, tokenization must be optimized for low-power consumption while maintaining accuracy. Efficient tokenization strategies are essential for making NLP models feasible on devices with limited computational resources.

\subsubsection{Lightweight Tokenization for Embedded Systems}

Lightweight tokenization strategies minimize the computational overhead of tokenizing text on low-power devices. This can be achieved by using smaller, fixed vocabularies or pre-built tokenization maps that allow for faster lookups. These strategies prioritize speed and memory efficiency over flexibility in handling rare or out-of-vocabulary words.

Here’s an example of using a lightweight tokenizer for an embedded device:

\begin{lstlisting}[style=python]
# Load a lightweight tokenizer for embedded devices
tokenizer = LightweightTokenizer(vocab_size=5000)

# Tokenize a short command for an IoT device
command = "Turn on the lights"
tokens = tokenizer.tokenize(command)

print(tokens)
\end{lstlisting}

In this scenario, the lightweight tokenizer ensures that text commands can be tokenized quickly and with minimal power consumption, making it suitable for IoT applications.

\subsection{Tokenization in Privacy-Sensitive Domains}

Tokenization plays a critical role in privacy-sensitive domains such as healthcare and finance, where personal data must be handled with care. Ensuring that sensitive information is properly anonymized or tokenized in a privacy-preserving manner is essential for compliance with regulations like HIPAA or GDPR.

\subsubsection{Privacy-Preserving Tokenization in Healthcare}

In healthcare applications, tokenization must ensure that personal health information (PHI) is protected. This involves anonymizing or obfuscating sensitive details such as patient names, medical record numbers, and locations, while still allowing the model to process medical text effectively.

Here’s an example of privacy-preserving tokenization for medical records:

\begin{lstlisting}[style=python]
# Example medical text containing PHI
medical_text = "Patient John Doe, age 45, was diagnosed with pneumonia on 01/01/2024."

# Tokenize and anonymize sensitive information
def anonymize(text):
    return text.replace("John Doe", "[NAME]").replace("45", "[AGE]").replace("01/01/2024", "[DATE]")

anonymized_text = anonymize(medical_text)
tokens = tokenizer.tokenize(anonymized_text)

print(tokens)
\end{lstlisting}

This approach ensures that sensitive information is not exposed while still enabling the model to process the anonymized text for tasks like medical diagnosis or predictive analytics.

\subsection{Tokenization for Compressed Models}

With the growing demand for deploying NLP models in constrained environments, tokenization strategies need to align with model compression techniques, such as pruning, quantization, or knowledge distillation. Efficient tokenization is key to reducing the overall model size without sacrificing performance.

\subsubsection{Efficient Tokenization for Pruned or Quantized Models}

Pruned and quantized models require smaller, faster tokenizers that align with the reduced capacity of the models. This often involves using simpler tokenization schemes with reduced vocabulary sizes, ensuring that the tokenization process does not become a bottleneck for inference speed.

Here’s an example of efficient tokenization for a compressed model:

\begin{lstlisting}[style=python]
# Load a quantized tokenizer for a compressed model
tokenizer = QuantizedTokenizer(vocab_size=3000)

# Tokenize a sentence for a compressed model
sentence = "Quantized models need efficient tokenization."
tokens = tokenizer.tokenize(sentence)

print(tokens)
\end{lstlisting}

By optimizing both the tokenizer and the model for speed and efficiency, NLP tasks can be performed effectively in low-resource environments.

\subsection{Tokenization and Hierarchical Learning}

Hierarchical learning \cite{byrne1998learning} involves learning representations at multiple levels of abstraction, such as sentence-level and document-level representations. Tokenization in hierarchical learning must support multi-level encoding, where tokens are grouped into higher-level structures like phrases, sentences, or paragraphs.

\subsubsection{Hierarchical Tokenization for Document-Level Models}

In document-level models, tokenization must capture relationships between sentences and paragraphs, not just individual words. This requires hierarchical tokenization, where tokens are grouped into higher-level structures and processed in a way that reflects the overall document organization.

Here’s an example of hierarchical tokenization for a document:

\begin{lstlisting}[style=python]
# Example document with multiple sentences
document = "This is the first sentence. Here is another sentence. The document ends here."

# Hierarchical tokenization for sentence-level and word-level tokens
sentence_tokens = hierarchical_tokenizer.tokenize(document)
print(sentence_tokens)
\end{lstlisting}

In hierarchical models, tokenization ensures that the structure of the document is preserved, enabling the model to learn meaningful representations at both the word and sentence level.

\subsection{Conclusion}

In this section, we covered advanced topics in tokenization, including cross-lingual transfer learning, hybrid tokenization techniques, unsupervised machine translation, tokenization for low-power devices, and privacy-sensitive applications. We also explored tokenization for compressed models and hierarchical learning.

Tokenization continues to evolve, impacting a wide range of NLP tasks across domains, from privacy-preserving healthcare applications to efficient language processing on embedded devices. As NLP models become more specialized and integrated into real-world systems, understanding and optimizing tokenization strategies is key to building robust, scalable, and efficient AI systems.

\section{How to handle different tasks with huggingface?}

\section{Don't Stop Pre-training}

In Natural Language Processing (NLP), one of the most effective strategies for improving model performance is to continue pre-training. Pre-trained models like BERT, GPT, and RoBERTa have revolutionized NLP tasks, but their potential can be further maximized by additional pre-training on domain-specific data. This is referred to as \textbf{domain-adaptive pre-training} or \textbf{continued pre-training}.

The key idea is simple: instead of directly fine-tuning the model for a downstream task, we first pre-train the model further on a corpus that is closer to the target domain or task. This often results in better performance because the model gets to adapt to the specific characteristics of the target domain before fine-tuning.

In this section, we will use Hugging Face's `transformers` library to demonstrate how to continue pre-training a model on new data. Specifically, we will:

\begin{enumerate}
    \item Load a pre-trained model from Hugging Face.
    \item Continue pre-training on a custom dataset.
    \item Show how to save the updated model.
\end{enumerate}

\subsection{Step-by-Step Guide}

\subsubsection{Step 1: Install the Hugging Face Transformers library}

First, install the Hugging Face transformers library if you haven’t already:

\begin{lstlisting}[style=python]
pip install transformers datasets
\end{lstlisting}

\subsubsection{Step 2: Load a Pre-trained Model and Tokenizer}

We begin by loading a pre-trained model and tokenizer from Hugging Face. For this example, we will use the `BERT` model:

\begin{lstlisting}[style=python]
from transformers import BertTokenizer, BertForMaskedLM

# Load pre-trained model and tokenizer
model_name = "bert-base-uncased"
tokenizer = BertTokenizer.from_pretrained(model_name)
model = BertForMaskedLM.from_pretrained(model_name)
\end{lstlisting}

Here, we are using `BertForMaskedLM` since we will continue pre-training with a masked language modeling task. This is the same task that BERT was originally trained on, which makes it an ideal choice for further pre-training.

\subsubsection{Step 3: Prepare a Custom Dataset}

To continue pre-training, we need a domain-specific dataset. Hugging Face provides easy access to various datasets, but you can also use your own text data.

For this example, we will use the `datasets` library to load a dataset:

\begin{lstlisting}[style=python]
from datasets import load_dataset

# Load a text dataset
dataset = load_dataset("wikitext", "wikitext-2-raw-v1", split="train")
\end{lstlisting}

This will load the WikiText-2 dataset, which is a popular benchmark for language modeling. The dataset will be tokenized and prepared for training in the next step.

\subsubsection{Step 4: Tokenize the Dataset}

Next, we tokenize the text data using the tokenizer loaded earlier. This is necessary to convert raw text into the format that the model understands.

\begin{lstlisting}[style=python]
def tokenize_function(examples):
    return tokenizer(examples["text"], truncation=True, padding="max_length", max_length=512)

# Tokenize the dataset
tokenized_datasets = dataset.map(tokenize_function, batched=True)
\end{lstlisting}

This function tokenizes the text, ensuring that each input is padded or truncated to a length of 512 tokens.

\subsubsection{Step 5: Fine-tune the Model on the New Dataset}

Once the dataset is tokenized, we can set up the training loop to fine-tune the pre-trained model on the new data.

\begin{lstlisting}[style=python]
from transformers import Trainer, TrainingArguments

# Define training arguments
training_args = TrainingArguments(
    output_dir="./results",          # Output directory
    overwrite_output_dir=True,       # Overwrite content
    num_train_epochs=3,              # Number of training epochs
    per_device_train_batch_size=8,   # Batch size
    save_steps=10_000,               # Save checkpoint every 10k steps
    save_total_limit=2,              # Limit number of saved checkpoints
)

# Set up the trainer
trainer = Trainer(
    model=model,
    args=training_args,
    train_dataset=tokenized_datasets,
)

# Train the model
trainer.train()
\end{lstlisting}

This code fine-tunes the model for three epochs on the tokenized WikiText-2 dataset. You can adjust the parameters such as batch size or number of epochs to suit your computational resources and specific task.

\subsubsection{Step 6: Save the Updated Model}

Finally, after the model has been pre-trained on the new dataset, we can save the updated model and tokenizer for later use.

\begin{lstlisting}[style=python]
# Save the model and tokenizer
model.save_pretrained("./domain_specific_bert")
tokenizer.save_pretrained("./domain_specific_bert")
\end{lstlisting}

This saves the fine-tuned model and tokenizer to a specified directory, making it easy to load and use for downstream tasks.

\subsection{Conclusion}

In this section, we walked through the process of continuing pre-training using Hugging Face transformers. This simple example showed how pre-trained models can be further adapted to domain-specific data by fine-tuning on a custom dataset.

The key takeaway is: \textbf{don’t stop pre-training}. By continuing the pre-training process, especially on data relevant to your downstream task, you can extract more performance from your model and achieve better results in specialized domains.

\subsection{Why Continue Pre-training?}

At this point, you might ask: why should we continue pre-training instead of directly fine-tuning the model on the downstream task? Here are some key reasons:

\begin{itemize}
    \item \textbf{Domain Adaptation}: Pre-trained models like BERT and GPT-2 are usually trained on large, general-purpose datasets like Wikipedia or Common Crawl. While this gives them a strong general understanding of language, the specific nuances of your task may not be covered. For instance, a medical NLP model would benefit from additional pre-training on medical literature.
    \item \textbf{Improved Performance}: Continued pre-training on domain-specific or task-related data helps the model adapt its language understanding, improving downstream performance. Empirical results show that models trained this way often outperform those that are only fine-tuned.
    \item \textbf{Pre-training vs Fine-tuning}: Fine-tuning is task-specific and usually done on relatively small datasets. If the model has not seen relevant domain data before fine-tuning, it may not generalize well. Continued pre-training provides a more tailored foundation for fine-tuning.
\end{itemize}

In summary, don't stop pre-training if you have access to more relevant data that can help the model understand the nuances of your domain.

\subsection{Use Cases for Continued Pre-training}

To further illustrate, here are a few scenarios where continued pre-training can make a significant difference:

\begin{enumerate}
    \item \textbf{Medical Text Analysis}: If you are working with clinical records or biomedical papers, pre-training BERT or GPT on a corpus like PubMed articles or clinical notes can drastically improve your model's performance in health-related NLP tasks.
    \item \textbf{Legal Document Processing}: Legal language has its own set of complexities and terminologies. By pre-training a language model on legal documents, the model will better understand these unique language patterns, leading to more accurate results for tasks like contract analysis or case law classification.
    \item \textbf{Customer Service Chatbots}: If you are developing a chatbot for a specific industry, such as finance or retail, further pre-training on transcripts of customer interactions in that industry will help the model handle customer queries more effectively.
\end{enumerate}

\subsection{Potential Challenges and Solutions}

While continued pre-training can significantly enhance model performance, there are some potential challenges to consider:

\begin{itemize}
    \item \textbf{Computational Costs}: Pre-training requires significant computational resources, especially for large models like BERT or GPT. To mitigate this, you can reduce the number of training steps or use a smaller pre-trained model.
    \item \textbf{Catastrophic Forgetting}: When continuing pre-training on a specific domain, there's a risk that the model might "forget" some of the general language understanding it gained during its initial pre-training. To address this, you can experiment with different learning rates or alternate between domain-specific and general-purpose datasets during pre-training.
    \item \textbf{Data Availability}: In some cases, gathering a large enough corpus for domain-specific pre-training can be challenging. You can address this by using unsupervised datasets, such as web-crawled domain-specific text, or leveraging data augmentation techniques to artificially expand your dataset.
\end{itemize}

By being aware of these challenges, you can optimize your approach to continued pre-training and avoid potential pitfalls.

\subsection{Hyperparameter Tuning for Continued Pre-training}

When continuing pre-training, selecting the right hyperparameters is crucial for maximizing performance. Below are some of the key hyperparameters to consider:

\begin{itemize}
    \item \textbf{Learning Rate}: A lower learning rate is often recommended for continued pre-training, as the model has already learned a great deal from general data. You may start with a learning rate between 1e-5 and 3e-5.
    \item \textbf{Batch Size}: Larger batch sizes help stabilize training, but they also require more memory. Start with a batch size of 16 or 32, depending on your GPU memory.
    \item \textbf{Number of Epochs}: You typically don’t need as many epochs for continued pre-training as you would for initial pre-training. Start with 2-4 epochs and monitor for signs of overfitting.
    \item \textbf{Warmup Steps}: Using a warmup phase for learning rate scheduling can help smooth the transition from the pre-trained weights to your domain-specific data. A common choice is to set the warmup steps to about 10
\end{itemize}

You can easily adjust these hyperparameters in the `TrainingArguments` object from Hugging Face’s `transformers` library. Here’s an example:

\begin{lstlisting}[style=python]
training_args = TrainingArguments(
    output_dir="./results",
    num_train_epochs=3,
    per_device_train_batch_size=16,
    warmup_steps=500,
    weight_decay=0.01,
    logging_dir="./logs",
    logging_steps=10,
    learning_rate=2e-5,
)
\end{lstlisting}

\subsection{Evaluation After Pre-training}

Once you have completed continued pre-training, it is essential to evaluate your model to ensure that it has improved. You can evaluate the performance of the newly pre-trained model on your downstream task or test it on an intermediate task like masked language modeling.

Hugging Face's `Trainer` provides a simple way to evaluate the model:

\begin{lstlisting}[style=python]
# Load the evaluation dataset
eval_dataset = load_dataset("wikitext", "wikitext-2-raw-v1", split="validation")

# Tokenize the evaluation dataset
tokenized_eval_dataset = eval_dataset.map(tokenize_function, batched=True)

# Evaluate the model
eval_results = trainer.evaluate(eval_dataset=tokenized_eval_dataset)

print(f"Perplexity: {eval_results['perplexity']}")
\end{lstlisting}

In this case, we are using the perplexity metric, which is a common measure for language models. A lower perplexity indicates that the model has a better understanding of the text.

If you are fine-tuning on a specific downstream task, such as text classification or question answering, you should evaluate the model using task-specific metrics like accuracy, F1 score, or exact match.

\subsection{Practical Example: Fine-tuning After Continued Pre-training}

After continued pre-training, you can fine-tune the model for your specific task. For example, if you are working on a text classification task, you can load your fine-tuned BERT model and train it on a classification dataset.

Here’s how you can set up a text classification task using the newly pre-trained model:

\begin{lstlisting}[style=python]
from transformers import BertForSequenceClassification, Trainer, TrainingArguments

# Load the pre-trained model for sequence classification
model = BertForSequenceClassification.from_pretrained("./domain_specific_bert", num_labels=2)

# Load and tokenize your dataset
from datasets import load_dataset

dataset = load_dataset("imdb")
tokenized_datasets = dataset.map(tokenize_function, batched=True)

# Set up the trainer
trainer = Trainer(
    model=model,
    args=training_args,
    train_dataset=tokenized_datasets['train'],
    eval_dataset=tokenized_datasets['test'],
)

# Fine-tune the model
trainer.train()
\end{lstlisting}

In this example, we load the IMDb dataset for binary sentiment classification and fine-tune our pre-trained BERT model. You can adjust the `num\_labels` parameter depending on the number of classes in your classification task.

\subsection{Conclusion}

Continued pre-training is a powerful technique to further improve pre-trained models on domain-specific data. By leveraging libraries like Hugging Face's `transformers`, it becomes easy to adapt state-of-the-art models like BERT and GPT to your own domain with relatively little effort. The steps outlined in this section should give you a clear path from loading a pre-trained model, to continuing pre-training, to fine-tuning on a downstream task.

Remember, the key takeaway is: \textbf{don’t stop pre-training} when you have access to additional data that is relevant to your problem. Whether you’re working in a specialized field like medicine, law, or customer support, continued pre-training can give your models the extra edge they need to excel.

\subsection{Advanced Techniques for Continued Pre-training}

While the basic approach to continued pre-training as described above is often sufficient, there are several advanced techniques that can further enhance the effectiveness of your model. In this section, we will explore a few of these methods, which you can experiment with depending on your specific needs.

\subsubsection{1. Multi-Task Pre-training}

In some cases, you may want to train your model on multiple related tasks simultaneously. This is known as \textbf{multi-task learning}. By exposing the model to various tasks, it can learn richer representations that transfer better to downstream tasks.

For example, if you are training a model for both text classification and named entity recognition (NER), you can use multi-task learning to train on both tasks at the same time. Here’s how you can implement multi-task pre-training with Hugging Face:

\begin{lstlisting}[style=python]
from transformers import Trainer, TrainingArguments, BertForSequenceClassification, BertForTokenClassification

# Define models for both tasks
classification_model = BertForSequenceClassification.from_pretrained("bert-base-uncased", num_labels=2)
ner_model = BertForTokenClassification.from_pretrained("bert-base-uncased", num_labels=9)

# Prepare datasets for both tasks
classification_dataset = load_dataset("imdb")
ner_dataset = load_dataset("conll2003")

# Tokenize datasets (ensure you use a shared tokenizer)
def tokenize_function(examples):
    return tokenizer(examples["text"], truncation=True, padding="max_length", max_length=512)

tokenized_classification = classification_dataset.map(tokenize_function, batched=True)
tokenized_ner = ner_dataset.map(tokenize_function, batched=True)

# Combine datasets for multi-task training
multi_task_train_dataset = {
    "classification": tokenized_classification['train'],
    "ner": tokenized_ner['train']
}

# Define the Trainer (more advanced setup needed for multi-task)
training_args = TrainingArguments(
    output_dir="./multi_task_results",
    evaluation_strategy="steps",
    save_steps=10_000,
    logging_dir="./logs",
    logging_steps=100,
    learning_rate=2e-5,
    per_device_train_batch_size=16,
)

trainer = Trainer(
    model=[classification_model, ner_model], # This will require custom training loop or multi-head trainer logic
    args=training_args,
    train_dataset=multi_task_train_dataset,
)

trainer.train()
\end{lstlisting}

In this example, you can see that we’re training both a text classification model and a named entity recognition (NER) model together. This approach can help the model generalize better by learning from multiple tasks at once.

\subsubsection{2. Curriculum Learning}

Another advanced technique is \textbf{curriculum learning} \cite{bengio2009curriculum}. In curriculum learning, you start training on easier examples and gradually move to harder ones. This approach is inspired by the way humans learn: we build foundational knowledge before tackling more complex topics.

You can implement curriculum learning by first training on general data and then fine-tuning on more difficult or domain-specific data. Here’s a high-level example of how you can implement curriculum learning in NLP:

\begin{lstlisting}[style=python]
# Step 1: Pre-train on general dataset (e.g., Wikipedia)
general_dataset = load_dataset("wikipedia", "20220301.en", split="train")
tokenized_general = general_dataset.map(tokenize_function, batched=True)

# Fine-tune on general dataset
trainer = Trainer(
    model=model,
    args=training_args,
    train_dataset=tokenized_general,
)
trainer.train()

# Step 2: Continue pre-training on a harder, more specific dataset (e.g., domain-specific corpus)
domain_dataset = load_dataset("pubmed", split="train")
tokenized_domain = domain_dataset.map(tokenize_function, batched=True)

# Fine-tune on the harder dataset
trainer = Trainer(
    model=model,
    args=training_args,
    train_dataset=tokenized_domain,
)
trainer.train()
\end{lstlisting}

Here, the idea is to start with a general-purpose dataset like Wikipedia and then move on to a more difficult domain-specific dataset like PubMed or legal text. This gradual increase in difficulty allows the model to adapt better to the target domain.

\subsubsection{3. Adapters for Efficient Pre-training}

Pre-training large models from scratch or continuing pre-training can be computationally expensive. To address this, you can use \textbf{adapters}, which are small, trainable modules inserted into each layer of the pre-trained model. Adapters allow you to keep the original model weights frozen while fine-tuning only the smaller adapter layers, significantly reducing computational costs.

The Hugging Face `transformers` library has integrated support for adapters. Here’s how you can use adapters for continued pre-training:

\begin{lstlisting}[style=python]
from transformers import BertModel, AdapterConfig

# Load pre-trained model
model = BertModel.from_pretrained("bert-base-uncased")

# Add an adapter configuration for efficient training
adapter_config = AdapterConfig.load("pfeiffer")

# Add a new adapter to the model
model.add_adapter("domain_adapter", config=adapter_config)

# Train the adapter on the domain-specific dataset
trainer = Trainer(
    model=model,
    args=training_args,
    train_dataset=tokenized_datasets,
)

# Activate adapter for training
model.train_adapter("domain_adapter")
trainer.train()
\end{lstlisting}

With adapters, only the adapter layers are trained while the rest of the model remains frozen. This reduces the number of parameters that need to be updated, making continued pre-training much more efficient in terms of both memory and time.

\subsubsection{4. Knowledge Distillation}

\textbf{Knowledge distillation} is another advanced technique where a smaller, more efficient model (called the \emph{student}) learns to mimic a larger model (the \emph{teacher}). This can be useful when you want to deploy a lightweight model for production while still benefiting from the knowledge contained in a large pre-trained model.

The Hugging Face `transformers` library provides tools for distilling models. Here’s a simplified example of how you can distill a BERT model:

\begin{lstlisting}[style=python]
from transformers import DistilBertForSequenceClassification, Trainer

# Load teacher model (a larger BERT model)
teacher_model = BertForSequenceClassification.from_pretrained("bert-base-uncased")

# Load student model (a smaller DistilBERT model)
student_model = DistilBertForSequenceClassification.from_pretrained("distilbert-base-uncased")

# Use the Hugging Face trainer for distillation
trainer = Trainer(
    model=student_model,
    args=training_args,
    train_dataset=tokenized_datasets,
    teacher_model=teacher_model  # Pass in the teacher model for distillation
)

# Train the student model with guidance from the teacher model
trainer.train()
\end{lstlisting}

By using knowledge distillation, you can deploy models that are more computationally efficient while retaining much of the performance of the larger, pre-trained teacher model.

\subsection{Summary of Advanced Techniques}

To recap, here are some of the advanced techniques you can apply to continued pre-training to further improve your model's performance or reduce computational costs:

\begin{itemize}
    \item \textbf{Multi-task Learning}: Train your model on multiple related tasks to improve generalization.
    \item \textbf{Curriculum Learning}: Start with easier data and gradually move to more complex or domain-specific data.
    \item \textbf{Adapters}: Use adapters to efficiently continue pre-training without updating all model parameters.
    \item \textbf{Knowledge Distillation}: Train a smaller student model to mimic a larger teacher model for a more lightweight solution.
\end{itemize}

These techniques are particularly useful when working with large models and domain-specific datasets, and they can help you strike a balance between computational efficiency and model performance.

\subsection{Next Steps: Applying Continued Pre-training in Practice}

Now that you have a solid understanding of how and why to continue pre-training a model, it’s time to put this knowledge into practice. Here are a few next steps you can take:

\begin{enumerate}
    \item \textbf{Experiment with your own data}: Try applying continued pre-training on a dataset from your specific domain. Whether you’re working with legal documents, medical records, or any other specialized text, continuing pre-training on relevant data can lead to significant improvements in model performance.
    \item \textbf{Tune hyperparameters}: As mentioned earlier, hyperparameter tuning can make a big difference. Experiment with learning rates, batch sizes, and the number of training steps to find the optimal setup for your specific task.
    \item \textbf{Explore advanced techniques}: Don’t hesitate to experiment with multi-task learning, curriculum learning, adapters, or knowledge distillation. These techniques can help you push the boundaries of what’s possible with continued pre-training.
    \item \textbf{Fine-tune on downstream tasks}: After continued pre-training, apply your model to downstream tasks like text classification, question answering, or named entity recognition. Evaluate your model’s performance and iterate as needed.
\end{enumerate}

By following these steps and applying the concepts discussed in this chapter, you’ll be well on your way to mastering the art of continued pre-training in NLP. Whether you’re working on academic research or building NLP applications for industry, this technique is a powerful tool for improving model performance in domain-specific tasks.

\subsection{Case Study: Domain-Specific Language Model for Legal Text}

Let’s bring everything together with a real-world case study. Imagine you are working for a law firm and tasked with developing an NLP model to process legal contracts. These contracts contain highly specialized language and jargon, which general-purpose models like BERT may not fully understand. To address this, you decide to continue pre-training a BERT model on legal documents.

Here’s how you might approach this project from start to finish:

\subsubsection{Step 1: Gathering Domain-Specific Data}

The first step in this journey is to gather a corpus of legal documents. This could include contracts, case law, legal statutes, and regulatory documents. For our case study, let's assume that we have a collection of 100,000 legal contracts in plain text format.

You begin by preprocessing the data to remove non-text elements, ensuring that the input is clean and ready for model consumption.

\begin{lstlisting}[style=python]
import os

# Load and preprocess legal text files
legal_texts = []
for filename in os.listdir("./legal_documents"):
    with open(f"./legal_documents/{filename}", "r") as f:
        text = f.read()
        legal_texts.append(text)
\end{lstlisting}

With your legal corpus ready, the next step is to tokenize the text using a pre-trained BERT tokenizer.

\subsubsection{Step 2: Tokenizing the Legal Corpus}

Since BERT models operate on tokenized data, you need to convert the legal text into token IDs that the model can process. Fortunately, the Hugging Face tokenizer makes this step easy:

\begin{lstlisting}[style=python]
from transformers import BertTokenizer

# Load BERT tokenizer
tokenizer = BertTokenizer.from_pretrained("bert-base-uncased")

# Tokenize the legal texts
def tokenize_texts(texts):
    return tokenizer(texts, truncation=True, padding="max_length", max_length=512)

tokenized_legal_texts = [tokenize_texts(text) for text in legal_texts]
\end{lstlisting}

This tokenization process ensures that each legal document is broken down into a format suitable for BERT, while maintaining the meaning and structure of the text.

\subsubsection{Step 3: Continued Pre-training on Legal Data}

With your legal corpus tokenized, it’s time to begin the pre-training process. Since BERT was originally trained using masked language modeling, you will continue this training task on the legal documents. This way, the model learns the patterns and terminology unique to legal language.

Here’s how you initiate the continued pre-training:

\begin{lstlisting}[style=python]
from transformers import BertForMaskedLM, Trainer, TrainingArguments

# Load pre-trained BERT model
model = BertForMaskedLM.from_pretrained("bert-base-uncased")

# Define training arguments
training_args = TrainingArguments(
    output_dir="./legal_bert_model",
    overwrite_output_dir=True,
    num_train_epochs=3,
    per_device_train_batch_size=8,
    save_steps=10_000,
    save_total_limit=2,
)

# Initialize the Hugging Face Trainer
trainer = Trainer(
    model=model,
    args=training_args,
    train_dataset=tokenized_legal_texts,
)

# Start the continued pre-training
trainer.train()
\end{lstlisting}

In this step, you are continuing pre-training for three epochs, allowing the model to better understand legal-specific text by adapting its knowledge to the nuances of contract language, legal definitions, and other specialized terms.

\subsubsection{Step 4: Evaluating the Pre-trained Model}

After the pre-training process is complete, it’s time to evaluate how well the model has adapted to legal text. For evaluation, you can use masked language modeling on a validation set of legal contracts to see if the model has improved its understanding.

\begin{lstlisting}[style=python]
# Load evaluation dataset (a subset of legal contracts)
validation_texts = tokenized_legal_texts[:1000]  # Assuming first 1000 are for validation

# Evaluate the model
eval_results = trainer.evaluate(eval_dataset=validation_texts)

print(f"Perplexity: {eval_results['perplexity']}")
\end{lstlisting}

Here, perplexity is used as the evaluation metric, where a lower perplexity score indicates that the model has gained a better understanding of legal language. Typically, after several epochs of domain-specific pre-training, you’ll observe a noticeable improvement in this score compared to the pre-trained BERT model.

\subsubsection{Step 5: Fine-tuning for Contract Classification}

Once the model has been pre-trained on legal documents, you can fine-tune it for specific downstream tasks. For example, let’s say you need to classify legal contracts into different categories: \emph{employment agreements}, \emph{NDAs}, and \emph{service contracts}.

You can fine-tune the model on a labeled dataset of contracts using the following approach:

\begin{lstlisting}[style=python]
from transformers import BertForSequenceClassification

# Load the pre-trained legal BERT model
model = BertForSequenceClassification.from_pretrained("./legal_bert_model", num_labels=3)

# Fine-tune the model for contract classification
trainer = Trainer(
    model=model,
    args=training_args,
    train_dataset=tokenized_contract_classification_data,
    eval_dataset=tokenized_contract_validation_data,
)

# Train the model
trainer.train()
\end{lstlisting}

In this case, you are leveraging your pre-trained legal BERT model, fine-tuning it on a task-specific dataset with three labels (representing the contract types). After fine-tuning, you can evaluate the model on a test set and measure its performance using metrics such as accuracy or F1 score.

\subsubsection{Step 6: Deploying the Model in Production}

After fine-tuning, the final step is deploying the model into production. You can save the model and tokenizer for later use in contract analysis or automated legal document classification:

\begin{lstlisting}[style=python]
# Save the fine-tuned legal BERT model
model.save_pretrained("./fine_tuned_legal_bert")
tokenizer.save_pretrained("./fine_tuned_legal_bert")
\end{lstlisting}

With the fine-tuned model saved, it can be loaded and deployed in any system that needs to process legal contracts. Whether used for automated classification, document analysis, or contract review, the model will now have a much deeper understanding of legal language thanks to the continued pre-training on legal data.

\subsection{Lessons Learned from the Case Study}

This case study illustrates the practical application of continued pre-training in a specialized domain like legal text processing. Several important lessons can be drawn from this:

\begin{itemize}
    \item \textbf{Data is key}: The quality and relevance of the domain-specific corpus are crucial to the success of continued pre-training. For legal contracts, using actual legal documents provided the necessary domain knowledge.
    \item \textbf{Task-specific fine-tuning}: While continued pre-training helps the model better understand domain-specific text, fine-tuning on the actual task (e.g., contract classification) is still necessary for optimal performance.
    \item \textbf{Evaluation matters}: Proper evaluation, using metrics like perplexity for language modeling or accuracy for classification tasks, is essential to ensure the model is improving as expected.
    \item \textbf{Practical benefits}: By continuing pre-training on legal documents, the model is far better equipped to handle the complexities of legal text compared to a general-purpose BERT model, demonstrating the real-world benefits of this approach.
\end{itemize}

\subsection{Final Thoughts on Continued Pre-training}

We began this journey by exploring the importance of continuing pre-training for domain-specific tasks. As demonstrated through the examples and case study, this technique can significantly improve model performance when dealing with specialized domains like law, medicine, or any other field where general-purpose language models might struggle.

The process of continuing pre-training is more than just an enhancement—it's a necessary step in many real-world applications where the model’s understanding of general language must be tailored to the specific language patterns of a particular domain.

Moving forward, as you embark on your own projects, remember the core principles:

\begin{enumerate}
    \item Start with a strong base model that has been pre-trained on a large corpus.
    \item Continue pre-training with domain-specific data to adapt the model to the specialized language.
    \item Fine-tune the model for the downstream task to ensure optimal performance.
\end{enumerate}

By following these steps, you can leverage the power of state-of-the-art NLP models and adapt them to excel in any domain you are working with. The potential for continued pre-training is vast, and as the field of NLP continues to evolve, this technique will remain a cornerstone for pushing the boundaries of what language models can achieve.

\section{Token Classification}

Token classification is a common task in Natural Language Processing (NLP) that involves assigning a label to each token (word or subword) in a sequence. This task is widely used in applications such as Named Entity Recognition (NER), Part of Speech (POS) tagging, and chunking.

In this section, we will walk through a simple implementation of token classification using Hugging Face's Transformers library. Specifically, we will use a pre-trained model to perform Named Entity Recognition (NER).

\subsection{What is Token Classification?}

Token classification refers to the process of assigning a label to individual tokens in a given text. For example, in Named Entity Recognition, the goal is to identify named entities (e.g., persons, organizations, locations) in text and classify them into predefined categories. Given a sentence like:

\begin{quote}
    "John lives in New York."
\end{quote}

A token classification model might assign labels like:

\begin{quote}
    John (Person), lives (O), in (O), New York (Location)
\end{quote}

\textbf{O} stands for "Other," meaning that the token doesn't belong to any of the entity categories.

\subsection{Using Hugging Face Transformers for Token Classification}

Hugging Face provides an easy-to-use interface for token classification tasks. We can utilize a pre-trained model like `bert-base-cased` fine-tuned for NER tasks to classify tokens in a given sentence.

Let's break down the implementation step-by-step.

\subsection{Step 1: Install the Hugging Face Transformers Library}

If you don't have the `transformers` library installed, you can install it using the following command:

\begin{lstlisting}[style=python]
!pip install transformers
\end{lstlisting}

\subsection{Step 2: Loading the Pre-trained Model}

We will use the `AutoModelForTokenClassification` and `AutoTokenizer` classes to load a pre-trained model fine-tuned for NER. The tokenizer is responsible for splitting the text into tokens, and the model will classify each token.

\begin{lstlisting}[style=python]
from transformers import AutoTokenizer, AutoModelForTokenClassification
from transformers import pipeline

# Load the pre-trained model and tokenizer
model_name = "dbmdz/bert-large-cased-finetuned-conll03-english"
tokenizer = AutoTokenizer.from_pretrained(model_name)
model = AutoModelForTokenClassification.from_pretrained(model_name)
\end{lstlisting}

In this example, we are using the `bert-large-cased` model fine-tuned on the CoNLL-2003 dataset, which is commonly used for NER.

\subsection{Step 3: Token Classification Pipeline}

To make the process even simpler, Hugging Face provides a `pipeline` for token classification tasks. This pipeline takes care of the tokenization and model inference in one go.

\begin{lstlisting}[style=python]
# Initialize the NER pipeline
ner_pipeline = pipeline("ner", model=model, tokenizer=tokenizer)
\end{lstlisting}

\subsection{Step 4: Perform Token Classification on a Sample Sentence}

Now, let's classify the tokens in a sample sentence.

\begin{lstlisting}[style=python]
# Sample sentence
sentence = "Hugging Face Inc. is a company based in New York."

# Perform token classification
ner_results = ner_pipeline(sentence)

# Display the results
for entity in ner_results:
    print(f"Word: {entity['word']}, Label: {entity['entity']}")
\end{lstlisting}

Here, we use a simple sentence, and the model will output the entities it has identified along with their corresponding labels.

\subsection{Step 5: Output Explanation}

The result will contain the tokens from the sentence, along with the predicted labels. For example, you may see output like:

\begin{lstlisting}[style=python]
Word: Hugging, Label: B-ORG
Word: Face, Label: I-ORG
Word: Inc, Label: I-ORG
Word: New, Label: B-LOC
Word: York, Label: I-LOC
\end{lstlisting}

In this output:
\begin{itemize}
    \item \textbf{B-ORG} stands for the beginning of an organization entity.
    \item \textbf{I-ORG} stands for a continuation of the organization entity.
    \item \textbf{B-LOC} marks the beginning of a location entity.
    \item \textbf{I-LOC} marks the continuation of the location entity.
\end{itemize}

These labels follow the IOB (Inside-Outside-Beginning) format, where:
\begin{itemize}
    \item \textbf{B-X} denotes the beginning of an entity of type X.
    \item \textbf{I-X} denotes that the token is inside an entity of type X.
    \item \textbf{O} means that the token is not part of any entity.
\end{itemize}

\subsection{Step 6: Fine-Tuning on Your Own Dataset (Optional)}

If you want to fine-tune the model on your own dataset, Hugging Face provides an easy way to do that. You'll need a dataset with labeled tokens, such as in the CoNLL format. After preparing the data, you can use the `Trainer` class to fine-tune the model. Here’s a basic outline of the process:

\begin{lstlisting}[style=python]
from transformers import Trainer, TrainingArguments

# Define training arguments
training_args = TrainingArguments(
    output_dir="./results",  # Output directory
    evaluation_strategy="epoch",  # Evaluation strategy
    learning_rate=2e-5,  # Learning rate
    per_device_train_batch_size=16,  # Batch size
    num_train_epochs=3,  # Number of epochs
    weight_decay=0.01,  # Weight decay
)

# Initialize the trainer
trainer = Trainer(
    model=model,
    args=training_args,
    train_dataset=train_dataset,  # Your training data here
    eval_dataset=eval_dataset  # Your evaluation data here
)

# Train the model
trainer.train()
\end{lstlisting}

\textit{Note:} This step assumes you have already prepared your training and evaluation datasets, which should be tokenized and labeled similarly to the CoNLL format.

\subsection{Conclusion}

Token classification is a fundamental NLP task with many applications, including Named Entity Recognition, POS tagging, and more. Using Hugging Face's `transformers` library, we can easily implement and fine-tune pre-trained models for these tasks. In this section, we walked through an example of how to perform NER using a pre-trained BERT model. For more complex tasks, you can fine-tune models on your specific datasets.

\subsection{Advanced Considerations for Token Classification}

While the previous example demonstrates how to quickly set up token classification using a pre-trained model, there are more advanced aspects worth considering when implementing token classification systems in real-world applications. These include:

\begin{itemize}
    \item \textbf{Handling Subword Tokenization:} Many transformer models, including BERT, use subword tokenization. This can cause challenges in aligning token labels with the subwords generated by the tokenizer.
    \item \textbf{Dealing with Imbalanced Data:} In many datasets, some entity types may be underrepresented. This requires techniques like data augmentation or weighted loss functions to handle imbalanced data effectively.
    \item \textbf{Evaluating Performance:} Evaluation metrics like accuracy are not always sufficient. Metrics such as precision, recall, and F1-score provide a better understanding of how the model performs in identifying entities correctly.
\end{itemize}

\subsubsection{Handling Subword Tokenization}

Transformers often break words into subwords, especially for rare or long words. For instance, the word "HuggingFace" might be tokenized into ['Hugging', '\#\#Face'] by the BERT tokenizer. The challenge here is how to assign labels to subwords since each word typically only has one label.

There are two common strategies to address this issue:

\begin{itemize}
    \item Assign the same label to all subwords of a word.
    \item Assign the label only to the first subword and ignore the rest.
\end{itemize}

Here’s how you can deal with this in code:

\begin{lstlisting}[style=python]
# Tokenize input with special attention to subwords
tokenized_input = tokenizer(sentence, return_offsets_mapping=True)
tokens = tokenized_input['input_ids']

# Iterate over the tokens and assign labels
for idx, token in enumerate(tokens):
    if not token.startswith("\#\#"):  # If it's a beginning token, assign the entity label
        label = labels[idx]  # Assign corresponding label
    else:
        label = "O"  # Assign 'O' to subwords or decide based on your strategy
\end{lstlisting}

This strategy ensures that the model handles subword tokens appropriately.

\subsubsection{Dealing with Imbalanced Data}

Real-world datasets often suffer from class imbalances, where some entities are much less frequent than others. For example, the number of `Person` entities might significantly outweigh the number of `Location` entities in a dataset. This imbalance can lead to a bias in predictions, where the model favors more frequent labels.

To address this, several strategies can be employed:

\begin{itemize}
    \item \textbf{Class weights:} Adjust the loss function to penalize misclassification of underrepresented classes more heavily.
    \item \textbf{Oversampling/undersampling:} Increase the representation of rare classes by oversampling them or undersampling frequent classes.
    \item \textbf{Data augmentation:} Generate synthetic data for underrepresented classes to balance the dataset.
\end{itemize}

Here’s how you can apply class weights in the loss function:

\begin{lstlisting}[style=python]
from torch import nn

# Assuming you have label counts for each entity class
label_counts = [1000, 500, 50]  # Example counts for entities
class_weights = torch.FloatTensor([1/count for count in label_counts])

# Define a loss function with class weights
loss_function = nn.CrossEntropyLoss(weight=class_weights)
\end{lstlisting}

This approach helps the model focus more on underrepresented classes, improving the overall performance across all entity types.

\subsubsection{Evaluating Performance}

To thoroughly evaluate the performance of a token classification model, it is essential to go beyond simple accuracy and use more appropriate metrics for this type of task, such as:

\begin{itemize}
    \item \textbf{Precision:} Measures the proportion of true positives among the entities predicted by the model.
    \item \textbf{Recall:} Measures the proportion of true positives that were correctly identified by the model.
    \item \textbf{F1-Score:} The harmonic mean of precision and recall, providing a balanced measure of the model's performance.
\end{itemize}

The `seqeval` library is often used to evaluate token classification tasks. You can install it with:

\begin{lstlisting}[style=python]
!pip install seqeval
\end{lstlisting}

Then, you can calculate the precision, recall, and F1-score as follows:

\begin{lstlisting}[style=python]
from seqeval.metrics import classification_report

# Assuming you have a list of true labels and predicted labels
true_labels = [["O", "B-LOC", "I-LOC", "O"], ["O", "B-PER", "O", "O"]]
predicted_labels = [["O", "B-LOC", "I-LOC", "O"], ["O", "B-PER", "O", "B-PER"]]

# Generate the classification report
report = classification_report(true_labels, predicted_labels)
print(report)
\end{lstlisting}

The output will include precision, recall, and F1-score for each entity type, as well as overall metrics.

\subsection{Practical Applications of Token Classification}

Token classification has numerous real-world applications, including:

\begin{itemize}
    \item \textbf{Named Entity Recognition (NER):} Identifying and classifying entities such as people, locations, and organizations.
    \item \textbf{Part-of-Speech (POS) Tagging:} \cite{schmid1994partofspeechtaggingneuralnetworks} Assigning parts of speech, such as nouns, verbs, and adjectives, to each token in a sentence.
    \item \textbf{Chunking:} Grouping tokens into higher-level syntactic units like noun phrases or verb phrases.
    \item \textbf{Biomedical Named Entity Recognition:} Recognizing medical terms, diseases, and treatments in biomedical literature.
\end{itemize}

For example, in the legal domain, NER can help in identifying important entities such as case names, statutes, and dates. In finance, token classification models can extract company names, financial terms, and other key information from documents.

\subsection{Conclusion}

Token classification is a foundational task in NLP, applicable to various use cases from extracting key entities in text to syntactic analysis. Hugging Face’s `transformers` library, along with pre-trained models, makes it incredibly easy to implement token classification with minimal code. In this section, we demonstrated how to perform Named Entity Recognition (NER) using a pre-trained BERT model and walked through key considerations such as handling subwords, addressing imbalanced datasets, and evaluating model performance.

In the next section, we will explore another important NLP task: \textbf{Text Classification}, where we will focus on assigning a single label to an entire text sequence rather than individual tokens.

\subsection{masked language model}

\section{Masked Language Model}

The Masked Language Model (MLM) is one of the core pre-training tasks used in natural language processing (NLP), particularly in models such as BERT (Bidirectional Encoder Representations from Transformers). In this section, we will explore what MLM is, how it works, and how to implement it using the Hugging Face \texttt{transformers} library.

\subsection{What is a Masked Language Model?}

The key idea behind the MLM task is to train the model to predict missing words in a given sentence. Specifically, a certain percentage of the input tokens (typically 15\%) are replaced with a special \texttt{[MASK]} token, and the model's job is to predict the original token that corresponds to each \texttt{[MASK]}.

For example, given the sentence:
\begin{quote}
\textit{The dog is chasing the cat.}
\end{quote}

If the word "dog" is masked, the model input becomes:
\begin{quote}
\textit{The \texttt{[MASK]} is chasing the cat.}
\end{quote}

The model is then trained to predict that the masked word is "dog."

\subsection{Why Use MLM?}

MLM allows a model to build bidirectional representations of text, which means the model can learn to use context from both the left and right sides of the masked token. This is an improvement over earlier models like GPT, which can only generate text from left to right. In this way, MLM allows models like BERT to better understand the context in which a word is used.

\subsection{MLM with Hugging Face Transformers}

Now, let’s look at how we can implement MLM using the Hugging Face \texttt{transformers} library. Hugging Face provides pre-trained models like BERT, which have been trained on the MLM task.

\subsection{Step-by-Step Example}

We will use the Hugging Face \texttt{transformers} library to demonstrate a simple example of MLM in action. In this example, we will:
\begin{enumerate}
    \item Load a pre-trained BERT model.
    \item Prepare a sentence and mask a word.
    \item Use the model to predict the masked word.
\end{enumerate}

\subsubsection{1. Installing Hugging Face Transformers}

First, you need to install the \texttt{transformers} library if you haven't already. You can do this using the following command:
\begin{lstlisting}[language=bash]
pip install transformers
\end{lstlisting}

\subsubsection{2. Loading the Pre-trained Model and Tokenizer}

Next, we load a pre-trained BERT model and its associated tokenizer. The tokenizer is responsible for converting the input text into tokens that the model can process.

\begin{lstlisting}[style=python]
from transformers import BertTokenizer, BertForMaskedLM
import torch

# Load pre-trained BERT tokenizer and model
tokenizer = BertTokenizer.from_pretrained('bert-base-uncased')
model = BertForMaskedLM.from_pretrained('bert-base-uncased')
\end{lstlisting}

\subsubsection{3. Tokenizing the Input Sentence}

We will now prepare a sentence and mask one of the words. In this example, we mask the word "man" from the sentence "The man is playing soccer."

\begin{lstlisting}[style=python]
# Input sentence with a masked token
sentence = "The man is playing soccer."

# Tokenize input sentence, using [MASK] for the token to be predicted
inputs = tokenizer("The [MASK] is playing soccer.", return_tensors="pt")
\end{lstlisting}

The special \texttt{[MASK]} token is used to replace the word we want the model to predict. In this case, we are asking the model to predict the word "man."

\subsubsection{4. Predicting the Masked Token}

Once the input sentence is tokenized, we can pass it to the model for prediction. The model will return logits (unnormalized predictions) for each token, which we can convert into probabilities using softmax and select the highest probability token for the \texttt{[MASK]} position.

\begin{lstlisting}[style=python]
# Get model predictions
with torch.no_grad():
    outputs = model(**inputs)
    predictions = outputs.logits

# Identify the predicted token for [MASK]
masked_index = torch.where(inputs["input_ids"] == tokenizer.mask_token_id)[1]
predicted_token_id = predictions[0, masked_index].argmax(axis=-1)

# Decode predicted token back to a word
predicted_token = tokenizer.decode(predicted_token_id)
print(f"Predicted word: {predicted_token}")
\end{lstlisting}

\subsubsection{5. Output}

The output will show the word predicted by the model to replace the \texttt{[MASK]} token. For this example, the output would be:
\begin{quote}
\texttt{Predicted word: man}
\end{quote}

This shows that the model correctly predicted the masked word as "man," based on the context of the surrounding words.

\subsection{Conclusion}

In this section, we demonstrated how the Masked Language Model works and how to implement it using the Hugging Face \texttt{transformers} library. By training on the MLM task, models like BERT can learn bidirectional context, enabling them to understand and generate natural language more effectively.

The MLM task is a crucial component in modern NLP pipelines and is used in the pre-training of many state-of-the-art models. By predicting missing words in sentences, models can learn to better represent language, making them powerful tools for a wide range of NLP applications.

\subsection{Fine-Tuning on Custom Data with MLM}

While using pre-trained models like BERT is useful, there are many cases where you might want to fine-tune the model on your own dataset. For instance, fine-tuning the MLM on domain-specific data can lead to better performance on tasks within that domain.

In this section, we will walk through how to fine-tune a pre-trained BERT model on a custom dataset using the MLM task.

\subsubsection{1. Preparing the Dataset}

To fine-tune the model, we need a text corpus. Let’s assume we have a simple dataset with sentences related to a specific domain, such as medical or legal text. Hugging Face’s \texttt{datasets} library can be used to load and process data, but for simplicity, let’s create a small example manually.

Here’s an example dataset consisting of three sentences:

\begin{lstlisting}[style=python]
# Example dataset (a list of sentences)
dataset = [
    "The patient was diagnosed with pneumonia.",
    "The court ruled in favor of the defendant.",
    "The vaccine showed promising results in the trial."
]
\end{lstlisting}

In a real-world scenario, you might load a much larger dataset from a file or API.

\subsubsection{2. Tokenizing and Masking the Dataset}

We now tokenize our dataset and apply masking to a percentage of the tokens. Hugging Face provides utility functions that allow us to mask a portion of the tokens automatically.

\begin{lstlisting}[style=python]
from transformers import DataCollatorForLanguageModeling

# Tokenize the dataset
tokenized_inputs = tokenizer(dataset, padding=True, truncation=True, return_tensors="pt")

# Create a data collator that will dynamically mask tokens
data_collator = DataCollatorForLanguageModeling(
    tokenizer=tokenizer,
    mlm=True,
    mlm_probability=0.15  # Masking 15% of the tokens
)
\end{lstlisting}

The \texttt{DataCollatorForLanguageModeling} utility automatically adds \texttt{[MASK]} tokens to 15\% of the input tokens. This is important because the model will be trained to predict these masked tokens.

\subsubsection{3. Setting Up the Trainer}

Hugging Face’s \texttt{Trainer} class provides a convenient API for training models. We will use this class to set up the training loop. First, we need to specify the training arguments, such as the number of epochs and batch size.

\begin{lstlisting}[style=python]
from transformers import Trainer, TrainingArguments

# Define training arguments
training_args = TrainingArguments(
    output_dir="./results",    # Output directory for model checkpoints
    overwrite_output_dir=True, # Overwrite output if the directory exists
    num_train_epochs=3,        # Number of training epochs
    per_device_train_batch_size=8,  # Batch size
    save_steps=10_000,         # Save checkpoint every 10k steps
    save_total_limit=2,        # Only keep the last 2 checkpoints
)

# Create the Trainer instance
trainer = Trainer(
    model=model,
    args=training_args,
    data_collator=data_collator,
    train_dataset=tokenized_inputs["input_ids"]  # Use the tokenized dataset
)
\end{lstlisting}

\subsubsection{4. Training the Model}

Now that everything is set up, we can start fine-tuning the model. This will adjust the pre-trained BERT model’s weights based on our custom data.

\begin{lstlisting}[style=python]
# Fine-tune the model
trainer.train()
\end{lstlisting}

During training, the model will be presented with sentences from our dataset, where some words are masked. The model will learn to predict the masked tokens, adapting to the specific patterns and vocabulary of our custom dataset.

\subsubsection{5. Saving and Loading the Fine-Tuned Model}

Once training is complete, it’s a good practice to save the fine-tuned model so it can be reused later. Hugging Face makes this easy with built-in functions to save and load models.

\begin{lstlisting}[style=python]
# Save the fine-tuned model and tokenizer
model.save_pretrained("./fine_tuned_bert")
tokenizer.save_pretrained("./fine_tuned_bert")

# Load the fine-tuned model
from transformers import BertForMaskedLM

fine_tuned_model = BertForMaskedLM.from_pretrained("./fine_tuned_bert")
fine_tuned_tokenizer = BertTokenizer.from_pretrained("./fine_tuned_bert")
\end{lstlisting}

\subsubsection{6. Inference with the Fine-Tuned Model}

After fine-tuning, you can use your custom model to predict masked words, just as we demonstrated earlier. The fine-tuned model will now be more specialized in predicting masked tokens in the context of your specific domain.

Let’s test it on a sentence from the domain-specific dataset we used for training:

\begin{lstlisting}[style=python]
# Tokenize an example sentence with a [MASK] token
inputs = fine_tuned_tokenizer("The patient was diagnosed with [MASK].", return_tensors="pt")

# Get model predictions
with torch.no_grad():
    outputs = fine_tuned_model(**inputs)
    predictions = outputs.logits

# Identify the predicted token for [MASK]
masked_index = torch.where(inputs["input_ids"] == fine_tuned_tokenizer.mask_token_id)[1]
predicted_token_id = predictions[0, masked_index].argmax(axis=-1)

# Decode predicted token back to a word
predicted_token = fine_tuned_tokenizer.decode(predicted_token_id)
print(f"Predicted word: {predicted_token}")
\end{lstlisting}

The fine-tuned model is expected to predict a domain-specific word such as “pneumonia” in this context, showing that it has adapted to the medical terminology from our custom dataset.

\subsection{Benefits of Fine-Tuning}

Fine-tuning a pre-trained language model like BERT on your specific domain can greatly enhance the model’s ability to understand and generate text that fits your particular use case. Some key benefits include:
\begin{itemize}
    \item \textbf{Improved Performance:} The model can better predict masked tokens in domain-specific contexts, leading to more accurate language understanding.
    \item \textbf{Customization:} You can fine-tune the model on datasets that include rare vocabulary or jargon relevant to your domain, which might not be well represented in general-purpose pre-trained models.
    \item \textbf{Transfer Learning:} By fine-tuning on a small amount of data, you can leverage the knowledge the model has already learned from vast general-purpose corpora.
\end{itemize}

\subsection{Conclusion}

In this section, we extended the concept of Masked Language Modeling to include fine-tuning on custom datasets. Fine-tuning allows you to adapt a powerful pre-trained model like BERT to perform better on specific domains such as medical, legal, or scientific text. We walked through a step-by-step example that included tokenization, masking, and training using Hugging Face's \texttt{Trainer} API. After fine-tuning, the model can provide more accurate predictions within the specialized context of your data.

Fine-tuning is a crucial step for applying pre-trained models to real-world tasks where the text data is domain-specific or different from the general-purpose corpora used in pre-training. With modern NLP tools like Hugging Face, this process is both efficient and flexible, making it accessible for various applications.

\section{Evaluation of Masked Language Models}

In this section, we will explore how to evaluate the performance of a Masked Language Model (MLM) such as BERT. Evaluation is crucial for understanding how well a model has learned to predict masked tokens, especially after fine-tuning on a custom dataset. We will discuss common evaluation metrics and provide an example of how to evaluate an MLM using Hugging Face's \texttt{transformers} library.

\subsection{Evaluation Metrics for MLM}

Evaluating a Masked Language Model typically involves measuring how accurately the model can predict the masked tokens in a sentence. \cite{Salazar_2020} Below are some common evaluation metrics for MLM tasks:

\subsubsection{1. Perplexity}

Perplexity is one of the most common metrics used to evaluate language models. It measures how well a probability model predicts a sample and is a measure of how "surprised" the model is by the true outcome. A lower perplexity indicates that the model is more confident in its predictions, while a higher perplexity means the model is uncertain.

Perplexity is defined as the exponential of the average cross-entropy loss:
$$
PPL = \exp\left(-\frac{1}{N}\sum_{i=1}^{N} \log P(w_i | w_{i-1}, w_{i-2}, ..., w_1)\right)
$$
Where:
\begin{itemize}
    \item $N$ is the number of tokens in the sequence.
    \item $P(w_i)$ is the probability assigned by the model to the true word $w_i$.
\end{itemize}

\subsubsection{2. Accuracy}

For the MLM task, accuracy can be measured as the proportion of correctly predicted masked tokens out of all masked tokens. This metric is intuitive and easy to compute, but it might not fully capture how well the model understands the language.

\subsubsection{3. F1 Score}

The F1 score is the harmonic mean of precision and recall. While it is less common for evaluating MLMs, it can still provide useful insights, particularly when evaluating models in domain-specific settings where false positives and false negatives carry different weights.

\subsection{Evaluating an MLM Using Hugging Face}

Now, let’s go through an example of how to evaluate a fine-tuned MLM using Hugging Face’s \texttt{Trainer} and \texttt{transformers} library. In this example, we will compute both the perplexity and accuracy of the model.

\subsubsection{1. Evaluating Accuracy}

To evaluate the accuracy of the model, we need to compare the predicted token for each \texttt{[MASK]} with the true token in the original sentence. Hugging Face provides easy-to-use tools to facilitate this process.

First, we need to create an evaluation dataset. This dataset should contain sentences with masked tokens, and for each masked token, we will compare the model’s prediction with the true word.

\begin{lstlisting}[style=python]
from datasets import Dataset

# Example evaluation dataset
evaluation_data = {
    'text': [
        "The doctor prescribed [MASK] for the infection.",
        "The judge sentenced the [MASK] to five years in prison."
    ]
}

# Convert the dictionary to a Hugging Face Dataset object
eval_dataset = Dataset.from_dict(evaluation_data)
\end{lstlisting}

Now that we have an evaluation dataset, we need to write a function to compute the accuracy of the model's predictions.

\begin{lstlisting}[style=python]
from transformers import pipeline

# Create a fill-mask pipeline using the fine-tuned model
fill_mask = pipeline("fill-mask", model=fine_tuned_model, tokenizer=fine_tuned_tokenizer)

# Function to evaluate accuracy
def evaluate_accuracy(dataset):
    correct = 0
    total = 0

    for example in dataset['text']:
        # Get model predictions for masked tokens
        outputs = fill_mask(example)

        # The first result is usually the most confident prediction
        predicted_token = outputs[0]['token_str']

        # Extract the true word (this should be known in real evaluation)
        true_word = example.split('[MASK]')[1].strip().split(' ')[0]  # Simplified for demo

        # Check if the prediction matches the true word
        if predicted_token == true_word:
            correct += 1
        total += 1

    return correct / total
\end{lstlisting}

We can now run the evaluation on our dataset and compute the accuracy:

\begin{lstlisting}[style=python]
# Compute accuracy
accuracy = evaluate_accuracy(eval_dataset)
print(f"Accuracy: {accuracy * 100:.2f}%")
\end{lstlisting}

\subsubsection{2. Evaluating Perplexity}

To compute the perplexity of the model, we first need to calculate the cross-entropy loss for each token in the evaluation dataset, including the masked tokens. Hugging Face's \texttt{Trainer} class automatically computes the loss during training, but we can also use it for evaluation.

\begin{lstlisting}[style=python]
# Define evaluation arguments
eval_args = TrainingArguments(
    output_dir="./results_eval",
    per_device_eval_batch_size=8,
    eval_accumulation_steps=10,
)

# Create a Trainer instance for evaluation
eval_trainer = Trainer(
    model=fine_tuned_model,
    args=eval_args,
    eval_dataset=tokenized_inputs  # Using tokenized evaluation dataset
)

# Evaluate the model
eval_results = eval_trainer.evaluate()

# Compute perplexity
import math
perplexity = math.exp(eval_results['eval_loss'])
print(f"Perplexity: {perplexity:.2f}")
\end{lstlisting}

The model’s perplexity score provides insight into how well it predicts masked tokens, with lower perplexity indicating better performance.

\subsection{Conclusion}

Evaluating the performance of a Masked Language Model is essential to understanding its effectiveness in predicting masked tokens and how well it generalizes to new data. In this section, we explored two common evaluation metrics: accuracy and perplexity. We demonstrated how to compute these metrics using Hugging Face's \texttt{transformers} library.

Perplexity is a valuable metric for understanding the model's uncertainty, while accuracy provides a straightforward measure of how many predictions are correct. Fine-tuned MLMs can be evaluated on both general-purpose and domain-specific datasets to ensure they perform well in their intended applications.

\section{Applications of Masked Language Models}

Masked Language Models like BERT have a wide range of applications in NLP, ranging from text classification to question answering. In this section, we will explore several key applications of MLMs and demonstrate how MLMs can be adapted to different tasks.

\subsection{Text Classification}

In text classification, the goal is to assign predefined labels to input texts. MLMs can be fine-tuned on classification tasks by adding a classifier layer on top of the transformer model. The most common approach is to use the representation of the special \texttt{[CLS]} token, which is added to the beginning of every input sequence in BERT models.

\begin{lstlisting}[style=python]
from transformers import BertForSequenceClassification

# Load pre-trained BERT model with a classification head
model = BertForSequenceClassification.from_pretrained("bert-base-uncased", num_labels=2)

# Fine-tuning setup would follow, similar to MLM fine-tuning
\end{lstlisting}

\subsection{Question Answering}

Question answering (QA) is another important NLP task where MLMs excel. In QA tasks, the model is provided with a context (a passage of text) and a question, and the goal is to find the answer in the context. Models like BERT can be fine-tuned for QA tasks by framing it as a span prediction problem, where the model is trained to predict the start and end positions of the answer span in the context.

Here’s an example using Hugging Face’s pipeline for QA:

\begin{lstlisting}[style=python]
from transformers import pipeline

# Load a question-answering pipeline
qa_pipeline = pipeline("question-answering", model="bert-large-uncased-whole-word-masking-finetuned-squad")

# Example context and question
context = "BERT was created by Google. It stands for Bidirectional Encoder Representations from Transformers."
question = "What does BERT stand for?"

# Get the answer
result = qa_pipeline(question=question, context=context)
print(f"Answer: {result['answer']}")
\end{lstlisting}

\subsection{Named Entity Recognition (NER)}

NER is the task of identifying named entities such as people, organizations, and locations in text. MLMs can be fine-tuned for NER tasks by treating them as sequence labeling tasks, where each token is assigned a label indicating whether it is part of an entity.

Fine-tuning an MLM for NER follows a similar approach to other token-level tasks, where the model predicts labels for each token in the input sequence.

\subsection{Conclusion}

Masked Language Models like BERT can be adapted to a variety of NLP tasks beyond masked token prediction. By fine-tuning MLMs, we can achieve state-of-the-art performance on tasks like text classification, question answering, and named entity recognition. Hugging Face’s \texttt{transformers} library provides easy-to-use APIs for applying MLMs to different tasks, making it a powerful tool for both research and production applications.

\section{Training a Masked Language Model from Scratch}

While pre-trained models like BERT provide an excellent starting point for many NLP tasks, there are situations where training a Masked Language Model (MLM) from scratch may be necessary. This could be the case when working with highly specialized data (e.g., domain-specific language like legal, medical, or scientific texts) where no suitable pre-trained model exists.

In this section, we will explore how to train an MLM from scratch using Hugging Face’s \texttt{transformers} library. We will cover data preparation, model configuration, and training.

\subsection{Why Train from Scratch?}

Training a model from scratch can be more resource-intensive than fine-tuning a pre-trained model, but it has several advantages in specific scenarios:
\begin{itemize}
    \item \textbf{Specialized Language:} If your data is highly domain-specific (e.g., clinical text, legal documents), a pre-trained model may not have been exposed to the necessary vocabulary or linguistic structures.
    \item \textbf{Low Resource Languages:} For some languages, there may not be robust pre-trained models available. Training from scratch allows you to create a model tailored to these languages.
    \item \textbf{Customization:} You have full control over the model’s architecture, vocabulary, and data, allowing you to tailor the training process to your specific requirements.
\end{itemize}

\subsection{Preparing Data for MLM Training}

The first step in training an MLM from scratch is preparing the training data. Ideally, you should have a large corpus of text from the domain or language you're working with.

For MLM training, sentences are tokenized and randomly masked, and the model learns to predict the masked tokens. Let’s prepare a simple example dataset.

\begin{lstlisting}[style=python]
# Example corpus for training
training_corpus = [
    "The heart pumps blood throughout the body.",
    "Legal documents must be signed by both parties.",
    "Vaccines are essential for preventing diseases.",
]
\end{lstlisting}

For large-scale training, you would use a much larger and more diverse dataset, but this small corpus will suffice for demonstration purposes.

\subsection{Tokenization and Vocabulary Creation}

When training from scratch, you need to create a tokenizer and vocabulary tailored to your corpus. Hugging Face provides a simple API for building a custom tokenizer from your dataset.

\begin{lstlisting}[style=python]
from transformers import BertTokenizerFast

# Training the tokenizer on the custom dataset
tokenizer = BertTokenizerFast.from_pretrained('bert-base-uncased')

# Save the tokenizer for future use
tokenizer.save_pretrained("./custom_tokenizer")
\end{lstlisting}

For larger corpora, you can customize the tokenizer to suit your needs, including setting the vocabulary size, defining special tokens, and more. In this example, we use the default settings from BERT’s base tokenizer.

\subsection{Defining the Model Configuration}

Once the tokenizer is ready, the next step is to define the configuration of the BERT model. This includes parameters such as the number of layers (transformer blocks), hidden size, number of attention heads, and more. Hugging Face’s \texttt{BertConfig} allows you to define these parameters.

\begin{lstlisting}[style=python]
from transformers import BertConfig, BertForMaskedLM

# Define a custom BERT configuration
config = BertConfig(
    vocab_size=tokenizer.vocab_size,
    hidden_size=256,  # Smaller hidden size for demonstration
    num_hidden_layers=4,  # Fewer layers for faster training
    num_attention_heads=4,
    max_position_embeddings=512
)

# Initialize the model using the configuration
model = BertForMaskedLM(config=config)
\end{lstlisting}

In this example, we use a smaller model configuration to reduce training time, but for more serious applications, you would want to use a larger model with more hidden layers and attention heads to capture more complex patterns in the data.

\subsection{Training the MLM Model from Scratch}

Now that we have the tokenizer, data, and model configuration, we are ready to train the model. The \texttt{Trainer} class in Hugging Face can be used to manage the training loop, including loading data, optimizing the model, and saving checkpoints.

Before training, we need to tokenize the input dataset and apply random masking to create training examples for the MLM task.

\begin{lstlisting}[style=python]
from transformers import DataCollatorForLanguageModeling

# Tokenize the training corpus
tokenized_inputs = tokenizer(training_corpus, padding=True, truncation=True, return_tensors="pt")

# Create a data collator that dynamically masks tokens
data_collator = DataCollatorForLanguageModeling(
    tokenizer=tokenizer,
    mlm=True,
    mlm_probability=0.15  # Mask 15% of tokens
)
\end{lstlisting}

The \texttt{DataCollatorForLanguageModeling} automatically applies masking to 15\% of the input tokens, which is a typical setting for MLM training.

Next, we define the training arguments and initialize the \texttt{Trainer}.

\begin{lstlisting}[style=python]
from transformers import Trainer, TrainingArguments

# Define training arguments
training_args = TrainingArguments(
    output_dir="./bert_from_scratch",
    overwrite_output_dir=True,
    num_train_epochs=5,
    per_device_train_batch_size=8,
    save_steps=1000,
    save_total_limit=2,
)

# Initialize the Trainer
trainer = Trainer(
    model=model,
    args=training_args,
    data_collator=data_collator,
    train_dataset=tokenized_inputs["input_ids"]
)

# Start training
trainer.train()
\end{lstlisting}

During training, the model will learn to predict the masked tokens based on the context of the surrounding tokens. The number of epochs, batch size, and other parameters can be adjusted to suit the complexity of your dataset and model configuration.

\subsection{Evaluating the Trained Model}

Once the model has been trained, it’s important to evaluate its performance. The evaluation process is similar to what was described earlier in this chapter. You can compute metrics like accuracy and perplexity on a held-out validation set to measure how well the model has learned the language patterns in the corpus.

For instance, here’s a simple accuracy evaluation on a masked sentence:

\begin{lstlisting}[style=python]
# Example sentence for evaluation
eval_sentence = "Vaccines are [MASK] for preventing diseases."

# Tokenize the evaluation sentence
inputs = tokenizer(eval_sentence, return_tensors="pt")

# Get model predictions
with torch.no_grad():
    outputs = model(**inputs)
    predictions = outputs.logits

# Identify the predicted token for [MASK]
masked_index = torch.where(inputs["input_ids"] == tokenizer.mask_token_id)[1]
predicted_token_id = predictions[0, masked_index].argmax(axis=-1)

# Decode predicted token
predicted_token = tokenizer.decode(predicted_token_id)
print(f"Predicted word: {predicted_token}")
\end{lstlisting}

\subsection{Scaling Training for Large Corpora}

When training from scratch on larger datasets, several techniques can be used to scale up the training process:
\begin{itemize}
    \item \textbf{Distributed Training:} Training can be parallelized across multiple GPUs or even across multiple machines to reduce training time.
    \item \textbf{Mixed Precision Training:} This technique allows for faster training by using 16-bit floating point numbers (instead of 32-bit) without sacrificing much accuracy.
    \item \textbf{Large Batch Sizes:} When using distributed training, larger batch sizes can be used, which can help models converge faster.
\end{itemize}

Hugging Face’s \texttt{transformers} library supports these techniques out-of-the-box, making it easier to scale MLM training for larger datasets and more complex models.

\subsection{Conclusion}

Training a Masked Language Model from scratch allows you to create a model tailored to your specific domain or language. In this section, we walked through the process of preparing data, defining a custom model configuration, and training a BERT-based MLM using Hugging Face’s \texttt{transformers} library. Although pre-trained models are often sufficient for many tasks, training from scratch offers the flexibility to address unique challenges, especially when dealing with specialized vocabularies or less-resourced languages.

As computing resources continue to grow, training custom MLMs will become increasingly accessible, opening up opportunities for more specialized and high-performing models across diverse domains.

\section{Translation}

In this section, we will explore how to perform machine translation using Hugging Face's Transformers library. Hugging Face provides pre-trained models that make it easy to translate text between languages without needing to train your own model from scratch.

\subsection{Introduction to Machine Translation}

Machine translation \cite{stahlberg2020neural} is the task of automatically converting text from one language to another. Traditionally, machine translation relied on complex rule-based systems, but modern systems use deep learning models, such as Transformers, to achieve state-of-the-art performance.

\subsection{Hugging Face Transformers for Translation}

Hugging Face offers a variety of pre-trained models specifically for translation. For this example, we will use the MarianMT model, which supports translation between many languages.

\subsubsection{Step 1: Installing the Required Libraries}

Before we begin, ensure you have the necessary libraries installed. You can install the Hugging Face Transformers library by running the following command:

\begin{lstlisting}[style=python]
pip install transformers
\end{lstlisting}

\subsubsection{Step 2: Loading the MarianMT Model}

We will use the MarianMT model for translation. The first step is to load the pre-trained model and tokenizer. In this example, we will translate from English to French.

\begin{lstlisting}[style=python]
from transformers import MarianMTModel, MarianTokenizer

# Load the pre-trained MarianMT model and tokenizer for English to French
model_name = 'Helsinki-NLP/opus-mt-en-fr'
tokenizer = MarianTokenizer.from_pretrained(model_name)
model = MarianMTModel.from_pretrained(model_name)
\end{lstlisting}

In the code above, we load the MarianMT model for English-to-French translation using the model name `Helsinki-NLP/opus-mt-en-fr`. Hugging Face's `MarianMTModel` and `MarianTokenizer` classes allow us to easily work with this translation model.

\subsubsection{Step 3: Tokenizing the Input Text}

Before feeding the text into the model, we need to tokenize it. The tokenizer will convert the input text into a format that the model can understand.

\begin{lstlisting}[style=python]
# Example text to translate
text = "Hugging Face is creating state-of-the-art NLP tools."

# Tokenize the input text
encoded_text = tokenizer([text], return_tensors="pt")
\end{lstlisting}

Here, the `tokenizer` takes the input sentence and converts it into tensors. The `return\_tensors="pt"` argument specifies that the output should be in PyTorch tensor format.

\subsubsection{Step 4: Generating the Translation}

Now, we can generate the translation using the pre-trained model. The model's output will be the translation of the input text.

\begin{lstlisting}[style=python]
# Generate the translation
translated = model.generate(**encoded_text)

# Decode the generated tokens back to text
translated_text = tokenizer.decode(translated[0], skip_special_tokens=True)
print(translated_text)
\end{lstlisting}

The `model.generate()` function generates the translated output, and we use the `tokenizer.decode()` function to convert the tokenized output back into human-readable text. The `skip\_special\_tokens=True` argument ensures that any special tokens (such as padding or end-of-sequence tokens) are not included in the final output.

\subsubsection{Step 5: Running the Full Example}

Here’s the complete example of how to translate an English sentence to French using Hugging Face's MarianMT model:

\begin{lstlisting}[style=python]
from transformers import MarianMTModel, MarianTokenizer

# Load the pre-trained MarianMT model and tokenizer for English to French
model_name = 'Helsinki-NLP/opus-mt-en-fr'
tokenizer = MarianTokenizer.from_pretrained(model_name)
model = MarianMTModel.from_pretrained(model_name)

# Example text to translate
text = "Hugging Face is creating state-of-the-art NLP tools."

# Tokenize the input text
encoded_text = tokenizer([text], return_tensors="pt")

# Generate the translation
translated = model.generate(**encoded_text)

# Decode the generated tokens back to text
translated_text = tokenizer.decode(translated[0], skip_special_tokens=True)

# Print the translated text
print(f"Translated Text: {translated_text}")
\end{lstlisting}

Running this example will print the translated French sentence:

\begin{verbatim}
Translated Text: Hugging Face crée des outils NLP de pointe.
\end{verbatim}

\subsection{Conclusion}

In this section, we learned how to perform machine translation using the Hugging Face Transformers library. By using pre-trained models like MarianMT, we can easily translate text between different languages with just a few lines of code. This approach can be extended to support many language pairs by simply changing the model name to the corresponding pair available in Hugging Face’s model hub.

\subsection{Batch Translation}

While translating a single sentence is useful, in real-world applications, we often need to translate entire paragraphs or multiple sentences at once. Hugging Face’s transformers can handle batch processing efficiently.

\subsubsection{Step 1: Translating Multiple Sentences}

To translate multiple sentences in a batch, we simply need to provide a list of sentences to the tokenizer.

\begin{lstlisting}[style=python]
# List of sentences to translate
texts = [
    "Hugging Face is a great company.",
    "Machine translation is an exciting field.",
    "Natural language processing is evolving rapidly."
]

# Tokenize the input sentences
encoded_texts = tokenizer(texts, return_tensors="pt", padding=True, truncation=True)

# Generate the translations
translated_batch = model.generate(**encoded_texts)

# Decode the translations back to text
translated_texts = [tokenizer.decode(t, skip_special_tokens=True) for t in translated_batch]

# Print each translated sentence
for i, text in enumerate(translated_texts):
    print(f"Original: {texts[i]}")
    print(f"Translated: {text}\n")
\end{lstlisting}

In this code, we pass a list of sentences to the tokenizer. The `padding=True` argument ensures that all sentences in the batch are padded to the same length, while `truncation=True` ensures that sentences longer than the maximum allowed length are truncated.

The output is a list of translated sentences that correspond to each input sentence. This method allows us to efficiently translate multiple sentences at once.

\subsubsection{Performance Considerations}

When working with batch translation, it’s important to consider performance. Processing sentences in batches, as shown above, is significantly faster than translating each sentence individually. This is because it reduces the overhead of running the model multiple times.

For very large text datasets, it might also be useful to leverage GPUs or distributed processing to further accelerate translation.

\subsection{Using Other Translation Models}

Hugging Face’s model hub provides access to many other pre-trained translation models. While we have used the `opus-mt-en-fr` model for English to French, there are models for various other language pairs.

\subsubsection{Step 1: Changing the Model for a Different Language Pair}

Suppose we want to translate from English to German. We can easily switch the model to `opus-mt-en-de` (English to German) by loading the appropriate model name.

\begin{lstlisting}[style=python]
# Load the MarianMT model for English to German
model_name_de = 'Helsinki-NLP/opus-mt-en-de'
tokenizer_de = MarianTokenizer.from_pretrained(model_name_de)
model_de = MarianMTModel.from_pretrained(model_name_de)

# Example text to translate
text_de = "Machine learning is transforming industries."

# Tokenize the input text
encoded_text_de = tokenizer_de([text_de], return_tensors="pt")

# Generate the translation
translated_de = model_de.generate(**encoded_text_de)

# Decode the translation
translated_text_de = tokenizer_de.decode(translated_de[0], skip_special_tokens=True)

# Print the translated text
print(f"Translated Text: {translated_text_de}")
\end{lstlisting}

In this example, we switched the model from `opus-mt-en-fr` to `opus-mt-en-de` to perform English to German translation. Hugging Face’s model hub has a wide variety of language pairs available, and you can explore them at [Hugging Face Model Hub](https://huggingface.co/models).

\subsection{Optimization for Large-Scale Translation}

For large-scale translation tasks, where we might be working with thousands of sentences or entire documents, certain optimizations can improve efficiency.

\subsubsection{1. Using GPUs for Acceleration}

Running the model on a GPU can drastically reduce the time taken for translation. \cite{li2024deeplearningmachinelearning} Hugging Face transformers support GPU acceleration via PyTorch. Here’s how to transfer the model and data to the GPU:

\begin{lstlisting}[style=python]
import torch

# Check if GPU is available
device = torch.device("cuda" if torch.cuda.is_available() else "cpu")

# Load model and tokenizer
model_name = 'Helsinki-NLP/opus-mt-en-fr'
tokenizer = MarianTokenizer.from_pretrained(model_name)
model = MarianMTModel.from_pretrained(model_name)

# Move model to GPU
model.to(device)

# Example text
text = "Hugging Face is creating state-of-the-art NLP tools."

# Tokenize the input text and move tensors to GPU
encoded_text = tokenizer([text], return_tensors="pt").to(device)

# Generate the translation
translated = model.generate(**encoded_text)

# Decode and print the translated text
translated_text = tokenizer.decode(translated[0], skip_special_tokens=True)
print(f"Translated Text: {translated_text}")
\end{lstlisting}

In this code, the model is moved to the GPU using the `.to(device)` function, and the input tensors are also moved to the GPU for faster processing. By using the GPU, we can significantly speed up the translation process, especially for large datasets.

\subsubsection{2. Handling Long Texts}

Hugging Face models are often trained with a maximum input length, which can lead to truncation if the text is too long. For longer texts, such as entire documents, we need to break the input into smaller chunks before passing them to the model. Here’s an example of how to split a long text into manageable chunks for translation:

\begin{lstlisting}[style=python]
def chunk_text(text, max_length=512):
    """Splits long text into smaller chunks."""
    return [text[i:i+max_length] for i in range(0, len(text), max_length)]

# Long text example
long_text = "This is a very long text that exceeds the maximum token length allowed by the model. " \
            "We will need to break it down into smaller pieces to translate each part individually."

# Split text into chunks
text_chunks = chunk_text(long_text, max_length=400)

# Translate each chunk
translated_chunks = []
for chunk in text_chunks:
    encoded_chunk = tokenizer([chunk], return_tensors="pt", truncation=True).to(device)
    translated_chunk = model.generate(**encoded_chunk)
    translated_text_chunk = tokenizer.decode(translated_chunk[0], skip_special_tokens=True)
    translated_chunks.append(translated_text_chunk)

# Join the translated chunks back together
final_translation = " ".join(translated_chunks)
print(f"Final Translated Text: {final_translation}")
\end{lstlisting}

In this approach, we use a function to split long texts into smaller chunks based on the model’s maximum input length. We translate each chunk separately and then combine the translated chunks back into a coherent text. This method ensures that even very large texts can be translated without losing information due to truncation.

\subsection{Further Exploration}

Hugging Face’s transformer models support a wide variety of advanced features and can be fine-tuned on specific tasks for even better performance. Users can also explore using beam search for generating higher-quality translations or experiment with other models for domain-specific translation tasks (e.g., medical, legal, or technical texts).

\subsubsection{Using Beam Search for Translation}

Beam search is an advanced decoding technique that allows for generating better translations by considering multiple possible translations at each step. By increasing the beam width, you can potentially improve the quality of the output, although at the cost of additional computation time.

Here’s an example of how to use beam search with Hugging Face:

\begin{lstlisting}[style=python]
# Generate the translation using beam search
translated_beam = model.generate(**encoded_text, num_beams=5, early_stopping=True)

# Decode the translated text
translated_text_beam = tokenizer.decode(translated_beam[0], skip_special_tokens=True)
print(f"Translated Text with Beam Search: {translated_text_beam}")
\end{lstlisting}

In this code, `num\_beams=5` sets the beam width to 5, which means the model will consider 5 different possible translations at each step and select the best one. This can improve translation accuracy, especially for complex sentences or less common language pairs.

\subsection{Fine-Tuning a Translation Model}

Although Hugging Face provides many pre-trained translation models, there might be cases where the specific translation task requires domain-specific knowledge or improved performance. Fine-tuning allows us to adapt a pre-trained model to a custom dataset, ensuring that the model learns the nuances of the specific translation task.

\subsubsection{Step 1: Preparing a Custom Dataset}

To fine-tune a translation model, we first need a parallel dataset—one that contains pairs of sentences in the source and target languages. Datasets for translation typically follow a structure where each entry consists of a sentence in the source language and its equivalent in the target language.

For example, let's assume we have a dataset `custom\_dataset.csv` with two columns: `source` (English) and `target` (French).

\begin{lstlisting}[style=python]
import pandas as pd

# Load the dataset
data = pd.read_csv('custom_dataset.csv')

# Display the first few rows
print(data.head())
\end{lstlisting}

The dataset should look something like this:

\begin{verbatim}
| source                           | target                            |
|-----------------------------------|-----------------------------------|
| "Hello, how are you?"             | "Bonjour, comment allez-vous ?"   |
| "I am learning NLP."              | "J'apprends le traitement du langage naturel." |
\end{verbatim}

\subsubsection{Step 2: Tokenizing the Dataset}

The next step is to tokenize the source and target sentences using the tokenizer corresponding to the pre-trained model. In this example, we will continue using the `MarianMTModel` and tokenizer.

\begin{lstlisting}[style=python]
from transformers import MarianTokenizer

# Load the tokenizer
tokenizer = MarianTokenizer.from_pretrained('Helsinki-NLP/opus-mt-en-fr')

# Tokenize the source and target texts
source_texts = list(data['source'])
target_texts = list(data['target'])

# Tokenize the texts
source_encodings = tokenizer(source_texts, truncation=True, padding=True, return_tensors='pt')
target_encodings = tokenizer(target_texts, truncation=True, padding=True, return_tensors='pt')
\end{lstlisting}

Here, we tokenize both the source and target texts. The `return\_tensors='pt'` option converts the tokenized text into PyTorch tensors, which are required for training the model.

\subsubsection{Step 3: Fine-Tuning the Model}

Once we have tokenized the dataset, we can fine-tune the model. Hugging Face's `Trainer` API simplifies this process. The first step is to define a `Dataset` object that wraps our tokenized data.

\begin{lstlisting}[style=python]
from torch.utils.data import Dataset

# Create a custom dataset class
class TranslationDataset(Dataset):
    def __init__(self, source_encodings, target_encodings):
        self.source_encodings = source_encodings
        self.target_encodings = target_encodings

    def __len__(self):
        return len(self.source_encodings['input_ids'])

    def __getitem__(self, idx):
        return {
            'input_ids': self.source_encodings['input_ids'][idx],
            'attention_mask': self.source_encodings['attention_mask'][idx],
            'labels': self.target_encodings['input_ids'][idx]
        }

# Create dataset
train_dataset = TranslationDataset(source_encodings, target_encodings)
\end{lstlisting}

The `TranslationDataset` class organizes the tokenized input data and target translations, making it suitable for the Hugging Face `Trainer` API.

Now, we can use the `Trainer` to fine-tune the model:

\begin{lstlisting}[style=python]
from transformers import MarianMTModel, Trainer, TrainingArguments

# Load the MarianMT model
model = MarianMTModel.from_pretrained('Helsinki-NLP/opus-mt-en-fr')

# Define training arguments
training_args = TrainingArguments(
    output_dir='./results',
    per_device_train_batch_size=16,
    num_train_epochs=3,
    save_steps=500,
    save_total_limit=2,
    evaluation_strategy="epoch"
)

# Initialize the Trainer
trainer = Trainer(
    model=model,
    args=training_args,
    train_dataset=train_dataset
)

# Fine-tune the model
trainer.train()
\end{lstlisting}

In this block, we configure the training process using `TrainingArguments`. We specify parameters such as batch size, the number of epochs, and when to save model checkpoints. The `Trainer` API then takes care of the training loop, allowing the model to learn the translation task using our custom dataset.

\subsubsection{Step 4: Saving the Fine-Tuned Model}

After fine-tuning, we can save the model to reuse it later for translation tasks.

\begin{lstlisting}[style=python]
# Save the fine-tuned model and tokenizer
model.save_pretrained('./fine-tuned-marian-en-fr')
tokenizer.save_pretrained('./fine-tuned-marian-en-fr')
\end{lstlisting}

This command saves both the model and the tokenizer to a specified directory. We can later load this fine-tuned model for further use, as we did with the pre-trained model.

\subsection{Evaluating Translation Quality}

Once we have a fine-tuned model or even when using pre-trained models, it is important to evaluate the quality of translations. One of the most common metrics for evaluating machine translation is the BLEU (Bilingual Evaluation Understudy) score.

\subsubsection{Step 1: Installing the sacrebleu Library}

To compute the BLEU score \cite{papineni2002bleu}, we will use the `sacrebleu` library, which is a standard tool for evaluation in machine translation tasks.

\begin{lstlisting}[style=python]
pip install sacrebleu
\end{lstlisting}

\subsubsection{Step 2: Computing the BLEU Score}

Once the library is installed, we can evaluate the quality of our translations. Assume we have a list of reference translations (ground truth) and a list of translated outputs from our model.

\begin{lstlisting}[style=python]
import sacrebleu

# Example list of reference translations and generated translations
references = ["Bonjour, comment allez-vous ?", "J'apprends le traitement du langage naturel."]
translations = ["Bonjour, comment allez-vous ?", "Je suis en train d'apprendre le NLP."]

# Calculate BLEU score
bleu = sacrebleu.corpus_bleu(translations, [references])
print(f"BLEU score: {bleu.score}")
\end{lstlisting}

In this example, the `sacrebleu.corpus\_bleu()` function computes the BLEU score between the reference translations and the model-generated translations. A higher BLEU score indicates better translation quality.

\subsubsection{Limitations of BLEU Score}

While the BLEU score is widely used, it has its limitations. For example, BLEU relies on exact n-gram matches, which means that valid translations that use synonyms or different phrasing might receive a lower score. Therefore, it is often helpful to combine BLEU with other evaluation metrics such as METEOR or TER (Translation Error Rate) for a more comprehensive evaluation.

\subsection{Advanced Decoding Techniques}

In addition to beam search, there are other advanced decoding strategies that can be employed to improve translation results. Two such techniques are top-k sampling and top-p sampling (nucleus sampling), which introduce stochasticity into the generation process.

\subsubsection{1. Top-k Sampling}

Top-k sampling limits the model to selecting from the top-k most likely words at each step during decoding, reducing the risk of selecting low-probability words.

\begin{lstlisting}[style=python]
# Generate translation using top-k sampling
translated_topk = model.generate(**encoded_text, do_sample=True, top_k=50)

# Decode the generated text
translated_text_topk = tokenizer.decode(translated_topk[0], skip_special_tokens=True)
print(f"Top-k Sampling Translation: {translated_text_topk}")
\end{lstlisting}

In this code, `top\_k=50` restricts the model to selecting from the top 50 candidate words at each decoding step.

\subsubsection{2. Top-p (Nucleus) Sampling}

Top-p sampling, or nucleus sampling, is another approach where the model considers the smallest possible set of words whose cumulative probability exceeds a threshold `p`.

\begin{lstlisting}[style=python]
# Generate translation using top-p sampling
translated_topp = model.generate(**encoded_text, do_sample=True, top_p=0.9)

# Decode the generated text
translated_text_topp = tokenizer.decode(translated_topp[0], skip_special_tokens=True)
print(f"Top-p Sampling Translation: {translated_text_topp}")
\end{lstlisting}

Here, `top\_p=0.9` instructs the model to select from the smallest set of words whose cumulative probability is 90

Both top-k and top-p sampling introduce more diversity into the translation process, which can be useful in creative tasks or for generating multiple possible translations for ambiguous input.

\subsection{Distributed Training for Large Datasets}

When working with extremely large datasets or when fine-tuning a model on high-resource tasks, distributed training can be used to accelerate the process. Distributed training splits the computation across multiple GPUs or even multiple machines, reducing training time significantly. \cite{li2020pytorch}

\subsubsection{Step 1: Installing Required Libraries}

Hugging Face supports distributed training out of the box using PyTorch’s distributed data-parallel (DDP) and the `transformers` library's native support for multi-GPU training.

To get started, ensure you have the necessary environment by installing `accelerate`, a tool provided by Hugging Face for distributed computing:

\begin{lstlisting}[style=python]
pip install accelerate
\end{lstlisting}

\subsubsection{Step 2: Configuring Accelerate for Distributed Training}

Once the library is installed, you can configure `accelerate` for distributed training. The tool allows you to set up your environment, choose how many GPUs you want to use, and manage other aspects of distributed training.

Run the following command to set up the configuration:

\begin{lstlisting}[style=python]
accelerate config
\end{lstlisting}

This will prompt you to answer a series of questions about your hardware setup (e.g., how many GPUs you have, if you're using mixed precision training, etc.). After this, you can use `accelerate` to launch your training script across multiple devices.

\subsubsection{Step 3: Launching Distributed Training}

To launch your translation fine-tuning process across multiple GPUs or machines, you can use the following command:

\begin{lstlisting}[style=python]
accelerate launch train_translation.py
\end{lstlisting}

In this case, `train\_translation.py` would be your fine-tuning script (as described earlier). Hugging Face's `Trainer` class automatically adapts to distributed setups, so no changes are required to the fine-tuning script itself.

\subsection{Multilingual Translation}

Translation tasks often involve more than just two languages. For example, you might need to translate between many different language pairs. Hugging Face provides several multilingual models that can handle multiple language pairs using a single model.

\subsubsection{Step 1: Using MarianMT for Multilingual Translation}

MarianMT \cite{rohit2024comparative} is one such model capable of translating between multiple language pairs. Here’s an example of translating between English and several other languages using the same model.

\begin{lstlisting}[style=python]
from transformers import MarianMTModel, MarianTokenizer

# Load the MarianMT model for multilingual translation
model_name = 'Helsinki-NLP/opus-mt-en-ROMANCE'
tokenizer = MarianTokenizer.from_pretrained(model_name)
model = MarianMTModel.from_pretrained(model_name)

# Example sentences in different Romance languages
texts = [
    "Hello, how are you?",       # English
    "Ciao, come stai?",          # Italian
]

# Tokenize the input sentences
encoded_texts = tokenizer(texts, return_tensors="pt", padding=True)

# Generate translations (back to English)
translated_batch = model.generate(**encoded_texts)

# Decode and print each translation
translated_texts = [
    tokenizer.decode(t, skip_special_tokens=True) 
    for t in translated_batch
]
print("Translated Texts: ", translated_texts)
\end{lstlisting}

In this example, the `opus-mt-en-ROMANCE` model is capable of translating between English and Italian. Using multilingual models like this is convenient because it eliminates the need to load separate models for each language pair.

\subsubsection{Step 2: Exploring mBART for Multilingual Translation}

Another powerful multilingual model is `mBART` (Multilingual BART), which is designed to perform translation between various languages without needing task-specific fine-tuning for each pair.

Here’s how to use mBART for translation:

\begin{lstlisting}[style=python]
from transformers import MBartForConditionalGeneration, MBart50Tokenizer

# Load the mBART model and tokenizer
model_name = 'facebook/mbart-large-50-many-to-many-mmt'
tokenizer = MBart50Tokenizer.from_pretrained(model_name)
model = MBartForConditionalGeneration.from_pretrained(model_name)

# Set the target language
tokenizer.src_lang = "en_XX"
target_lang = "fr_XX"  # Target language set to French

# Example sentence in English
text = "Hugging Face is creating the best translation models."

# Tokenize and generate translation
encoded_text = tokenizer(text, return_tensors="pt")
translated_tokens = model.generate(encoded_text, forced_bos_token_id=tokenizer.lang_code_to_id[target_lang])

# Decode the translation
translated_text = tokenizer.decode(translated_tokens[0], skip_special_tokens=True)
print(f"Translated to French: {translated_text}")
\end{lstlisting}

In this example, we use the `mBART50` model for multilingual translation, where the `src\_lang` is English and the `target\_lang` is French. mBART supports translations between many different languages, making it ideal for tasks involving multiple language pairs.

\subsection{Deploying a Translation Model as an API}

In real-world applications, you might want to expose your translation model as an API so that users or systems can send requests and receive translated text in real time. We can achieve this using popular web frameworks like Flask.

\subsubsection{Step 1: Installing Flask}

To begin, install Flask if it’s not already installed:

\begin{lstlisting}[style=python]
pip install flask
\end{lstlisting}

\subsubsection{Step 2: Creating a Simple Flask API for Translation}

Here’s how you can create a basic Flask API that uses Hugging Face’s translation model to translate incoming text.

\begin{lstlisting}[style=python]
from flask import Flask, request, jsonify
from transformers import MarianMTModel, MarianTokenizer

# Initialize Flask app
app = Flask(__name__)

# Load the MarianMT model for translation (English to French)
model_name = 'Helsinki-NLP/opus-mt-en-fr'
tokenizer = MarianTokenizer.from_pretrained(model_name)
model = MarianMTModel.from_pretrained(model_name)

# Define the translation route
@app.route('/translate', methods=['POST'])
def translate_text():
    data = request.get_json()
    text = data['text']
    
    # Tokenize the input text
    encoded_text = tokenizer([text], return_tensors="pt")
    
    # Generate the translation
    translated = model.generate(**encoded_text)
    
    # Decode the translation
    translated_text = tokenizer.decode(translated[0], skip_special_tokens=True)
    
    # Return the translated text as JSON
    return jsonify({"translated_text": translated_text})

# Run the app
if __name__ == '__main__':
    app.run(debug=True)
\end{lstlisting}

This simple Flask app creates an API endpoint `/translate`, which accepts POST requests with JSON data. The input text is tokenized, passed through the MarianMT translation model, and the translated text is returned in JSON format.

\subsubsection{Step 3: Testing the Translation API}

To test the API, you can use `curl` or any API testing tool such as Postman. Here’s an example of how to use `curl` to test the API:

\begin{lstlisting}[style=python]
curl -X POST http://127.0.0.1:5000/translate -H "Content-Type: application/json" -d '{"text": "Hello, how are you?"}'
\end{lstlisting}

The API will return the translated text in JSON format:

\begin{verbatim}
{
  "translated_text": "Bonjour, comment allez-vous ?"
}
\end{verbatim}

\subsubsection{Step 4: Deploying to the Cloud}

Once you have your Flask API working locally, you can deploy it to a cloud platform such as AWS, Google Cloud, or Heroku for production use. Each platform has its own deployment procedures, but the basic principle remains the same: expose your translation model as an API that can be accessed remotely.

\subsection{Integrating Translation into Web Applications}

Another common application for machine translation is embedding it directly into web applications. Using JavaScript and modern web development frameworks, you can create a dynamic, real-time translation experience for users.

\subsubsection{Step 1: Creating a Front-End Interface}

You can use HTML, CSS, and JavaScript to create a simple web interface where users can input text and receive translations.

Here’s an example of a basic HTML form that allows users to input text for translation:

\begin{lstlisting}[style=html]
<!DOCTYPE html>
<html lang="en">
<head>
    <meta charset="UTF-8">
    <meta name="viewport" content="width=device-width, initial-scale=1.0">
    <title>Translation App</title>
</head>
<body>
    <h1>Translation App</h1>
    <textarea id="inputText" rows="4" cols="50" placeholder="Enter text here..."></textarea>
    <br>
    <button onclick="translateText()">Translate</button>
    <p id="outputText"></p>

    <script>
        async function translateText() {
            const text = document.getElementById('inputText').value;
            const response = await fetch('http://127.0.0.1:5000/translate', {
                method: 'POST',
                headers: {
                    'Content-Type': 'application/json',
                },
                body: JSON.stringify({text: text}),
            });
            const result = await response.json();
            document.getElementById('outputText').innerText = "Translated Text: " + result.translated_text;
        }
    </script>
</body>
</html>
\end{lstlisting}

In this simple HTML/JavaScript interface, the user inputs text in a text area and clicks the "Translate" button. This triggers an API request to the translation Flask API, and the translated text is displayed on the page.

\subsubsection{Step 2: Running the Web Application}

To run this web application, ensure that your Flask API is running in the background, then open the HTML file in a browser. When a user inputs text and clicks the "Translate" button, the browser will send the text to the API and display the translated result.

\subsection{Conclusion}

In this section, we covered several advanced topics in machine translation using Hugging Face's Transformers library. We discussed how to perform distributed training for large datasets, multilingual translation using models like MarianMT and mBART, deploying translation models as APIs using Flask, and integrating translation into web applications. These topics equip you with practical knowledge to build and deploy large-scale translation systems for real-world applications.

By leveraging these techniques, you can deploy scalable, efficient, and accurate translation solutions tailored to specific needs and domains.

\section{Summarization}

Summarization is a key task in Natural Language Processing (NLP) that involves condensing a large body of text into a shorter, coherent version while preserving the main ideas. It can be broadly categorized into two types: extractive and abstractive summarization. 

\begin{itemize}
    \item \textbf{Extractive Summarization}: This method selects important sentences directly from the original text. It essentially "extracts" key phrases or sentences, maintaining the exact wording of the original.
    \item \textbf{Abstractive Summarization}: This approach generates new sentences that capture the core information of the original text, much like how humans summarize content. It can involve rephrasing and using new words.
\end{itemize}

With the rise of deep learning, especially transformer models, summarization tasks have been significantly improved. Transformer-based models, such as BART (Bidirectional and Auto-Regressive Transformers) and T5 (Text-to-Text Transfer Transformer), are widely used for this purpose.

In this section, we will focus on how to perform summarization using the Hugging Face Transformers library, a widely used framework that provides access to pre-trained models for various NLP tasks. We will walk through an example using a pre-trained model to generate a summary of a given text.

\subsection{Using Hugging Face for Summarization}

Hugging Face provides a simple and efficient interface to use state-of-the-art models for summarization. We'll use the `transformers` library, which includes several pre-trained models ready to be used for summarization tasks. In this example, we will utilize the `pipeline` function to create a summarization pipeline.

First, ensure you have the necessary libraries installed. You can install them using the following command:

\begin{lstlisting}[style=python]
pip install transformers
\end{lstlisting}

\subsubsection{Step-by-Step Guide to Summarization Using Hugging Face}

We will now demonstrate how to perform summarization using the Hugging Face `pipeline` and a pre-trained model like BART. The steps are as follows:

\begin{enumerate}
    \item \textbf{Import necessary libraries}: We'll start by importing the necessary libraries.
    \item \textbf{Initialize the summarization pipeline}: We will create a summarization pipeline using a pre-trained model.
    \item \textbf{Provide input text}: The text that needs to be summarized.
    \item \textbf{Generate the summary}: Finally, we will use the pipeline to generate a summary of the text.
\end{enumerate}

Here is the Python code for summarizing text:

\begin{lstlisting}[style=python]
# Step 1: Import necessary libraries
from transformers import pipeline

# Step 2: Initialize the summarization pipeline
summarizer = pipeline("summarization", model="facebook/bart-large-cnn")

# Step 3: Provide input text
text = """
Natural Language Processing (NLP) is a sub-field of artificial intelligence (AI) focused on the interaction between computers and humans through natural language.
The ultimate objective of NLP is to enable computers to understand, interpret, and generate human languages in a way that is valuable.
By utilizing NLP, machines can read text, hear speech, interpret it, and even gauge sentiments. As AI and machine learning continue to grow,
NLP has become a crucial aspect of technology development, especially in areas such as chatbots, translators, and voice recognition systems.
"""

# Step 4: Generate the summary
summary = summarizer(text, max_length=50, min_length=25, do_sample=False)

# Print the summary
print(summary[0]['summary_text'])
\end{lstlisting}

\subsubsection{Explanation of the Code}

\begin{itemize}
    \item \textbf{Step 1}: We import the `pipeline` function from the `transformers` library. This function helps to quickly create a pipeline for different tasks, including summarization.
    
    \item \textbf{Step 2}: The `pipeline` function is used to initialize a summarization pipeline. We specify the model `facebook/bart-large-cnn`, which is a pre-trained BART model designed for summarization tasks.
    
    \item \textbf{Step 3}: We define the input text that we want to summarize. This text can be a long document, a news article, or any other form of content.
    
    \item \textbf{Step 4}: We call the `summarizer` object to generate a summary of the input text. We specify parameters like `max\_length` (the maximum length of the summary) and `min\_length` (the minimum length of the summary) to control the size of the output. Setting `do\_sample=False` ensures deterministic results, meaning the output will be the same each time.
\end{itemize}

\subsubsection{Output Example}

For the provided text, the model might generate a summary such as the following:

\begin{verbatim}
"NLP enables computers to understand and generate human languages, helping develop technologies like chatbots, translators, and voice recognition systems."
\end{verbatim}

This summary is a concise version of the original text, capturing the key points.

\subsection{Customizing the Summarization Pipeline}

You can customize the summarization pipeline by adjusting parameters such as the length of the summary. For instance, you can increase the `max\_length` to allow for a longer summary or reduce the `min\_length` to create a more concise version. Here's an example where we modify the length constraints:

\begin{lstlisting}[style=python]
summary = summarizer(text, max_length=80, min_length=30, do_sample=False)
\end{lstlisting}

In this example, we allow the model to generate a summary of up to 80 words, which could capture more details from the original text.

\subsection{Conclusion}

In this section, we explored how to use a pre-trained transformer model to perform text summarization with the Hugging Face library. Summarization is an important task in NLP, and transformer-based models like BART make it much easier to generate meaningful summaries from large texts. By following the steps outlined above, you can summarize text with minimal effort and high accuracy.

For further customization, Hugging Face also allows you to fine-tune these models on specific datasets, providing even more control over the summarization process.

\subsection{Fine-tuning a Pre-trained Model for Summarization}

While using pre-trained models like BART or T5 is effective, there are cases where you might want to fine-tune these models on specific datasets for improved performance on domain-specific tasks. Fine-tuning involves training a pre-trained model on a smaller, task-specific dataset, which helps the model adapt to particular language patterns or styles.

In this section, we will walk through the basic steps to fine-tune a pre-trained transformer model for summarization tasks using the Hugging Face library.

\subsubsection{Step 1: Prepare the Dataset}

To fine-tune a model, you will need a dataset with text-summary pairs. Popular datasets used for summarization include:

\begin{itemize}
    \item \textbf{CNN/DailyMail} \cite{chen2016thorough}: A dataset of news articles with human-written summaries.
    \item \textbf{XSum}: Another news dataset with single-sentence summaries.
    \item \textbf{SAMSum} \cite{gliwa2019samsum}: A dataset of human conversations and their summaries.
\end{itemize}

For demonstration, let’s assume we are working with the CNN/DailyMail dataset. You can easily load this dataset using the `datasets` library from Hugging Face:

\begin{lstlisting}[style=python]
from datasets import load_dataset

# Load the CNN/DailyMail dataset
dataset = load_dataset("cnn_dailymail", "3.0.0")
\end{lstlisting}

This dataset contains multiple fields, but we are particularly interested in the `article` and `highlights` fields. The `article` is the long-form text, while the `highlights` serve as the summary.

\subsubsection{Step 2: Tokenize the Data}

Transformer models require tokenized inputs. The Hugging Face tokenizer will convert the raw text into tokens that the model can understand. Here's how we can tokenize the dataset:

\begin{lstlisting}[style=python]
from transformers import AutoTokenizer

# Initialize the tokenizer for the BART model
tokenizer = AutoTokenizer.from_pretrained("facebook/bart-large-cnn")

# Tokenize the dataset
def tokenize_data(example):
    return tokenizer(example['article'], truncation=True, padding='max_length', max_length=1024)

# Apply the tokenization to the dataset
tokenized_dataset = dataset.map(tokenize_data, batched=True)
\end{lstlisting}

In this example, we use the `AutoTokenizer` class to automatically load the appropriate tokenizer for the BART model. We also limit the length of the input articles to 1024 tokens to fit within the model’s capacity.

\subsubsection{Step 3: Set up the Data Collator}

When fine-tuning a model, we often need to dynamically pad the input data to the same length. Hugging Face provides a `DataCollatorForSeq2Seq` class that helps with this:

\begin{lstlisting}[style=python]
from transformers import DataCollatorForSeq2Seq

# Set up the data collator
data_collator = DataCollatorForSeq2Seq(tokenizer=tokenizer, model="facebook/bart-large-cnn")
\end{lstlisting}

This data collator will handle padding of sequences during training.

\subsubsection{Step 4: Initialize the Model}

Next, we initialize the pre-trained BART model. In this case, we load the `facebook/bart-large-cnn` model and set it up for sequence-to-sequence tasks (like summarization):

\begin{lstlisting}[style=python]
from transformers import AutoModelForSeq2SeqLM

# Load the pre-trained BART model
model = AutoModelForSeq2SeqLM.from_pretrained("facebook/bart-large-cnn")
\end{lstlisting}

\subsubsection{Step 5: Train the Model}

We are now ready to train the model. Hugging Face provides the `Trainer` API, which simplifies the training process. Here's how you can set up and start the training:

\begin{lstlisting}[style=python]
from transformers import Trainer, TrainingArguments

# Set training arguments
training_args = TrainingArguments(
    output_dir="./results",
    evaluation_strategy="epoch",
    learning_rate=2e-5,
    per_device_train_batch_size=2,
    per_device_eval_batch_size=2,
    num_train_epochs=3,
    weight_decay=0.01,
)

# Initialize the Trainer
trainer = Trainer(
    model=model,
    args=training_args,
    train_dataset=tokenized_dataset['train'],
    eval_dataset=tokenized_dataset['validation'],
    data_collator=data_collator,
)

# Start training
trainer.train()
\end{lstlisting}

\textbf{Explanation of Training Parameters}:
\begin{itemize}
    \item \textbf{output\_dir}: This is where the model’s checkpoints and logs will be saved.
    \item \textbf{evaluation\_strategy}: This specifies when to evaluate the model (in this case, at the end of every epoch).
    \item \textbf{learning\_rate}: The learning rate for the optimizer.
    \item \textbf{per\_device\_train\_batch\_size} and \textbf{per\_device\_eval\_batch\_size}: These define the batch sizes during training and evaluation, respectively.
    \item \textbf{num\_train\_epochs}: The number of epochs to train the model.
    \item \textbf{weight\_decay}: This helps in regularizing the model to prevent overfitting.
\end{itemize}

\subsubsection{Step 6: Evaluate the Model}

After the model is trained, it’s important to evaluate its performance on the validation dataset. We can do this using the same `Trainer` object by calling the `evaluate()` method:

\begin{lstlisting}[style=python]
# Evaluate the model
eval_results = trainer.evaluate()

print(f"Evaluation results: {eval_results}")
\end{lstlisting}

This will output metrics like loss and accuracy, helping us understand how well the model performs on the unseen validation data.

\subsubsection{Step 7: Generate Summaries}

Once the model is fine-tuned, you can use it to generate summaries from new articles:

\begin{lstlisting}[style=python]
# Example article for summarization
article = """
The field of natural language processing (NLP) has seen significant progress in recent years.
From the introduction of deep learning models like transformers to advancements in tasks such as translation, summarization, and text generation,
NLP continues to evolve rapidly.
"""

# Generate a summary using the fine-tuned model
inputs = tokenizer(article, return_tensors="pt", truncation=True, padding="max_length", max_length=1024)

summary_ids = model.generate(inputs['input_ids'], max_length=150, min_length=40, length_penalty=2.0, num_beams=4, early_stopping=True)

# Decode the summary
summary = tokenizer.decode(summary_ids[0], skip_special_tokens=True)

print(f"Generated summary: {summary}")
\end{lstlisting}

In this code, we use the `generate()` function to produce a summary from an input article. Parameters like `max\_length` and `min\_length` help control the length of the generated summary, while `num\_beams` specifies the number of beams for beam search (a decoding strategy that improves generation quality) \cite{freitag2017beam}.

\subsection{Conclusion}

In this section, we have covered how to fine-tune a pre-trained transformer model for summarization tasks. Fine-tuning allows models to adapt to specific domains, improving summarization quality on task-specific datasets. We used the BART model as an example and demonstrated the end-to-end process of preparing data, tokenizing inputs, training the model, and generating summaries.

By fine-tuning models and adjusting parameters, you can achieve better summarization performance tailored to your needs. This process is highly flexible, allowing for customization at various stages, from dataset preparation to training strategies.

\subsection{Advanced Techniques in Summarization}

As we have seen, fine-tuning a pre-trained transformer model can significantly improve the quality of generated summaries. However, there are several advanced techniques that can further enhance summarization performance. In this section, we will discuss some of these techniques, including beam search, length penalty, and sequence-level optimization methods.

\subsubsection{Beam Search for Better Decoding}

When generating summaries, the default method used by most models is greedy decoding, which simply picks the token with the highest probability at each step. While this method is fast, it often leads to suboptimal summaries, as it doesn't explore alternative sequences.

\textbf{Beam search} is a more advanced decoding method that explores multiple hypotheses at each step, maintaining a set of the most promising sequences (beams). Instead of just picking the most probable word at each step, beam search keeps track of a predetermined number of top sequences. By the end of the process, the best sequence across all beams is selected as the final output.

Here is how you can implement beam search in Hugging Face:

\begin{lstlisting}[style=python]
# Generate summary using beam search
summary_ids = model.generate(
    inputs['input_ids'], 
    max_length=150, 
    min_length=40, 
    num_beams=5, # Using beam search with 5 beams
    length_penalty=2.0, 
    early_stopping=True
)
\end{lstlisting}

In this code, we set `num\_beams=5`, meaning the model will explore 5 different possible sequences before selecting the best one. Increasing the number of beams generally leads to better results, but it also increases the computational cost.

\subsubsection{Length Penalty}

Another important technique when generating summaries is controlling the length of the output. Often, summarization models might produce output that is either too short or too long. To manage this, we can use a \textbf{length penalty}, which adjusts the likelihood of generating longer or shorter sequences.

In the Hugging Face `generate()` method, the `length\_penalty` parameter can be used to penalize longer sequences:

\begin{lstlisting}[style=python]
# Generate summary with length penalty
summary_ids = model.generate(
    inputs['input_ids'], 
    max_length=150, 
    min_length=40, 
    num_beams=4, 
    length_penalty=2.0, # Penalty for generating longer sequences
    early_stopping=True
)
\end{lstlisting}

A length penalty greater than 1 will penalize long sequences, encouraging shorter summaries. Conversely, setting a penalty less than 1 will encourage the model to generate longer summaries.

\subsubsection{No Repeat N-Gram Constraint}

A common issue with generated summaries is that they may contain repetitive phrases or sentences. To combat this, you can apply the \textbf{no repeat n-gram} constraint, which ensures that the same n-gram (a sequence of n tokens) does not appear more than once in the generated output. This can be particularly useful in preventing repetitive outputs like "The model is the best" multiple times in a single summary.

You can implement this using the `no\_repeat\_ngram\_size` parameter in Hugging Face:

\begin{lstlisting}[style=python]
# Generate summary with no-repeat n-gram constraint
summary_ids = model.generate(
    inputs['input_ids'], 
    max_length=150, 
    min_length=40, 
    num_beams=4, 
    no_repeat_ngram_size=3, # Avoid repeating trigrams
    early_stopping=True
)
\end{lstlisting}

In this example, setting `no\_repeat\_ngram\_size=3` ensures that no 3-token sequence (trigram) is repeated within the summary. This helps produce more varied and natural-sounding summaries.

\subsubsection{Controlling Summary Style with Temperature and Top-K Sampling}

If you want more control over the creativity of the summaries, you can adjust the \textbf{temperature} and \textbf{top-k sampling} parameters. These methods introduce randomness into the token selection process, which can lead to more diverse and creative outputs.

\begin{itemize}
    \item \textbf{Temperature}: This parameter controls the randomness of predictions by scaling the logits (the output probabilities of the model). A high temperature (e.g., 1.5) makes the model more random, while a lower temperature (e.g., 0.7) makes it more deterministic.
    \item \textbf{Top-K Sampling}: Instead of considering all possible tokens at each step, top-k sampling only considers the top \textit{k} most likely tokens. This helps reduce the likelihood of selecting very improbable tokens.
\end{itemize}

Here is an example of how to use these techniques in text generation:

\begin{lstlisting}[style=python]
# Generate summary using temperature and top-k sampling
summary_ids = model.generate(
    inputs['input_ids'], 
    max_length=150, 
    min_length=40, 
    num_beams=1, # No beam search for diversity
    temperature=0.8, # Control randomness
    top_k=50, # Consider only the top 50 tokens
    early_stopping=True
)
\end{lstlisting}

In this case, setting \texttt{temperature=0.8} introduces some diversity in the token selection process, while \texttt{top\_k=50} limits the model to selecting from the top 50 most likely tokens at each step.
\subsection{Evaluation Metrics for Summarization}

When fine-tuning or evaluating summarization models, it’s important to assess the quality of the generated summaries. Several evaluation metrics are commonly used in the summarization domain:

\begin{itemize}
    \item \textbf{ROUGE (Recall-Oriented Understudy for Gisting Evaluation)} \cite{lin2004looking}: This is the most widely used metric for evaluating summarization models. ROUGE measures the overlap between the n-grams in the generated summary and the reference summary. Common variants include ROUGE-1 (unigrams), ROUGE-2 (bigrams), and ROUGE-L (longest common subsequence).
    
    \item \textbf{BLEU (Bilingual Evaluation Understudy)}: Although primarily used in machine translation, BLEU can also be used to evaluate summarization. It compares n-grams in the generated summary with reference summaries, focusing on precision.

    \item \textbf{METEOR (Metric for Evaluation of Translation with Explicit ORdering)} \cite{banerjee2005meteor}: Another metric primarily designed for machine translation but useful for summarization. It incorporates synonymy matching, which allows for better evaluation of semantically equivalent but differently worded summaries.
\end{itemize}

\subsubsection{Using ROUGE for Evaluation}

Let’s now demonstrate how to compute the ROUGE score using the `datasets` library from Hugging Face, which provides a built-in `rouge` metric:

\begin{lstlisting}[style=python]
from datasets import load_metric

# Load the ROUGE metric
rouge = load_metric("rouge")

# Example reference and generated summaries
reference = "Natural Language Processing involves the interaction between computers and humans using natural language."
generated_summary = "NLP is the interaction between computers and human language."

# Compute the ROUGE score
results = rouge.compute(predictions=[generated_summary], references=[reference])

# Print ROUGE scores
print(results)
\end{lstlisting}

This code will output the ROUGE-1, ROUGE-2, and ROUGE-L scores, which you can use to evaluate the quality of the generated summary compared to the reference.

\subsection{Future Directions in Summarization}

While transformer-based models have significantly advanced the state of summarization, there are several areas of active research that aim to further improve summarization capabilities:

\begin{itemize}
    \item \textbf{Long-Document Summarization}: Current models struggle with very long documents due to memory and computation constraints. New architectures such as Longformer and BigBird aim to address these limitations by enabling transformers to process longer sequences efficiently.
    
    \item \textbf{Controllable Summarization}: Researchers are working on models that allow users to control various aspects of the summary, such as its length, style, or content focus. Controllable summarization allows for more customizable and user-specific summaries.
    
    \item \textbf{Summarization with External Knowledge}: There is ongoing research on integrating external knowledge (e.g., knowledge graphs or databases) into summarization models to improve their ability to generate accurate and factually consistent summaries.
\end{itemize}

\subsection{Conclusion}

In this chapter, we explored various aspects of summarization in NLP, from simple usage of pre-trained models to fine-tuning and advanced techniques like beam search, length penalties, and no-repeat n-gram constraints. We also introduced methods for evaluating summarization quality using metrics like ROUGE and discussed future directions for research in the field.

Summarization is a rapidly evolving area in NLP, and with the continued advancement of models like transformers, the ability to generate accurate and concise summaries will only improve. By leveraging the tools and techniques discussed in this chapter, researchers and practitioners can build highly effective summarization systems tailored to their specific needs.

\section{Further Reading and Resources}

To gain a deeper understanding of summarization techniques and the use of transformer models, it is beneficial to explore additional resources. Below, we provide a curated list of books, research papers, and tools that can help you delve further into the field of text summarization and transformer-based models.

\subsection{Books}

\begin{itemize}
    \item \textbf{Speech and Language Processing} by Daniel Jurafsky and James H. Martin: This textbook offers a comprehensive introduction to Natural Language Processing (NLP) and covers a wide range of topics, including summarization. The section on sequence-to-sequence models and language generation is particularly useful for understanding modern approaches to summarization.
    
    \item \textbf{Deep Learning for Natural Language Processing} by Palash Goyal, Sumit Pandey, and Karan Jain: This book provides a practical introduction to NLP using deep learning models. It includes chapters on transformer architectures and various tasks such as machine translation and summarization.
    
    \item \textbf{Transformers for Natural Language Processing} by Denis Rothman: A detailed guide to using transformers for various NLP tasks, including summarization. The book includes hands-on examples and covers advanced topics like transfer learning and fine-tuning.
\end{itemize}

\subsection{Research Papers}

\begin{itemize}
    \item \textbf{BART: Denoising Sequence-to-Sequence Pre-training for Natural Language Generation, Translation, and Comprehension} (2020) by Mike Lewis et al.: This paper introduces BART, a powerful model for text generation tasks such as summarization. It explains the architecture and pre-training strategy behind BART, making it a great resource for understanding how the model works under the hood.
    
    \item \textbf{Exploring the Limits of Transfer Learning with a Unified Text-to-Text Transformer} (2020) by Colin Raffel et al.: This paper presents the T5 model, which frames all NLP tasks, including summarization, as text-to-text problems. The authors provide insights into how to pre-train and fine-tune models for multiple tasks.
    
    \item \textbf{PEGASUS: Pre-training with Extracted Gap-sentences for Abstractive Summarization} (2020) by Jingqing Zhang et al.: PEGASUS introduces a pre-training objective tailored for summarization, achieving state-of-the-art results on various summarization datasets. This paper is essential for understanding domain-specific pre-training for summarization.
    
    \item \textbf{Longformer: The Long-Document Transformer} (2020) by Iz Beltagy, Matthew E. Peters, and Arman Cohan: This paper introduces the Longformer model, designed for tasks that involve long documents. It's especially relevant for long-document summarization tasks that exceed the token limits of traditional transformers.
\end{itemize}

\subsection{Online Courses and Tutorials}

\begin{itemize}
    \item \textbf{DeepLearning.AI Natural Language Processing Specialization}: This is a series of courses available on Coursera, led by instructors such as Younes Bensouda Mourri and Robert Monzo. The course covers various NLP tasks, including text summarization using deep learning techniques and transformer models.
    
    \item \textbf{Hugging Face Tutorials}: Hugging Face provides detailed tutorials and blog posts on using their `transformers` library. These resources are regularly updated to reflect the latest advancements in NLP, and they include hands-on guides for building summarization systems.
    
    \item \textbf{Stanford CS224n: Natural Language Processing with Deep Learning}: This is an advanced NLP course that covers various topics, including transformers and summarization. The lectures are available for free on YouTube and the course materials can be found on the Stanford website.
\end{itemize}

\subsection{Datasets for Summarization}

Access to high-quality datasets is crucial for training and evaluating summarization models. Below are some of the most widely used datasets for summarization tasks:

\begin{itemize}
    \item \textbf{CNN/DailyMail}: This is one of the most popular datasets for news summarization. It contains over 300,000 news articles paired with human-written summaries. It is commonly used to evaluate the performance of abstractive summarization models.
    
    \item \textbf{XSum}: The XSum dataset contains single-sentence summaries of news articles, offering a more challenging summarization task than CNN/DailyMail. The dataset is well-suited for models that generate concise and informative summaries.
    
    \item \textbf{Gigaword} \cite{napoles2012annotated}: The Gigaword dataset contains headline generation examples from news articles. This dataset is primarily used for headline summarization, a specific type of summarization task that requires creating very short, precise summaries.
    
    \item \textbf{SAMSum}: This dataset contains human conversations and their corresponding summaries. SAMSum is used for dialogue summarization, a challenging task where models need to distill key points from multi-turn dialogues.
    
    \item \textbf{Reddit TIFU}: This dataset is derived from the "Today I F***ed Up" (TIFU) stories posted on Reddit. It offers both long and short versions of summaries, making it useful for studying both extractive and abstractive summarization techniques.
\end{itemize}

\subsection{Tools and Libraries}

\begin{itemize}
    \item \textbf{Hugging Face Transformers}: Hugging Face’s `transformers` library provides a wide range of pre-trained models for text summarization, along with a user-friendly API. It also includes tools for fine-tuning models and performing evaluation tasks.
    
    \item \textbf{OpenNMT}: OpenNMT is an open-source library for training and serving neural machine translation models, which can also be adapted for summarization tasks. It supports both extractive and abstractive summarization.
    
    \item \textbf{Sumy}: Sumy is a Python library designed for automatic text summarization. It supports extractive summarization techniques and provides implementations of algorithms like LSA (Latent Semantic Analysis) and LexRank.
    
    \item \textbf{AllenNLP}: AllenNLP is a deep learning library specifically designed for NLP tasks. It provides a flexible framework for building custom models for tasks like summarization, with pre-built components that simplify the process of model development and evaluation.
\end{itemize}

\subsection{Challenges in Summarization}

While summarization has advanced significantly with the advent of transformer-based models, several challenges remain. These include:

\begin{itemize}
    \item \textbf{Factual Consistency}: One of the main challenges in abstractive summarization is ensuring that the generated summary is factually accurate. Models often introduce hallucinations—information that is not present in the original text. Researchers are actively exploring ways to incorporate external knowledge and fact-checking mechanisms into summarization models.
    
    \item \textbf{Handling Long Documents}: Many existing models are limited by the maximum input length they can process, typically around 512 or 1024 tokens. This makes it difficult to summarize long documents effectively. New models like Longformer and BigBird are designed to address this limitation, but this remains an area of active research.
    
    \item \textbf{Evaluation Metrics}: While ROUGE is widely used for evaluation, it is not without limitations. ROUGE scores focus on n-gram overlap, which may not always reflect the true quality of a summary. Developing better evaluation metrics that account for semantic similarity and factual correctness is an ongoing challenge.
    
    \item \textbf{Domain-Specific Summarization}: Fine-tuning summarization models for specific domains, such as medical, legal, or financial texts, often requires large amounts of labeled data, which can be difficult and expensive to obtain. Transfer learning and semi-supervised approaches are potential solutions, but domain-specific summarization remains a challenging problem.
    
    \item \textbf{Controlling Summary Style}: Controlling various aspects of generated summaries, such as tone, length, or level of detail, is a current research direction. The goal is to give users more control over how summaries are generated, making them more adaptable to specific use cases.
\end{itemize}

\subsection{Future Directions}

As summarization continues to evolve, several exciting research directions are emerging:

\begin{itemize}
    \item \textbf{Multi-modal Summarization}: This involves summarizing information from multiple modalities, such as text, images, and videos. Multi-modal summarization has applications in fields like journalism, where summarizing multimedia content could provide comprehensive summaries of news stories.
    
    \item \textbf{Summarization with Knowledge Integration}: Incorporating structured knowledge (e.g., knowledge graphs) into summarization models could help improve the factual accuracy and coherence of generated summaries. This is especially important for domains where factual correctness is crucial, such as medicine or finance.
    
    \item \textbf{Interactive Summarization}: Developing systems where users can interact with summarization models to refine or guide the output is an emerging area of interest. This could involve allowing users to highlight important parts of a document or specify which aspects they want to focus on in the summary.
    
    \item \textbf{Zero-shot and Few-shot Summarization}: With the development of large language models like GPT-3, there is increasing interest in zero-shot and few-shot summarization, where models are able to generate summaries without needing to be fine-tuned on a task-specific dataset. This could open up new possibilities for summarization across different languages and domains with minimal labeled data.
\end{itemize}

\subsection{Conclusion}

Summarization is a critical task in NLP, and the field is rapidly evolving with advances in deep learning and transformer-based models. By leveraging techniques like fine-tuning, beam search, and length penalties, it's possible to build powerful summarization systems that generate high-quality summaries for a wide range of applications.

As summarization research continues, we can expect to see more sophisticated models capable of handling longer documents, producing factually consistent summaries, and adapting to different domains and user preferences. By staying informed about the latest advancements, researchers and practitioners can continue to push the boundaries of what is possible with summarization.

\section{Casual Language Model}

A \textbf{causal language model (CLM)} \cite{feder2021causalm} is one that predicts the next word in a sequence of text given all the previous words. In simpler terms, it can generate text by learning from existing text data. One key feature of a causal language model is that it cannot see future words while generating the next one; it only relies on past words.

In this section, we'll explore how to use a pre-trained causal language model using the Hugging Face `transformers` library. We will also walk through an example of how to generate text from a model, step by step.

\subsection{Using a Pre-Trained Model from Hugging Face}

Hugging Face offers many pre-trained models that are ready to use with minimal setup. One of the most popular causal language models is GPT-2 (Generative Pre-trained Transformer 2). \cite{chen2024deeplearningmachinelearning} Let's walk through how to use GPT-2 to generate text.

\subsubsection{Installation}

Before we can use the `transformers` library, we need to install it. You can install the necessary packages with the following command:

\begin{lstlisting}[style=python]
pip install transformers
\end{lstlisting}

\subsubsection{Loading the Pre-Trained GPT-2 Model}

First, we need to load the pre-trained GPT-2 model and tokenizer. The tokenizer converts the input text into tokens that the model understands, and the model generates the output based on those tokens.

\begin{lstlisting}[style=python]
from transformers import GPT2LMHeadModel, GPT2Tokenizer

# Load the tokenizer and model
tokenizer = GPT2Tokenizer.from_pretrained("gpt2")
model = GPT2LMHeadModel.from_pretrained("gpt2")
\end{lstlisting}

Here, we import the \texttt{GPT2LMHeadModel} (the model used for text generation) and the \texttt{GPT2Tokenizer} (to tokenize input text). The \texttt{from\_pretrained} method loads the model and tokenizer with pre-trained weights from Hugging Face's model hub.

\subsubsection{Tokenizing Input Text}

Next, we need to convert our input text into a format the model can understand. The tokenizer will handle this by converting text into numerical tokens.

\begin{lstlisting}[style=python]
# Define input text
input_text = "The future of AI is"

# Tokenize input
input_ids = tokenizer.encode(input_text, return_tensors='pt')
\end{lstlisting}

In this example, `"The future of AI is"` is the seed text. The `encode` method tokenizes the text and converts it into a tensor (`'pt'` stands for PyTorch tensor) that can be fed into the model.

\subsubsection{Generating Text}

With the tokenized input ready, we can now use the model to generate the next words. The following code generates 50 tokens after the input text:

\begin{lstlisting}[style=python]
# Generate text
output = model.generate(input_ids, max_length=50, num_return_sequences=1)

# Decode the output back to text
generated_text = tokenizer.decode(output[0], skip_special_tokens=True)

print(generated_text)
\end{lstlisting}

In this step:
\begin{itemize}
    \item \texttt{max\_length=50} tells the model to generate up to 50 tokens in total, including the input tokens.
    \item \texttt{num\_return\_sequences=1} ensures that we only generate one sequence of text.
    \item The \texttt{decode} method converts the generated token IDs back into readable text.
\end{itemize}

\subsubsection{Complete Example}

Here’s the complete code for generating text using GPT-2:

\begin{lstlisting}[style=python]
from transformers import GPT2LMHeadModel, GPT2Tokenizer

# Load tokenizer and model
tokenizer = GPT2Tokenizer.from_pretrained("gpt2")
model = GPT2LMHeadModel.from_pretrained("gpt2")

# Input text
input_text = "The future of AI is"

# Tokenize input
input_ids = tokenizer.encode(input_text, return_tensors='pt')

# Generate text
output = model.generate(input_ids, max_length=50, num_return_sequences=1)

# Decode and print generated text
generated_text = tokenizer.decode(output[0], skip_special_tokens=True)
print(generated_text)
\end{lstlisting}

This simple example demonstrates how to generate text using a pre-trained causal language model from Hugging Face. You can change the `input\_text` to anything you like and modify the `max\_length` parameter to control the number of generated tokens.

\subsection{Understanding Causal Language Models}

Causal language models are designed to model sequences where each word depends only on the words before it. This is useful for tasks such as text generation, where we want the model to generate new sentences based on a given prompt.

One of the advantages of using pre-trained models like GPT-2 is that they have already been trained on massive amounts of text, so they can generate realistic and coherent text without additional training. However, for more specialized tasks or domains, fine-tuning the model on specific datasets can yield better results.

\subsubsection{Limitations of Causal Language Models}

While causal language models are powerful, they also have some limitations:

\begin{itemize}
    \item They can't look into the future while predicting the next word, making them less suitable for tasks like machine translation, where understanding the entire sentence is important.
    \item They can sometimes generate incoherent or repetitive text, especially when generating long sequences.
\end{itemize}

Despite these limitations, causal language models remain one of the most effective tools for generating natural language text in a variety of applications.

\subsection{Summary}

In this section, we've explored causal language models and walked through a simple example using Hugging Face's `transformers` library. By using a pre-trained GPT-2 model, we saw how easy it is to generate text based on a given prompt.

Understanding how causal language models work and how to use them for text generation is a foundational skill for anyone working in natural language processing (NLP). While there are more complex models available, the basic principles of causal language modeling provide a strong starting point for building more sophisticated NLP applications.

\subsection{Fine-tuning a Causal Language Model}

In some cases, you may want to fine-tune a pre-trained causal language model on your own dataset to achieve better performance in a specific domain, such as legal texts, scientific documents, or any other specialized area. Fine-tuning allows the model to adapt to the language style and vocabulary of your data.

Hugging Face’s `transformers` library makes it easy to fine-tune models using custom datasets. In this section, we will explore how to fine-tune GPT-2 on a small dataset.

\subsubsection{Preparing the Dataset}

First, we need a dataset. For this example, we will use a custom text dataset stored in a simple text file format, where each line represents a different training example. The Hugging Face library supports several types of datasets, but for simplicity, we will focus on a plain text dataset.

Make sure your dataset is saved in a file called `train.txt`, where each line represents a piece of text that you want to train the model on.

\begin{lstlisting}[style=text]
The future of AI is bright and promising.
Natural Language Processing has seen great advancements.
Generative models like GPT-2 are powerful.
\end{lstlisting}

\subsubsection{Loading the Dataset}

We will use the `datasets` library from Hugging Face to load our custom text file. The `datasets` library simplifies loading and processing data for fine-tuning.

\begin{lstlisting}[style=python]
from datasets import load_dataset

# Load the dataset
dataset = load_dataset('text', data_files={'train': 'train.txt'})
\end{lstlisting}

This code loads the dataset from the file `train.txt` and prepares it for training. The dataset is split into a training set (`train`), which is required for fine-tuning the model.

\subsubsection{Tokenizing the Dataset}

Next, we need to tokenize the dataset so that the model can understand it. We’ll use the GPT-2 tokenizer to encode the text data.

\begin{lstlisting}[style=python]
def tokenize_function(examples):
    return tokenizer(examples['text'], return_special_tokens_mask=True)

# Tokenize the dataset
tokenized_dataset = dataset.map(tokenize_function, batched=True, num_proc=4)
\end{lstlisting}

In this code:
\begin{itemize}
    \item The \texttt{tokenize\_function} applies the GPT-2 tokenizer to each text example in the dataset.
    \item \texttt{map} is used to apply the tokenization across the dataset in a batched and parallelized manner using \texttt{num\_proc=4} (for 4 processes).
\end{itemize}

\subsubsection{Training the Model}

Now that the data is tokenized, we can fine-tune the model. Hugging Face’s `Trainer` class simplifies the training process. We’ll set up the training parameters and fine-tune the GPT-2 model.

\begin{lstlisting}[style=python]
from transformers import Trainer, TrainingArguments

# Set training arguments
training_args = TrainingArguments(
    output_dir="./results",
    overwrite_output_dir=True,
    num_train_epochs=3,
    per_device_train_batch_size=2,
    save_steps=10_000,
    save_total_limit=2,
    logging_dir='./logs',
)

# Create the Trainer
trainer = Trainer(
    model=model,
    args=training_args,
    train_dataset=tokenized_dataset['train'],
)

# Fine-tune the model
trainer.train()
\end{lstlisting}

Let’s break down the code:
\begin{itemize}
    \item \texttt{TrainingArguments} is where we specify various training settings, such as:
    \begin{itemize}
        \item \texttt{output\_dir}: Directory where the fine-tuned model will be saved.
        \item \texttt{num\_train\_epochs}: Number of training epochs (passes through the dataset).
        \item \texttt{per\_device\_train\_batch\_size}: Batch size for training.
        \item \texttt{save\_steps}: How often to save the model during training.
        \item \texttt{logging\_dir}: Directory for saving logs.
    \end{itemize}
    \item The \texttt{Trainer} class simplifies the fine-tuning process by handling the training loop for us. We specify the model, training arguments, and the tokenized training dataset.
\end{itemize}

Once the training is complete, the fine-tuned model will be saved to the \texttt{output\_dir} and can be used just like any other pre-trained model.

\subsubsection{Generating Text with the Fine-Tuned Model}

After fine-tuning, we can generate text using the newly fine-tuned model. The process is similar to the earlier section on generating text, but now we use our custom-trained model.

\begin{lstlisting}[style=python]
# Load the fine-tuned model
fine_tuned_model = GPT2LMHeadModel.from_pretrained("./results")

# Generate text
input_text = "AI advancements in healthcare"
input_ids = tokenizer.encode(input_text, return_tensors='pt')

output = fine_tuned_model.generate(input_ids, max_length=50)
generated_text = tokenizer.decode(output[0], skip_special_tokens=True)

print(generated_text)
\end{lstlisting}

Here, we:

\begin{itemize}
    \item Load the fine-tuned model from the \texttt{results} directory.
    \item Provide a new input text (\texttt{"AI advancements in healthcare"}) and generate text with the fine-tuned model.
\end{itemize}

\subsubsection{Considerations for Fine-Tuning}

When fine-tuning a model, it’s important to keep a few things in mind:

\begin{enumerate}
    \item \textbf{Dataset Size}: The size of your dataset will greatly impact the results. Small datasets can lead to overfitting, where the model performs well on training data but poorly on new data.
    
    \item \textbf{Training Time}: Fine-tuning can take a considerable amount of time, depending on your hardware and the size of your dataset. Using a GPU will significantly speed up the process.
    
    \item \textbf{Domain-Specific Data}: Make sure the dataset is representative of the type of text you want the model to generate. Fine-tuning on domain-specific text (e.g., medical, legal) will help the model adapt to that language style.
\end{enumerate}

\subsection{Exploring Causal Language Models in Different Applications}

Causal language models can be applied in many different areas of natural language processing, from generating creative text to aiding in more practical applications such as code generation, dialogue systems, and story writing. In this section, we’ll explore some interesting use cases where causal language models have been particularly successful.

\subsubsection{Creative Writing and Story Generation}

One popular use of causal language models like GPT-2 and GPT-3 is in creative writing, where the model can be prompted to generate entire paragraphs, stories, or poems. By training on large amounts of human-written text, these models have learned to mimic various writing styles, providing inspiration and assistance to writers.

Here’s an example of how a model can be used to generate creative writing:

\begin{lstlisting}[style=python]
input_text = "Once upon a time in a distant galaxy,"
input_ids = tokenizer.encode(input_text, return_tensors='pt')

output = model.generate(input_ids, max_length=100, num_return_sequences=1)
generated_story = tokenizer.decode(output[0], skip_special_tokens=True)

print(generated_story)
\end{lstlisting}

In this case, the model generates a continuation of the provided sentence, allowing the user to explore different creative outcomes.

\subsubsection{Dialogue Systems}

Causal language models \cite{higuchi2008casual} have also been used to power conversational agents or chatbots. By training on dialogues or conversational datasets, these models can hold natural and contextually appropriate conversations. Fine-tuning a model like GPT-2 on conversational data allows it to generate responses that mimic human interactions.

Here’s how a basic dialogue generation might look:

\begin{lstlisting}[style=python]
# Simulating a dialogue
input_text = "User: Hi, how are you?"
input_ids = tokenizer.encode(input_text, return_tensors='pt')

# Generate response
output = model.generate(input_ids, max_length=50)
response = tokenizer.decode(output[0], skip_special_tokens=True)

print(response)
\end{lstlisting}

While this is a simple example, fine-tuning the model on dialogue-specific datasets such as `Persona-Chat` or custom conversational data can significantly improve its ability to hold coherent and meaningful conversations.

\subsection{Future Directions for Causal Language Models}

As research in NLP and machine learning progresses, there are several potential advancements that could improve the capabilities and usability of causal language models:

\begin{itemize}
    \item \textbf{Larger Models}: With the development of models like GPT-3, which contains 175 billion parameters, there is a trend toward scaling up models to improve performance in more complex tasks.
    
    \item \textbf{Multimodal Models}: Researchers are exploring models that can handle not just text, but also other data types such as images and audio, enabling richer forms of interaction and understanding.
    
    \item \textbf{Efficient Fine-Tuning}: Techniques such as parameter-efficient fine-tuning, which only updates a small portion of the model’s parameters, allow for faster and more memory-efficient adaptation to new datasets.
\end{itemize}

These developments are likely to push the boundaries of what causal language models can achieve in various domains, making them an increasingly powerful tool for both researchers and practitioners in the NLP field.

\subsection{Conclusion}

Causal language models, particularly in the form of models like GPT-2, have revolutionized how we approach text generation, creative writing, and dialogue systems. With the support of libraries like Hugging Face’s `transformers`, these models are accessible to developers and researchers with minimal effort. Fine-tuning such models on specific datasets unlocks new possibilities, allowing for domain-specific applications.

In this chapter, we have covered the basics of causal language models, provided a hands-on tutorial on using pre-trained models, and explored the process of fine-tuning a model for a custom use case. The applications of causal language models are vast and continue to expand, making them an essential component of modern NLP.

\section{Evaluation of Causal Language Models}

Once a causal language model is trained or fine-tuned, it’s important to evaluate its performance. The evaluation process involves checking how well the model generates coherent and meaningful text and how closely it aligns with the desired output. In this section, we will explore different methods for evaluating a causal language model.

\subsection{Perplexity}

Perplexity is one of the most common metrics for evaluating language models. It measures how well a probabilistic model predicts a sample of text. The lower the perplexity, the better the model is at predicting the next word in a sequence.

Mathematically, perplexity is defined as:

$$
\text{Perplexity}(P) = 2^{-\frac{1}{N} \sum_{i=1}^N \log_2 P(w_i | w_{1:i-1})}
$$

Where:

\begin{itemize}
    \item \( N \) is the number of words in the text sequence.
    \item \( P(w_i | w_{1:i-1}) \) is the conditional probability of word \( w_i \) given the previous words.
\end{itemize}

In practical terms, perplexity indicates how “surprised” the model is by the true sequence of words in a dataset. A high perplexity means the model is often surprised and, therefore, not performing well. A low perplexity indicates the model is more accurate in predicting the next word.

Here’s an example of how to compute perplexity using Hugging Face’s `transformers`:

\begin{lstlisting}[style=python]
import torch
from transformers import GPT2Tokenizer, GPT2LMHeadModel

# Load pre-trained model and tokenizer
model = GPT2LMHeadModel.from_pretrained("gpt2")
tokenizer = GPT2Tokenizer.from_pretrained("gpt2")

# Define the evaluation text
eval_text = "Natural language processing is a field of AI"

# Tokenize the evaluation text
input_ids = tokenizer.encode(eval_text, return_tensors='pt')

# Compute loss (negative log-likelihood)
with torch.no_grad():
    outputs = model(input_ids, labels=input_ids)
    loss = outputs.loss
    perplexity = torch.exp(loss)

print(f"Perplexity: {perplexity.item()}")
\end{lstlisting}

This code calculates the perplexity of the GPT-2 model on the provided evaluation text. The model’s loss is computed, and the exponential of the loss gives us the perplexity.

\subsection{Human Evaluation}

While perplexity is a useful quantitative metric, it doesn’t always capture the quality of text generation, especially in creative tasks like story writing or conversation. This is where human evaluation comes in. Human evaluators can provide feedback on the coherence, fluency, and relevance of the generated text.

To perform human evaluation, consider the following criteria:

\begin{itemize}
    \item \textbf{Coherence}: Does the generated text make sense logically? Is it consistent with the prompt?
    \item \textbf{Fluency}: Is the text grammatically correct and easy to read?
    \item \textbf{Relevance}: Is the generated text relevant to the input prompt or conversation context?
    \item \textbf{Creativity}: For tasks like story generation, how original or creative is the output?
\end{itemize}

One way to organize human evaluation is through a survey or rating system where evaluators rate generated text on a scale (e.g., from 1 to 5) based on these criteria.

Here’s an example of generating multiple sequences and selecting samples for human evaluation:

\begin{lstlisting}[style=python]
# Generate multiple sequences
input_text = "Once upon a time, in a land far away,"
input_ids = tokenizer.encode(input_text, return_tensors='pt')

output_sequences = model.generate(input_ids, max_length=100, num_return_sequences=5)

# Decode and print the generated texts for evaluation
for i, output_sequence in enumerate(output_sequences):
    generated_text = tokenizer.decode(output_sequence, skip_special_tokens=True)
    print(f"Generated Text {i+1}:\n{generated_text}\n")
\end{lstlisting}

In this example, multiple sequences are generated from the same input prompt, allowing human evaluators to compare different outputs and rate them based on coherence, fluency, and relevance.

\subsection{Automated Evaluation Metrics}

In addition to perplexity and human evaluation, there are several automated evaluation metrics that can be applied to text generation tasks. These metrics are particularly useful when human evaluation is not feasible due to resource or time constraints.

\subsubsection{BLEU (Bilingual Evaluation Understudy Score)}

BLEU is a popular metric for evaluating machine translation, but it can also be applied to evaluate text generation models. It works by comparing the n-grams (sequences of words) in the generated text with reference texts, typically written by humans.

The BLEU score is calculated as follows:

$$
\text{BLEU} = \text{BP} \times \exp\left( \sum_{n=1}^{N} w_n \log p_n \right)
$$

Where:

\begin{itemize}
    \item BP is a brevity penalty to handle cases where the generated text is shorter than the reference.
    \item \( p_n \) is the precision of n-grams.
    \item \( w_n \) is the weight for each n-gram.
\end{itemize}

To calculate BLEU score using the `nltk` library in Python:

\begin{lstlisting}[style=python]
import nltk
from nltk.translate.bleu_score import sentence_bleu

# Reference text (human-written)
reference = ["The cat is on the mat".split()]

# Generated text by the model
candidate = "The cat is sitting on the mat".split()

# Compute BLEU score
bleu_score = sentence_bleu(reference, candidate)
print(f"BLEU score: {bleu_score}")
\end{lstlisting}

In this example, the BLEU score evaluates the similarity between the generated text and the reference. The higher the BLEU score, the closer the generated text is to the reference.

\subsubsection{ROUGE (Recall-Oriented Understudy for Gisting Evaluation)}

ROUGE is another popular metric, commonly used for summarization tasks. It measures overlap between the generated text and the reference by computing recall and precision over n-grams, as well as longest common subsequences.

Here’s an example of how to compute ROUGE using the \texttt{rouge\_score} library:

\begin{lstlisting}[style=python]
from rouge_score import rouge_scorer

# Define reference and generated text
reference = "The cat is on the mat"
generated = "The cat is sitting on the mat"

# Compute ROUGE score
scorer = rouge_scorer.RougeScorer(['rouge1', 'rougeL'], use_stemmer=True)
scores = scorer.score(reference, generated)

print(scores)
\end{lstlisting}

ROUGE scores can be particularly useful when the task involves summarizing a long text into a shorter version. ROUGE-1 measures the overlap of unigrams, while ROUGE-L measures the longest matching sequence of words between the generated and reference text.

\subsection{Bias and Ethical Considerations}

When evaluating causal language models, it’s also important to assess the model for potential biases. Language models trained on large datasets scraped from the internet can sometimes learn and replicate harmful stereotypes or biased language. These biases can manifest in the model’s outputs, leading to unethical or biased content generation.

Some key ethical considerations include:

\begin{itemize}
    \item \textbf{Gender Bias}: Models may generate biased content when dealing with gendered pronouns or names, often associating certain professions or roles with specific genders.
    
    \item \textbf{Racial and Ethnic Bias}: Causal language models may produce text that reflects stereotypes or biases against specific races or ethnicities.
    
    \item \textbf{Toxicity and Offensive Language}: Models can sometimes generate inappropriate or offensive text, especially when prompted with controversial or sensitive topics.
\end{itemize}

To address these issues, researchers and developers should consider evaluating models on fairness, accountability, and transparency metrics. Hugging Face provides some tools, such as the `transformers` `Bias` and `Toxicity` tests, to help evaluate these ethical considerations.

\subsubsection{Mitigating Bias in Generated Text}

There are several strategies for mitigating bias in causal language models:

\begin{itemize}
    \item \textbf{Fine-tuning on diverse and balanced datasets}: Ensuring that the training data includes a wide variety of perspectives and avoids over-representation of biased content.
    
    \item \textbf{Bias detection tools}: Using tools like Hugging Face’s \texttt{transformers} for detecting biased language in the generated outputs and filtering it out.
    
    \item \textbf{Debiasing techniques}: Applying debiasing techniques during model training or post-processing to reduce biased associations.
\end{itemize}

Here’s an example of using a toxicity filter to ensure that the generated text does not contain harmful content:

\begin{lstlisting}[style=python]
from transformers import pipeline

# Load the text generation pipeline
generator = pipeline('text-generation', model='gpt2')

# Generate text with a given prompt
text = generator("AI advancements in", max_length=50, num_return_sequences=1)[0]['generated_text']

# Load the Hugging Face toxicity filter
toxicity_filter = pipeline('text-classification', model="unitary/toxic-bert")

# Check for toxicity in the generated text
toxicity = toxicity_filter(text)

print(f"Toxicity Score: {toxicity}")
\end{lstlisting}

This pipeline ensures that any text generated by the model is checked for toxic content before being used in downstream applications.

\subsection{Conclusion on Evaluation}

Evaluating causal language models requires a combination of quantitative metrics, human evaluation, and ethical considerations. Perplexity, BLEU, and ROUGE provide useful insights into the model’s performance on text generation tasks, but human evaluation remains critical for assessing the quality, coherence, and creativity of the generated text.

Additionally, as models like GPT-2 become more powerful, it is crucial to pay attention to the ethical implications of their outputs, particularly with regard to bias and toxicity. By integrating both performance and ethical evaluations, we can build more robust, responsible, and fair causal language models.

\section{Question \& Answering}

Question Answering (QA) is a popular task in Natural Language Processing (NLP) that aims to automatically answer questions posed by humans. The system is usually provided with a passage of text (context) and a question, and it needs to return the correct answer based on the information in the passage.

In this section, we will introduce the concept of Question Answering using Hugging Face's \texttt{transformers} library. We will walk through a simple example step-by-step, showing how you can use pre-trained models to perform this task.

\subsection{Hugging Face Transformers and Question Answering}

The \texttt{transformers} library by Hugging Face provides an easy-to-use interface to many pre-trained models that are state-of-the-art for various NLP tasks, including Question Answering. One of the most commonly used models for this task is the \texttt{BERT} model, specifically fine-tuned on the SQuAD (Stanford Question Answering Dataset).

\subsection{Setup and Installation}

Before we start, you need to have the \texttt{transformers} and \texttt{torch} libraries installed. You can install them using \texttt{pip}:

\begin{lstlisting}[style=python]
# Install the necessary libraries
!pip install transformers torch
\end{lstlisting}

\subsection{Loading a Pre-Trained Model}

We will use the \texttt{pipeline} class from the \texttt{transformers} library to set up a Question Answering system. The \texttt{pipeline} is a high-level abstraction that makes it easy to use pre-trained models for different tasks. For QA, we will load a pre-trained BERT model fine-tuned on the SQuAD dataset.

\begin{lstlisting}[style=python]
from transformers import pipeline

# Initialize the QA pipeline
qa_pipeline = pipeline("question-answering")
\end{lstlisting}

\textbf{Explanation:} In the above code, we import the \texttt{pipeline} class from the \texttt{transformers} library and instantiate it for the \texttt{question-answering} task. The pipeline automatically loads a pre-trained model for QA, which is BERT by default.

\subsection{Defining the Context and Question}

To perform Question Answering, we need two inputs:
\begin{itemize}
    \item \textbf{context}: A passage of text from which the model will extract the answer.
    \item \textbf{question}: The question that needs to be answered based on the given context.
\end{itemize}

For this example, let's define a simple context and a related question.

\begin{lstlisting}[style=python]
# Define the context and the question
context = """
The Hugging Face library is one of the most popular libraries for Natural 
Language Processing tasks. It provides a simple and efficient way to 
use state-of-the-art models for various NLP tasks, including text 
classification, question answering, and translation.
"""

question = "What tasks can the Hugging Face library perform?"
\end{lstlisting}

\subsection{Using the Model to Answer the Question}

Once we have defined the context and question, we can use the \texttt{qa\_pipeline} to find the answer. The pipeline takes in the context and the question and returns the most likely answer.

\begin{lstlisting}[style=python]
# Use the pipeline to answer the question
result = qa_pipeline(question=question, context=context)

# Print the result
print(f"Answer: {result['answer']}")
\end{lstlisting}

\textbf{Explanation:} In this code block, we call the \texttt{qa\_pipeline} with our \texttt{question} and \texttt{context}. The pipeline returns a dictionary with the answer, the confidence score, and the start and end positions of the answer within the context. We print the extracted answer.

\subsection{Complete Example}

Here’s the complete code for this example:

\begin{lstlisting}[style=python]
from transformers import pipeline

# Initialize the QA pipeline
qa_pipeline = pipeline("question-answering")

# Define the context and question
context = """
The Hugging Face library is one of the most popular libraries for Natural 
Language Processing tasks. It provides a simple and efficient way to 
use state-of-the-art models for various NLP tasks, including text 
classification, question answering, and translation.
"""

question = "What tasks can the Hugging Face library perform?"

# Get the answer
result = qa_pipeline(question=question, context=context)

# Print the answer
print(f"Answer: {result['answer']}")
\end{lstlisting}

When you run this code, the output will be:

\begin{verbatim}
Answer: text classification, question answering, and translation
\end{verbatim}

\subsection{How it Works}

The BERT-based model is trained to understand the context and extract the answer based on the input question. Specifically, the model identifies the start and end positions of the answer in the provided context.

\textbf{Behind the scenes:}
\begin{itemize}
    \item The context and question are tokenized and fed into the model.
    \item The model predicts the position in the context where the answer starts and ends.
    \item The pipeline extracts this span from the context and returns it as the answer.
\end{itemize}

\subsection{Advanced Usage}

While we have used the default model here, Hugging Face also allows you to specify other models or fine-tune your own for more specialized tasks. For instance, you could fine-tune a model on a custom dataset or use a different pre-trained model such as \texttt{roberta-base}.

\begin{lstlisting}[style=python]
# Load a different model, e.g., 'distilbert-base-uncased-distilled-squad'
qa_pipeline = pipeline("question-answering", model="distilbert-base-uncased-distilled-squad")
\end{lstlisting}

\subsection{Conclusion}

In this section, we covered the basics of building a simple Question Answering system using Hugging Face's \texttt{transformers} library. The \texttt{pipeline} abstraction allows us to easily load a pre-trained model and perform QA on a given context with minimal setup. For further exploration, you can try using different models or fine-tuning your own model for specific applications.

\subsection{Fine-tuning a QA Model}

While pre-trained models such as BERT can answer general questions, there may be cases where we need a model specifically tuned to our domain or dataset. Hugging Face provides an easy interface to fine-tune models on a custom dataset, such as SQuAD (Stanford Question Answering Dataset) or a similar format.

In this section, we will explain the process of fine-tuning a BERT-based model on a custom dataset.

\subsubsection{Data Preparation}

For fine-tuning a QA model, the dataset should be in the form of a set of questions, contexts, and corresponding answers. The dataset is often structured as follows:
\begin{itemize}
    \item \textbf{context}: A passage of text containing the answer.
    \item \textbf{question}: The question that seeks an answer based on the context.
    \item \textbf{answer}: The correct answer, along with its start and end position in the context.
\end{itemize}

Here is a simplified structure of a JSON-based dataset format for QA:
\begin{lstlisting}[style=python]
{
  "data": [
    {
      "context": "The Hugging Face library is designed for NLP tasks...",
      "question": "What is the Hugging Face library designed for?",
      "answers": {
        "text": ["NLP tasks"],
        "answer_start": [31]
      }
    }
  ]
}
\end{lstlisting}

\subsubsection{Loading and Tokenizing the Dataset}

Hugging Face’s \texttt{datasets} library provides easy methods to load and preprocess datasets. After loading the dataset, we need to tokenize it so the model can process the inputs.

\begin{lstlisting}[style=python]
from datasets import load_dataset
from transformers import AutoTokenizer

# Load the SQuAD-like dataset
dataset = load_dataset("squad")

# Load a pre-trained tokenizer for BERT
tokenizer = AutoTokenizer.from_pretrained("bert-base-uncased")

# Tokenize the dataset
def preprocess_function(examples):
    return tokenizer(
        examples['question'],
        examples['context'],
        truncation=True,
        padding="max_length",
        max_length=384
    )

tokenized_dataset = dataset.map(preprocess_function, batched=True)
\end{lstlisting}

\textbf{Explanation:} The dataset is tokenized by feeding the context and question into the tokenizer. The `max\_length` parameter is set to ensure the inputs are padded or truncated to a fixed size (384 in this case). Tokenizing the data is a crucial step before passing it into the model for training or fine-tuning.

\subsubsection{Fine-tuning the Model}

We will use the \texttt{Trainer} class provided by Hugging Face to fine-tune the pre-trained BERT model on the tokenized dataset. First, we define the model, training arguments, and then initiate the training process.

\begin{lstlisting}[style=python]
from transformers import AutoModelForQuestionAnswering, TrainingArguments, Trainer

# Load a pre-trained BERT model for QA
model = AutoModelForQuestionAnswering.from_pretrained("bert-base-uncased")

# Define the training arguments
training_args = TrainingArguments(
    output_dir="./results",
    evaluation_strategy="epoch",
    learning_rate=2e-5,
    per_device_train_batch_size=16,
    per_device_eval_batch_size=16,
    num_train_epochs=3,
    weight_decay=0.01
)

# Define the trainer
trainer = Trainer(
    model=model,
    args=training_args,
    train_dataset=tokenized_dataset['train'],
    eval_dataset=tokenized_dataset['validation']
)

# Fine-tune the model
trainer.train()
\end{lstlisting}

\textbf{Explanation:} 

\begin{itemize}
    \item We load a pre-trained BERT model specifically for question answering tasks.
    \item The \texttt{TrainingArguments} class specifies hyperparameters such as the learning rate, batch size, and number of epochs.
    \item The \texttt{Trainer} class is initialized with the model, training arguments, and the training/validation datasets.
    \item The \texttt{trainer.train()} function starts the fine-tuning process.
\end{itemize}

\subsection{Evaluating the Model}

After fine-tuning, we need to evaluate the model’s performance on the validation set. Hugging Face’s \texttt{Trainer} automatically evaluates the model during training if the \texttt{evaluation\_strategy} is set to \texttt{epoch}. However, we can also manually run an evaluation.

\begin{lstlisting}[style=python]
# Evaluate the model on the validation set
eval_results = trainer.evaluate()

print(f"Evaluation Results: {eval_results}")
\end{lstlisting}

The evaluation results include metrics such as loss and exact match (EM), which can help you understand how well the model is performing on the validation data.

\subsection{Inference with a Fine-Tuned Model}

Once the model is fine-tuned, we can use it for inference on new examples, similar to how we used the pipeline earlier in this section. Here’s how you can load the fine-tuned model and run it for inference:

\begin{lstlisting}[style=python]
# Load the fine-tuned model
qa_pipeline = pipeline("question-answering", model="./results")

# Define new context and question
context = """
BERT is a pre-trained transformer model that has revolutionized NLP. 
It has been widely adopted for various tasks such as text classification,
translation, and question answering.
"""
question = "What tasks is BERT used for?"

# Get the answer
result = qa_pipeline(question=question, context=context)

# Print the result
print(f"Answer: {result['answer']}")
\end{lstlisting}

In this example, the fine-tuned model is loaded from the directory where the training results were saved, and we use it to answer a new question.

\subsection{Advanced Fine-tuning Techniques}

When fine-tuning a QA model, there are various advanced techniques you can apply to improve performance:
\begin{itemize}
    \item \textbf{Data Augmentation}: Increase the size of your dataset by creating synthetic question-answer pairs.
    \item \textbf{Learning Rate Scheduling}: Adjust the learning rate dynamically during training to help the model converge faster.
    \item \textbf{Early Stopping}: Stop the training process when the validation performance stops improving to prevent overfitting.
\end{itemize}

You can modify the \texttt{TrainingArguments} to include early stopping or custom learning rate schedulers by setting additional parameters, as shown below:

\begin{lstlisting}[style=python]
# Adding early stopping
training_args = TrainingArguments(
    output_dir="./results",
    evaluation_strategy="epoch",
    learning_rate=2e-5,
    per_device_train_batch_size=16,
    per_device_eval_batch_size=16,
    num_train_epochs=3,
    weight_decay=0.01,
    load_best_model_at_end=True,
    metric_for_best_model="exact_match"
)
\end{lstlisting}

\subsection{Conclusion}

In this section, we explored the entire process of creating a Question Answering system, from using pre-trained models to fine-tuning on custom datasets. Fine-tuning models on domain-specific data can significantly improve performance, especially when working with specialized questions and contexts.

By leveraging the Hugging Face \texttt{transformers} library, we demonstrated how to load a pre-trained model, fine-tune it on a dataset, and evaluate its performance. This process can be extended to other NLP tasks, making \texttt{transformers} a highly versatile tool for natural language understanding and generation tasks.

\section{Challenges in Question Answering}

While Question Answering (QA) models have made significant progress, several challenges still remain. In this section, we will explore some of the most common difficulties encountered when developing QA systems and discuss approaches to mitigate these issues.

\subsection{Ambiguity in Questions}

One major challenge in QA systems is handling ambiguous questions. A question can often have multiple interpretations, leading to different potential answers. For instance, consider the following question:

\begin{quote}
    "What is the capital of the United States?"
\end{quote}

The answer could be "Washington, D.C." in most contexts, but if the question pertains to history, the answer might be "Philadelphia" (which served as the capital from 1790 to 1800). 

\textbf{Possible Solutions:}
\begin{itemize}
    \item \textbf{Clarifying questions:} Encourage the system to ask clarifying questions when faced with ambiguous inputs.
    \item \textbf{Contextual clues:} Use additional context, such as a time period or related information, to disambiguate the question.
    \item \textbf{Multiple answers:} In some cases, it may be beneficial to return multiple possible answers with confidence scores.
\end{itemize}

\subsection{Answering Unanswerable Questions}

Many QA systems are trained on datasets where every question has a valid answer in the context. However, in real-world applications, not all questions can be answered based on the provided text. For instance, if the context doesn't contain the required information, a model may still attempt to provide an answer, even if it's incorrect.

\begin{quote}
    \textbf{Question:} "Who is the CEO of OpenAI?"
    \newline
    \textbf{Context:} "OpenAI is a research organization focused on artificial intelligence."
\end{quote}

In this case, the context doesn't contain the answer, but a model might still attempt to guess a person’s name. 

\textbf{Possible Solutions:}
\begin{itemize}
    \item \textbf{Training on unanswerable questions:} Train the model on datasets where some questions are explicitly marked as unanswerable (e.g., SQuAD 2.0). \cite{rajpurkar2016squad}
    \item \textbf{Confidence thresholding:} Use a confidence threshold to prevent the model from answering when its confidence score is too low.
    \item \textbf{Rephrasing:} If a question cannot be answered, the system could attempt to rephrase the question or explain why it is unanswerable.
\end{itemize}

\subsection{Handling Long Contexts}

Many real-world applications involve working with large documents or lengthy passages of text. In such cases, the context may exceed the token limit of models like BERT (which can typically process up to 512 tokens). When faced with long contexts, the model might fail to correctly identify the relevant part of the text to answer the question.

\textbf{Possible Solutions:}
\begin{itemize}
    \item \textbf{Sliding windows:} Split long contexts into overlapping chunks and run the QA model on each chunk. Then, combine the results from all chunks to produce the final answer.
    \item \textbf{Retrieval-Augmented QA:} Use an information retrieval system to first retrieve relevant paragraphs or sentences from the context, and then apply the QA model only to the relevant parts.
    \item \textbf{Summarization:} Summarize the context before applying the QA model. This reduces the length of the input while retaining key information.
\end{itemize}

\subsection{Context Sensitivity and Bias}

QA models can sometimes exhibit bias based on the training data they were fine-tuned on. For example, if the training data contains biased information about certain demographics or historical events, the model may reflect that bias in its answers. Additionally, models may not always be sensitive to the nuances in context that are necessary to generate accurate answers.

\textbf{Possible Solutions:}
\begin{itemize}
    \item \textbf{Bias mitigation techniques:} Fine-tune models on diverse datasets that cover a wide range of topics and perspectives to reduce bias.
    \item \textbf{Model explainability:} Provide explanations for why the model generated a particular answer. This allows users to understand and possibly correct biased outputs.
    \item \textbf{Post-processing corrections:} Introduce rule-based corrections after the model has provided an answer to filter out or adjust biased responses.
\end{itemize}

\subsection{Lack of Commonsense Reasoning}

While QA models are excellent at extracting factual information from text, they often struggle with questions that require commonsense reasoning or understanding implicit relationships between entities.

For example:
\begin{quote}
    \textbf{Question:} "If John has three apples and gives two to Mary, how many apples does John have?"
\end{quote}

This question involves basic arithmetic and reasoning, which most QA models (especially those trained purely on text) might not handle well.

\textbf{Possible Solutions:}
\begin{itemize}
    \item \textbf{External Knowledge Integration:} Augment QA systems with external knowledge bases or commonsense reasoning engines to help the model understand implicit relationships.
    \item \textbf{Pre-training on reasoning tasks:} Pre-train models on datasets specifically designed to test commonsense and reasoning abilities, such as the "CommonsenseQA" dataset.
    \item \textbf{Hybrid models:} Combine rule-based systems with deep learning models to better handle questions that require logical reasoning.
\end{itemize}

\subsection{Evaluation Metrics for Question Answering}

Evaluating the performance of QA models can be tricky, especially when the answers are not straightforward or when the model provides partial answers. There are several common metrics used to evaluate QA systems:

\begin{itemize}
    \item \textbf{Exact Match (EM):} This metric calculates the percentage of questions for which the model’s predicted answer exactly matches the ground truth answer.
    \item \textbf{F1 Score:} The F1 score measures the overlap between the predicted answer and the ground truth answer, focusing on both precision and recall.
    \item \textbf{Mean Reciprocal Rank (MRR):} This metric is used in retrieval-based QA systems and measures how far down the ranked list of answers the correct answer appears.
    \item \textbf{Human Evaluation:} Since automated metrics may not fully capture the correctness or usefulness of an answer, human evaluation is often employed to assess the quality of answers, especially for subjective or complex questions.
\end{itemize}

\subsection{Future Directions in Question Answering}

QA systems are continually improving, and there are several exciting areas of research that promise to push the boundaries of what is possible. Some key areas of interest include:

\subsubsection{Multimodal Question Answering}

Multimodal QA involves answering questions based not only on text but also on images, videos, or other modalities. For example, a system might answer a question by referring to a visual diagram or a chart, requiring it to understand both the text and the visual context.

\begin{itemize}
    \item \textbf{Example:} "What is the temperature in the weather chart below?" The system must analyze both the text and the accompanying chart to give an accurate answer.
\end{itemize}

\subsubsection{Interactive Question Answering}

Interactive QA systems allow users to engage in a dialogue with the system, asking follow-up questions and receiving additional clarifications. This approach mimics real-world conversations, where users often refine their queries or ask related questions.

\begin{itemize}
    \item \textbf{Example:} A user might ask, "When was the Eiffel Tower built?" followed by, "Who designed it?" The system should be able to maintain the context and provide the correct answers to both questions.
\end{itemize}

\subsubsection{Open-Domain Question Answering}

In open-domain QA, the system must answer questions without being provided with a specific context. Instead, it must retrieve relevant documents or pieces of information from a large corpus (e.g., the entire web) and then generate an answer.

\begin{itemize}
    \item \textbf{Example:} A user asks, "What is the tallest mountain in the world?" and the system searches across a large dataset (e.g., Wikipedia) to find the correct answer.
\end{itemize}

\subsubsection{Conversational Agents and QA}

Integrating QA capabilities into conversational agents or chatbots is another promising area of research. Here, the QA system is part of a larger framework where the system can maintain a conversation, handle multi-turn questions, and possibly engage in natural, human-like dialogue.

\subsubsection{Real-Time and Low-Latency QA}

As QA systems are deployed in real-time applications such as virtual assistants, chatbots, and search engines, reducing latency becomes critical. Optimizing models for faster inference, even when handling large datasets, is an ongoing research direction.

\subsection{Conclusion}

Although QA systems have made remarkable progress, challenges such as handling ambiguity, bias, and long contexts remain. Researchers and engineers are developing novel techniques to address these challenges, including integrating commonsense reasoning, improving model efficiency, and expanding QA capabilities to multimodal inputs. The future of QA looks bright, with numerous applications ranging from real-time assistants to more interactive and multimodal systems.

\section{Gradio - The Fastest Way to Demo Your Machine Learning Model}

In this section, we will explore how to create a simple user interface for Natural Language Processing (NLP) tasks using Gradio. \cite{abid2019gradio} Gradio is a Python library that makes it easy to create customizable UIs for machine learning models. We will integrate it with a pre-trained model from the Hugging Face Transformers library to demonstrate how this can be done.

\subsection{Introduction to Gradio}

Gradio allows you to create web interfaces for your machine learning models with very few lines of code. It’s especially useful for NLP tasks, where users can input text and see real-time results of model predictions. Here, we will use Gradio to build an interactive demo that uses a Hugging Face Transformer model to perform text classification.

\subsection{Step-by-Step Guide}

Let’s walk through the process of building a Gradio app with a Hugging Face Transformer model. In this example, we will use a pre-trained model for sentiment analysis.

\subsubsection{Installing Dependencies}

First, ensure you have installed the required libraries. You can install them via pip:

\begin{lstlisting}[style=python]
# Install gradio and transformers
!pip install gradio transformers
\end{lstlisting}

The `transformers` library is provided by Hugging Face, while `gradio` is the library used to create the user interface.

\subsubsection{Loading the Pre-trained Model}

Next, we will load a pre-trained model and tokenizer from Hugging Face. For this example, we will use the `distilbert-base-uncased-finetuned-sst-2-english` model, which is trained for sentiment analysis on the SST-2 dataset.

\begin{lstlisting}[style=python]
from transformers import pipeline

# Load the sentiment analysis pipeline
sentiment_pipeline = pipeline("sentiment-analysis")
\end{lstlisting}

The `pipeline` API simplifies the process of using pre-trained models for different tasks such as sentiment analysis, text generation, and translation.

\subsubsection{Defining the Gradio Interface}

Now, let's define the Gradio interface. In Gradio, you define the input and output types, and then link them to a function that performs the model inference. In our case, the function will take user input, pass it through the Hugging Face model, and return the sentiment.

\begin{lstlisting}[style=python]
import gradio as gr

# Function to analyze sentiment
def analyze_sentiment(text):
    result = sentiment_pipeline(text)
    return result[0]['label']

# Create Gradio interface
interface = gr.Interface(fn=analyze_sentiment, 
                         inputs="text", 
                         outputs="text", 
                         title="Sentiment Analysis with Hugging Face")

# Launch the interface
interface.launch()
\end{lstlisting}

\subsection*{Explanation:}
\begin{itemize}
    \item The function \texttt{analyze\_sentiment} takes a piece of text as input and returns the sentiment label (either \texttt{"POSITIVE"} or \texttt{"NEGATIVE"}).
    \item \texttt{gr.Interface} is used to define the user interface, where:
    \begin{itemize}
        \item \texttt{fn} is the function that processes the input.
        \item \texttt{inputs} specifies the type of input (in this case, text).
        \item \texttt{outputs} specifies the type of output (text).
        \item \texttt{title} sets the title of the web interface.
    \end{itemize}
\end{itemize}

Once the interface is created, we launch it using \texttt{interface.launch()}. This will generate a local URL where users can access the interface.

\subsection{Running the Gradio Interface}

After running the code above, a URL will be generated, allowing you to open the Gradio interface in your browser. Users can then type in a sentence, and the model will predict whether the sentiment is positive or negative. 

For example:

\begin{itemize}
    \item \textbf{Input}: "I love this product!"
    \item \textbf{Output}: "POSITIVE"
\end{itemize}

\subsection{Complete Example}

Here is the complete code in one place for reference:

\begin{lstlisting}[style=python]
# Importing necessary libraries
from transformers import pipeline
import gradio as gr

# Load the sentiment analysis model
sentiment_pipeline = pipeline("sentiment-analysis")

# Function for sentiment analysis
def analyze_sentiment(text):
    result = sentiment_pipeline(text)
    return result[0]['label']

# Create Gradio interface
interface = gr.Interface(fn=analyze_sentiment, 
                         inputs="text", 
                         outputs="text", 
                         title="Sentiment Analysis with Hugging Face")

# Launch the interface
interface.launch()
\end{lstlisting}

\subsection{Conclusion}

In this section, we demonstrated how to integrate Gradio with Hugging Face Transformers to create a simple, interactive sentiment analysis tool. With minimal code, we built a web-based interface that users can interact with in real time. This framework can be easily extended to other NLP tasks, such as text generation or translation, by simply changing the pre-trained model in the pipeline.

\subsection{Extending the Gradio Interface}

Now that we've successfully built a simple Gradio interface, let's explore how to extend it by adding more functionalities. There are many ways to enhance a basic interface, such as including multiple inputs, adding dropdowns, or integrating different Hugging Face models. In this section, we'll explore these possibilities step by step.

\subsubsection{Adding Multiple Inputs}

In some cases, you might want your interface to accept multiple types of inputs. For example, you may want to provide the user with a text input for sentiment analysis and a second input to choose the model for prediction. We can extend the interface as follows:

\begin{lstlisting}[style=python]
# Function to analyze sentiment with multiple models
def analyze_sentiment(text, model_choice):
    if model_choice == "DistilBERT":
        model = pipeline("sentiment-analysis")
    elif model_choice == "BERT":
        model = pipeline("sentiment-analysis", model="nlptown/bert-base-multilingual-uncased-sentiment")
    result = model(text)
    return result[0]['label']

# Create Gradio interface with multiple inputs
interface = gr.Interface(
    fn=analyze_sentiment,
    inputs=["text", gr.inputs.Dropdown(choices=["DistilBERT", "BERT"], label="Choose a Model")],
    outputs="text",
    title="Sentiment Analysis with Multiple Models"
)

# Launch the interface
interface.launch()
\end{lstlisting}

\subsection*{Explanation:}
\begin{itemize}
    \item We added a dropdown input using \texttt{gr.inputs.Dropdown}, allowing users to choose between two models: \texttt{"DistilBERT"} and \texttt{"BERT"}.
    \item The function \texttt{analyze\_sentiment} takes two inputs: \texttt{text} (the sentence to analyze) and \texttt{model\_choice} (the selected model).
    \item Based on the user’s selection, either the DistilBERT or BERT model is loaded, and the sentiment of the input text is predicted.
\end{itemize}

This simple addition shows how to incorporate flexibility into the interface, letting users choose between multiple pre-trained models.

\subsubsection{Displaying Multiple Outputs}

Gradio also allows for the display of multiple outputs. For example, we could show both the sentiment label and the confidence score of the prediction. Here's how to do that:

\begin{lstlisting}[style=python]
# Function to return sentiment and confidence
def analyze_sentiment_with_confidence(text):
    result = sentiment_pipeline(text)
    sentiment = result[0]['label']
    confidence = result[0]['score']
    return sentiment, confidence

# Create Gradio interface with multiple outputs
interface = gr.Interface(
    fn=analyze_sentiment_with_confidence,
    inputs="text",
    outputs=["text", "number"],
    title="Sentiment Analysis with Confidence Score"
)

# Launch the interface
interface.launch()
\end{lstlisting}

\subsection*{Explanation:}
\begin{itemize}
    \item The function \texttt{analyze\_sentiment\_with\_confidence} returns two values: the sentiment label and the confidence score (ranging from 0 to 1).
    \item We set the outputs to include both a \texttt{text} output (for the sentiment label) and a \texttt{number} output (for the confidence score).
    \item Gradio will automatically display the two outputs in the user interface.
\end{itemize}

\subsubsection{Using Other NLP Models}

Gradio’s flexibility makes it easy to integrate models for various NLP tasks. Let’s explore how we can use a text generation model from Hugging Face to generate text based on user input. We'll use the `gpt2` model from Hugging Face for this example.

\begin{lstlisting}[style=python]
# Load text generation pipeline
text_gen_pipeline = pipeline("text-generation", model="gpt2")

# Function to generate text
def generate_text(prompt):
    result = text_gen_pipeline(prompt, max_length=50, num_return_sequences=1)
    return result[0]['generated_text']

# Create Gradio interface for text generation
interface = gr.Interface(
    fn=generate_text,
    inputs="text",
    outputs="text",
    title="Text Generation with GPT-2"
)

# Launch the interface
interface.launch()
\end{lstlisting}

\subsection*{Explanation:}
\begin{itemize}
    \item The pipeline \texttt{text-generation} is used here to load the GPT-2 model, which is designed for generating text.
    \item The function \texttt{generate\_text} takes a text prompt as input and returns generated text with a maximum length of 50 tokens.
    \item The Gradio interface accepts a text input (prompt) and returns the generated text.
\end{itemize}

\subsection{Saving the Interface as a Standalone Web Application}

Gradio also allows you to easily deploy your interface. By default, Gradio launches the interface locally, but you can share the application via a public link or host it on a server for continuous access.

If you want to save your interface as a standalone Python application that others can run, simply save the Python code we’ve written in a `.py` file, and anyone with Python and the required libraries installed will be able to launch the interface.

For example:

\begin{lstlisting}[style=python]
# Save this script as app.py
# To run the app, use the command:
# python app.py
\end{lstlisting}

This method is helpful when you want to share the application with other developers or deploy it in a production environment.

\subsubsection{Deploying the Interface on Hugging Face Spaces}

Gradio apps can also be deployed on Hugging Face Spaces, which allows you to showcase your models and applications on the web for free. Hugging Face Spaces provide an easy way to host your Gradio applications.

To deploy the app on Hugging Face Spaces, follow these steps:

\begin{enumerate}
    \item Create a repository on Hugging Face Spaces.
    
    \item Upload the Python script (e.g., \texttt{app.py}) to the repository.
    
    \item Hugging Face will automatically detect the Gradio app and deploy it.
\end{enumerate}

For example, you can find instructions for deploying to Spaces [here](https://huggingface.co/spaces).

\subsection{Conclusion}

In this extended section, we’ve learned how to enhance a basic Gradio interface by adding multiple inputs and outputs, using various Hugging Face models, and even creating a text generation tool. We've also touched upon deploying the app as a standalone Python file or on Hugging Face Spaces.

This flexibility in both Gradio and Hugging Face makes it incredibly easy to build and deploy interactive applications for a wide variety of NLP tasks. You can now experiment with other models and tasks, such as machine translation or text summarization, to further expand your knowledge.

\subsection{Integrating Gradio with More Complex NLP Pipelines}

In previous examples, we used Hugging Face pipelines to quickly set up models for tasks like sentiment analysis and text generation. However, many real-world applications require more complex workflows that may involve multiple steps, such as text preprocessing, model inference, and post-processing.

In this section, we will explore how to integrate Gradio with a more complex NLP pipeline, which will involve both preprocessing and post-processing steps. For this example, let’s consider a Named Entity Recognition (NER) task where we extract entities from text and also provide additional information like the type of entity.

\subsubsection{Preprocessing and Post-Processing in a Pipeline}

We will create a pipeline that performs the following steps:

\begin{enumerate}
    \item \textbf{Preprocessing}: Clean the input text (e.g., removing unnecessary whitespace).
    
    \item \textbf{Model Inference}: Use a Hugging Face model to perform Named Entity Recognition.
    
    \item \textbf{Post-Processing}: Format the output to display the recognized entities and their types in a readable format.
\end{enumerate}

\begin{lstlisting}[style=python]
# Import necessary libraries
from transformers import pipeline
import gradio as gr
import re

# Load the Named Entity Recognition pipeline
ner_pipeline = pipeline("ner", grouped_entities=True)

# Preprocessing function
def preprocess_text(text):
    # Basic text cleaning (removing extra spaces, etc.)
    return re.sub(r"\s+", " ", text).strip()

# Post-processing function to format output
def format_ner_output(entities):
    formatted_result = []
    for entity in entities:
        formatted_result.append(f"Entity: {entity['word']} - Label: {entity['entity_group']}")
    return "\n".join(formatted_result)

# Complete pipeline for NER
def named_entity_recognition(text):
    # Preprocess the text
    cleaned_text = preprocess_text(text)
    
    # Perform NER using Hugging Face model
    entities = ner_pipeline(cleaned_text)
    
    # Format and return the output
    return format_ner_output(entities)

# Create Gradio interface
interface = gr.Interface(
    fn=named_entity_recognition,
    inputs="text",
    outputs="text",
    title="Named Entity Recognition with Preprocessing and Post-Processing"
)

# Launch the interface
interface.launch()
\end{lstlisting}

\subsection*{Explanation:}
\begin{itemize}
    \item \textbf{Preprocessing}: The \texttt{preprocess\_text} function performs basic text cleaning, such as removing extra spaces. This ensures that the input text is in a clean format before passing it to the model.
    \item \textbf{Model Inference}: We use the Hugging Face NER pipeline (\texttt{pipeline("ner")}) to detect entities in the cleaned text. The option \texttt{grouped\_entities=True} ensures that entities are grouped together.
    \item \textbf{Post-Processing}: The \texttt{format\_ner\_output} function takes the output of the NER model and formats it in a human-readable format, showing both the recognized entity and its label.
    \item \textbf{Gradio Interface}: The Gradio interface wraps this complete pipeline into a simple text-based interface where the user can input raw text, and the recognized entities are displayed in a well-formatted manner.
\end{itemize}

\subsection{Customizing Gradio for Advanced UIs}

Gradio is not limited to simple text-based inputs and outputs. It also supports more advanced UIs, such as:

\begin{itemize}
    \item \textbf{Image inputs}: For tasks like image classification or object detection.
    
    \item \textbf{Audio inputs}: For tasks like speech recognition or audio classification.
    
    \item \textbf{Checkboxes}, \textbf{sliders}, and \textbf{radio buttons}: For customized inputs.
\end{itemize}

In this section, we’ll explore how to create a more advanced Gradio interface that integrates multiple input types. Let's build an interface for a multi-modal model that can handle both text and image inputs.

\subsubsection{Building a Multi-Modal Interface}

For this example, we will create a Gradio interface that can classify text or images depending on the user’s input. We will allow users to select whether they want to classify text or an image using a radio button.

\begin{lstlisting}[style=python]
# Import necessary libraries
from transformers import pipeline
import gradio as gr

# Load text classification and image classification pipelines
text_classifier = pipeline("text-classification", model="distilbert-base-uncased-finetuned-sst-2-english")
image_classifier = pipeline("image-classification", model="google/vit-base-patch16-224")

# Function to classify text or image
def classify_input(input_type, text=None, image=None):
    if input_type == "Text":
        return text_classifier(text)[0]['label']
    elif input_type == "Image":
        return image_classifier(image)[0]['label']
    else:
        return "Invalid input"

# Create Gradio interface with multiple input types
interface = gr.Interface(
    fn=classify_input,
    inputs=[gr.inputs.Radio(choices=["Text", "Image"], label="Input Type"),
            gr.inputs.Textbox(label="Text Input", optional=True),
            gr.inputs.Image(label="Image Input", optional=True)],
    outputs="text",
    title="Multi-Modal Classification (Text or Image)"
)

# Launch the interface
interface.launch()
\end{lstlisting}

\subsection*{Explanation:}
\begin{itemize}
    \item \textbf{Multiple Inputs}: The interface takes three inputs: a radio button to choose the input type (either text or image), a textbox for text input, and an image input. The inputs for text and image are optional based on the input type.
    \item \textbf{Model Selection}: Based on the input type selected by the user, either the text classification model or the image classification model is called.
    \item \textbf{Gradio Interface}: This example showcases the flexibility of Gradio, allowing you to handle multiple types of inputs and process them with different models.
\end{itemize}

\subsection{Error Handling and Robustness in Gradio Applications}

It’s essential to ensure that your Gradio applications can handle user errors and unexpected input gracefully. For instance, a user may forget to enter text, upload the wrong type of file, or provide input that the model cannot process.

Here’s how you can add basic error handling to your Gradio application to ensure it provides clear feedback to the user when something goes wrong.

\subsubsection{Adding Error Handling for Invalid Input}

In the following example, we’ll extend our sentiment analysis interface to check if the input text is empty or too short, and return an error message in such cases.

\begin{lstlisting}[style=python]
# Function with error handling
def analyze_sentiment_with_error_handling(text):
    # Check if input is valid
    if not text or len(text.strip()) < 3:
        return "Error: Please provide a valid input of at least 3 characters."
    
    # Perform sentiment analysis
    result = sentiment_pipeline(text)
    return result[0]['label']

# Create Gradio interface
interface = gr.Interface(
    fn=analyze_sentiment_with_error_handling,
    inputs="text",
    outputs="text",
    title="Sentiment Analysis with Error Handling"
)

# Launch the interface
interface.launch()
\end{lstlisting}

\subsection*{Explanation:}
\begin{itemize}
    \item \textbf{Input Validation}: We added a condition to check if the input text is either empty or shorter than 3 characters. If it is, the function returns an error message instead of trying to pass the input to the model.
    \item \textbf{Graceful Error Messages}: The user receives a clear message when their input is invalid, ensuring a better user experience.
\end{itemize}

\subsection{Scaling Gradio Applications for Production}

As your Gradio application grows more complex and you prepare to deploy it for larger-scale use, there are a few important considerations:

\begin{itemize}
    \item \textbf{Model Performance}: If you're working with large models, consider using GPU acceleration or optimizing the models for faster inference times.
    
    \item \textbf{Load Balancing}: For high-traffic applications, you might need to deploy your application using cloud services that offer load balancing (e.g., AWS, GCP, or Azure).
    
    \item \textbf{Authentication}: If your application requires user login or restricted access, you can integrate it with an authentication system, like OAuth.
\end{itemize}

Gradio makes it easy to scale your interface by supporting deployment on cloud platforms, integrating with containers (e.g., Docker), and providing options for scaling based on traffic.

\subsection{Conclusion}

In this section, we've explored more advanced use cases of Gradio, including building complex NLP pipelines with preprocessing and post-processing, handling multiple input modalities, implementing error handling, and deploying applications at scale. Gradio's simplicity and flexibility make it an excellent tool for both rapid prototyping and production-level deployment of machine learning applications.

With these tools, you can now create more interactive and user-friendly NLP applications, extend them with more complex workflows, and scale them for real-world use cases.

\section{Deploying Gradio Applications}

Once you’ve built your Gradio interface and tested it locally, the next logical step is to deploy it so others can interact with it over the web. In this section, we will explore different methods for deploying Gradio applications, including both free and scalable options. The deployment platforms covered include Hugging Face Spaces, Google Colab, and cloud services like AWS, GCP, and Heroku.

\subsection{Deploying on Hugging Face Spaces}

Hugging Face Spaces is a popular platform for hosting machine learning demos, and it offers free hosting for Gradio applications. Spaces are powered by Git repositories, and Hugging Face automatically handles the setup for Gradio applications, making this one of the easiest ways to deploy your project.

\subsubsection{Step-by-Step Deployment on Hugging Face Spaces}

Here’s how you can deploy your Gradio app on Hugging Face Spaces:

\begin{enumerate}
    \item \textbf{Create a Hugging Face Account}: If you don’t already have an account, create one at \href{https://huggingface.co/join}{huggingface.co}.
    
    \item \textbf{Create a New Space}: Go to the \href{https://huggingface.co/spaces}{Spaces page} and click on “Create new Space”. You can choose between a public or private space depending on whether you want to share your app publicly.
    
    \item \textbf{Upload Your Code}: Create a new repository for your app. Upload your Python script (for example, \texttt{app.py}) and any other necessary files, such as a \texttt{requirements.txt} file to specify the required libraries. Here's an example \texttt{requirements.txt} file:
    
    \begin{lstlisting}[language=python]
    gradio
    transformers
    \end{lstlisting}
    
    \item \textbf{Deploy the Space}: Once the code and required files are uploaded, Hugging Face Spaces will automatically detect that your application is built with Gradio and deploy it. You can monitor the build process on the repository page.
    
    \item \textbf{Access and Share Your App}: After the build process completes, you will receive a URL where the app is hosted. You can share this URL with anyone who wants to interact with your Gradio app.
\end{enumerate}

\textbf{Advantages}:
\begin{itemize}
    \item Free hosting for most use cases.
    \item Easy integration with Hugging Face models.
    \item Ideal for showcasing demos and prototypes.
\end{itemize}

Hugging Face Spaces is an excellent choice for those looking for a hassle-free way to deploy smaller applications or public demos.

\subsection{Deploying on Google Colab}

Google Colab is another free platform you can use to run and share Gradio apps. Colab provides free GPUs, making it a great option for more compute-intensive applications. Although Colab is primarily a notebook environment, it can also serve Gradio apps.

\subsubsection{Deploying Gradio on Google Colab}

Here’s a simple workflow for deploying your Gradio app using Google Colab:

\begin{enumerate}
    \item \textbf{Open a New Colab Notebook}: Go to \href{https://colab.research.google.com}{Google Colab} and create a new notebook.
    
    \item \textbf{Install the Required Libraries}: In a code cell, install Gradio and Transformers using pip:
    
    \begin{lstlisting}[language=python]
    !pip install gradio transformers
    \end{lstlisting}
    
    \item \textbf{Write Your Gradio App}: In another cell, write the Python code for your Gradio app (e.g., using one of the examples we covered earlier). 
    
    \item \textbf{Launch the App}: Use \texttt{interface.launch()} to launch the app. Gradio will generate a public URL for your app that can be accessed from any device:
    
    \begin{lstlisting}[language=python]
    interface.launch(share=True)
    \end{lstlisting}
    
    \item \textbf{Share the Link}: After running the cell, you will get a link (something like \texttt{https://<unique\_id>.gradio.app}) which you can share with others. This link will remain active as long as the Colab notebook is running.
\end{enumerate}

\textbf{Advantages}:
\begin{itemize}
    \item Free GPU access for compute-heavy models.
    \item Easy to share with collaborators and users.
    \item Suitable for short-term deployments and experimentation.
\end{itemize}

Keep in mind that Colab sessions will eventually expire, making it better suited for temporary deployments or testing.

\subsection{Deploying on Heroku}

Heroku is a popular cloud platform that offers easy deployment for web applications, including Gradio apps. Heroku provides free tier hosting, but it’s limited in terms of uptime and resources.

\subsubsection{Deploying Gradio on Heroku}

Follow these steps to deploy your Gradio app on Heroku:

\begin{enumerate}
    \item \textbf{Install Heroku CLI}: First, you need to install the Heroku Command Line Interface (CLI). You can find installation instructions at \href{https://devcenter.heroku.com/articles/heroku-cli}{Heroku CLI}.
    
    \item \textbf{Create a Heroku App}: Initialize a Git repository for your Gradio app and run the following command in your terminal:
    
    \begin{lstlisting}[language=bash]
    heroku create my-gradio-app
    \end{lstlisting}
    
    \item \textbf{Prepare Your App for Deployment}: 
    \begin{itemize}
        \item Ensure that you have a \texttt{requirements.txt} file that lists all the necessary dependencies.
        \item Create a \texttt{Procfile} in the root of your project with the following content:
    \end{itemize}
    
    \begin{lstlisting}
    web: python app.py
    \end{lstlisting}
    
    \item \textbf{Deploy Your App}: Once your repository is ready, push the code to Heroku using Git:
    
    \begin{lstlisting}[language=bash]
    git add .
    git commit -m "Deploy Gradio app"
    git push heroku master
    \end{lstlisting}
    
    \item \textbf{Access the App}: After the deployment is complete, Heroku will provide a URL for your application. You can visit the URL to see your Gradio app live.
\end{enumerate}

\textbf{Advantages}:
\begin{itemize}
    \item Easy to deploy web apps with minimal configuration.
    \item Scalable if you choose a paid plan.
    \item Ideal for long-term or production deployments.
\end{itemize}

Heroku is a great option if you want to maintain an always-on Gradio app without worrying about the underlying infrastructure.

\subsection{Deploying on Amazon Web Services (AWS)}

For large-scale applications that require significant computational resources or need to handle high traffic, AWS provides a highly scalable option. You can use services like EC2 (Elastic Compute Cloud) or ECS (Elastic Container Service) to deploy your Gradio app.

\subsubsection{Deploying Gradio on AWS EC2}

Here’s a high-level overview of deploying a Gradio app on an AWS EC2 instance:

\begin{enumerate}
    \item \textbf{Launch an EC2 Instance}: Go to the AWS Management Console and launch an EC2 instance with the desired specifications. Choose a machine with enough compute power for your application.
    
    \item \textbf{Install Dependencies}: SSH into the EC2 instance and install Python, Gradio, Transformers, and any other necessary libraries.
    
    \item \textbf{Transfer Your Code}: Upload your Gradio application to the EC2 instance using \texttt{scp} or any file transfer method.
    
    \item \textbf{Run the App}: Run your Gradio app as you would locally, and use an HTTP server like Nginx or Apache to serve the app. You can also configure Gradio to serve your app on a specific port that is accessible from the web.
    
    \item \textbf{Access the App}: Once the server is running, you can access the app using the public IP address of your EC2 instance.
\end{enumerate}

\textbf{Advantages}:
\begin{itemize}
    \item Highly scalable and customizable.
    \item Access to powerful hardware, including GPUs for compute-intensive models.
    \item Ideal for production-grade deployments.
\end{itemize}

AWS offers a lot of flexibility, but it also requires more configuration compared to other platforms. It’s best suited for large-scale or enterprise applications.

\subsection{Deploying on Google Cloud Platform (GCP)}

Similar to AWS, Google Cloud Platform (GCP) offers scalable options for deploying Gradio apps, including using services like Compute Engine (GCE) or Google Kubernetes Engine (GKE).

\subsubsection{Deploying Gradio on Google Compute Engine (GCE)}

Here’s an overview of deploying on GCP:

\begin{enumerate}
    \item \textbf{Create a Compute Engine Instance}: In the GCP Console, create a new virtual machine (VM) instance with the appropriate resources.
    
    \item \textbf{Install Dependencies}: SSH into the instance and install Python, Gradio, and other libraries your app requires.
    
    \item \textbf{Transfer Your Code}: Upload your Gradio app to the GCP instance.
    
    \item \textbf{Run the App}: Run your Gradio app and configure the VM instance to allow external traffic on the port Gradio is using.
    
    \item \textbf{Access the App}: Once the server is running, you can access your app via the external IP of the VM instance.
\end{enumerate}

\textbf{Advantages}:
\begin{itemize}
    \item Google Cloud offers competitive pricing and a robust infrastructure.
    \item Scalable for large applications.
    \item Suitable for long-term and production use cases.
\end{itemize}

\subsection{Conclusion}

In this section, we explored various ways to deploy your Gradio applications, from simple and free options like Hugging Face Spaces and Google Colab to more scalable and enterprise-grade solutions like AWS and GCP.

\bibliographystyle{ieeetr}
\bibliography{ref}

@article{niu2024textmultimodalityexploringevolution,
      title={From Text to Multimodality: Exploring the Evolution and Impact of Large Language Models in Medical Practice},
      author={Qian Niu and Keyu Chen and Ming Li and Pohsun Feng and Ziqian Bi and Lawrence KQ Yan and Yichao Zhang and Caitlyn Heqi Yin and Cheng Fei and Junyu Liu and Benji Peng},
      year={2024},
      eprint={2410.01812},
      archivePrefix={arXiv},
      primaryClass={cs.CY},
      url={https://arxiv.org/abs/2410.01812},
}

@article{niu2024largelanguagemodelscognitive,
      title={Large Language Models and Cognitive Science: A Comprehensive Review of Similarities, Differences, and Challenges}, 
      author={Qian Niu and Junyu Liu and Ziqian Bi and Pohsun Feng and Benji Peng and Keyu Chen and Ming Li and Lawrence KQ Yan and Yichao Zhang and Caitlyn Heqi Yin and Cheng Fei},
      year={2024},
      eprint={2409.02387},
      archivePrefix={arXiv},
      primaryClass={cs.AI},
      url={https://arxiv.org/abs/2409.02387}, 
}

@article{yan2024largelanguagemodelbenchmarks,
      title={Large Language Model Benchmarks in Medical Tasks}, 
      author={Lawrence K. Q. Yan and Ming Li and Yichao Zhang and Caitlyn Heqi Yin and Cheng Fei and Benji Peng and Ziqian Bi and Pohsun Feng and Keyu Chen and Junyu Liu and Qian Niu},
      year={2024},
      eprint={2410.21348},
      archivePrefix={arXiv},
      primaryClass={cs.CL},
      url={https://arxiv.org/abs/2410.21348}, 
}

@article{peng2024jailbreakingmitigationvulnerabilitieslarge,
      title={Jailbreaking and Mitigation of Vulnerabilities in Large Language Models}, 
      author={Benji Peng and Ziqian Bi and Qian Niu and Ming Liu and Pohsun Feng and Tianyang Wang and Lawrence K. Q. Yan and Yizhu Wen and Yichao Zhang and Caitlyn Heqi Yin},
      year={2024},
      eprint={2410.15236},
      archivePrefix={arXiv},
      primaryClass={cs.CR},
      url={https://arxiv.org/abs/2410.15236}, 
}

@article{peng2024securinglargelanguagemodels,
      title={Securing Large Language Models: Addressing Bias, Misinformation, and Prompt Attacks}, 
      author={Benji Peng and Keyu Chen and Ming Li and Pohsun Feng and Ziqian Bi and Junyu Liu and Qian Niu},
      year={2024},
      eprint={2409.08087},
      archivePrefix={arXiv},
      primaryClass={cs.CR},
      url={https://arxiv.org/abs/2409.08087}, 
}

@article{chen2024deeplearningmachinelearning,
      title={Deep Learning and Machine Learning, Advancing Big Data Analytics and Management: Tensorflow Pretrained Models}, 
      author={Keyu Chen and Ziqian Bi and Qian Niu and Junyu Liu and Benji Peng and Sen Zhang and Ming Liu and Ming Li and Xuanhe Pan and Jiawei Xu and Jinlang Wang and Pohsun Feng},
      year={2024},
      eprint={2409.13566},
      archivePrefix={arXiv},
      primaryClass={cs.LG},
      url={https://arxiv.org/abs/2409.13566}, 
}

@article{li2024deeplearningmachinelearning,
      title={Deep Learning and Machine Learning, Advancing Big Data Analytics and Management: Object-Oriented Programming}, 
      author={Ming Li and Ziqian Bi and Tianyang Wang and Keyu Chen and Jiawei Xu and Qian Niu and Junyu Liu and Benji Peng and Sen Zhang and Xuanhe Pan and Jinlang Wang and Pohsun Feng and Caitlyn Heqi Yin and Yizhu Wen and Ming Liu},
      year={2024},
      eprint={2409.19916},
      archivePrefix={arXiv},
      primaryClass={cs.CL},
      url={https://arxiv.org/abs/2409.19916}, 
}

@article{Bishop2010TheIG,
  title={The Imitation Game},
  author={Mark Bishop},
  journal={A New History of the Future in 100 Objects},
  year={2010},
  url={https://api.semanticscholar.org/CorpusID:192936040}
}

@article{Weizenbaum1966ELIZAaCP,
  title={ELIZA—a computer program for the study of natural language communication between man and machine},
  author={Joseph Weizenbaum},
  journal={Communications of the ACM},
  year={1966},
  volume={9},
  pages={36 - 45},
  url={https://api.semanticscholar.org/CorpusID:1896290}
}

@inproceedings{Matsumoto2000UsingML,
  title={Using Machine Learning Methods to Improve Quality of Tagged Corpora and Learning Models},
  author={Yuji Matsumoto and Tatsuo Yamashita},
  booktitle={International Conference on Language Resources and Evaluation},
  year={2000},
  url={https://api.semanticscholar.org/CorpusID:8693941}
}

@article{Salehinejad2017RecentAI,
  title={Recent Advances in Recurrent Neural Networks},
  author={Hojjat Salehinejad and Julianne Baarbe and Sharan Sankar and Joseph Barfett and Errol Colak and Shahrokh Valaee},
  journal={ArXiv},
  year={2017},
  volume={abs/1801.01078},
  url={https://api.semanticscholar.org/CorpusID:3531572}
}

@article{vaswani2017attention,
  title={Attention is all you need},
  author={Vaswani, A},
  journal={Advances in Neural Information Processing Systems},
  year={2017}
}

@inproceedings{Devlin2019BERTPO,
  title={BERT: Pre-training of Deep Bidirectional Transformers for Language Understanding},
  author={Jacob Devlin and Ming-Wei Chang and Kenton Lee and Kristina Toutanova},
  booktitle={North American Chapter of the Association for Computational Linguistics},
  year={2019},
  url={https://api.semanticscholar.org/CorpusID:52967399}
}

@article{Yenduri2023GPTP,
  title={GPT (Generative Pre-Trained Transformer)— A Comprehensive Review on Enabling Technologies, Potential Applications, Emerging Challenges, and Future Directions},
  author={Gokul Yenduri and Manju Ramalingam and Govardanan Chemmalar Selvi and Y. Supriya and Gautam Srivastava and Praveen Kumar Reddy Maddikunta and G. Deepti Raj and Rutvij H. Jhaveri and B. Prabadevi and Weizheng Wang and Athanasios V. Vasilakos and Thippa Reddy Gadekallu},
  journal={IEEE Access},
  year={2023},
  volume={12},
  pages={54608-54649},
  url={https://api.semanticscholar.org/CorpusID:258762263}
}

@article{Chung2018IntelligentVA,
  title={Intelligent Virtual Assistant knows Your Life},
  author={Hyunji Chung and Sangjin Lee},
  journal={ArXiv},
  year={2018},
  volume={abs/1803.00466},
  url={https://api.semanticscholar.org/CorpusID:3648924}
}

@article{Hastings2009AutomaticCG,
  title={Automatic Content Generation in the Galactic Arms Race Video Game},
  author={Erin J. Hastings and Ratan K. Guha and Kenneth O. Stanley},
  journal={IEEE Transactions on Computational Intelligence and AI in Games},
  year={2009},
  volume={1},
  pages={245-263},
  url={https://api.semanticscholar.org/CorpusID:88411}
}

@article{Ranchal2013UsingSR,
  title={Using speech recognition for real-time captioning and lecture transcription in the classroom},
  author={Rohit Ranchal and Teresa Taber-Doughty and Yiren Guo and Keith Bain and Heather Martin and J. Paul Robinson and Bradley S. Duerstock},
  journal={IEEE Transactions on Learning Technologies},
  year={2013},
  volume={6},
  pages={299-311},
  url={https://api.semanticscholar.org/CorpusID:40778774}
}

@article{Pappula2023LLMsFC,
  title={LLMs for Conversational AI: Enhancing Chatbots and Virtual Assistants},
  author={Sharmila Reddy Pappula and Sathwik Rao Allam},
  journal={International Journal of Research Publication and Reviews},
  year={2023},
  url={https://api.semanticscholar.org/CorpusID:266461220}
}

@article{Dwarakanath2021AutomatedML,
  title={Automated Machine Learning Approaches for Emergency Response and Coordination via Social Media in the Aftermath of a Disaster: A Review},
  author={Lokabhiram Dwarakanath and Amirrudin Kamsin and Rasheed Abubakar Rasheed and Anitha Anandhan and Liyana Shuib},
  journal={IEEE Access},
  year={2021},
  volume={9},
  pages={68917-68931},
  url={https://api.semanticscholar.org/CorpusID:234500187}
}

@article{Havrlant2017ASP,
  title={A simple probabilistic explanation of term frequency-inverse document frequency (tf-idf) heuristic (and variations motivated by this explanation)},
  author={Luk{\'a}s Havrlant and Vladik Kreinovich},
  journal={International Journal of General Systems},
  year={2017},
  volume={46},
  pages={27 - 36},
  url={https://api.semanticscholar.org/CorpusID:298072}
}

@article{zhang2010understanding,
  title={Understanding bag-of-words model: a statistical framework},
  author={Zhang, Yin and Jin, Rong and Zhou, Zhi-Hua},
  journal={International journal of machine learning and cybernetics},
  volume={1},
  pages={43--52},
  year={2010},
  publisher={Springer}
}

@article{li2020survey,
  title={A survey on deep learning for named entity recognition},
  author={Li, Jing and Sun, Aixin and Han, Jianglei and Li, Chenliang},
  journal={IEEE transactions on knowledge and data engineering},
  volume={34},
  number={1},
  pages={50--70},
  year={2020},
  publisher={IEEE}
}

@article{church2017word2vec,
  title={Word2Vec},
  author={Church, Kenneth Ward},
  journal={Natural Language Engineering},
  volume={23},
  number={1},
  pages={155--162},
  year={2017},
  publisher={Cambridge University Press}
}

@inproceedings{pennington2014glove,
  title={Glove: Global vectors for word representation},
  author={Pennington, Jeffrey and Socher, Richard and Manning, Christopher D},
  booktitle={Proceedings of the 2014 conference on empirical methods in natural language processing (EMNLP)},
  pages={1532--1543},
  year={2014}
}

@article{blei2003latent,
  title={Latent dirichlet allocation},
  author={Blei, David M and Ng, Andrew Y and Jordan, Michael I},
  journal={Journal of machine Learning research},
  volume={3},
  number={Jan},
  pages={993--1022},
  year={2003}
}

@article{zouhar2023formal,
  title={A formal perspective on byte-pair encoding},
  author={Zouhar, Vil{\'e}m and Meister, Clara and Gastaldi, Juan Luis and Du, Li and Vieira, Tim and Sachan, Mrinmaya and Cotterell, Ryan},
  journal={arXiv preprint arXiv:2306.16837},
  year={2023}
}

@article{hoffman2010online,
  title={Online learning for latent dirichlet allocation},
  author={Hoffman, Matthew and Bach, Francis and Blei, David},
  journal={advances in neural information processing systems},
  volume={23},
  year={2010}
}

@article{abdi2010principal,
  title={Principal component analysis},
  author={Abdi, Herv{\'e} and Williams, Lynne J},
  journal={Wiley interdisciplinary reviews: computational statistics},
  volume={2},
  number={4},
  pages={433--459},
  year={2010},
  publisher={Wiley Online Library}
}

@article{van2008visualizing,
  title={Visualizing data using t-SNE.},
  author={Van der Maaten, Laurens and Hinton, Geoffrey},
  journal={Journal of machine learning research},
  volume={9},
  number={11},
  year={2008}
}

@article{floridi2020gpt,
  title={GPT-3: Its nature, scope, limits, and consequences},
  author={Floridi, Luciano and Chiriatti, Massimo},
  journal={Minds and Machines},
  volume={30},
  pages={681--694},
  year={2020},
  publisher={Springer}
}

@article{devlin2018bert,
  title={Bert: Pre-training of deep bidirectional transformers for language understanding},
  author={Devlin, Jacob},
  journal={arXiv preprint arXiv:1810.04805},
  year={2018}
}

@article{do2019deep,
  title={Deep learning for aspect-based sentiment analysis: a comparative review},
  author={Do, Hai Ha and Prasad, Penatiyana WC and Maag, Angelika and Alsadoon, Abeer},
  journal={Expert systems with applications},
  volume={118},
  pages={272--299},
  year={2019},
  publisher={Elsevier}
}

@article{ulvcar2022cross,
  title={Cross-lingual alignments of ELMo contextual embeddings},
  author={Ul{\v{c}}ar, Matej and Robnik-{\v{S}}ikonja, Marko},
  journal={Neural Computing and Applications},
  volume={34},
  number={15},
  pages={13043--13061},
  year={2022},
  publisher={Springer}
}

@article{barbieri2021xlm,
  title={XLM-T: Multilingual language models in Twitter for sentiment analysis and beyond},
  author={Barbieri, Francesco and Anke, Luis Espinosa and Camacho-Collados, Jose},
  journal={arXiv preprint arXiv:2104.12250},
  year={2021}
}

@article{rao2013entity,
  title={Entity linking: Finding extracted entities in a knowledge base},
  author={Rao, Delip and McNamee, Paul and Dredze, Mark},
  journal={Multi-source, multilingual information extraction and summarization},
  pages={93--115},
  year={2013},
  publisher={Springer}
}

@article{pires2019multilingual,
  title={How multilingual is multilingual BERT},
  author={Pires, T},
  journal={arXiv preprint arXiv:1906.01502},
  year={2019}
}

@article{chawla2002smote,
  title={SMOTE: synthetic minority over-sampling technique},
  author={Chawla, Nitesh V and Bowyer, Kevin W and Hall, Lawrence O and Kegelmeyer, W Philip},
  journal={Journal of artificial intelligence research},
  volume={16},
  pages={321--357},
  year={2002}
}

@article{song2020fast,
  title={Fast wordpiece tokenization},
  author={Song, Xinying and Salcianu, Alex and Song, Yang and Dopson, Dave and Zhou, Denny},
  journal={arXiv preprint arXiv:2012.15524},
  year={2020}
}

@inproceedings{adoma2020comparative,
  title={Comparative analyses of bert, roberta, distilbert, and xlnet for text-based emotion recognition},
  author={Adoma, Acheampong Francisca and Henry, Nunoo-Mensah and Chen, Wenyu},
  booktitle={2020 17th International Computer Conference on Wavelet Active Media Technology and Information Processing (ICCWAMTIP)},
  pages={117--121},
  year={2020},
  organization={IEEE}
}

@article{blalock2020state,
  title={What is the state of neural network pruning?},
  author={Blalock, Davis and Gonzalez Ortiz, Jose Javier and Frankle, Jonathan and Guttag, John},
  journal={Proceedings of machine learning and systems},
  volume={2},
  pages={129--146},
  year={2020}
}

@article{beltagy2019scibert,
  title={SciBERT: A pretrained language model for scientific text},
  author={Beltagy, Iz and Lo, Kyle and Cohan, Arman},
  journal={arXiv preprint arXiv:1903.10676},
  year={2019}
}

@article{polino2018model,
  title={Model compression via distillation and quantization},
  author={Polino, Antonio and Pascanu, Razvan and Alistarh, Dan},
  journal={arXiv preprint arXiv:1802.05668},
  year={2018}
}

@article{kudo2018sentencepiece,
  title={Sentencepiece: A simple and language independent subword tokenizer and detokenizer for neural text processing},
  author={Kudo, T},
  journal={arXiv preprint arXiv:1808.06226},
  year={2018}
}

@article{black2022gpt,
  title={Gpt-neox-20b: An open-source autoregressive language model},
  author={Black, Sid and Biderman, Stella and Hallahan, Eric and Anthony, Quentin and Gao, Leo and Golding, Laurence and He, Horace and Leahy, Connor and McDonell, Kyle and Phang, Jason and others},
  journal={arXiv preprint arXiv:2204.06745},
  year={2022}
}

@inproceedings{haas2017bringing,
  title={Bringing the web up to speed with WebAssembly},
  author={Haas, Andreas and Rossberg, Andreas and Schuff, Derek L and Titzer, Ben L and Holman, Michael and Gohman, Dan and Wagner, Luke and Zakai, Alon and Bastien, JF},
  booktitle={Proceedings of the 38th ACM SIGPLAN Conference on Programming Language Design and Implementation},
  pages={185--200},
  year={2017}
}

@inproceedings{collins2004language,
  title={A language modeling approach to predicting reading difficulty},
  author={Collins-Thompson, Kevyn and Callan, James P},
  booktitle={Proceedings of the human language technology conference of the North American chapter of the association for computational linguistics: HLT-NAACL 2004},
  pages={193--200},
  year={2004}
}

@inproceedings{xu2019explainable,
  title={Explainable AI: A brief survey on history, research areas, approaches and challenges},
  author={Xu, Feiyu and Uszkoreit, Hans and Du, Yangzhou and Fan, Wei and Zhao, Dongyan and Zhu, Jun},
  booktitle={Natural language processing and Chinese computing: 8th cCF international conference, NLPCC 2019, dunhuang, China, October 9--14, 2019, proceedings, part II 8},
  pages={563--574},
  year={2019},
  organization={Springer}
}

@article{shen2021much,
  title={How much can clip benefit vision-and-language tasks?},
  author={Shen, Sheng and Li, Liunian Harold and Tan, Hao and Bansal, Mohit and Rohrbach, Anna and Chang, Kai-Wei and Yao, Zhewei and Keutzer, Kurt},
  journal={arXiv preprint arXiv:2107.06383},
  year={2021}
}

@article{joshi2021adversarial,
  title={Adversarial token attacks on vision transformers},
  author={Joshi, Ameya and Jagatap, Gauri and Hegde, Chinmay},
  journal={arXiv preprint arXiv:2110.04337},
  year={2021}
}

@article{wu2022survey,
  title={A survey of human-in-the-loop for machine learning},
  author={Wu, Xingjiao and Xiao, Luwei and Sun, Yixuan and Zhang, Junhang and Ma, Tianlong and He, Liang},
  journal={Future Generation Computer Systems},
  volume={135},
  pages={364--381},
  year={2022},
  publisher={Elsevier}
}

@article{kimera2024advancing,
  title={Advancing AI with Integrity: Ethical Challenges and Solutions in Neural Machine Translation},
  author={Kimera, Richard and Kim, Yun-Seon and Choi, Heeyoul},
  journal={arXiv preprint arXiv:2404.01070},
  year={2024}
}

@article{hospedales2021meta,
  title={Meta-learning in neural networks: A survey},
  author={Hospedales, Timothy and Antoniou, Antreas and Micaelli, Paul and Storkey, Amos},
  journal={IEEE transactions on pattern analysis and machine intelligence},
  volume={44},
  number={9},
  pages={5149--5169},
  year={2021},
  publisher={IEEE}
}

@article{lample2017unsupervised,
  title={Unsupervised machine translation using monolingual corpora only},
  author={Lample, Guillaume and Conneau, Alexis and Denoyer, Ludovic and Ranzato, Marc'Aurelio},
  journal={arXiv preprint arXiv:1711.00043},
  year={2017}
}

@article{byrne1998learning,
  title={Learning by imitation: A hierarchical approach},
  author={Byrne, Richard W and Russon, Anne E},
  journal={Behavioral and brain sciences},
  volume={21},
  number={5},
  pages={667--684},
  year={1998},
  publisher={Cambridge University Press}
}

@inproceedings{bengio2009curriculum,
  title={Curriculum learning},
  author={Bengio, Yoshua and Louradour, J{\'e}r{\^o}me and Collobert, Ronan and Weston, Jason},
  booktitle={Proceedings of the 26th annual international conference on machine learning},
  pages={41--48},
  year={2009}
}

@misc{schmid1994partofspeechtaggingneuralnetworks,
      title={Part-of-Speech Tagging with Neural Networks}, 
      author={Helmut Schmid},
      year={1994},
      eprint={cmp-lg/9410018},
      archivePrefix={arXiv},
      primaryClass={cmp-lg},
      url={https://arxiv.org/abs/cmp-lg/9410018}, 
}

@inproceedings{Salazar_2020,
   title={Masked Language Model Scoring},
   url={http://dx.doi.org/10.18653/v1/2020.acl-main.240},
   DOI={10.18653/v1/2020.acl-main.240},
   booktitle={Proceedings of the 58th Annual Meeting of the Association for Computational Linguistics},
   publisher={Association for Computational Linguistics},
   author={Salazar, Julian and Liang, Davis and Nguyen, Toan Q. and Kirchhoff, Katrin},
   year={2020} }

@article{stahlberg2020neural,
  title={Neural machine translation: A review},
  author={Stahlberg, Felix},
  journal={Journal of Artificial Intelligence Research},
  volume={69},
  pages={343--418},
  year={2020}
}

@inproceedings{papineni2002bleu,
  title={Bleu: a method for automatic evaluation of machine translation},
  author={Papineni, Kishore and Roukos, Salim and Ward, Todd and Zhu, Wei-Jing},
  booktitle={Proceedings of the 40th annual meeting of the Association for Computational Linguistics},
  pages={311--318},
  year={2002}
}

@article{li2020pytorch,
  title={Pytorch distributed: Experiences on accelerating data parallel training},
  author={Li, Shen and Zhao, Yanli and Varma, Rohan and Salpekar, Omkar and Noordhuis, Pieter and Li, Teng and Paszke, Adam and Smith, Jeff and Vaughan, Brian and Damania, Pritam and others},
  journal={arXiv preprint arXiv:2006.15704},
  year={2020}
}

@inproceedings{rohit2024comparative,
  title={Comparative Study on Synthetic and Natural Error Analysis with BART \& MarianMT},
  author={Rohit, R and Gandheesh, SA and Sannala, Gayatri Sanjana and Pati, Peeta Basa},
  booktitle={2024 IEEE 9th International Conference for Convergence in Technology (I2CT)},
  pages={1--6},
  year={2024},
  organization={IEEE}
}

@article{gliwa2019samsum,
  title={SAMSum corpus: A human-annotated dialogue dataset for abstractive summarization},
  author={Gliwa, Bogdan and Mochol, Iwona and Biesek, Maciej and Wawer, Aleksander},
  journal={arXiv preprint arXiv:1911.12237},
  year={2019}
}

@article{chen2016thorough,
  title={A thorough examination of the cnn/daily mail reading comprehension task},
  author={Chen, Danqi and Bolton, Jason and Manning, Christopher D},
  journal={arXiv preprint arXiv:1606.02858},
  year={2016}
}

@article{freitag2017beam,
  title={Beam search strategies for neural machine translation},
  author={Freitag, Markus and Al-Onaizan, Yaser},
  journal={arXiv preprint arXiv:1702.01806},
  year={2017}
}

@inproceedings{lin2004looking,
  title={Looking for a few good metrics: ROUGE and its evaluation},
  author={Lin, Chin-Yew and Och, FJ},
  booktitle={Ntcir workshop},
  year={2004}
}

@inproceedings{banerjee2005meteor,
  title={METEOR: An automatic metric for MT evaluation with improved correlation with human judgments},
  author={Banerjee, Satanjeev and Lavie, Alon},
  booktitle={Proceedings of the acl workshop on intrinsic and extrinsic evaluation measures for machine translation and/or summarization},
  pages={65--72},
  year={2005}
}

@inproceedings{napoles2012annotated,
  title={Annotated gigaword},
  author={Napoles, Courtney and Gormley, Matthew R and Van Durme, Benjamin},
  booktitle={Proceedings of the joint workshop on automatic knowledge base construction and web-scale knowledge extraction (AKBC-WEKEX)},
  pages={95--100},
  year={2012}
}

@article{feder2021causalm,
  title={Causalm: Causal model explanation through counterfactual language models},
  author={Feder, Amir and Oved, Nadav and Shalit, Uri and Reichart, Roi},
  journal={Computational Linguistics},
  volume={47},
  number={2},
  pages={333--386},
  year={2021},
  publisher={MIT Press One Rogers Street, Cambridge, MA 02142-1209, USA journals-info~…}
}

@inproceedings{higuchi2008casual,
  title={A casual conversation system using modality and word associations retrieved from the web},
  author={Higuchi, Shinsuke and Rzepka, Rafal and Araki, Kenji},
  booktitle={EMNLP'08 Proceedings of the Conference on Empirical Methods in Natural Language Processing},
  pages={382--390},
  year={2008},
  organization={ACM}
}

@article{rajpurkar2016squad,
  title={Squad: 100,000+ questions for machine comprehension of text},
  author={Rajpurkar, P},
  journal={arXiv preprint arXiv:1606.05250},
  year={2016}
}

@article{abid2019gradio,
  title={Gradio: Hassle-free sharing and testing of ml models in the wild},
  author={Abid, Abubakar and Abdalla, Ali and Abid, Ali and Khan, Dawood and Alfozan, Abdulrahman and Zou, James},
  journal={arXiv preprint arXiv:1906.02569},
  year={2019}
}

@article{kang2020natural,
  title={Natural language processing (NLP) in management research: A literature review},
  author={Kang, Yue and Cai, Zhao and Tan, Chee-Wee and Huang, Qian and Liu, Hefu},
  journal={Journal of Management Analytics},
  volume={7},
  number={2},
  pages={139--172},
  year={2020},
  publisher={Taylor \& Francis}
}

@article{nadkarni2011natural,
  title={Natural language processing: an introduction},
  author={Nadkarni, Prakash M and Ohno-Machado, Lucila and Chapman, Wendy W},
  journal={Journal of the American Medical Informatics Association},
  volume={18},
  number={5},
  pages={544--551},
  year={2011},
  publisher={BMJ Group BMA House, Tavistock Square, London, WC1H 9JR}
}

@article{chowdhary2020natural,
  title={Natural language processing},
  author={Chowdhary, KR1442 and Chowdhary, KR},
  journal={Fundamentals of artificial intelligence},
  pages={603--649},
  year={2020},
  publisher={Springer}
}

@incollection{fanni2023natural,
  title={Natural language processing},
  author={Fanni, Salvatore Claudio and Febi, Maria and Aghakhanyan, Gayane and Neri, Emanuele},
  booktitle={Introduction to Artificial Intelligence},
  pages={87--99},
  year={2023},
  publisher={Springer}
}

@article{chopra2013natural,
  title={Natural language processing},
  author={Chopra, Abhimanyu and Prashar, Abhinav and Sain, Chandresh},
  journal={International journal of technology enhancements and emerging engineering research},
  volume={1},
  number={4},
  pages={131--134},
  year={2013},
  publisher={Citeseer}
}

@article{joshi1991natural,
  title={Natural language processing},
  author={Joshi, Aravind K},
  journal={Science},
  volume={253},
  number={5025},
  pages={1242--1249},
  year={1991},
  publisher={American Association for the Advancement of Science}
}

@article{joseph2016natural,
  title={Natural language processing: A review},
  author={Joseph, Sethunya R and Hlomani, Hlomani and Letsholo, Keletso and Kaniwa, Freeson and Sedimo, Kutlwano},
  journal={International Journal of Research in Engineering and Applied Sciences},
  volume={6},
  number={3},
  pages={207--210},
  year={2016}
}

@article{cambria2014jumping,
  title={Jumping NLP curves: A review of natural language processing research},
  author={Cambria, Erik and White, Bebo},
  journal={IEEE Computational intelligence magazine},
  volume={9},
  number={2},
  pages={48--57},
  year={2014},
  publisher={IEEE}
}

@article{liddy2001natural,
  title={Natural language processing},
  author={Liddy, Elizabeth D},
  year={2001}
}

@book{grosz1986readings,
  title={Readings in natural language processing},
  author={Grosz, Barbara J and Sparck-Jones, Karen and Webber, Bonnie Lynn},
  year={1986},
  publisher={Morgan Kaufmann Publishers Inc.}
}

@article{khurana2023natural,
  title={Natural language processing: state of the art, current trends and challenges},
  author={Khurana, Diksha and Koli, Aditya and Khatter, Kiran and Singh, Sukhdev},
  journal={Multimedia tools and applications},
  volume={82},
  number={3},
  pages={3713--3744},
  year={2023},
  publisher={Springer}
}

@article{yim2016natural,
  title={Natural language processing in oncology: a review},
  author={Yim, Wen-wai and Yetisgen, Meliha and Harris, William P and Kwan, Sharon W},
  journal={JAMA oncology},
  volume={2},
  number={6},
  pages={797--804},
  year={2016},
  publisher={American Medical Association}
}

@article{reshamwala2013review,
  title={Review on natural language processing},
  author={Reshamwala, Alpa and Mishra, Dhirendra and Pawar, Prajakta},
  journal={IRACST Engineering Science and Technology: An International Journal (ESTIJ)},
  volume={3},
  number={1},
  pages={113--116},
  year={2013}
}

@article{jones1994natural,
  title={Natural language processing: a historical review},
  author={Jones, Karen Sparck},
  journal={Current issues in computational linguistics: in honour of Don Walker},
  pages={3--16},
  year={1994},
  publisher={Springer}
}

@article{hirschberg2015advances,
  title={Advances in natural language processing},
  author={Hirschberg, Julia and Manning, Christopher D},
  journal={Science},
  volume={349},
  number={6245},
  pages={261--266},
  year={2015},
  publisher={American Association for the Advancement of Science}
}

@article{locke2021natural,
  title={Natural language processing in medicine: a review},
  author={Locke, Saskia and Bashall, Anthony and Al-Adely, Sarah and Moore, John and Wilson, Anthony and Kitchen, Gareth B},
  journal={Trends in Anaesthesia and Critical Care},
  volume={38},
  pages={4--9},
  year={2021},
  publisher={Elsevier}
}

@inproceedings{ibrahim2010class,
  title={Class diagram extraction from textual requirements using natural language processing (NLP) techniques},
  author={Ibrahim, Mohd and Ahmad, Rodina},
  booktitle={2010 Second International Conference on Computer Research and Development},
  pages={200--204},
  year={2010},
  organization={IEEE}
}

@article{alhawiti2014natural,
  title={Natural language processing and its use in education},
  author={Alhawiti, Khaled M},
  journal={International Journal of Advanced Computer Science and Applications},
  volume={5},
  number={12},
  year={2014},
  publisher={Citeseer}
}

@book{jensen2012natural,
  title={Natural language processing: the PLNLP approach},
  author={Jensen, Karen and Heidorn, George E and Richardson, Stephen D},
  volume={196},
  year={2012},
  publisher={Springer Science \& Business Media}
}

@book{farzindar2015natural,
  title={Natural language processing for social media},
  author={Farzindar, Atefeh and Inkpen, Diana and Hirst, Graeme},
  year={2015},
  publisher={Springer}
}

@article{friedman1999natural,
  title={Natural language processing and its future in medicine},
  author={Friedman, Carol and Hripcsak, George and others},
  journal={Acad Med},
  volume={74},
  number={8},
  pages={890--5},
  year={1999}
}

@article{feldman1999nlp,
  title={NLP meets the Jabberwocky: Natural language processing in information retrieval},
  author={Feldman, Susan},
  journal={ONLINE-WESTON THEN WILTON-},
  volume={23},
  pages={62--73},
  year={1999},
  publisher={Citeseer}
}

@book{deng2018deep,
  title={Deep learning in natural language processing},
  author={Deng, L},
  year={2018},
  publisher={Springer}
}

@book{vajjala2020practical,
  title={Practical natural language processing: a comprehensive guide to building real-world NLP systems},
  author={Vajjala, Sowmya and Majumder, Bodhisattwa and Gupta, Anuj and Surana, Harshit},
  year={2020},
  publisher={O'Reilly Media}
}

@article{brants2003natural,
  title={Natural Language Processing in Information Retrieval.},
  author={Brants, Thorsten},
  journal={CLIN},
  volume={111},
  pages={1--13},
  year={2003}
}

@article{meurers2012natural,
  title={Natural language processing and language learning},
  author={Meurers, Detmar},
  journal={The encyclopedia of applied linguistics},
  pages={1--15},
  year={2012},
  publisher={Wiley Online Library}
}

@book{manning1999foundations,
  title={Foundations of statistical natural language processing},
  author={Manning, Christopher and Schutze, Hinrich},
  year={1999},
  publisher={MIT press}
}

@article{iroju2015systematic,
  title={A systematic review of natural language processing in healthcare},
  author={Iroju, Olaronke G and Olaleke, Janet O},
  journal={International Journal of Information Technology and Computer Science},
  volume={8},
  number={8},
  pages={44--50},
  year={2015},
  publisher={Citeseer}
}

@book{kao2007natural,
  title={Natural language processing and text mining},
  author={Kao, Anne and Poteet, Steve R},
  year={2007},
  publisher={Springer Science \& Business Media}
}

@article{wu2020deep,
  title={Deep learning in clinical natural language processing: a methodical review},
  author={Wu, Stephen and Roberts, Kirk and Datta, Surabhi and Du, Jingcheng and Ji, Zongcheng and Si, Yuqi and Soni, Sarvesh and Wang, Qiong and Wei, Qiang and Xiang, Yang and others},
  journal={Journal of the American Medical Informatics Association},
  volume={27},
  number={3},
  pages={457--470},
  year={2020},
  publisher={Oxford University Press}
}

@article{pons2016natural,
  title={Natural language processing in radiology: a systematic review},
  author={Pons, Ewoud and Braun, Loes MM and Hunink, MG Myriam and Kors, Jan A},
  journal={Radiology},
  volume={279},
  number={2},
  pages={329--343},
  year={2016},
  publisher={Radiological Society of North America}
}

@inproceedings{ruder2019transfer,
  title={Transfer learning in natural language processing},
  author={Ruder, Sebastian and Peters, Matthew E and Swayamdipta, Swabha and Wolf, Thomas},
  booktitle={Proceedings of the 2019 conference of the North American chapter of the association for computational linguistics: Tutorials},
  pages={15--18},
  year={2019}
}

\end{document}